\documentclass{report}
\usepackage{appendix}
\usepackage[utf8]{inputenc}
\usepackage{graphicx}
\usepackage{caption}
\usepackage{subcaption}
\usepackage{wrapfig}
\usepackage{tikz}
\usepackage[outline]{contour}
\usepackage{listofitems}
\usepackage{microtype}
\usepackage{etoolbox}
\usepackage{hyperref}

\usepackage{amsmath, amssymb}
\usepackage{bm}
\usepackage{nicematrix}
\usepackage{mathtools}
\usepackage{chemformula}
\usepackage{siunitx}
\usepackage{enumitem}
\usepackage{breqn}
\usepackage{algorithm}
\usepackage{algpseudocode}

\usepackage{times}
\usepackage{setspace}
\usepackage[maxbibnames = 5, maxcitenames = 2, mincitenames = 1, sorting = none]{biblatex}
\usepackage[margin=1.25in]{geometry}
\usetikzlibrary{positioning}
\usetikzlibrary{fit}
\usetikzlibrary{graphs, graphs.standard, quotes}

\newcommand\numberthis{\addtocounter{equation}{1}\tag{\theequation}}

\newcommand{\vast}{\bBigg@{3}}
\makeatother

\usepackage{color}

\DeclareMathOperator*{\Extr}{Extr}
\DeclareMathOperator*{\argmax}{argmax}

\DeclareMathOperator*{\sign}{sign}

\DeclareMathOperator{\real}{Re}

\DeclareMathOperator{\Prob}{P}

\DeclareMathOperator{\diag}{diag}

\begin{document}

\begin{titlepage}
\begin{center}
    \vspace*{1cm}
    \includegraphics[width=0.4\textwidth]{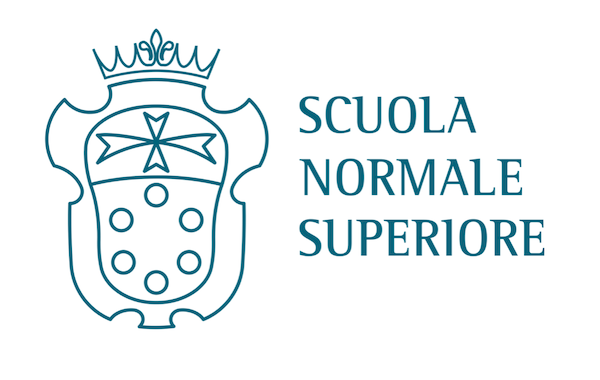}

    \vspace{0.5cm}

    \Large
    Faculty of Science\\
    PhD course in\\
    Computational methods and mathematical models for sciences and finance\\
    \vspace{0.5cm}
    Cycle XXXVII\\

    \vspace{1.5cm}
    \Huge
    \textbf{Explaining Machine Learning and Memorization with Statistical Mechanics}

    \vspace{1cm}
    \Large
    Scientific Disciplinary Sector INF/01

    \vspace{1.5cm}
    
    \textbf{Candidate: Robin Thériault}
    
    Supervisor: Prof. Daniele Tantari\\
    Coordinator: Prof. Michele Benzi
    
    \vspace{0.5cm}
            
    Academic year 2024/2025
\end{center}
\end{titlepage}

\section*{Acknowledgments}
I would like to thank a few people without whom this thesis would not have been possible. My PhD supervisor, Daniele, for guiding me throughout the thesis with impeccable patience and rigor while also giving me a lot of freedom to choose the topics. My PhD coordinator, Michele, for kindly advising me and overseeing my progress. My colleague Francesco for our fruitful collaboration. My partner, Daniela, for supporting me unconditionally and encouraging me to go out of my comfort zone. My parents, Lyne and Charles, for fostering my curiosity and encouraging me to pursue my goals. My brother, Lucas, for keeping in touch and playing games with me despite the physical distance and the time constraints. My McGill supervisor, Paul, for having introduced me to scientific research.

\begin{abstract}
    Artificial neural networks (NNs) and machine learning (ML) algorithms are poorly understood from a theoretical perspective, which makes it difficult to fully realize their potential and overcome their weaknesses. For instance, ML algorithms train NN weights by moving them along a low-dimensional subspace of their allowed values, but this implicitly low-dimensional learning structure is not properly exploited to improve training because its nature is not well understood. Moreover, trained NNs are easily confused by pervasive adversarial attacks whose theoretical underpinnings are still unclear. This thesis aims to improve our theoretical understanding of NNs and ML, with a particular focus on adversarial attacks and implicitly low-dimensional learning. For this purpose, we use mathematical tools from statistical mechanics to study different types of NNs and ways in which they can fit the data. In particular, we study two classes of models that fit the data with various degrees of learning and memorization: dense associative memory (DAM) and restricted Boltzmann machines (RBM). In the process, we investigate connections between different versions of these models that are useful to make analytical investigations more efficient.

    First, we study a type of DAM called dense Hopfield network (dense HN) in the teacher-student setting where it is trained using data generated by another dense HN. On the Nishimori line, we show that the phase where dense HNs in the teacher-student setting are able to learn data coincides with the spin-glass phase of dense HNs with random memorized patterns. Outside the Nishimori line, we investigate the noise tolerance and adversarial robustness of dense HNs. In particular, we derive an exact formula for the adversarial robustness of the student at zero temperature, and we clarify why the adversarial robustness of dense HNs changes as a function of the learning regime.

    Second, we study RBMs in the teacher-student setting. When the teacher's weights are uncorrelated, we validate the conjecture that the performance of the student in learning them is independent of the number of hidden units. Moreover, we show that a student that is larger than necessary to learn the teacher's weights adopts a low-dimensional learning strategy in which only a subset of its hidden units end up correlated with those of the teacher, which we argue can be used as a toy model for studying the lottery ticket hypothesis. When the teacher's weights are correlated together rather than purely random, we show that the student crosses multiple regimes of data representation where it learns them in increasingly detailed ways as the number of samples in its training dataset increases.

    Finally, we study a type of RBM that belongs to the class of DAMs and is capable of both supervised and unsupervised classification. As before, our methods are based on statistical mechanics calculations in the teacher-student setting. We propose a novel regularization scheme inspired by these calculations, which we find to make training on real data significantly more stable. Moreover, we show that the weights learned by relatively small DAMs trained on both real and synthetic data are saddle points of larger DAMs, and we implement an algorithm that uses this hierarchy to significantly accelerate training on real data.
\end{abstract}

\tableofcontents

\chapter{Introduction}
\tikzstyle{netnode}=[thick, draw = black, circle, minimum size = 22]
\tikzset{
>=latex,
connect/.style={thick},
connect arrow/.style={-{Latex[length=4,width=3.5]}, thick, shorten <=0.5, shorten >=1}
}
\contourlength{1.4pt}

\section{Background}
\label{sec:background}
Artificial Neural Networks (NNs) and Machine Learning (ML) algorithms have, with stunning speed, gone from niche tools to being the state-of-the-art in solving numerous fundamental and practical problems. To name a few of their achievements, they have long surpassed humans in image recognition tasks \cite{dan2011committee}, they are able to emulate written language so well that it is possible to have an open-ended conversation with them \cite{openai2024gpt4}, and they can predict the structure of complex molecules such as proteins \cite{jumper2021highly, Abramson2024accurate}, which has promising applications in drug discovery and in medicine as a whole \cite{yang2023alpha}. In the process, they have also become much larger and more sophisticated, to the point where even models with hundreds of millions of parameters such as latent diffusion \cite{Rombach2022high} are considered compact by today's standards. This level of complexity makes their behavior very challenging to explain theoretically. However, the general idea behind them is relatively simple.

In essence, Machine Learning (ML) is a set of computer algorithms for fitting high-dimensional data $\mathbf{x}$ with a function $F_{\mathbf{W}}$ by adjusting its parameters $\mathbf{W}$, which are also known as the function's \textit{weights}. This process is commonly refer to as \textit{training} or \textit{learning} the weights. Here, the term high-dimensional data means data with many components $x_1, x_2, ..., x_N$, which are usually concrete \textit{features} of the data. For example, digital grayscale images are a type of high-dimensional data whose components are the intensities of their pixels.
An artificial Neural Network (NN) is a function $F_{\mathbf{W}}$ whose weights $\mathbf{W}$ can be interpreted as the strengths of connections in a graph or network. For example, the linear model $y = \sum_{i = 1}^N w_i x_i$ of $\mathbf{x} = \left( x_1, x_2, ..., x_N \right)$ is an NN whose weights $w_1, w_2, ..., w_N$ can be ``machine learned'' from a set of high-dimensional data points $\left( \mathbf{x}^1, y^1 \right), \left( \mathbf{x}^2, y^2 \right), ..., \left( \mathbf{x}^M, y^M \right)$ using linear regression (see Fig. \ref{fig:linear_regression}). Intuitively, each connection $i$ is like a pipe with a width of $w_i$ that controls how much of the input component $x_i$ flows to the output component $y$ (see Fig. \ref{fig:NN_diagrams}, left panel). In general, NNs have a more sophisticated network of connections, some of which perform non-linear operations (see Fig. \ref{fig:NN_diagrams}, right panel), which is what makes them much more complicated and powerful than the linear model \cite{HORNIK1989feedforward}. The nodes at the junctions of these NN connections are typically called \textit{neurons} or \textit{units}.

\begin{figure}
    \centering
    \includegraphics[width=0.75\linewidth]{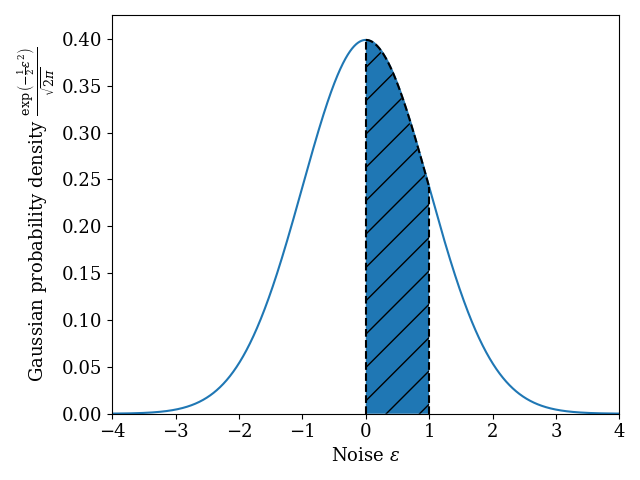}
    \caption{Illustration of the probability density distribution $\frac{\exp \left( -\frac{1}{2 \sigma^2} \varepsilon^2 \right)}{\sqrt{2\pi \sigma^2}}$ of Gaussian noise $\varepsilon$ with $\sigma^2 = 1$. The probability that $\varepsilon$ falls between two values is equal to the area under the curve between these two values. For example, the probability that $\varepsilon$ ends up between $0$ and $1$ is given by the area of the hatched blue region. Reducing $\sigma^2$ makes the central peak of the distribution narrower and higher, and thus the noise $\varepsilon$ more likely to be small.}
    \label{fig:gaussian_distribution}
\end{figure}

ML algorithms are particularly useful for learning probability distribution functions that describe how data points are placed relative to each other. The goal of such models is usually to generate new data points that are close enough to the existing ones to appear realistic, such as coherent sentences in the case of (large) language models \cite{openai2024gpt4}. Simpler NNs are also used to predict a noisy output from a deterministic input, such as for categorizing an image into one of many classes \cite{lecun1998gradient}. Learning a generic function and a probability distribution are two sides of the same coin. For example, linear regression is arguably equivalent to fitting data $\left( \mathbf{x}, y_{\text{data}} \right)$ with the ``deterministic'' distribution $p \left( y \mid \mathbf{x} \right) = \delta \left( y - \sum_{i = 1}^N w_i x_i \right)$ such that $y$ is equal to $\sum_{i = 1}^N w_i x_i$ with probability $1$ and cannot take any other values (see Fig. \ref{fig:linear_regression}, right panel). On the flip side, we can also fit the data with a proper probability distribution such as the Gaussian distribution $p \left( y \mid \mathbf{x} \right) = \frac{1}{\sqrt{2\pi \sigma^2}} \exp \left( -\frac{1}{2 \sigma^2} \left[ y - \sum_{i = 1}^N w_i x_i \right]^2 \right)$ of $y$ given $\mathbf{x}$ (illustrated in Fig. \ref{fig:gaussian_distribution}), then use the resulting $w_i$ as the weights of the linear fit. Generating data points amounts to \textit{sampling} the probability distribution fit to the data. In the case of the linear model, it consists of calculating $y_{\text{model}} = \sum_{i = 1}^N w_i x_i$ for a given $\mathbf{x}$ and adding some random noise $\varepsilon$ to the result, for example according to the Gaussian distribution mentioned above (see Figs. \ref{fig:gaussian_distribution} and \ref{fig:linear_regression}). In this context, the \textit{variance} $\sigma^2$ represents the level of noise injected into $y$. When it is sufficiently small, the generated values of $y$ are practically indistinguishable from the $y_{\text{model}}$ obtained without adding noise.
\begin{figure}
    \centering
    \includegraphics[width=\linewidth]{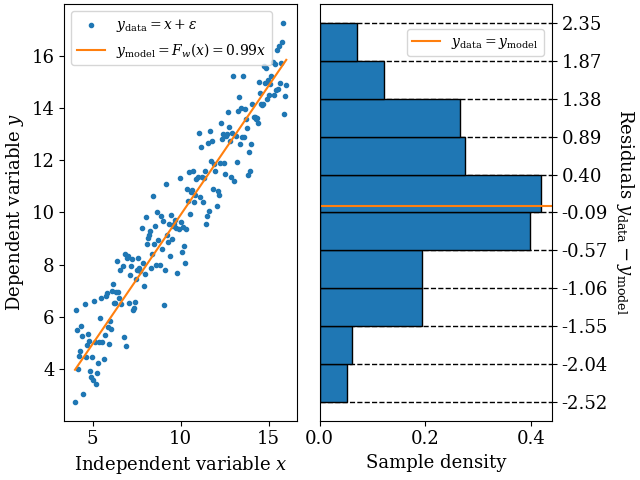}
    \caption{In the left panel, a linear function $y_{\text{model}} = F_w \left( x \right) = w x$ fit to data $y_{\text{data}} = x + \varepsilon$, where $\varepsilon$ is Gaussian noise with variance $\sigma^2 = 1$ (see Fig. \ref{fig:gaussian_distribution}). The learned weight is $w = 0.99$. In the right panel, the residual distribution of the data around the fit. The orange line $y_{\text{data}} = y_{\text{model}}$ can also be interpreted as a ``deterministic'' distribution fit to the data.}
    \label{fig:linear_regression}
\end{figure}
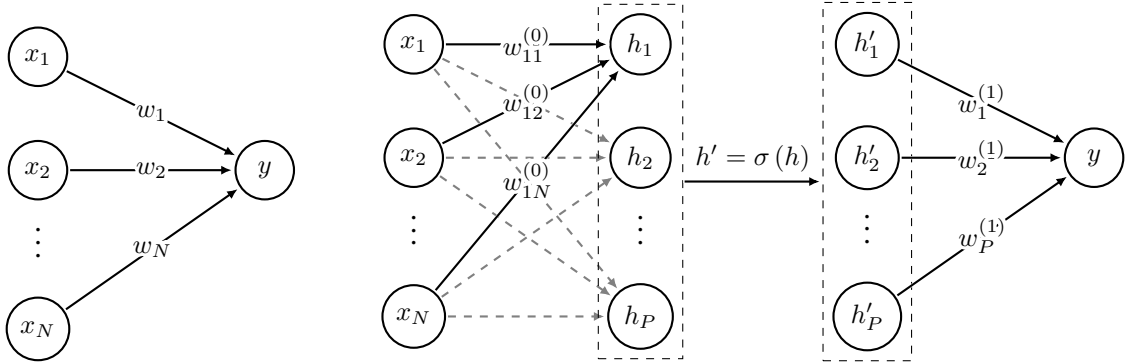
\begin{figure}
    \centering
    \begin{tikzpicture}[x = 3cm, y = 1.5cm]
        \def\NI{3} 
        \def\NO{1} 
        \def\yshift{0.4} 
      
        \foreach \i [evaluate={\c=int(\i==\NI); \y=\NI/2-\i-\c*\yshift; \index=(\i<\NI?int(\i):"N");}] in {1,...,\NI}{ 
            \node[netnode, outer sep = 0.6] (NI-\i) at (0, \y) {$x_{\index}$};
        }
    
        \foreach \i [evaluate={\c=int(\i==\NO); \y=\NO/2-\i; \index=(\i<\NO?int(\i):"P");}] in {\NO,...,1}{ 
        \node[netnode] (NO-\i) at (1,\y) {$y$};
            \foreach \j [evaluate={\index=(\j<\NI?int(\j):"N");}] in {1,...,\NI}{ 
            \draw[connect arrow, white, line width = 1.2] (NI-\j) -- (NO-\i);
            \draw[connect arrow] (NI-\j) -- (NO-\i)
            node[pos = 0.50] {\contour{white}{$w_{\index}$}};
            }
        }
        
        \path (NI-\NI) --++ (0,1+\yshift) node[midway,scale=1.2] {$\vdots$};
    \end{tikzpicture}\hspace{30pt}
    \begin{tikzpicture}[x = 3cm, y = 1.5cm]
        \def\NI{3} 
        \def\NH{3} 
        \def\NP{3} 
        \def\NO{1} 
        \def\yshift{0.4} 
      
        \foreach \i [evaluate={\c=int(\i==\NI); \y=\NI/2-\i-\c*\yshift; \index=(\i<\NI?int(\i):"N");}] in {1,...,\NI}{ 
            \node[netnode, outer sep = 0.6] (NI-\i) at (0, \y) {$x_{\index}$};
        };

        \def\fitlist{}
        \foreach \i [evaluate={\c=int(\i==\NH); \y=\NH/2-\i-\c*\yshift; \index=(\i<\NH?int(\i):"P");}] in {\NH,...,1}{ 
        \ifnum\i=1 
            \node[netnode]
            (NH-\i) at (1,\y) {$h_{\index}$};
            \foreach \j [evaluate={\index=(\j<\NI?int(\j):"N");}] in {1,...,\NI}{ 
            \draw[connect, white, line width=1.2] (NI-\j) -- (NH-\i);
            \draw[connect arrow] (NI-\j) -- (NH-\i)
            node[pos=0.50] {\contour{white}{$w_{1 \index}^{(0)}$}};
            }
        \else 
            \node[netnode]
            (NH-\i) at (1,\y) {$h_{\index}$};
            \foreach \j in {1,...,\NI}{ 
            \draw[connect arrow, dashed, gray] (NI-\j) -- (NH-\i);
            }
            \fi
        \xdef\fitlist{\fitlist(NH-\i)}
        };
        \node[draw, dashed, black, fit = \fitlist](NH){};

        \def\fitlist{}
        \foreach \i [evaluate={\c=int(\i==\NP); \y=\NP/2-\i-\c*\yshift; \index=(\i<\NP?int(\i):"P");}] in {1,...,\NP}{ 
            \node[netnode, outer sep = 0.6] (NP-\i) at (2,\y) {$h_{\index}^\prime$};
            \xdef\fitlist{\fitlist(NP-\i)}
        }
        \node[draw, dashed, black, fit = \fitlist](NP){};

        \draw[connect, white, line width=1.2] (NH) -- (NP);
        \draw[connect arrow] (NH) -- (NP) node[above, pos=0.50] {\contour{white}{$h^\prime = \sigma \left( h \right)$}
        };
    
        \foreach \i [evaluate={\c=int(\i==\NO); \y=\NO/2-\i; \index=(\i<\NO?int(\i):"1");}] in {\NO,...,1}{ 
        \node[netnode] (NO-\i) at (3,\y) {$y$};
            \foreach \j [evaluate={\index=(\j<\NP?int(\j):"P");}] in {1,...,\NP}{ 
            \draw[connect arrow, white, line width = 1.2] (NP-\j) -- (NO-\i);
            \draw[connect arrow] (NP-\j) -- (NO-\i)
            node[pos = 0.50] {\contour{white}{$w^{(1)}_{\index}$}};
            }
        }
        \path (NI-\NI) --++ (0,1+\yshift) node[midway,scale=1.2] {$\vdots$};
        \path (NH-\NH) --++ (0,1+\yshift) node[midway,scale=1.2] {$\vdots$};
        \path (NP-\NP) --++ (0,1+\yshift) node[midway,scale=1.2] {$\vdots$};
    \end{tikzpicture}
    \caption{In the left panel, the network representation of the linear model $y = \sum_{i = 1}^N w_i x_i$. In the right panel, the network representation of $y = \sum_{\mu = 1}^P w_{\mu}^{(1)} \sigma \left( \sum_{i = 1}^N w_{\mu i}^{(0)} x_i \right)$, where $\sigma$ is a non-linear operation, also known as an activation function. Both models are Neural Networks (NNs), but the right-panel one can fit much more complicated data.}
    \label{fig:NN_diagrams}
\end{figure}

Due to the difficulty in studying them theoretically and how quickly they have evolved, ML and NNs are still poorly understood, making it difficult to fully realize their potential and overcome their weaknesses. On the one hand, it was observed that ML algorithms train NN weights by moving them along
a low-dimensional subspace, or \textit{manifold}, of their allowed values \cite{frankle2018lottery, mao2024training}, which could potentially be used to design more efficient algorithms if it were better understood. To use an analogy, it is more efficient to locate objects on the surface of the Earth with the two-dimensional latitude-longitude coordinate system than with the three-dimensional Cartesian coordinate system. For simplicity, we call this phenomenon \textit{implicitly low-dimensional learning} for the rest of the Introduction.

\begin{figure}
    \centering
    \includegraphics[width=0.75\linewidth]{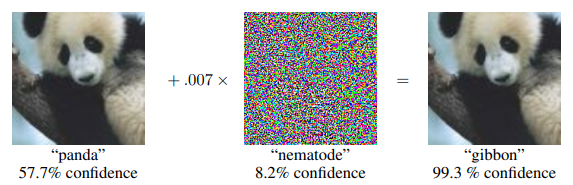}
    \caption{An adversarial attack making an Neural Network (NN) trained on the ImageNet dataset \cite{deng2009imagenet} recognize a panda as a gibbon. This plot comes from \cite{goodfellow2015explaining}, which contains additional details about the adversarial attack and NN and in question.}
    \label{fig:panda_adversarial}
\end{figure}

On the other hand, NNs that classify data with high accuracy can often be fooled by modifying the data with carefully crafted perturbations that are either meaningless or completely invisible to humans \cite{biggio2013evasion, szegedy2013intriguing} (see Fig. \ref{fig:panda_adversarial} for an example). This data corruption process, called an \textit{adversarial attack}, can be dangerous in some cases. For example, it can confuse the traffic sign recognition algorithms of self-driving vehicles \cite{pavlitska2023adversarial}. Adversarial attacks are not well understood, which makes them hard to circumvent.

In the same way that a bridge model made of ice cream sticks can give a rough idea of the physics of a real bridge, many aspects of ML and NNs can be understood by studying simpler high-dimensional models. For example, \cite{goldt2020modelling, gerace2021generalization} modeled high-dimensional data as lying on a low-dimensional manifold and \cite{tanner2024high, vilucchio2024geometry, vilucchio2025existence} studied adversarial attacks in linear models. The calculations and theoretical modeling in both of these two lines of work are based on statistical mechanics, and this is not a coincidence.
In fact, the field of statistical physics (also known as statistical mechanics) was originally developed to study phase transitions in materials with a large number of particles, but it can also be used to describe data points with many components, datasets with many data points, NNs with many neurons, etc. In particular, when we change their properties and those of their training data, NNs can undergo phase transitions between different regimes of data representation (uninformative, universal, specialized, etc.) \cite{barbier2025optimal} analogous to those between the states of matter (gaseous, liquid, solid, etc.).
The scientific literature on the topic is rich and flourishing \cite{charbonneau2023spin}, recent highlights being the 2024 Nobel Prize awarded to John Hopfield and Geoffrey Hinton for their highly influential work on two models of NNs inspired by statistical physics: the Hopfield networks \cite{hopfield1982neural} and restricted Boltzmann machines \cite{hinton2002training}, respectively.

As suggested in the previous paragraph, Restricted Boltzmann Machines (RBMs) and Hopfield Networks (HNs) are simple enough to be studied with analytical calculations, yet still exhibit some of the most intriguing characteristics of modern ML and NNs. For example, RBMs and generalizations of HNs called \textit{dense Hopfield networks} \cite{chen1986high, psaltis1986nonlinear, baldi1987number, gardner1987multiconnected, horn1988capacities, krotov2016dense} can fit data with a rich underlying structure by undergoing a multistage learning process whose first stages were found to be low dimensional \cite{decelle2017spectral, decelle2018thermodynamics, bachtis2025cascade, boukacem2024waddington}, a property that was then exploited to accelerate RBM training \cite{bereux2025fast}. Moreover, dense HNs were observed to be either vulnerable or robust to adversarial attacks in different controlled scenarios \cite{krotov2018dense}, suggesting that it could be possible to improve our understanding of adversarial attacks by studying them.

RBMs and HNs fit data $\mathbf{x}$ using an energy function $H$, which is also known as a \textit{Hamiltonian}. In ML terminology, they are energy-based models. The energy function has weights analogous to those of other NNs, which we will denote as $\mathbf{J}$, and it is designed so that more realistic data points typically have lower energy than less realistic ones once the weights $\mathbf{J}$ have been determined. Intuitively, a physical object leaking energy with the environment, such as a ball rolling down a mountain, eventually converges to a configuration where its energy cannot decrease anymore, which in the case of the ball corresponds to standing still at the bottom of a valley. Such a state is called an \textit{equilibrium configuration} or a \textit{local minimum} of the energy. The local minima $\mathbf{x}$ of an energy-based model with given weights $\mathbf{J}$ can be found analogously, and these states generally correspond to realistic data provided that the model and the strategy used to find $\mathbf{J}$ are appropriate to represent its underlying structure \cite{hopfield1982neural, hinton2002training, song2021score}. For example, the minimum energy configurations of HNs are typically preexisting data points stored in the weights by a ``Machine Memorization'' (MM) procedure (described in Section \ref{sec:hopfield_networks}), and RBMs can learn to generate new data on which they were not trained.
\begin{figure}
    \centering
    \includegraphics[width=0.75\linewidth]{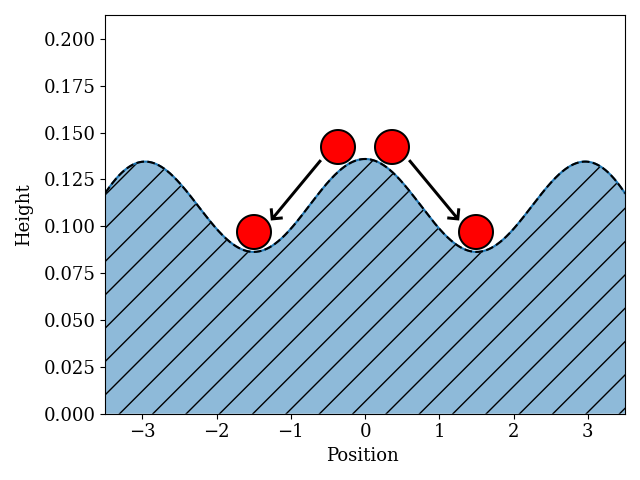}
    \caption{Sketch of two balls sliding down a mountain. At equilibrium, they stand motionless at the bottom of the valleys surrounding the mountain, which are local minima of the gravitational energy.}
    \label{fig:energy_landscape}
\end{figure}

This thesis studies the regimes of data representation and the phase transitions of RBMs and dense HNs \cite{chen1986high, psaltis1986nonlinear, baldi1987number, gardner1987multiconnected, horn1988capacities, krotov2016dense, ramsauer2020hopfield} to advance our understanding of MM, ML and NNs, with a focus on implicitly low-dimensional learning and adversarial attacks. Our methods are based on statistical mechanics calculations that are enabled by the relative simplicity of these energy-based models. On the way towards this objective, we investigate deep relationships between different versions of these models and their phase transitions \cite{barra2017phase, decelle2021inverse, albanese2022replica, hou2019minimal} that considerably simplify some of the calculations \cite{alemanno2023hopfield, manzan2025effect}.

The rest of this Chapter aims to introduce readers with a background in probability theory and physics to the models that we study and the tools that we use to do so. In Section \ref{sec:hopfield_networks}, we use HNs as an example to present the general idea behind energy-based models. In Section \ref{sec:models}, we build upon this foundation to describe RBMs and dense HNs. In Section \ref{sec:tools}, we present the teacher-student setting that we use in our analytical calculations and explain what kind of results can be derived with it.
Finally, Section \ref{sec:outline} uses the notions presented in Sections \ref{sec:hopfield_networks}, \ref{sec:models} and \ref{sec:tools} to put together a comprehensive outline of the other Chapters of the thesis, which present the results of our research.

Here is a high-level preview of these results. In Chapter \ref{chap:dense_HN_paper}, we study the adversarially vulnerable and adversarially robust regimes of dense HNs using a formal relationship between MM and ML that we investigate in detail \cite{theriault2024dense}. In Chapter \ref{chap:RBM_paper}, we study the various learning strategies of RBMs as a function of the properties of the training data \cite{theriault2025modeling}. In particular, we show how RBMs that are larger than necessary to learn the data adopt an implicitly low-dimensional learning strategy. In Chapter \ref{chap:DAM_paper}, we develop a theory of implicitly low-dimensional learning in a kind of RBM that belongs to the class of \textit{Dense Associative Memory} (DAM), which is a generalization of dense HNs \cite{theriault2026saddle}. We then use our theory to greatly accelerate training.

\section{The Hopfield network: a simple example of energy-based model}
\label{sec:hopfield_networks}
In this Section, we present HNs as simple examples of energy-based models. This exercise is useful for building intuition before tackling RBMs and dense HNs, which are more complicated.

\begin{figure}
    \centering
    \includegraphics[width=0.25\linewidth]{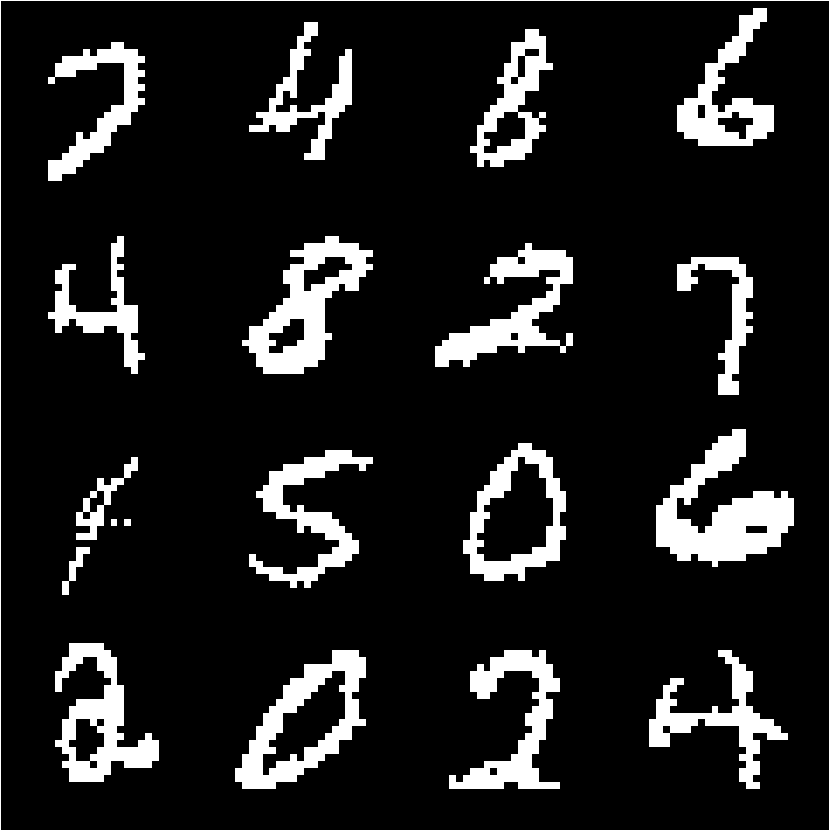}
    \hspace{40pt}
    \begin{tikzpicture}
        \node[netnode, outer sep = 0.6] (1) at (2, 0) {$\sigma_1$};
        \node[netnode, outer sep = 0.6] (2) at (0.61803399, 1.90211303) {$\sigma_2$};
        \node[netnode, outer sep = 0.6] (3) at (-1.61803399, 1.17557050) {$\sigma_3$};
        \node[netnode, outer sep = 0.6] (4) at (-1.61803399, -1.17557050) {$\sigma_4$};
        \node[netnode, outer sep = 0.6] (5) at (0.61803399, -1.90211303) {$\sigma_5$};

        \foreach \j [evaluate={\index=int(\j);}] in {2,...,5}{
            \draw[connect, white, line width=1.2] (\j) -- (1);
            \draw[connect] (\j) -- (1)
            node[pos=0.50] {\contour{white}{$J_{1 \index}$}};
        }
        
        \foreach \i in {2,...,4}{
            \pgfmathtruncatemacro\result{\i+1}
            \foreach \j in {\result,...,5}
                \draw[connect, dashed, gray] (\j) -- (\i);
            }
    \end{tikzpicture}
    \caption{In the left panel, images from the binarized version \cite{salakhutdinov2008quantitative} of the MNIST dataset of handwrittern digits \cite{lecun1998gradient}. The pixels of these images have two possible values: one for white and the other for black. In the right panel, the network representation of the Hopfield Network (HN) Hamiltonian $H \left[ \boldsymbol{\sigma} ; \mathbf{J} \right] = -\sum_{i = 1}^N \sum_{j = i + 1}^N J_{i j} \sigma_i \sigma_j$ with $N = 5$. The connections $J_{i j}$ are bidirectional, so they are drawn as ordinary lines rather than arrows.}
    \label{fig:Hopfield_diagram}
\end{figure}
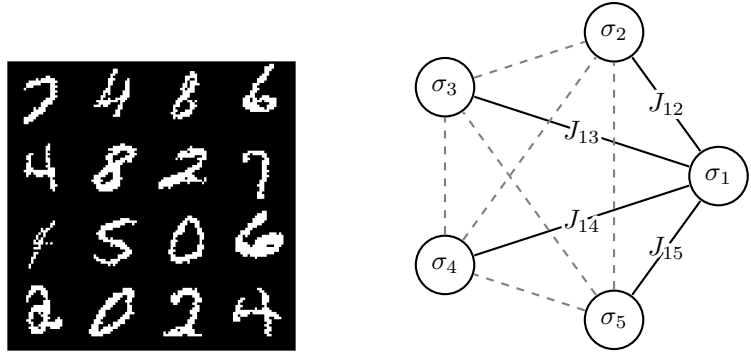

HNs use their units $\boldsymbol{\sigma} = \left\{ \sigma_i \right\}_{i = 1}^N = \left( \sigma_1, \sigma_2, ..., \sigma_N \right)$ to represent data whose components can take only two values, such as binary black-and-white images (see Fig. \ref{fig:Hopfield_diagram}, left panel). Without loss of generality, we take these two values to be $+1$ and $-1$. In this case, the HN Hamiltonian takes the form
\begin{equation}
    \label{eq:Hopfield_Hamiltonian}
    H_{\text{HN}} \left[ \boldsymbol{\sigma} ; \mathbf{J} \right] = -\sum_{i = 1}^N \sum_{j = i + 1}^N J_{i j} \sigma_i \sigma_j = -\sum_{i < j = 1}^N J_{i j} \sigma_i \sigma_j,
\end{equation}
where the weights $J_{i j}$ are the strengths of the connections between different components $\sigma_i$ and $\sigma_j$. Contrary to those of the NNs shown in Fig. (\ref{fig:NN_diagrams}), these connections are bidirectional: they are meant to represent how $\sigma_i$ and $\sigma_j$ are correlated rather than how one influences the other (see Fig. \ref{fig:Hopfield_diagram}, right panel). The sums in the Hamiltonian (Eq. \ref{eq:Hopfield_Hamiltonian}) are restricted to $j > i$ so that the contribution of every connection to the total energy is counted exactly once.

There are many ways to fit the weights $J_{i j}$ to data \cite{kanter1987associative, storkey1997increasing, barra2012equivalence, serricchio2025daydreaming}, the simplest one being Hebb's rule \cite{hopfield1982neural, hebb2005organization}
\begin{align*}
    \label{eq:Hebb}
    J_{i j} &= R_{i j} \left( \boldsymbol{\xi} \right) = \frac{1}{N} \sum_{\mu = 1}^P \xi^\mu_i \xi^\mu_j, \numberthis
\end{align*}
where $i < j$ and each pattern $\boldsymbol{\xi}^\mu = \left\{ \xi^\mu_i \right\}_{i = 1}^N$ in the set $\boldsymbol{\xi} = \left\{ \boldsymbol{\xi}^\mu \right\}_{\mu = 1}^P$ is a data point given to the model. This rule establishes strong positive connections between neurons $\sigma_i$ and $\sigma_j$ that consistently take the same value ($\sigma_i = \sigma_j$), strong negative connections between neurons that consistently take opposite values ($\sigma_i = -\sigma_j$) and weak connections between neurons that do not synchronize in either of these ways. It is often summarized by the mantra ``neurons that fire together, wire together. Neurons that do not sync, fail to link''.

HNs can retrieve the patterns stored in their weights, thus acting as models of associative memory \cite{hopfield1982neural} and making Hebb's rule a form of ``Machine Memorization'' (MM).
To be more precise, once its weights have been determined, an HN can be driven to local minima of the energy as described in \cite{hopfield1982neural}. When the weights are given by Hebb's rule with a sufficiently small number $P$ of patterns \cite{hopfield1982neural, amit1985storing}, the units $\boldsymbol{\sigma}$ usually converge to one of the patterns $\boldsymbol{\xi}^\mu$ or to a mixture of them \cite{amit1987statistical, amit1987information}. Conversely, when the number $P$ of patterns exceeds a specific threshold proportional to the number $N$ of components per data point, the HN typically no longer converges to meaningful results. In other words, it has limited storage capacity \cite{hopfield1982neural, amit1985storing}.

As we argued in Section \ref{sec:background}, learning a generic function and a probability distribution are two sides of the same coin. This observation suggests that we could learn the patterns $\boldsymbol{\xi}^\mu$ by fitting a probability distribution of the energy $H_{\text{HN}} \left[ \boldsymbol{\sigma} ; R \left( \boldsymbol{\xi} \right) \right] = -\frac{1}{N} \sum_{i < j = 1}^N \sum_{\mu = 1}^P \xi^\mu_i \xi^\mu_j \sigma_i \sigma_j$ (see Eqs. \ref{eq:Hopfield_Hamiltonian} and \ref{eq:Hebb}) to the data, as studied in \cite{decelle2021inverse, alemanno2023hopfield, barra2012equivalence}, rather than setting the patterns directly equal to the data. For this purpose, we turn to statistical mechanics.

In statistical mechanics, a physical system exchanging energy with the environment often does not converge to a single well-defined equilibrium configuration, but rather to an \textit{equilibrium distribution} of configurations such that more probable configurations have lower energy. The ambient temperature $T$ controls the level of noise in the distribution in the same way as the variance $\sigma^2$ in the Gaussian distribution (see Section \ref{sec:background}). In particular, as $T$ decreases, minimum-energy configurations become more probable. Let us illustrate this point using water. At high temperature, water molecules are randomly distributed in space as vapor. With decreasing temperature, they eventually condense into ice, which is their minimum-energy configuration. Formally, the equilibrium distribution of configurations $c$ with energy $H \left[ c \right]$ is proportional to $\exp \left( -\frac{1}{T} H \left[ c \right] \right)$ at temperature $T$. We say that they follow the \textit{(Boltzmann)-Gibbs} distribution, a property that is assumed to hold analogously for energy-based models. In HNs, this choice leads to the distribution
\begin{equation}
    \label{eq:prelim_Hopfield_distribution}
    \Prob \left( \boldsymbol{\sigma} \mid \boldsymbol{\xi} \right) = \frac{\exp \left( -\frac{1}{T} H_{\text{HN}} \left[ \boldsymbol{\sigma} ; R \left( \boldsymbol{\xi} \right) \right] \right)}{\sum_{\boldsymbol{\sigma}^\prime \in \left\{ -1, 1 \right\}^N} \exp \left( -\frac{1}{T} H_{\text{HN}} \left[ \boldsymbol{\sigma}^\prime ; R \left( \boldsymbol{\xi} \right) \right] \right)}
\end{equation}
for the state $\boldsymbol{\sigma}$ given the patterns $\boldsymbol{\xi}$, where $\left\{ -1, 1 \right\}^N$ is the set of all possible values of $\boldsymbol{\sigma}$, which we recall contain only $+1$'s and $-1$'s.
The denominator of Eq. (\ref{eq:prelim_Hopfield_distribution}) normalizes the distribution so that the total probability $\sum_{\boldsymbol{\sigma} \in \left\{ -1, 1 \right\}^N} \Prob \left( \boldsymbol{\sigma} \mid \boldsymbol{\xi} \right)$ is equal to $1$, and the temperature $T$ is an abstract quantity representing the level of noise injected into $\boldsymbol{\sigma}$. Following statistical mechanics conventions, we define the \textit{partition function} $Z \left( \boldsymbol{\xi} \right) = \sum_{\boldsymbol{\sigma}^\prime \in \left\{ -1, 1 \right\}^N} \exp \left( -\frac{1}{T} H_{\text{HN}} \left[ \boldsymbol{\sigma}^\prime ; R \left( \boldsymbol{\xi} \right) \right] \right)$ and the \textit{inverse temperature} $\beta = 1/T$. We then write the Gibbs distribution as
\begin{equation}
    \label{eq:Hopfield_distribution}
    \Prob_\beta \left( \boldsymbol{\sigma} \mid \boldsymbol{\xi} \right) = Z_\beta \left( \boldsymbol{\xi} \right)^{-1} \exp \left( -\beta H_{\text{HN}} \left[ \boldsymbol{\sigma} ; R \left( \boldsymbol{\xi} \right) \right] \right),
\end{equation}
where we use the subscript $\beta$ to explicitly mark the dependence of $\Prob \left( \boldsymbol{\sigma} \mid \boldsymbol{\xi} \right)$ and $Z \left( \boldsymbol{\xi} \right)$ on $\beta$. Now that we have derived this probability distribution, we explain how to learn the patterns. Note that the inverse temperature $\beta$ and the number of patterns $P$ are not learned. We call such parameters \textit{hyperparameters}.

A relatively intuitive way to fit Eq. (\ref{eq:Hopfield_distribution}) to a dataset $\mathcal{D} = \left\{ \boldsymbol{\sigma}^a \right\}_{a = 1}^M = \left( \boldsymbol{\sigma}^1, \boldsymbol{\sigma}^2, ..., \boldsymbol{\sigma}^M \right)$ is to find the patterns that make the dataset the most probable, a procedure known as maximum likelihood estimation. The probability that the Gibbs distribution (Eq. \ref{eq:Hopfield_distribution}) produces a specific data point $\boldsymbol{\sigma}^a$ from such a dataset is $\Prob_\beta \left( \boldsymbol{\sigma}^a \mid \boldsymbol{\xi} \right)$. Similarly, the probability that it generates the whole dataset is $\Prob_\beta \left( \mathcal{D} \mid \boldsymbol{\xi} \right) = \prod_{a = 1}^M \Prob_\beta \left( \boldsymbol{\sigma}^a \mid \boldsymbol{\xi} \right) = \Prob_\beta \left( \boldsymbol{\sigma}^1 \mid \boldsymbol{\xi} \right) \times \Prob_\beta \left( \boldsymbol{\sigma}^2 \mid \boldsymbol{\xi} \right) \times ... \times \Prob_\beta \left( \boldsymbol{\sigma}^M \mid \boldsymbol{\xi} \right)$. Maximum likelihood estimation amounts to finding the patterns $\boldsymbol{\xi}$ that maximize $\Prob_\beta \left( \mathcal{D} \mid \boldsymbol{\xi} \right)$. It is usually formulated as maximizing the \textit{log likelihood}
\begin{equation}
    \label{eq:log-likelihood}
    L \left( \boldsymbol{\xi} \right) = \frac{1}{M} \log \Prob_\beta \left( \mathcal{D} \mid \boldsymbol{\xi} \right) = \frac{1}{M} \sum_{a = 1}^M \log \Prob_\beta \left( \boldsymbol{\sigma}^a \mid \boldsymbol{\xi} \right),
\end{equation}
which is mathematically equivalent and easier to handle numerically. Maximum likelihood estimation is popular for training energy-based models and NNs because there are various efficient ways to perform it numerically \cite{hinton2002training, rosenblatt1958perceptron, dempster1977maximum, ruder2017overview}, which we will not explain in detail here. In particular, it was used to train HNs in \cite{decelle2021inverse}. For historical reasons, maximum likelihood training of NNs is generally phrased as minimizing a \textit{loss function} equal to minus the log likelihood, and the shape of the loss as a function of the learnable parameters is called the \textit{loss landscape}.

Alternatively, we can also train the patterns $\boldsymbol{\xi}$ by sampling the \textit{posterior} distribution $\Prob_\beta \left( \boldsymbol{\xi} \mid \mathcal{D} \right)$, which represents the probability that an HN with patterns $\boldsymbol{\xi}$ generates the data $\mathcal{D}$. Although this algorithm is generally less efficient from a numerical point of view, it can be studied analytically to characterize the fundamental limits of NN training \cite{zdeborova2016statistical}, which has been applied to HNs \cite{alemanno2023hopfield}, RBMs \cite{hou2019minimal, manzan2025effect}, linear models \cite{gardner1989unfinished, gyorgyi1990first-order} and far beyond \cite{charbonneau2023spin}. The analytical calculations presented in this thesis follow this approach, which also ends up providing information about the loss landscape in Chapter \ref{chap:DAM_paper} \cite{theriault2026saddle}. To understand how to calculate $\Prob_\beta \left( \boldsymbol{\xi} \mid \mathcal{D} \right)$ from $\Prob_\beta \left( \mathcal{D} \mid \boldsymbol{\xi} \right)$, we first study a similar problem in a more familiar setting. Suppose that we know the probability $\Prob \left( \text{the ground is wet} \mid \text{it rains} \right)$ that the ground is wet when it rains. We want to know the probability $\Prob \left( \text{it rains} \mid \text{the ground is wet} \right)$ that it rains when the ground is wet to predict the weather. The probability that it rains \textit{and} that the ground is wet is equal to the probability that it rains when the ground is wet \textit{multiplied by} the probability that the ground is wet. Formally, this relationship is written as
\begin{align*}
    \Prob \left( \text{it rains}, \text{the ground is wet} \right) &= \Prob \left( \text{it rains} \mid \text{the ground is wet} \right) \Prob \left( \text{the ground is wet} \right).
\end{align*}
Crucially, it also works in the other direction, that is
\begin{align*}
    \Prob \left( \text{it rains}, \text{the ground is wet} \right) &= \Prob \left( \text{the ground is wet} \mid \text{it rains} \right) \Prob \left( \text{it rains} \right).
\end{align*}
Combining both of these equalities, we find
\begin{align*}
    \Prob \left( \text{it rains} \mid \text{the ground is wet} \right) &= \frac{\Prob \left( \text{the ground is wet} \mid \text{it rains} \right) \Prob \left( \text{it rains} \right)}{\Prob \left( \text{the ground is wet} \right)},
\end{align*}
which is known as Bayes' theorem. In the case of HNs, it gives
\begin{equation}
    \label{eq:Hopfield_posterior}
    \Prob_\beta \left( \boldsymbol{\xi} \mid \mathcal{D} \right) = \frac{\Prob_\beta \left( \mathcal{D} \mid \boldsymbol{\xi} \right) \Prob_\beta \left( \boldsymbol{\xi} \right)}{\Prob_\beta \left( \mathcal{D} \right)},
\end{equation}
where the \textit{prior} $\Prob_\beta \left( \boldsymbol{\xi} \right)$ represents the knowledge that we have about the distribution of the patterns before training and $\Prob_\beta \left( \mathcal{D} \right) = \sum_{\boldsymbol{\xi} \in \left\{ -1, 1 \right\}^{N \times P}} \Prob_\beta \left( \mathcal{D} \mid \boldsymbol{\xi} \right) \Prob_\beta \left( \boldsymbol{\xi} \right)$ normalizes the distribution in the same way as $Z_\beta \left( \boldsymbol{\xi} \right)$ in the Gibbs distribution (Eq. \ref{eq:Hopfield_distribution}). By analogy, we call the normalization constant $\Prob_\beta \left( \mathcal{D} \right)$ the \textit{posterior partition function} and write it as $\mathcal{Z}_\beta \left( \mathcal{D} \right)$.

\section{Restricted Boltzmann machines and dense Hopfield networks}
\label{sec:models}
As for HNs, RBMs and dense HNs are energy-based models whose corresponding Hamiltonians and Gibbs distributions can be fit on data using maximum likelihood estimation or posterior sampling. Here, we summarize the differences between them.

The generalized HNs that are nowadays called \textit{dense} were developed shortly after the original to improve upon its storage capacity \cite{chen1986high, psaltis1986nonlinear, baldi1987number, gardner1987multiconnected, horn1988capacities}. These models have been receiving renewed attention under the name \textit{dense associative memory} (DAM) since they were used for pattern recognition in \cite{krotov2016dense}. In the process, the meaning of the term dense associative was also broadened to encompass more general models with large storage capacities \cite{krotov2021large, krotov2025modern}, which were in turn related to various other ML paradigms \cite{ramsauer2020hopfield, hoover2023memory, ota2023attention, ambrogioni2024search, karakida2024hierarchical}. This evolution in terminology is the reason why we adopt the convention of calling the generalized HNs first considered in \cite{chen1986high, psaltis1986nonlinear, baldi1987number, gardner1987multiconnected, horn1988capacities} dense HNs rather than DAMs. The dense HN Hamiltonian is
\begin{gather*}
    \label{eq:DHN_Hamiltonian}
    H_{\text{DHN}} \left[ \boldsymbol{\sigma} ; \mathbf{J} \right] = -\sum_{i_1 < ... < i_p = 1}^N J_{i_1 ... i_p} \sigma_{i_1} ... \sigma_{i_p}, \numberthis \\
    \text{where} \quad J_{i_1 ... i_p} = R_{i_1 ... i_p} \left( \boldsymbol{\xi} \right) = \frac{p!}{N^{p-1}} \sum_{\mu = 1}^M \xi_{i_1}^\mu ... \xi_{i_p}^\mu.
\end{gather*}
The weights of Eq. (\ref{eq:DHN_Hamiltonian}) connect the units $\boldsymbol{\sigma} = \left\{ \sigma_i \right\}_{i = 1}^N = \left( \sigma_1, \sigma_2, ..., \sigma_N \right)$ within groups of size $p$. These connections, which are called \textit{$p$-body interactions}, reduce to those of the original HN when $p = 2$.\footnote{Except for a factor of two that can be removed by rescaling the inverse temperature.}

\begin{figure}
    \centering
    \includegraphics[width=0.35\linewidth]{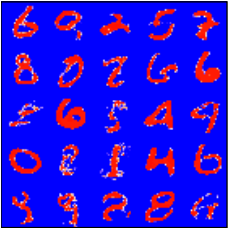}
    \caption{Patterns $\boldsymbol{\xi}^\mu$ learned by the dense Hopfield network (HN) studied in \cite{krotov2016dense} when trained on the MNIST dataset of handwritten digits. This plot comes from \cite{krotov2016dense}, which contains additional details about the dense HN in question.}
    \label{fig:prototypes}
\end{figure}

Dense HNs bridge the gap between ML and MM in the sense that they learn prototypes of the data \cite{krotov2016dense} when trained with an ML algorithm. For example, their patterns become prototypes of digits when trained on the MNIST dataset of handwritten digits (see Fig. \ref{fig:prototypes}). These dense HNs were observed to have a variable level of adversarial robustness as a function of $p$ \cite{krotov2018dense}, offering an interesting opportunity to study adversarial attacks and how to mitigate them.

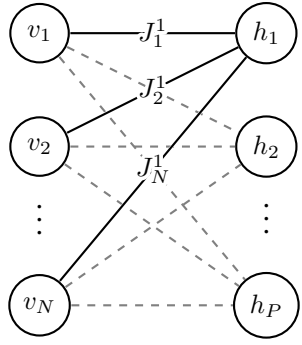
\begin{figure}
    \centering
    \begin{tikzpicture}[x = 3cm, y = 1.5cm]
        \def\NI{3} 
        \def\NH{3} 
        \def\yshift{0.4} 
      
        \foreach \i [evaluate={\c=int(\i==\NI); \y=\NI/2-\i-\c*\yshift; \index=(\i<\NI?int(\i):"N");}] in {1,...,\NI}{ 
            \node[netnode, outer sep = 0.6] (NI-\i) at (0, \y) {$v_{\index}$};
        };

        \def\fitlist{}
        \foreach \i [evaluate={\c=int(\i==\NH); \y=\NH/2-\i-\c*\yshift; \index=(\i<\NH?int(\i):"P");}] in {\NH,...,1}{ 
        \ifnum\i=1 
            \node[netnode]
            (NH-\i) at (1,\y) {$h_{\index}$};
            \foreach \j [evaluate={\index=(\j<\NI?int(\j):"N");}] in {1,...,\NI}{ 
            \draw[connect, white, line width=1.2] (NI-\j) -- (NH-\i);
            \draw[connect] (NI-\j) -- (NH-\i)
            node[pos=0.50] {\contour{white}{$J_{\index}^1$}};
            }
        \else 
            \node[netnode]
            (NH-\i) at (1,\y) {$h_{\index}$};
            \foreach \j in {1,...,\NI}{ 
            \draw[connect, dashed, gray] (NI-\j) -- (NH-\i);
            }
            \fi
        \xdef\fitlist{\fitlist(NH-\i)}
        };
        \path (NI-\NI) --++ (0,1+\yshift) node[midway,scale=1.2] {$\vdots$};
        \path (NH-\NH) --++ (0,1+\yshift) node[midway,scale=1.2] {$\vdots$};
    \end{tikzpicture}
    \caption{The network representation of the restricted Boltzmann machine (RBM) Hamiltonian $H \left[ \mathbf{v}, \mathbf{h} ; \mathbf{J} \right] = -\sum_{i = 1}^N \sum_{\mu = 1}^P J^{\mu}_i v_i  h_\mu$. The connections $J_i^\mu$ are bidirectional, so they are represented using ordinary lines instead of arrows.}
    \label{fig:RBM_diagram}
\end{figure}

Although they have $2$-body interactions like the original HN, RBMs differ from it in that they represent data on two levels of abstraction. Intuitively, these two levels are roughly analogous to how we see objects and how we describe them mentally (for example, thinking of a fire as ``hot'', ``red'' and ``bright''). In practice, however, the abstract concepts that RBMs associate with data are different from our own mental representations. Formally, the RBM Hamiltonian is
\begin{equation}
    \label{eq:RBM_Hamiltonian}
    H_{\text{RBM}} \left[ \mathbf{v}, \mathbf{h} ; \mathbf{J} \right] = -\sum_{i = 1}^N \sum_{\mu = 1}^P J^{\mu}_i v_i  h_\mu,
\end{equation}
where the visible layer $\mathbf{v} = \left\{ v_i \right\}_{i = 1}^N = \left( v_1, v_2, ..., v_N \right)$ is the data expressed in terms of its components, the hidden layer $\mathbf{h} = \left\{ h_\mu \right\}_{\mu = 1}^P$ is a high-level representation of the data in terms of abstract concepts, and the learnable weights $\mathbf{J} = \left\{ J^\mu_i \right\}_{1 \leq i \leq N}^{1 \leq \mu \leq P}$ are connections that the RBM makes between $\mathbf{v}$ and $\mathbf{h}$. As in HNs, these connections are bidirectional, which means that they represent how $\mathbf{v}$ and $\mathbf{h}$ are correlated rather than how one influences the other (see Fig. \ref{fig:RBM_diagram}).
The units $v_i$ and $h_\mu$ of the two layers can be bound to $\pm1$ as those of HNs, but various other choices are also possible. For example, they can be integers within an interval \cite{Salakhutdinov2007restricted} or arbitrary real numbers, thus defining multiple types of RBMs \cite{barra2012equivalence}. Every such RBM has a Gibbs distribution of the form
\begin{equation}
    \label{eq:RBM_distribution}
    \Prob_\beta \left( \mathbf{v}, \mathbf{h} \mid \mathbf{J} \right) = Z_\beta \left( \mathbf{J} \right)^{-1} \Prob_0 \left( \mathbf{v} \right) \Prob_0 \left( \mathbf{h} \right) \exp \left( -\beta H \left[ \mathbf{v}, \mathbf{h} ; \mathbf{J} \right] \right),
\end{equation}
where the inverse temperature $\beta$ regulates the strength of the interactions that $\mathbf{v}$ and $\mathbf{h}$ exchange through $H$, the priors $\Prob_0 \left( \mathbf{v} \right)$ and $\Prob_0 \left( \mathbf{h} \right)$ are the distributions of $\mathbf{v}$ and $\mathbf{h}$ when they are decoupled, i.e. $\beta = 0$, and the partition function $Z_\beta \left( \mathbf{J} \right)$ is again defined so that the total probability is equal to $1$. The priors on both layers, which represent the knowledge that we have on the data before training, usually have a simple form such that their units are (a priori) independent and identically distributed (i.i.d.) with respect to each other. For example, neurons restricted to $\pm1$ usually have a \textit{Rademacher} prior
\begin{align*}
    \Prob_0 \left( v_i \right) = \begin{cases}
        1/2 &\quad \text{if $v_i = \pm1$} \\
        0 &\quad \text{otherwise},
    \end{cases}
\end{align*}
and units that can be arbitrary real numbers usually have a Gaussian prior $\Prob_0 \left( v_i \right) = \frac{\exp \left( -\frac{1}{2 \sigma^2} v_i^2 \right)}{\sqrt{2\pi \sigma^2}}$. Simple priors with non-i.i.d. units are also possible. For example, RBM layers whose units are restricted to always be at a distance of $\sqrt{\sum_{i = 1}^N v_i^2} = 1$ from the origin can be described using the \textit{hyperspherical} prior
\begin{align*}
    \Prob_0 \left( \mathbf{v} \right) = \begin{cases}
        \frac{1}{\Omega_N \left( \beta \right)} &\quad \text{if $\sqrt{\sum_{i = 1}^N v_i^2} = 1$} \\
        0 &\quad \text{otherwise},
    \end{cases}
\end{align*}
where $\Omega_N \left( 0 \right)$ is the size of the region of space where $\sqrt{\sum_{i = 1}^N v_i^2} = 1$, which is called the unit \textit{hypersphere}. Additionally, the categorical (or \textit{Potts} \cite{potts1952some, wu1982potts}) prior
\begin{equation*}
    \Prob_0 \left( \mathbf{v} \right) = \begin{cases}
        \frac{1}{N + 1} &\quad \text{if $\mathbf{v} \in \left\{ 0, 1 \right\}^N$ and $\sum_{i = 1}^N v_i \in \left\{ 0, 1 \right\}$} \\
        0 &\quad \text{otherwise},
    \end{cases}
\end{equation*}
is used in RBM layers where at most one unit $v_i$ can be activated at once (equal to $1$), while all the others are turned off (equal to $0$).

Given that the priors have a simple form, using a higher temperature $T = 1/\beta$ encourages the RBM to learn simpler weights $J^\mu_i$ during training, which is known as \textit{regularization} in ML. Intuitively, regularization encourages NNs not to overthink problems with relatively simple solutions, which can help them perform better.

RBMs are trained by fitting to the data their marginal distribution $\Prob_\beta \left( \mathbf{v} \mid \mathbf{J} \right) = \int_{\mathbf{h}} d \mathbf{h} \Prob_\beta \left( \mathbf{v}, \mathbf{h} \mid \mathbf{J} \right)$, which embodies how they represent the data with all their hidden units simultaneously. The marginal distribution often does not have a simple closed form and must be approximated, such as by repeatedly sampling the conditional distributions $\Prob_\beta \left( \mathbf{h} \mid \mathbf{v}, \mathbf{J} \right)$ and $\Prob_\beta \left( \mathbf{v} \mid \mathbf{h}, \mathbf{J} \right)$ following the contrastive divergence algorithm introduced in \cite{hinton2002training}. In some cases, the marginal distributions of RBMs with specific priors reduce to the Gibbs distributions of other energy models \cite{Decelle2021restricted}. For example, the marginal distribution of an RBM with a Rademacher prior on its visible units and a Gaussian prior on its hidden units simplifies to
\begin{equation*}
    \Prob_\beta \left( \mathbf{v} \mid \mathbf{J} \right) = Z_\beta \left( \mathbf{J} \right)^{-1} \exp \left( \beta \sum_{i < j = 1}^N \sum_{\mu = 1}^P J^\mu_i J^\mu_j v_i v_j \right),
\end{equation*}
which is the Gibbs distribution of a Hopfield network with inverse temperature $\beta N$, memorized patterns $\mathbf{J}^\mu = \left\{ J^\mu_i \right\}_{i = 1}^N$ and units $\mathbf{v}$. In other words, RBMs with these priors are formally equivalent to HNs \cite{barra2012equivalence}. By analogy, the rows $\mathbf{J}^\mu$ of the weights are occasionally called patterns and written as $\boldsymbol{\xi}^\mu$ even for other choices of priors.

RBMs with a hyperspherical prior on their visible units and a categorical prior on their hidden units become \textit{Dense Associative Memory} (DAM) models that learn prototypes of the data like dense HNs. Formally, the marginal distribution of such an RBM simplifies to
\begin{equation}
    \label{eq:DAM_distribution}
    \Prob_\beta \left( \mathbf{v} \mid \mathbf{J} \right) = \frac{1}{P + 1} \sum_{\mu = 1}^P \frac{\exp \left( \beta \sum_{i = 1}^N J^\mu_i v_i \right)}{\Omega_N \left( \beta \right)} + \frac{1}{P + 1} \frac{1}{\Omega_N \left( 0 \right)},
\end{equation}
where the normalization constant $\Omega_N \left( \beta \right)$ is defined in Chapter \ref{chap:DAM_paper}. In that Chapter, we discuss such DAMs and their affinity for prototype learning in more detail.

\section{Statistical mechanics of networks in the teacher-student setting }
\label{sec:tools}
We now present the main theoretical tool that we use to study the statistical mechanics of RBMs and dense HNs: the \textit{teacher-student setting} \cite{gardner1989unfinished}. This tool has been used numerous times to solve various fundamental problems in the field of ML and NNs \cite{charbonneau2023spin}.

In the teacher-student setting, a \textit{teacher} NN with weights $\mathbf{J}^*$ generates a sample dataset $\mathcal{D}$ on which a \textit{student} NN then trains its own weights $\mathbf{J}$. In other words, the teacher shows the dataset $\mathcal{D}$ to the student to teach it how to mimic its contents. This framework provides a controlled environment for studying NN behavior with analytical calculations. Although NNs with suitable weights can represent arbitrarily complex data \cite{HORNIK1989feedforward}, the teacher weights $\mathbf{J}^*$ are usually determined according to a simple criterion to simplify calculations. For example, they can be fixed samples from a simple probability distribution. Moreover, in the case of dense HNs, we assume that the weights $\mathbf{J}^*$ and $\mathbf{J}$ decompose into patterns $\boldsymbol{\xi}^{* \mu}$ and $\boldsymbol{\xi}^{\nu}$ following the second line of Eq. (\ref{eq:DHN_Hamiltonian}).

In our work, the student has limited knowledge of the architecture of the teacher. To be more precise, it knows whether the teacher is a dense HN or an RBM and, in the latter case, the priors on the RBM's neurons. However, unless explicitly stated otherwise, it does not know the hyperparameters of the teacher, such as its inverse temperature $\beta^*$, its number of patterns $P^*$, its interaction order $p^*$, etc. Therefore, it adopts the same form as the teacher, but with hyperpameters that are generally different, which we write without an asterisk (*) to distinguish them from those of the teacher. Following statistical mechanics conventions, we call the special case where the student has the same hyperparameters as the teacher the \textit{Nishimori line} \cite{nishimori1980exact, nishimori2001statistical, contucci2009spin, iba1999nishimori}.

Given that the teacher and the student are two instances of the same type of model, the degree of similarity between their weights is a good indication of the performance of the student in learning the data. We measure this degree of similarity using the \textit{overlaps} $Q \left( \boldsymbol{\xi}^{* \mu}, \boldsymbol{\xi}^{\nu} \right) = \frac{1}{N} \sum_{i = 1}^N \xi^{* \mu}_i \xi^{\nu}_i$ between the patterns of the student and those of the teacher, which we recall are the rows $\mathbf{J}^\nu$ and $\mathbf{J}^{* \mu}$ of the weights for RBMs. In the case where all components $\xi^{* \mu}_i$ and $\xi^{\nu}_i$ of the patterns $\boldsymbol{\xi}^{* \mu}$ and $\boldsymbol{\xi}^{\nu}$ are equal to $\pm1$, $Q \left( \boldsymbol{\xi}^{* \mu}, \boldsymbol{\xi}^{\nu} \right) = 1$ means that the student learns the teacher pattern $\boldsymbol{\xi}^{* \mu}$ perfectly. We use other types of overlaps to probe various other aspects of learning. For example, the overlaps $Q \left( \boldsymbol{\xi}^{1 \mu}, \boldsymbol{\xi}^{2 \nu} \right) = \frac{1}{N} \sum_{i = 1}^N \xi^{1 \mu}_i \xi^{2 \nu}_i$ between two learning attempts $\boldsymbol{\xi}^{1} = \left\{ \xi^{1 \mu}_i \right\}_{1 \leq i \leq N}^{1 \leq \mu \leq P}$ and $\boldsymbol{\xi}^{2} = \left\{ \xi^{2 \mu}_i \right\}_{1 \leq i \leq N}^{1 \leq \mu \leq P}$ of the student represent its tendency to stay frozen in specific pattern configurations rather than visiting all possible values of $\boldsymbol{\xi}$. $Q \left( \boldsymbol{\xi}^{1 \mu}, \boldsymbol{\xi}^{2 \nu} \right)$ can sometimes be relatively large even when $Q \left( \boldsymbol{\xi}^{* \mu}, \boldsymbol{\xi}^{\nu} \right)$ is close to $0$, which represents the situation where the student mistakenly learns a very different pattern from that of the teacher, or in other words misunderstands what the teacher is trying to teach it. In general, the overlaps change with the hyperparameters of the teacher-student setting and the number of samples $M$ in the training dataset $\mathcal{D}$, defining different regimes of data representation.

Our theoretical analyses are based on computing mean overlaps, which are also known as \textit{order parameters}, using the \textit{replica method} (explained at length in \cite{nishimori2001statistical}). This method is well-established in the statistical mechanics community, where it is used to study a type of physical system called \textit{spin glass} \cite{charbonneau2023spin, nishimori2001statistical}. Spin glasses are characterized by disordered components that are perpetually frozen, or \textit{quenched}, as the system converges to equilibrium. In the case of NNs in the teacher-student setting, the quenched disorder is the patterns of the teacher and the dataset that it generates, which are kept fixed during learning. Conversely, the student patterns, which are allowed to evolve throughout training, are \textit{not} quenched. Although they can sometimes end up freezing in disordered \textit{spin-glass} states characterized by $Q \left( \boldsymbol{\xi}^{* \mu}, \boldsymbol{\xi}^{\nu} \right) \approx 0$ and $Q \left( \boldsymbol{\xi}^{1 \mu}, \boldsymbol{\xi}^{2 \nu} \right) > 0$, it is a consequence of the learning task rather than a hard constraint.

\begin{figure}
    \centering
    \includegraphics[width=0.6\linewidth]{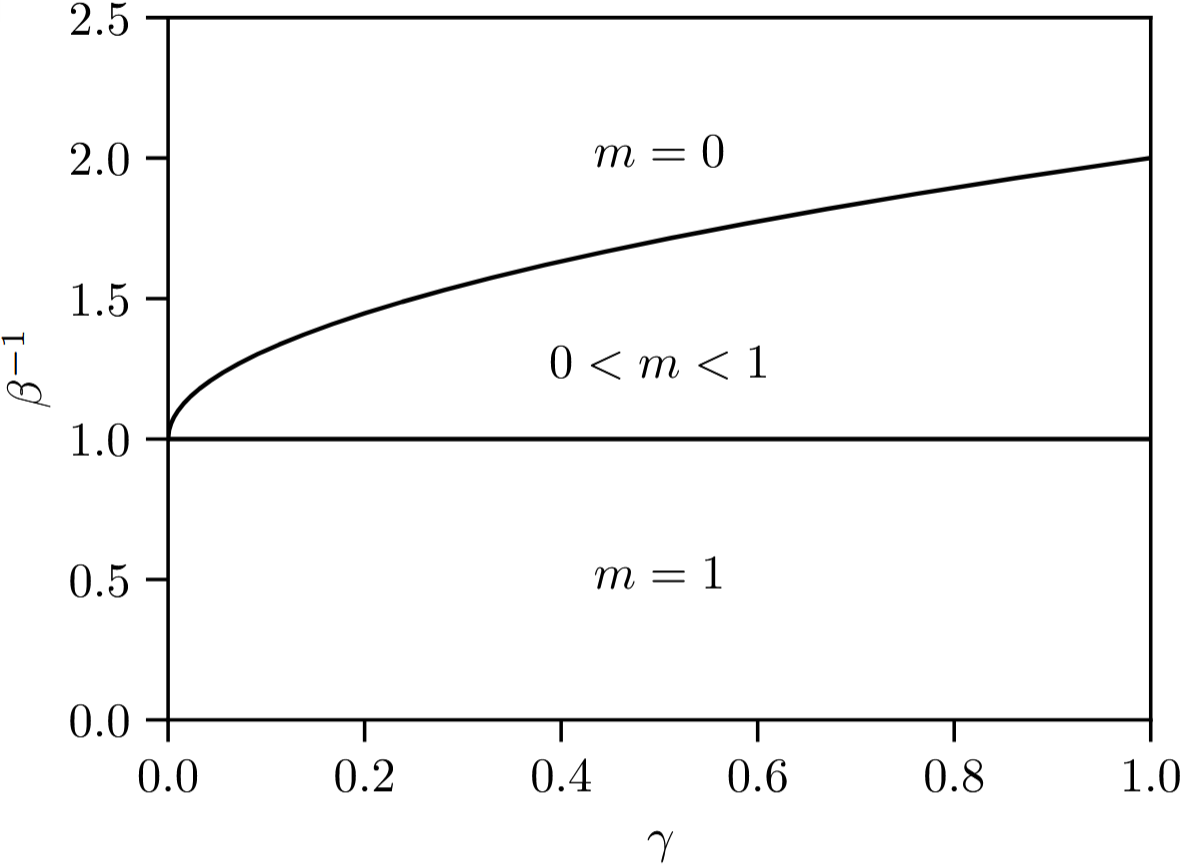}
    \caption{Phase diagram of Hopfield networks (HNs) in the teacher-student setting when the student and the teacher have a single pattern and the same inverse temperature $\beta = \beta^*$. The axes measure the amount of data $\gamma = M/N$ and the temperature $T = 1/\beta$. The phases, separated by black lines, are explained in the text. This plot comes from \cite{alemanno2023hopfield}.}
    \label{fig:phase_diagram}
\end{figure}

We use these calculations to characterize the learning phases of RBMs and dense HNs as a function of the number of samples $M$ in the training dataset $\mathcal{D}$ and the hyperparameters of the teacher-student setting. In particular, we calculate phase diagrams representing the different phases and phase transitions of RBMs and dense HNs as a function of the hyperparameters and the size of the training dataset. For example, Fig. (\ref{fig:phase_diagram}), which is taken from \cite{alemanno2023hopfield}, is the phase diagram of HNs in the teacher-student setting when the student and the teacher have a single pattern and the same inverse temperature $\beta = \beta^*$. We see three phases and two phase transitions (black lines) as a function of the amount of data $\gamma = M/N$ and the temperature $T = 1/\beta$, which we will now explain in terms of the mean teacher-student overlap (subsequently $m$) and the mean overlap between the teacher pattern and the samples that it generates (subsequently $r$). Below the first phase transition, at the bottom of the plot, the temperature $T = 1/\beta$ is low enough for the teacher to generate highly informative training samples ($r > 0$) and for the student HN to perfectly learn the teacher pattern through them ($m = 1$). Between the two phase transitions, the training dataset is much noisier ($r = 0$), but the student can still learn the teacher pattern imperfectly ($0 < m < 1$). Finally, above the second phase transition, the temperature $T = 1/\beta$ is too high and the dataset too small for the student HN to learn anything about the teacher pattern ($m = 0$).

\section{Outline of the thesis}
\label{sec:outline}
In Chapter \ref{chap:dense_HN_paper} (based on a work with Daniele Tantari, published as \cite{theriault2024dense}), we study dense HNs with a single pattern in the teacher-student setting. In particular, we investigate the relationship between a dense HN with a single pattern trained using $M$ data points (the inverse model) and a dense HN with $M$ random memorized patterns (the direct model), the latter of which is well understood thanks to previous studies \cite{chen1986high, psaltis1986nonlinear, baldi1987number, gardner1987multiconnected, horn1988capacities, krotov2016dense, albanese2022replica}. To be more precise, we show analytically that the phase where the inverse model with $\beta = \beta^*$ and $p = p^*$ is capable of learning coincides with the phase where the direct model freezes in random states (called the spin-glass phase), refining the well-established \textit{quiet planting} bound that the learning phase of the inverse model extends at least as far as the spin-glass phase of the direct model \cite{zdeborova2016statistical, achlioptas2008algorithmic, krzakala2009hiding, zdeborova2011quiet}. Pushing our analysis of this relationship further, we study noise tolerance and adversarial attacks in dense HNs memorizing as patterns the $M$ examples generated by the teacher. In particular, we derive a mathematical formula quantifying the adversarial robustness of dense HNs at zero temperature, and we clarify why the adversarial robustness of the dense HNs studied in \cite{krotov2018dense} changes as a function of the interaction order $p$. Our results lay the groundwork for a broader statistical mechanics theory of adversarial attacks, contemporaneously with complementary studies of adversarial attacks in linear models \cite{tanner2024high, vilucchio2024geometry, vilucchio2025existence}.

In Chapter \ref{chap:RBM_paper} (based on a work with Francesco Tosello and Daniele Tantari, published as \cite{theriault2025modeling}), we study RBMs in the teacher-student setting where the numbers of hidden units $P^*$ and $P$ are free to be any natural numbers much smaller than the number of samples in the training dataset. This work generalizes the results of \cite{hou2019minimal}, which studied the teacher-student setting with $P^* = P = 2$ hidden units. First, we investigate the relationship between our framework and RBMs in the teacher-student setting with $P^* = P = 1$ hidden units. To be more precise, we show that an RBM with $P$ hidden units learns data generated by a teacher with $P^* = P$ random patterns with the same overlaps as $P$ RBMs with one hidden unit each, thus validating a conjecture formulated in \cite{barra2017phase}. Going one step further, we study the scenario where the student has more hidden units than necessary to learn the training data, i.e. $P > P^*$. We show that, in this context, only $P^*$ of its patterns end up learning the data, while the others freeze in disordered states. Empirically, we investigate the connection between this low-dimensional learning strategy and the lottery ticket hypothesis according to which a generic randomly initialized overparameterized NN contains subnetworks that fit data with similar accuracy as the entire trained network when they are extracted from it and trained independently \cite{frankle2018lottery}. Finally, we show that, when the teacher patterns are correlated together rather than purely random, the student undergoes successive phase transitions where it learns the data in increasingly detailed ways as the size of the training dataset increases.

In Chapter \ref{chap:DAM_paper} (based on a work with Daniele Tantari, published as \cite{theriault2026saddle}), we study the teacher-student setting of an RBM that fits into the class of \textit{Dense Associative Memory} (DAM) models. This DAM generalizes Eq. (\ref{eq:DAM_distribution}), which is itself a close relative of the DAM studied in \cite{ramsauer2020hopfield, ota2023attention, lucibello2024exponential}, to a form capable of both supervised and unsupervised pattern classification. As in Chapter \ref{chap:RBM_paper}, we study a model with a generic number of hidden units. Based on our results, we propose a novel regularization scheme that makes training significantly more stable. Inspired by the observation that dense HNs cross multiple saddle points of the loss landscape during training \cite{boukacem2024waddington}, we study the saddle points of our DAM.
We demonstrate that the weights learned by DAMs with a relatively small number of hidden units are saddle points of the loss landscape of larger DAMs (see Eq. \ref{eq:log-likelihood}), thus extending studies of an analogous property in related models \cite{Decelle2021restricted, kappen1993using, KAPPEN1995deterministic, nijman1997symmetry, rose1990statistical, kloppenburg1997deterministic, akaho2000nonmonotonic}. We exploit this hierarchy by letting our DAM grow during learning \cite{wu2019splitting, wang2019energy}, which makes it significantly faster to train as a function of its final size.

\chapter{Dense Hopfield networks in the teacher-student setting}
\label{chap:dense_HN_paper}
\begin{table}[h!]
    \centering
    \begin{tabular}{cc}
        Based on the article \cite{theriault2024dense}, & doi: \href{https://doi.org/10.21468/SciPostPhys.17.2.040}{10.21468/SciPostPhys.17.2.040} \\ available under the CC BY 4.0 license \includegraphics[width=0.1\linewidth]{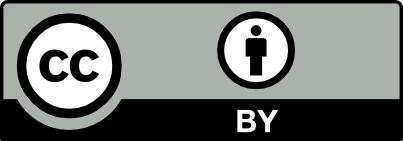} & \href{https://crossmark.crossref.org/dialog/?doi=10.21468/SciPostPhys.17.2.040&amp;domain=pdf&amp;date_stamp=2024-08-08}{\includegraphics[width=0.2\linewidth]{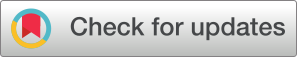}}
    \end{tabular}
\end{table}
\apptocmd{\sloppy}{\hbadness 10000\relax}{}{}
\section{Introduction}
\label{sec:dense_HN_introduction}

Hopfield networks are artificial neural networks that model associative memory \cite{hopfield1982neural}. In the Hopfield model, examples $\sigma \in \left\{ -1, 1 \right\}^N$ of memories $\xi^{\mu} \in \left\{ -1, 1 \right\}^N$, $\mu=1,\ldots,M$, are retrieved by sampling the Gibbs distribution of
a $2$-body Hamiltonian $H \left[ \sigma \mid \xi \right]$ at a given temperature $T$ \cite{amit1985spin}. Hopfield networks can be trained in a biologically plausible way using Hebb's rule \cite{hopfield1982neural, hebb2005organization}, which leads to $H \left[ \sigma \mid \xi \right] = -\frac{1}{N} \sum_{\mu = 1}^M \left( \sum_{i = 1}^{N} \xi^{\mu}_i \sigma_i \right)^2$. However, they can only store up to $M \sim \mathcal{O} \left( N \right)$ i.i.d. random memories in the limit of large $N$ \cite{hopfield1982neural, cover1965geometrical, amit1985storing}. One way to find this scaling is to study the phase diagram of $H \left[ \sigma \mid \xi \right]$ as a function of the temperature $T$ and load $\alpha = \frac{M}{N}$ \cite{amit1985storing}, where the so-called ferromagnetic phase, which extends up to $\alpha \approx 0.14$, corresponds to accurate retrieval.

Since Hopfield's seminal work, several generalizations have been investigated in relation to their critical storage capacity and retrieval capabilities. For example, parallel retrieval has been studied in relation to pattern sparsity \cite{agliari2012multitasking,agliari2013immunemedium,agliari2013immune,sollich2014extensive,agliari2017retrieving} or hierarchical interactions \cite{agliari2015retrieval,agliari2018non,agliari2015hierarchical,agliari2014metastable,agliari2015topological}, and non-universality has been shown with respect to
more general pattern entries and unit priors \cite{barra2017phase,barra2018phase,barra2015multi,agliari2017neural,barra2012glassy,genovese2020legendre,rocchi2017high}. Efforts to overcome the $\mathcal{O} \left( N \right)$ limitation of the capacity led to the development of a novel class of modern Hopfield networks \cite{ramsauer2020hopfield, widrich2020modern, krotov2021large}, which are sometimes called dense due to their faculty to store much more memories than the original Hopfield model \cite{krotov2016dense}. These neural networks surpass $\mathcal{O} \left( N \right)$ storage capacity by using higher-order interactions instead of the original $2$-body couplings \cite{chen1986high, psaltis1986nonlinear, baldi1987number, gardner1987multiconnected, abbott1987storage, horn1988capacities}.
In particular, Gardner \cite{gardner1987multiconnected} calculated the Replica-Symmetric (RS) phase diagram of the Hamiltonian $H \left[ \sigma \mid \xi \right] = -\sum_{i_1 < ... < i_p = 1}^{N} J_{i_1 ... i_p} \sigma_{i_1} ... \sigma_{i_p}$ with $p$-body interactions $J_{i_1 ... i_p} = \frac{p!}{N^{p - 1}} \sum_{\mu=1}^{M} \xi^{\mu}_{i_1} ... \xi^{\mu}_{i_p}$ conditioned on i.i.d. random memories $\xi^{\mu} \in \left\{ -1, 1 \right\}^N$, finding a $M=\mathcal{O} \left( N^{p-1} \right)$ storage capacity. These calculations were later extended to include the effects of $1$-step Replica Symmetry Breaking (1RSB) \cite{albanese2022replica}.

Although they draw a rather detailed picture of the retrieval of individual i.i.d. random memories, these results are not the end of the story. First of all, 1RSB calculations allegedly struggle to find the paramagnetic to spin-glass phase transition accurately at large $p$ because of numerical instability issues \cite{albanese2022replica}.
Second of all, dense Hopfield networks have been rapidly gaining a renewed attention for reasons other than their storage capacity since a recent paper \cite{krotov2016dense} by Krotov and Hopfield (K \& H), where they were used as a trainable machine learning architecture.
For instance, they have been related to transformers \cite{ramsauer2020hopfield, hoover2023energy} and diffusion models \cite{hoover2023memory, ambrogioni2024search}, and they were found to be significantly more explainable and adversarially robust than feedforward neural networks with ReLU activation functions \cite{krotov2016dense, krotov2018dense}.

One such aspect of dense Hopfield networks that is still poorly understood is their performance as generative models for unsupervised learning, where they are trained over some given dataset to reproduce its probability distribution.
As far as we are aware, this problem has not yet been studied theoretically for $p$-body models with $p \geq 3$. However, it was studied for the original $2$-body Hopfield network by using the teacher-student setting \cite{alemanno2023hopfield} first described in \cite{barra2017phase,barra2018phase,decelle2021inverse}.
In the teacher-student setting, which is also called inverse problem in opposition to the direct problem of random pattern retrieval,
a student model $H \left[ \xi \mid \sigma \right]$ is trained
with $M$ teacher examples $\sigma^a \sim H \left[ \sigma^a \mid \xi^* \right]$ conditioned on the planted pattern $\xi^*$. In other words, the student tries to infer the pattern $\xi^*$ of the teacher using a structured set of examples $\sigma^a$.

At finite load $\alpha = \frac{M}{N}$, two regimes of pattern retrieval were found: example Retrieval ($eR$) and signal Retrieval ($sR$).
In the $eR$ phase, the student tries to reconstruct $\xi^*$ by directly retrieving the examples $\sigma^a$, which is a good strategy provided that they are strongly correlated with $\xi^*$. In the $sR$ phase, on the other hand, retrieval is done by extracting subtle cues from weakly correlated examples. The two types of examples used in these two retrieval strategies are respectively called prototypes and features of $\xi^*$ \cite{krotov2016dense}.
Interestingly, a prototype regime and a feature regime were also observed by K \& H in dense Hopfield networks trained to classify real data \cite{krotov2016dense}, where it was found that the prototype regime is significantly more adversarially robust than the feature regime.
In other words, the prototype regime is more resistant than the feature regime to small data perturbations that are specifically designed to cause incorrect classification \cite{biggio2013evasion, szegedy2013intriguing}. This prototype approach is arguably a big step towards designing adversarially robust neural networks, a long-standing problem that still lacks a fully satisfying solution \cite{goodfellow2015explaining, madry2018towards, muhammad2022survey}.

In this work, we study the performance of $p$-body Hopfield networks in the teacher-student setting, revealing a prototype regime and a feature regime as in the $2$-body model. In Section \ref{sec:gardner_overview}, we review Gardner's main results in studying $p$-body Hopfield models and summarize what the rest of the literature on spin-glass models with $p$-body interactions tell us about the paramagnetic to spin-glass phase transition in $p$-body Hopfield models. In Section \ref{sec:dense_HN_teacher-student}, we compute the phase diagram of these $p$-body models in the teacher-student setting.
In Section \ref{sec:retrieval}, we discuss the transition to the retrieval phase in the inverse problem. In Section \ref{sec:trans}, we compare this retrieval transition against the transition to the spin-glass phase in the direct problem.
Despite their different nature, we show that these two transitions are equivalent on the Nishimori line where the teacher and the student have the same $p$ and $T$ \cite{nishimori1980exact, nishimori2001statistical, contucci2009spin, iba1999nishimori}.
In Section \ref{sec:match_discussion}, we discuss the phase diagram on the Nishimori line in more details.
In Section \ref{sec:mismatch_discussion} and Section \ref{sec:tol}, we discuss the phase diagram outside of the Nishimori line. First of all, we investigate the effect of using an inference temperature different from the dataset noise.
Second of all, we reveal that using a larger $p$ for the student than the teacher gives the student an extensive tolerance against both teacher noise and pattern interference. Finally, in Section \ref{sec:adv}, we derive a closed-form expression that measures the adversarial robustness of the student at zero temperature and explain what our results reveal about the nature of adversarial attacks.

\section{Overview of Gardner's results}
\label{sec:gardner_overview}

Consider the $p$-body Hamiltonian
\begin{equation}\label{eq:H}
    H \left[ \sigma \mid \xi \right] = -\sum_{i_1 < ... < i_p = 1}^{N} J_{i_1 ... i_p} \sigma_{i_1} ... \sigma_{i_p}= -\frac{p!}{N^{p - 1}}\sum_{i_1 < ... < i_p = 1}^{N}  \sum_{\mu=1}^{M} \xi^{\mu}_{i_1} ... \xi^{\mu}_{i_p} \sigma_{i_1} ... \sigma_{i_p} \,,
\end{equation}
conditioned on a set of $M=\frac{\alpha N^{p-1}}{p!}$ quenched memories $\xi^\mu\in\{-1,1\}^N$, $\mu=1,\ldots,M$, sampled i.i.d. from the Rademacher distribution $\frac{1}{2} \left[ \delta \left( \xi^\mu_i-1 \right) + \delta \left(\xi^\mu_i +1 \right) \right]$. In the \textit{direct model}, patterns $\sigma$ are in turn sampled from the equilibrium Gibbs distribution $P(\sigma \mid \xi)=Z^{-1}e^{-\beta H[\sigma \mid \xi]}$, where $\beta\geq 0$ is the inverse temperature and $Z=\sum_\sigma e^{-\beta H[\sigma \mid \xi]}$ is the system's partition function. The so-called \textit{direct problem} studied by Gardner \cite{gardner1987multiconnected} consists of quantifying the performance of this model as a method of memory retrieval.
In that context, the overlap $\frac{1}{N} \sum_i \xi^{\mu}_i \sigma_i$ is a good measure of retrieval accuracy, and its expected value can be derived from the quenched free entropy
$f = \frac{1}{N} \left\langle \log Z \right\rangle_{\xi}$ in the thermodynamic limit $N\to\infty$.
At finite $p$, Gardner used the (non-rigorous) replica trick \cite{charbonneau2022replica} to evaluate the RS approximation of $f$ (see also Appendix \ref{app:direct_cumulant})
in terms of a variational principle of the form
\begin{align*}
    f = \lim_{N\to\infty}\frac{1}{N} \left\langle \log Z \right\rangle_\xi & = \lim_{N\to\infty, L \rightarrow 0} \left( \frac{\partial}{\partial L} \left[ \frac{1}{N} \log \left\langle Z^L \right\rangle_\xi \right] \right)=\Extr_{m,k,q,k,r} \ f(m,k,q,r)\,,
\end{align*}
whose solution is
\begin{align}\label{eq:saddlepointdirect}
    q & = \int_{\mathbb{R}} dx \frac{1}{\sqrt{2\pi}} \exp \left( -\frac{1}{2} x^2 \right) \tanh^2 \left( \beta \left[ \sqrt{\alpha r} x + k \right] \right) \nonumber \,, \\
    m & = \int_{\mathbb{R}} dx \frac{1}{\sqrt{2\pi}} \exp \left( -\frac{1}{2} x^2 \right) \tanh \left( \beta \left[ \sqrt{\alpha r} x + k \right] \right) \nonumber   \,, \\
    r & = p q^{p-1} \,,\nonumber                                                                                                                                          \\
    k & = p m^{p-1}\,,
\end{align}
and the order parameters $m$ and $q$ are to be interpreted as expected overlaps. To be more precise, $m$ can be shown to be the expected overlap of a retrieval attempt $\sigma$ against one memory in the thermodynamic limit, i.e. $m=\lim_{N\to\infty} \left\langle \frac 1 N \sum_{i} \xi^\mu_i \sigma_i \right\rangle_{\xi,\sigma}$. Similarly, $q$ is the expected overlap between two retrieval attempts $\sigma^1$ and $\sigma^2$, i.e.
$q =\lim_{N\to\infty} \left\langle \frac{1}{N} \sum_i \sigma^{1}_i \sigma^{2}_i \right\rangle_{\xi,\sigma}$ or equivalently $q=\lim_{N\to\infty}\left\langle\frac 1 N \sum_i \left\langle\sigma_i\right\rangle^2_\sigma \right\rangle_\xi$.
Intuitively, $q$ measures the tendency of the system to stay frozen in specific configurations rather than visiting all possible values of $\sigma$.

\begin{figure}[t!]
    \centering
    \includegraphics[width = 0.495\textwidth]{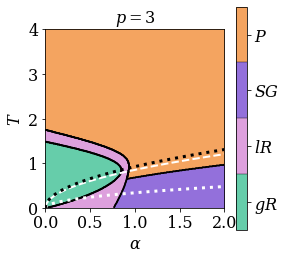}
    \includegraphics[width = 0.495\textwidth]{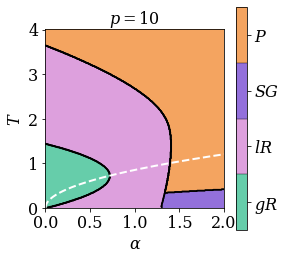}
    \caption{RS phase diagrams of the direct models with $p = 3$ on the left and $p = 10$ on the right. Accurate pattern retrieval is not possible in the Paramagnetic phase ($P$) or in the Spin-Glass phase ($SG$), but it is possible in the local Retrieval phase ($lR$) and in the global Retrieval phase ($gR$).
        The ferromagnetic fixed point corresponding to accurate pattern retrieval is globally stable in the $gR$ phase, but locally stable in the $lR$ phase.
        The phase diagrams are inexact below the white dashed line where the total entropy of the paramagnetic phase becomes negative. The black dotted line overlaying the $p = 3$ diagram is the (exact) 1RSB $P$-$SG$ transition temperature $T_s \left( \alpha, 3 \right)$, which is obtained by rescaling by $\sqrt{2 \alpha}$ the corresponding transition temperature of the spin-glass model with $p$-body Gaussian interactions. The d1RSB transition $T_d \left( \alpha, 3 \right)$ is very close to $T_s \left( \alpha, 3 \right)$ throughout the displayed range of $\alpha$. The white dotted line in the $p = 3$ plot is the temperature $T_G \left( \alpha, 3 \right)$ below which multiple steps of RSB are required to compute the free entropy. It is also obtained by rescaling by $\sqrt{2 \alpha}$ the corresponding transition temperature of the Gaussian spin-glass model.}
    \label{fig:direct_phase_diagrams}
\end{figure}

The resulting RS phase diagram (see Fig. \ref{fig:direct_phase_diagrams}) are derived from the value of the order parameters as a function of
three \textit{hyperparameters}:
the interaction order $p$, temperature $T=1/\beta$ and load $\alpha = \frac{M p!}{N^{p-1}}$. There are four different phases:
\begin{itemize}
    \item In the Paramagnetic phase ($P$), the overlaps $m$ and $q$ both vanish. The network does not retrieve any specific pattern: sampled configurations are completely random.
    \item In the Spin-Glass phase ($SG$), $m$ vanishes but $q > 0$. In other terms, the network does not retrieve individual stored memories but rather converges to spurious patterns depending on all the memories in a nontrivial way.
    \item In the signal Retrieval phases ($lR$ and $gR$), $m \neq 0$ and $q > 0$, which means that the network is able to retrieve the stored memories. $lR$ and $gR$ are respectively locally stable and globally stable. In other words, local retrieval $lR$ is only attainable from initial conditions in a limited neighborhood of a memory $\xi^\mu$, while global retrieval $gR$ is accessible from any initial conditions given enough time. These two phases are said to be ferromagnetic.
\end{itemize}
Gardner also calculated the exact $p \rightarrow \infty$ phase diagram without making any assumptions about replica symmetry \cite{gardner1987multiconnected}. In this limit, the resulting Paramagnetic to Spin-Glass ($P$-$SG$) phase transition occurs at a temperature $T_E(\alpha)$ that coincides with the boundary of the region where the total entropy of the paramagnetic phase becomes negative, given by $\beta^2 \alpha = 2 \log 2$ (white dashed line in Fig. \ref{fig:direct_phase_diagrams}).

At finite $p$, Gardner's results only tell us that the model cannot be in the paramagnetic phase below $T_E(\alpha)$. Therefore, a spin-glass transition should occur at a temperature $T_s\left(\alpha,p\right) \geq T_E(\alpha)$. Since the RS spin glass solution of Eqs. (\ref{eq:saddlepointdirect}) exists only below $T_E(\alpha)$ (violet region in Fig. \ref{fig:direct_phase_diagrams}), the spin-glass transition must be towards a RSB spin-glass phase.

Outside of the signal retrieval phases, the free entropy of the direct model is the
same as for the spin-glass model with $p$-body Gaussian interactions where the temperature is rescaled by a factor of $\sqrt{2 \alpha}$ \cite{gross1984simplest, gardner1985spin}. Therefore, the spin-glass and paramagnetic solutions are the same in the direct model as in this Gaussian spin-glass model, and we expect the exact phase diagrams of both models to be identical when the direct model is not in its signal retrieval phases.
According to previous work on the Gaussian model with finite $p$ \cite{gardner1985spin}, a 1RSB solution with $m = k = 0$ exists and is globally stable throughout a whole phase below $T_s(\alpha,p)\geq T_E(\alpha)$ (see Fig. \ref{fig:direct_phase_diagrams}). This solution becomes unstable at a lower transition temperature $T_G(\alpha, p)$ (see Fig. \ref{fig:inverse_phase_diagrams}), below which multiple steps of RSB are required. In the limit of $p\to\infty$, it holds that $T_s(\alpha,p)\to T_E(\alpha)$ and $T_G(\alpha,p)\to 0$. In other terms, the direct model becomes 1RSB, which is consistent with the fact that it is converging to a random energy model with temperature rescaled by $\sqrt{2 \alpha}$ \cite{derrida1980random, gross1984simplest, gardner1987multiconnected}. Finally, we mention that this type of models exhibits a random first order transition
phenomenology \cite{monasson1995structural, montanari2003nature, crisanti2005complexity, franz2017universality}: there is in fact a range of temperatures $T_s(\alpha,p)\leq T\leq T_d(\alpha,p)$ where the dynamics get trapped in an exponential number of metastable clusters, with an emerging RSB structure that does not affect the free energy (see Fig. \ref{fig:inverse_phase_diagrams}). This range of temperatures thus defines a so-called dynamical 1RSB (d1RSB) phase. Below $T_s(\alpha,p)$, the number of clusters is no longer exponential, and the system undergoes
the thermodynamic 1RSB phase transition that we mentioned previously. The critical temperatures $T_G(\alpha,p)$, $T_s(\alpha,p)$ and $T_d(\alpha,p)$ can all be obtained by standard RSB methods, but the resulting saddle-point equations can be prone to numerical instability at large $p$ \cite{albanese2022replica}. In Sections \ref{sec:trans} and \ref{sec:match_discussion}, we discuss an alternative way to obtain $T_s(\alpha,p)$ and $T_d(\alpha,p)$.

\section{Teacher-student setting}
\label{sec:dense_HN_teacher-student}
On our end, we study a dense Hopfield network with Hamiltonian (\ref{eq:H}) as a generative model for unsupervised learning.
In that context, the memories $\xi$ are model parameters that have to be trained in such a way that the examples of a given dataset $\{\sigma^a\}_{a=1}^M$ result as typical network configurations.

In particular, we study a controlled teacher-student setting in which the examples are sampled from the probability distribution $P \left( \sigma^a \mid \xi^* \right)$ of a so-called \textit{teacher} dense Hopfield network conditioned on a single
\textit{planted} pattern $\xi^* \in \left\{ -1, 1 \right\}^N$
whose entries are quenched Rademacher random variables.
A \textit{student} dense Hopfield network, also known as the \textit{inverse model}, then samples its own student pattern $\xi$ from the posterior distribution
\begin{align*}\label{eq:post}
    P \left( \xi \mid \sigma \right)
     & = \frac{P(\xi)\prod_{a=1}^{M} P \left( \sigma^a \mid \xi \right)}{P(\sigma)}= \frac{P(\xi)}{P(\sigma)} \prod_{a=1}^M Z^{-1}\exp\left(-\beta H[\sigma^a \mid \xi]\right)\,,
\end{align*}
where $P \left( \sigma^a \mid \xi \right)$ is the Gibbs distribution of the direct model with a single memory $\xi$, and $P \left( \xi \right)$ is the prior on $\xi$ that is chosen to be uniform. Since the direct model
has only a single pattern, $Z$ does not depend on $\xi$ (see Appendix \ref{app:teacher-student_partition}), and the posterior simplifies to
\begin{align*}
    P \left( \xi \mid \sigma \right) & = \mathcal{Z}^{-1}(\sigma) \exp\left(-\beta H[\xi \mid \sigma]\right)\,.
\end{align*}
In sum, the student posterior distribution
is that of a dense Hopfield network where $\xi$ plays the role of the
sampled pattern and the examples $\sigma$ act like the $M$ quenched memories.
Our task, called the \textit{inverse problem}, consists of quantifying the student's capability to infer the teacher pattern, which we will also call the \textit{signal}.
Like Gardner,
we calculate
a free entropy of the form
$f = \frac{1}{N} \left\langle \log \mathcal{Z} \right\rangle_{\sigma}$ in the thermodynamic limit $N\to\infty$. This time, however, the average $\left\langle \cdot\right\rangle_{\sigma}$ is over a structured set of examples $\sigma$.
In fact, we recall that, unlike the i.i.d. memories studied by Gardner, the examples $\sigma^a$ are sampled from the teacher distribution $P \left( \sigma^a \mid \xi^* \right)$.

In general, the student does not have access to the teacher generative model.
In our controlled teacher-student setting, the student knows that the correct model for $P \left( \sigma^a \mid \xi \right)$ is a dense Hopfield network.
Nevertheless, it does not necessarily have access to the interaction order $p^*$ and inverse temperature $\beta^*$ used by the teacher.
Therefore, we denote the student hyperparameters by $p$ and $\beta$ and emphasize that
they are not necessarily equal to $p^*$ and $\beta^*$. As previously stated, we calculate the free entropy
\begin{align}
    f = \frac{1}{N} \left\langle \log \mathcal{Z} \right\rangle_{\sigma}
     & = 2^{-N}\sum_{\xi^*}\sum_{\sigma} \left[ Z^* \right]^{-M} \exp\left( \beta^* \frac{p^*!}{N^{p^*-1}} \sum_{a=1}^M \sum_{i_1 < ... < i_{p^*}} \xi^*_{i_1} ... \xi^*_{i_{p^*}} \sigma^{a}_{i_1} ... \sigma^{a}_{i_p} \right)\nonumber \\
     & \quad\times \log \sum_{\xi} \exp\left( \beta \frac{p!}{N^{p-1}} \sum_{a =1}^M \sum_{i_1 < ... < i_p} \xi^b_{i_1} ... \xi^b_{i_p} \sigma^{a}_{i_1} ... \sigma^{a}_{i_p} \right)\,,
\end{align}
in the thermodynamic limit $N\to\infty$. We then draw phase diagrams of the inverse problem as a function of $p^*$, $T^* = 1/\beta^*$, $p$, $T = 1/\beta$ and $\alpha$, where $\alpha$ is $M$ normalized to $\mathcal{O} \left( 1 \right)$. Unless explicitly specified otherwise, we use $\alpha = \frac{M p!}{N^{p - 1}}$.

\subsection{Matched interaction orders}
\label{sec:matched_student}
We first consider the case where $p^*=p$ and the only possible mismatch between the teacher and student networks is in the inverse temperature, i.e. $\beta^*\neq\beta$. At low $T^*$, the student's task is easy.
In fact, below the critical temperature $T_ {\text{crit}}$ of the direct problem with one pattern (see Fig. \ref{fig:direct_phase_diagrams}, $\alpha = 0$ axis), the teacher produces examples $\sigma^a$ that cluster around $\xi^*$.
Therefore, the student can infer $\xi^*$ by aligning its pattern $\xi$ with the examples $\sigma^a$. This retrieval strategy works even when using a very small amount of examples (see \cite{alemanno2023hopfield}). Since the size of our dataset is extensive, the retrieval accuracy is maximum in the thermodynamic limit. We call this region the (accurate) example Retrieval phase ($eR$).

Conversely, when $T^*$ is above $T_{\text{crit}}$, the examples in the training set are very noisy and we do not observe a finite overlap between $\sigma^a$ and $\xi^*$ (see Fig. \ref{fig:direct_phase_diagrams}, $\alpha = 0$ axis). In this regime, we find that the RS approximation of the $p^* = p$ free entropy can be computed (see Appendix \ref{app:teacher-student_free_entropy}) in terms of the variational principle
\begin{align}
    \label{eq:tsfree}
    \nonumber f & = \Extr_{m,k,q,r,q^*,r^*} \Bigg\{ \beta^* \beta \alpha \left[ q^{*} \right]^p - \frac{1}{2} \beta^2 \alpha q^p + \beta m^p - \beta^* \beta \alpha r^{*} q^{*}                                                               \\
                & \quad+ \frac{1}{2} \beta^2 \alpha r q - \frac{1}{2} \beta^2 \alpha r - \beta m k + \frac{1}{2} \beta^2 \alpha + \log 2 \nonumber                                                                                            \\
                & \quad+ \int dx \frac{1}{\sqrt{2\pi}} \exp \left\{ -\frac{1}{2} x^2 \right\} \Big\langle \log \left[ \cosh \left( \beta \left[ \sqrt{\alpha r} x + \beta^* \alpha r^* + k z \right] \right) \right] \Big\rangle_z \Bigg\}\,,
\end{align}
whose solution is the saddle-point equations
\begin{align*}
    \label{eqn:saddle-point_p_s=p}
    q^* & = \int_{\mathbb{R}} dx \frac{1}{\sqrt{2\pi}} \exp \left( -\frac{1}{2} x^2 \right) \Big\langle \tanh \left( \beta \left[ \sqrt{\alpha r} x + \beta^* \alpha r^* + k z \right] \right) \Big\rangle_z   \,,            \\
    q   & = \int_{\mathbb{R}} dx \frac{1}{\sqrt{2\pi}} \exp \left( -\frac{1}{2} x^2 \right) \Big\langle \tanh^2 \left( \beta \left[ \sqrt{\alpha r} x + \beta^* \alpha r^* + k z \right] \right) \Big\rangle_z   \,,          \\
    m   & = \int_{\mathbb{R}} dx \frac{1}{\sqrt{2\pi}} \exp \left( -\frac{1}{2} x^2 \right) \Big\langle z \tanh \left( \beta \left[ \sqrt{\alpha r} x + \beta^* \alpha r^* + k z \right] \right) \Big\rangle_z \numberthis\,, \\
    r^* & = p \left[ q^{*} \right]^{p-1}    \,,                                                                                                                                                                               \\
    r   & = p q^{p-1}   \,,                                                                                                                                                                                                   \\
    k   & = p m^{p-1}\,,
\end{align*}
where $z$ is a Rademacher random variable and $\alpha = \frac{M p!}{N^{p - 1}}$. As in the direct model described in Section \ref{sec:gardner_overview}, the order parameters $m$ and $q$ have a clear interpretation in terms of expected overlaps. $m=\lim_{N\to\infty}\left\langle \frac 1 N \sum_{i} \xi_i \sigma^a_i \right\rangle_{\xi^*,\sigma,\xi}$ is the expected overlap of a retrieval attempt with an example $\sigma^a$, and $q=\lim_{N\to\infty}\left\langle \frac 1 N \sum_{i} \left\langle\xi_i\right\rangle^2_{\xi}  \right\rangle_{\xi^*,\sigma}$ is the expected overlap between two retrieval attempts. Similarly, $q^*$ is the expected overlap between the teacher and student patterns, i.e. $q^*=\lim_{N\to\infty}\left\langle \frac 1 N \sum_{i} \xi^*_i\xi_i  \right\rangle_{\xi^*,\sigma,\xi}$. Therefore, it is a good measure of inference performance.
The free entropy (Eq. \ref{eq:tsfree}) is expected to be exact in absence of mismatch between the teacher and the student, i.e. $\beta^*=\beta$. This condition is known as the Nishimori line \cite{nishimori1980exact, nishimori2001statistical, contucci2009spin, iba1999nishimori}. Outside of the Nishimori region, RSB corrections are expected. Like the direct problem, the inverse problem with $T^* > T_{\text{crit}}$ has different phases characterized by the values of the order parameters:
\begin{itemize}
    \item In the Paramagnetic phase ($P$), the overlaps $m$, $q^* $ and $q$ all vanish.
    \item In the signal Retrieval phases ($lR$ and $gR$), $m = 0$ but $q^*\neq 0$ and $q>0$. $lR$ and $gR$ are respectively
          locally stable and globally stable. In other words, local retrieval $lR$ is only attainable from initial conditions in a limited neighborhood of $\xi^*$, while global retrieval $gR$ is accessible from any initial conditions given enough time. These two phases are also said to be ferromagnetic.
    \item In the (inaccurate) example Retrieval phase ($eR$), $m\neq 0$ and $q>0$ but $q^*=0$.
    \item In the Spin-Glass phase (SG), $q>0$ but $q^*$ and $m$ vanish.
\end{itemize}
In sum, when $T^*$ is above $T_{\text{crit}}$, the student can only learn the teacher pattern in the signal retrieval phases. In all the other phases, the student pattern is uncorrelated with the signal, being either a random
guess ($P$ phase), aligned with a noisy example (inaccurate $eR$ phase), or aligned with a spurious low energy state ($SG$
phase). We stress that we cannot have $m \neq 0$ and $q^* \neq 0$ at the same time (accurate $eR$ phase) when $T^* > T_{\text{crit}}$ because  $\lim_{N\to\infty}\left\langle \frac 1 N \sum_{i} \xi^*_i \sigma^a_i \right\rangle_{\xi^*,\sigma} = 0$ in that regime (see Fig. \ref{fig:direct_phase_diagrams}, $\alpha = 0$ axis).

\subsection{Mismatched interaction orders}
\label{sec:mismatched_student}
We also investigate the $T^* > T_{\text{crit}}$ regime in the presence of a mismatch between the interaction orders of the teacher and student networks, i.e.  $p^*\neq p$.
We focus on the case of $p^* = 2$ and even $p \geq 3$ to study the consequences of fitting the teacher of \cite{alemanno2023hopfield} using a student with higher order interactions.
We find
two different scaling regimes of the training set size $M$ and inverse temperature $\beta^*$ that make retrieval possible (see Appendix D):
\begin{itemize}
    \item a large-noise scaling where $\beta^* \sim \mathcal{O} \left( N^{2/p - 1} \right)$ and $M \sim \mathcal{O} \left( N^{p - 1} \right)$,  such that $\alpha = \frac{M p!}{N^{p-1}}$ and $\lambda = \frac{\left[ \beta^* \right]^{p/2}}{(p/2)!} N^{p/2 - 1}$ are finite;
    \item a finite-noise scaling where $\beta^* \sim \mathcal{O} \left( 1 \right)$ and  $M \sim \mathcal{O} \left( N^{p/2} \right)$,  such that $\alpha = \frac{M (p/2 + 1)!}{N^{p/2}}$ is finite.
\end{itemize}
In the large-noise scaling, we obtain saddle point equations similar to Eqs. (\ref{eqn:saddle-point_p_s=p}) but with $\beta^*$ replaced by $\lambda$ (see Appendix \ref{app:teacher-student_free_entropy}). Conversely, the finite noise scaling leads to
\begin{align*}
    \label{eqn:saddle-point_p_s=2}
    q^* & = \Big\langle \tanh \left( \beta \left[ \eta \alpha r^* + k z \right] \right) \Big\rangle_z        \,,       \\
    m   & = \Big\langle z \tanh \left( \beta \left[ \eta \alpha r^* + k z \right] \right) \Big\rangle_z \numberthis\,, \\
    r^* & = p \left[ q^{*} \right]^{p-1}   \,,                                                                         \\
    k   & = p m^{p-1}\,,
\end{align*}
where $\eta$ generally depends on $\beta^*$ and $p$ in a nontrivial way, but we find that $\eta = \frac{2 \left[ \beta^* \right]^2}{\left( 1 - 2\beta^* \right)^2}$ when $p = 4$ (see Appendix \ref{app:teacher-student_free_entropy}). These equations can also be derived by extrapolating the large-noise equations to $\alpha_{\text{large noise}} \rightarrow 0$ and $\lambda \rightarrow \infty$ with fixed $\lambda \alpha_{\text{large noise}} = \eta \alpha_{\text{finite noise}}$.

\section{Results and Discussion}
\subsection{Retrieval transition at large interaction order}\label{sec:retrieval}
The paramagnetic solution of Eqs. (\ref{eqn:saddle-point_p_s=p}) always exists and is globally stable in the part of the phase diagram where the temperature $T$ is relatively large and $\alpha = \frac{M p!}{N^{p - 1}}$ is relatively small. On the other hand,
the $gR$ phase exists when $\beta^2 \alpha p$ and $\beta^* \beta \alpha p$ are both large. In fact, in that limit,
$q^* = q = 1$ is a fixed point of Eqs. (\ref{eqn:saddle-point_p_s=p}).
The critical line where $gR$ becomes globally stable instead of $P$ is not clear from this analysis alone, but we can at least find it analytically in the limit of infinite $p$. As for the direct model, the free entropy and the total entropy of the paramagnetic phase are respectively $\frac{1}{2} \beta^2 \alpha + \log 2$ and $-\frac{1}{2} \beta^2 \alpha + \log 2$ \cite{gardner1987multiconnected}. At the same time, the $p \rightarrow \infty$ free entropy takes the form
\begin{align*}
    f & = \Extr \Bigg\{ \beta^* \beta \alpha \ \theta\left( q^{*} - 1 \right) - \frac{1}{2} \beta^2 \alpha \ \theta\left( q - 1 \right) - \beta^* \beta \alpha r^{*} q^{*} + \frac{1}{2} \beta^2 \alpha r q - \frac{1}{2} \beta^2 \alpha r + \frac{1}{2} \beta^2 \alpha \\
      & \quad+ \log 2 + \int dx \frac{1}{\sqrt{2\pi}} \exp \left\{ -\frac{1}{2} x^2 \right\} \log \left[ \cosh \left( \sqrt{\beta^2 \alpha r} x + \beta^* \beta \alpha r^* \right) \right] \Bigg\}\,,
\end{align*}
where $\theta \left( q - 1 \right) := \lim_{p \rightarrow \infty} q^p$, $q \in \left[ 0, 1 \right]$, is the Heaviside step function jumping at $q=1$, i.e. $\theta(1)=1$ and $\theta(q)=0$ $\forall q\in[0,1)$.
In this limit, the ferromagnetic phase is characterized by $q=q^*=1$, and its free entropy is then
\begin{align*}
    f & = \beta^* \beta \alpha - \beta^* \beta \alpha p + \int dx \frac{1}{\sqrt{2\pi}} \exp \left\{ -\frac{1}{2} x^2 \right\} \log \left[ 2 \cosh \left( \sqrt{\beta^2 \alpha p} x + \beta^* \beta \alpha p \right) \right] \\
      & \approx \beta^* \beta \alpha - \beta^* \beta \alpha p + \int dx \frac{1}{\sqrt{2\pi}} \exp \left\{ -\frac{1}{2} x^2 \right\} \left( \sqrt{\beta^2 \alpha p} x + \beta^* \beta \alpha p \right)                       \\
      & = \beta^* \beta \alpha\,.
\end{align*}
The corresponding total entropy is $s = f - \beta \frac{\partial f}{\partial \beta} = 0$, as expected from a ferromagnetic phase with $q^* = q = 1$. On the Nishimori line, $f = \beta^* \beta \alpha$ becomes larger than the free entropy of the paramagnetic phase, which triggers a phase transition, if and only if
\begin{align}\label{eq:crlinelargep}
    T & < \sqrt{\frac{\alpha}{2 \log 2}}\,,
\end{align}
where $T_E= \sqrt{\frac{\alpha}{2 \log 2}}$ is also the temperature below which the total entropy of the paramagnetic phase becomes negative. Outside of the Nishimori line, this inequality generalizes to $\beta^* \beta \alpha > \frac{1}{2} \beta^2 \alpha + \log 2 $, leading to
\begin{align*}
    \beta^* - \sqrt{\left[ \beta^* \right]^2 - \frac{2 \log 2}{\alpha}} < \beta & < \beta^* + \sqrt{\left[ \beta^* \right]^2 - \frac{2 \log 2}{\alpha}}\,,
\end{align*}
while the temperature where the paramagnetic total entropy becomes negative stays the same.

\subsection{Transition to the ordered phases: Universality}\label{sec:trans}
In the $p \rightarrow \infty$ limit, the transition towards $gR$ of the inverse model on the Nishimori line is identical to the exact $P$-$SG$ transition of the direct model \cite{gardner1987multiconnected}.
We claim that these two critical lines are actually closely related for any $p$. In the Hopfield model with $p=2$, they were already shown to be identical \cite{alemanno2023hopfield}.
We will now argue that they overlap for any $p$ and $\beta$ such that
$T > T_{\text{crit}}$ (see Figs. \ref{fig:inverse_phase_diagrams} and \ref{fig:direct_phase_diagrams}).
In the case of $p=2$, both lines can be obtained exactly from the RS approximation of either the direct model or the inverse model, so there is no obvious advantage to using this equivalence in calculations. In general, while the inverse problem on the Nishimori line is
replica symmetric, the direct problem is not, and the $p\geq 3$ replica symmetric $P$-$SG$ transition is not exact. Moreover, even the
critical line calculated using 1RSB may be inaccurate due to numerical instability \cite{albanese2022replica}. In this situation, the knowledge of the $gR$ transition in the replica-symmetric inverse problem can be used
to locate the exact $P$-$SG$ transition of the direct problem, where symmetry breaking occurs.

For that purpose, we will argue that, given $T > T_{\text{crit}}$, \textit{the direct model is in the paramagnetic phase if and only if the inverse model is in the paramagnetic phase.}

The converse implication comes from the fact that since (see Appendix \ref{app:teacher-student_partition})
\begin{equation}
    P \left( \sigma \right) = \frac{1}{2^{M N}} \frac{\mathcal{Z}(\sigma)}{\left\langle \mathcal{Z} \right\rangle}\,,
\end{equation}
the example distribution $P \left( \sigma \right)$ of the inverse problem is contiguous \cite{roussas1972contiguity} to the uniform distribution, i.e. the memory distribution of the direct problem, when
\begin{equation}
    \lim_{N \rightarrow \infty} \left\{ \frac{\log  \mathcal{Z} - \log \left\langle \mathcal{Z} \right\rangle}{N} \right\} = 0\,.
\end{equation}
As determined in Appendix \ref{app:teacher-student_partition} and  \ref{app:teacher-student_free_entropy}, the annealed expression $\frac 1 N \log \left\langle \mathcal{Z} \right\rangle$ is equal to the free entropy of the paramagnetic phase. Therefore, when the inverse model is in the paramagnetic phase, $P \left( \sigma \right)$ is contiguous to the uniform distribution. This property is called quiet planting and is known to occur more generally in mean-field paramagnets \cite{achlioptas2008algorithmic, krzakala2009hiding, zdeborova2011quiet, zdeborova2016statistical}.
In our problem setting, it means that if the inverse model is in the paramagnetic phase, then it is equivalent to the direct model. In particular, if the inverse model is in the paramagnetic phase, then so is the direct model.
In more intuitive terms, the $gR$ transition temperature of the inverse model must be greater than or equal to the $P$-$SG$ transition temperature of the direct model because the ensemble of examples $\sigma^a$ generated by the teacher model is on average at least as structured as the set of i.i.d. random memories stored in the direct model.

For the direct implication, notice that the average replicated partition function of the direct model in the paramagnetic phase can be approximated as (see Appendix \ref{app:rsb_ansatz})
\begin{align*}
    \left\langle Z^L \right\rangle & \approx \frac{1}{\left\langle Z \right\rangle} \Bigg\langle \sum_{\sigma} \exp \Bigg( \beta N \sum_{\gamma} \sum_{\mu \in \Gamma_{\gamma}} \left[ \frac{1}{N} \sum_i \xi^{\mu}_i \sigma^{\gamma}_i \right]^p    \\
                                   & \quad+ \beta \sum_{\gamma} \sum_{\mu \in \Bar{\Gamma}} \frac{p!}{N^{p-1}} \sum_{i_1 < ... < i_p} \xi^{\mu}_{i_1} ... \xi^{\mu}_{i_p} \sigma^{\gamma}_{i_1} ... \sigma^{\gamma}_{i_p} \Bigg)                     \\
                                   & \quad \sum_{\sigma_0} \exp \left( \beta \sum_{\mu \in \Bar{\Gamma}} \frac{p!}{N^{p-1}} \sum_{i_1 < ... < i_p} \xi^{\mu}_{i_1} ... \xi^{\mu}_{i_p} \sigma^{0}_{i_1} ... \sigma^{0}_{i_p} \right) \Bigg\rangle\,.
\end{align*}
This expression is identical to the replicated partition function of the inverse model with $T > T_{\text{crit}}$, which therefore must also be in the paramagnetic phase.

As a consequence, when $T > T_{\text{crit}}$, the $P$-$SG$ transition line of the direct model must be identical to the $gR$ transition line of the inverse model on the Nishimori line.

\subsection{Phase diagram on the Nishimori line}
\label{sec:match_discussion}

\begin{figure}[t!]
    \centering
    \includegraphics[width = 0.325\textwidth]{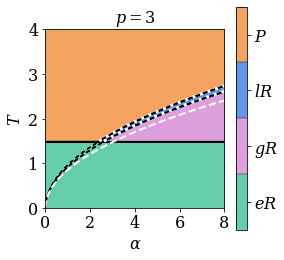}
    \includegraphics[width = 0.325\textwidth]{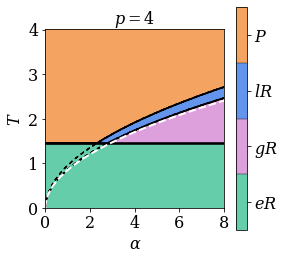}
    \includegraphics[width = 0.325\textwidth]{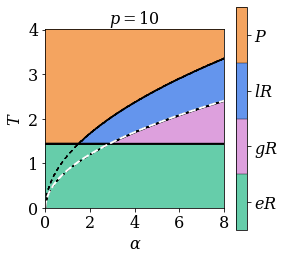}
    \caption{Exact RS phase diagrams of inverse models on the Nishimori line, i.e. $p^* = p$ and $\beta^* = \beta$. the left, center and right plots respectively have $p = 3$, $p = 4$ and $p = 10$.
        Accurate pattern retrieval is not possible in the Paramagnetic phase ($P$), but it is possible in the local Retrieval phase ($lR$), in the global Retrieval phase ($gR$) and in the example Retrieval phase ($eR$).
        The ferromagnetic fixed point corresponding to accurate pattern retrieval is globally stable in the $gR$ phase, but locally stable in the $lR$ phase.
        The critical temperature of the $eR$ phase is the critical temperature $T_{\text{crit}}$ of the direct problem with one pattern (see Fig. \ref{fig:direct_phase_diagrams}, $\alpha = 0$ axis).
        The black dashed lines mark the spurious continuation of the $lR$ and $gR$ phase boundaries through the $eR$ phase.
        The white dashed line is the $p \rightarrow \infty$ $gR$ critical line calculated analytically in Section \ref{sec:retrieval}. It matches the corresponding numerical phase boundary increasingly well as $p$ grows larger. The white dotted lines on the $p = 3$ plot mark the 1RSB and d1RSB critical temperatures $T_s \left( \alpha, 3 \right)$ and $T_d \left( \alpha, 3 \right)$ of the direct model (see Section \ref{sec:gardner_overview}). We truncated them below $T_{\text{crit}}$ for improved visibility. $T_s \left( \alpha, 3 \right)$ and $T_d \left( \alpha, 3 \right)$ are obtained by rescaling the corresponding critical temperatures found in \cite{montanari2003nature} by $\sqrt{2 \alpha}$.}
    \label{fig:inverse_phase_diagrams}
\end{figure}

On the Nishimori line, the student is fully informed about the teacher generative model and uses $\beta=\beta^*$ and $p=p^*$.
In this scenario, thanks to the Nishimori identities \cite{nishimori2001statistical}, it is well known that $\xi^*$ and $\xi$ play symmetric roles and that
$q^* = q$.
For the same reason, the overlaps $\frac 1 N \sum_i \xi^*_i\xi_i$ and $\frac 1 N \sum_i \xi^1_i\xi^2_i$ have the same distribution. From the self-averaging of $\frac 1 N \sum_i \xi^*_i\xi_i$, it follows that the system is expected to be replica symmetric, and Eqs. (\ref{eq:tsfree}) and (\ref{eqn:saddle-point_p_s=p}) are expected to hold.
Fig. (\ref{fig:inverse_phase_diagrams}) shows the phase diagrams obtained by solving the saddle-point equations numerically on the Nishimori line.
Both $q^*=q$ and the replica symmetry condition are verified. In particular, numerical solutions of a few values of $p \geq 3$ show that the $gR$ transition occurs at a higher $T$ than the line $\beta^2 \alpha = 2 \log 2$
where the total entropy of the paramagnetic phase becomes negative. In other terms, the phase transition towards $gR$ prevents the total entropy from becoming negative when $T$ decreases below $\sqrt{\frac{\alpha}{2\log2}}$, which is consistent with the RS solution being exact on the Nishimori line.

At low $T$, the student can learn efficiently within the accurate $eR$ regime. In this phase, learning is possible ($q^*\neq0$) because the examples are correlated with the signal and the student can retrieve it by simply being aligned with them ($m\neq 0$).

At high $T$, learning is possible only if the amount of examples, i.e. the size of the dataset, is sufficiently large.
When $\alpha$ is too small, Eqs. (\ref{eqn:saddle-point_p_s=p}) have only a paramagnetic fixed point because the amount of information carried by the dataset is not large enough. Numerical solutions suggest that the paramagnetic fixed point always exist and it is actually locally stable in the whole high-temperature regime. When $\alpha$ is sufficiently large, the signal retrieval fixed point appears as a locally stable attractor ($lR$ phase). It becomes globally stable ($gR$ phase) as the size of the dataset is increased further or the student temperature decreases.

As per the previous Section, the critical boundary of the $gR$ phase obtained by solving Eqs. \ref{eqn:saddle-point_p_s=p} is identical to the 1RSB $P$-$SG$ transition temperature $T_s(\alpha,p)$ of the direct model. Similarly, we observe that the metastable $lR$ phase coincides with the d1RSB phase of the direct model (see Fig. \ref{fig:inverse_phase_diagrams}). Our results are also consistent with the fact that $T_s(\alpha,p)\to T_E(\alpha)$ in the $p \rightarrow \infty$ limit. In fact, we find that the analytical limit boundary closely agrees with the numerical solution of the saddle-point equations with $p^* = p = 10$ and remains a good approximation even down to $p^* = p = 4$.

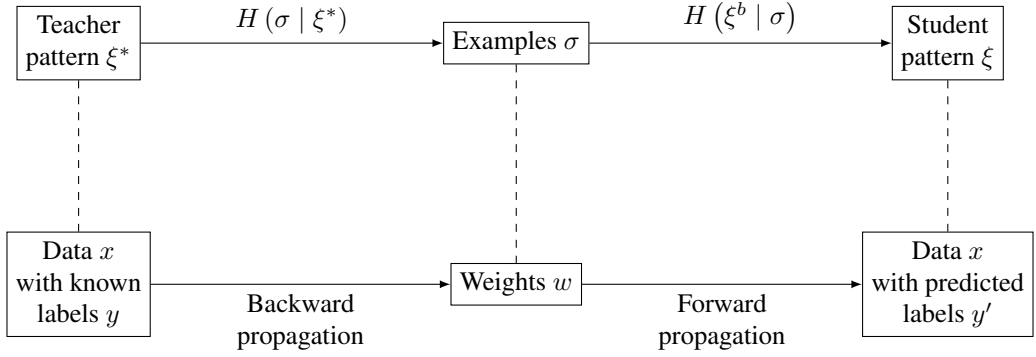
\begin{figure}[t!]
    \centering

    \begin{tikzpicture}[align = center, node distance = 2cm and 4cm]

        \node[draw](xi^*){Teacher \\ pattern $\xi^*$};

        \node[draw](sigma)[right = of xi^*]{Examples $\sigma$};

        \node[draw](xi^b)[right = of sigma]{Student \\ pattern $\xi$};

        \node[draw](v)[below = of xi^*]{Data $x$ \\ with known \\ labels $y$};

        \node[draw](xi) at (sigma |- v){Weights $w$};

        \node[draw](c)[below = of xi^b]{Data $x$ \\
            with predicted \\ labels $y'$};

        \draw[draw, ->](xi^*.east)--(sigma.west) node[midway, above]{$H \left( \sigma \mid  \xi^* \right)$};

        \draw[draw, ->](sigma.east)--(xi^b.west) node[midway, above]{$H \left( \xi^b  \mid  \sigma \right)$};

        \draw[draw, ->](v.east)--(xi.west) node[midway, below]{Backward \\ propagation};

        \draw[draw, ->](xi.east)--(c.west) node[midway, below]{Forward \\ propagation};

        \draw[draw, dashed](xi^*.south)--(v.north);

        \draw[draw, dashed](sigma.south)--(xi.north);

        \draw[draw, dashed](xi^b.south)--(c.north);

    \end{tikzpicture}

    \caption{The first row of this diagram sketches how a $p$-body Hopfield network in the teacher-student setting can reconstruct an incomplete pattern $\xi^b$ to match the teacher pattern $\xi^*$ by relying on the examples $\sigma$ obtained from $\xi^*$. The second row summarizes how a dense neural network trained by K \& H can recover the labels $y'$ of the data $x$ given the weights $w$ learned from $x$ \cite{krotov2016dense}.
        Both models tackle similar tasks using an approach where $\sigma$ and $\xi^b$ respectively play the same roles as $w$ and $(x, y')$.
        The forward propagation algorithm used to generate $y'$ is similar to the update rule of the student (see \cite{krotov2016dense} and Appendix \ref{app:hamiltonians}), but the backpropagation algorithm used to learn $w$ is very different from the update rule of the teacher.}
    \label{fig:gardner-krotov_comparison_diagram}
\end{figure}

In the student model, $\sigma$ plays a similar role as the weights of the trainable dense Hopfield network model that K \& H designed for classification of data \cite{krotov2016dense}. In that context, $\xi$ is analogous to the test data whose labels are being predicted (see Fig. \ref{fig:gardner-krotov_comparison_diagram}). In fact, the computation performed by K \& H's model to recover labels is similar to the update rule used by the student to infer the teacher pattern (see Appendix \ref{app:hamiltonians}).
Moreover, the $eR$ and $gR$ phases are respectively reminiscent of the prototype and feature regimes of K \& H's networks. Therefore, we believe that the student can act as a toy model of label prediction in these two regimes.

Comparing instead the phase diagrams of our inverse model with that of the inverse $2$-body Hopfield model, we see that the $eR$ and $gR$ phases of the inverse $p$-body model with $p \geq 3$ are respectively analogous to the $eR$ and $sR$ (signal Retrieval) phases presented in \cite{alemanno2023hopfield}.
One of the key differences between $p = 2$ and $p \geq 3$ is that the paramagnetic to signal retrieval phase transition of the $p$-body model is second order for $p = 2$ but first order for $p \geq 3$.
On the one hand, the second order phase transition of $p = 2$ indicates that its paramagnetic fixed point is never locally stable and sets an unambiguous boundary between the $sR$ phase where $\xi^*$ can be
recovered starting from any initial conditions and the paramagnetic phase where pattern retrieval is impossible \cite{zdeborova2016statistical}.
On the other hand, the first order phase transition of $p \geq 3$ allows the retrieval and paramagnetic regimes to coexist.
The $lR$ phase is locally stable precisely because it coexists with the paramagnetic phase and has a lower free entropy. Meanwhile, the $gR$ phase also coexists with the paramagnetic phase, but has a larger free entropy.
In the presence of phase coexistence, an algorithm trying to retrieve $\xi^*$ starting from random initial conditions can get stuck in the paramagnetic phase instead.
In fact, it has been conjectured that there is no algorithm with random initial conditions that can find such a ferromagnetic fixed point in a tractable amount of time
\cite{zdeborova2016statistical, antenucci2019glassy}. That kind of metastable region was thus given the name \textit{hard phase} \cite{zdeborova2016statistical, zdeborova2007phase}.
In summary, we expect that $p \geq 3$ models in the $gR$ phase can only recover partially corrupted patterns whereas $p = 2$ can recover them entirely.

\begin{figure}[t!]
    \centering
    \includegraphics[width = 0.325\textwidth]{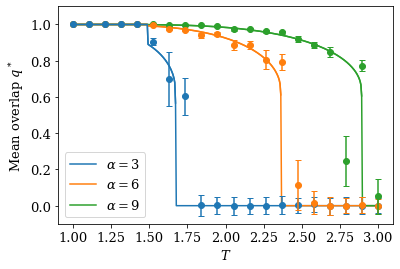}
    \includegraphics[width = 0.325\textwidth]{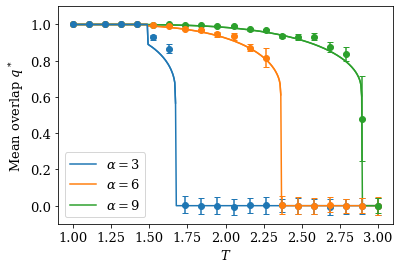}
    \includegraphics[width = 0.325\textwidth]{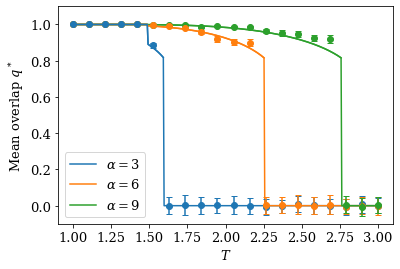}
    \caption{Monte Carlo simulations of the $p = 3$ inverse model compared against RS saddle-point solutions. The $lR$ phase is included on the left and central plots, but not on the right one. The left plot has $\varepsilon = 0$, and the two other ones have a handpicked $\varepsilon$ such that the simulations are initalized near the saddle-point solutions. The dots are simulation data at a few values of $\alpha$, and the lines are slices of the saddle-point solutions at the same $\alpha$. The teacher generates $M = \frac{\alpha N^{p-1}}{p!}$ examples $\sigma^a$ with $N = 512$ components each, and the simulation results are then averaged over $L = 100$ student patterns. The simulation data is sometimes systematically shifted up with respect to the saddle-point solution. This difference is notably visible on the central plot, right after the fall from $eR$ to $gR$ when $\alpha = 3$.}
    \label{fig:inverse_monte_carlo_simulations}
\end{figure}

Fig. (\ref{fig:inverse_monte_carlo_simulations}) shows results from Monte Carlo simulations with $p = 3$, where $L$ replicas of the student pattern $\{\xi^b\}_{b=1}^L$ are initialized to the teacher pattern $\xi^*$ corrupted by some Rademacher noise $\varepsilon$. In other words, the initial values of $\xi^b_i$ are sampled from
the distribution $\left( 1 - \varepsilon \right) \delta \left(\xi_i - \xi^*_i \right) + \frac{\varepsilon}{2} \left[ \delta \left( \xi_i+1 \right) + \delta \left( \xi_i - 1 \right) \right]$ with $\varepsilon \in [0, 1]$.
The value of $\varepsilon$ is tuned so that the simulations start relatively close to the saddle-point solutions. As explained previously, $gR$ is a hard phase, so this initialization is necessary to make $\xi^b$ converge to $gR$ in a reasonable amount of time. Additionally, it is also used to make $\xi^b$ converge to the $lR$ phase rather than the $P$ phase when desired.
Once the simulations are over, the overlaps are averaged over all $L$ replicas.
If we fix $\varepsilon = 0$, then the simulations generally converge to the $lR$ phase when it is a fixed point.
If instead we initialize them to the saddle-point solutions by handpicking $\varepsilon$, then they stay near the initial overlaps. In either case, the simulations converge to $eR$ when it is globally stable.
Some simulation data points might be systematically shifted up with respect to the saddle-point solutions. However, this difference decreases with the system size $N$, so finite size effects seem sufficient to explain it (see Fig. \ref{fig:finite_size_effects} in Appendix \ref{app:monte_carlo_simulations}). Overall, the Monte Carlo simulations are in very good agreement with the $p = 3$ overlap landscape obtained by solving the saddle-point equations numerically.

\subsection{Inference temperature vs dataset noise}
\label{sec:mismatch_discussion}

In the two next Sections, we will discuss the phase diagram when the student is only partially informed about the teacher generative model, i.e. when the Nishimori conditions do not hold. We start with the case where $p=p^*$ but $\beta\neq\beta^*$, i.e. the inference temperature $T$ is different from the dataset noise $T^*$. As we argued in Section \ref{sec:matched_student}, the student accurately retrieves $\xi^*$ when $T^* < T_ {\text{crit}}$. On the other hand, we must solve the saddle-points equations (see Eqs. \ref{eqn:saddle-point_p_s=p}) to study $T^* > T_ {\text{crit}}$.

\begin{figure}[t!]
    \centering
    \includegraphics[width = 0.325\textwidth]{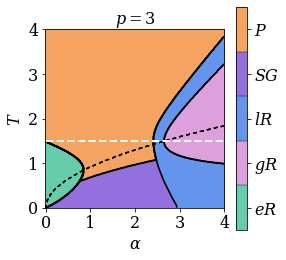}
    \includegraphics[width = 0.325\textwidth]{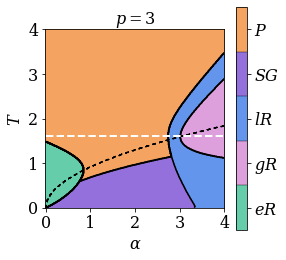}
    \includegraphics[width = 0.325\textwidth]{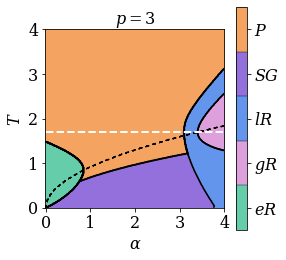}
    \includegraphics[width = 0.325\textwidth]{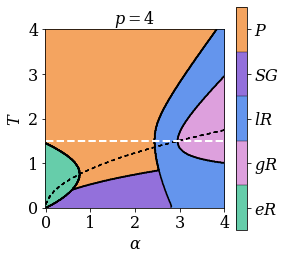}
    \includegraphics[width = 0.325\textwidth]{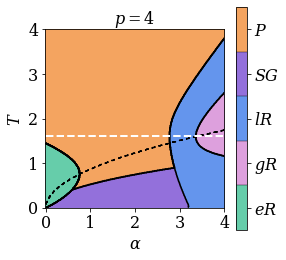}
    \includegraphics[width = 0.325\textwidth]{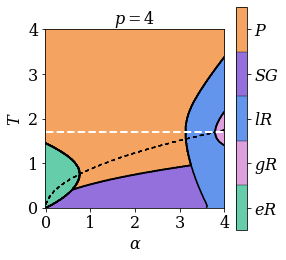}
    \caption{RS phase diagrams of inverse models with $p^* = p$ and fixed $\beta^*$. The top and bottom rows of plots respectively have $p^* = p = 3$ and $p^* = p = 4$. In the same way, the left, central and right columns correspond to $T^* = 1.5$, $T^* = 1.6$ and $T^* = 1.7$. Accurate pattern retrieval is not possible in the Paramagnetic phase ($P$), in the Spin-Glass phase ($SG$) or in the example Retrieval phase ($eR$), but it is possible in the local Retrieval phase ($lR$) and in the global Retrieval phase ($gR$). The ferromagnetic fixed point corresponding to accurate pattern retrieval is globally stable in the $gR$ phase, but locally stable in the $lR$ phase. Conversely, the $SG$ fixed point is always locally stable and leads the student to a frozen spurious signal.
        The white dashed line indicates the Nishimori line $\beta^* = \beta$. The black dashed lined is the $gR$ phase boundary on the Nishimori line. As explained in Section \ref{sec:match_discussion}, we expect it to overlap the exact $SG$ phase transition.}
    \label{fig:inverse_phase_diagrams_beta_s=/=beta}
\end{figure}

We show the phase diagram of this region on Fig. (\ref{fig:inverse_phase_diagrams_beta_s=/=beta}). At high inference temperature $T$, the situation is similar to Fig. (\ref{fig:inverse_phase_diagrams}): retrieval is possible if the data load
$\alpha$ is sufficiently large, but the paramagnetic phase is always locally stable. The situation is different when the inference temperature is low. In that case, there are two phases that we did not see
for $\beta = \beta^*$: the inaccurate $eR$ phase and the $SG$ phase. When $\alpha$ is relatively small, the student falls in the inaccurate $eR$ phase. In this regime, it has finite overlap with one of the noisy examples and cannot retrieve the signal $\xi^*$. When $\alpha$ is larger, the interference among the noisy examples prevents the student to be aligned with them. In this regime, the $SG$ phase, the student locally converge to spurious patterns that are uncorrelated with the signal.

Accurate pattern retrieval is only possible in the $lR$ and $gR$ phases where $\alpha$ is so large that the student can gather enough information from the dataset to become very close to $\xi^*$.
The phase diagrams indicate that pattern retrieval is optimal on the Nishimori line in the sense that $\beta = \beta^*$ is the inverse temperature where the student needs the least examples to recover $\xi^*$. In other words, the student's performance is nonmonotonic in $T$ and peaks at $T = T^*$. These properties were also observed in the teacher-student setting of the $p = 2$ Hopfield network \cite{alemanno2023hopfield}.

\pagebreak

Contrary to what one would expect to see on the exact phase diagram \cite{nishimori1980exact, nishimori2001statistical}, the Nishimori line $T=T^*$ does not to cross a triple point on the RS phase diagram.
The issue is that the RS phase diagram is not exact outside of the Nishimori line. In particular, the $SG$ phase boundary is not exact.
Outside of the retrieval regime, the free entropy of the inverse model is the same as the direct model.
Since the transition towards $gR$ of the inverse model on the Nishimori line overlaps the exact $P$-$SG$ transition of the direct model (see Section \ref{sec:match_discussion}), we deduce that it must also overlap the exact $P$-$SG$ transition of the \textit{inverse} model outside of the $gR$ phase.
Plotting it on the RS phase diagrams, we see that it indeed crosses the Nishimori line and the $gR$ phase boundary at the same point, which therefore becomes a triple point, as expected.

\subsection{Interaction order and noise tolerance}\label{sec:tol}
So far, we assumed that the student is informed about the interaction order used by the teacher, i.e. $p=p^*$. In this Section, we investigate the role of the student's choice of $p$ when the task is to learn from a dataset sampled by a $2$-body Hopfield network, i.e. $p^*=2$. We study two different non trivial scalings regimes of $M$ and $\beta^*$ that make pattern inference possible (see Appendix \ref{app:teacher-student_free_entropy}).

\subsubsection{Large noise scaling}
\begin{figure}[t!]
    \centering
    \includegraphics[width = 0.495\textwidth]{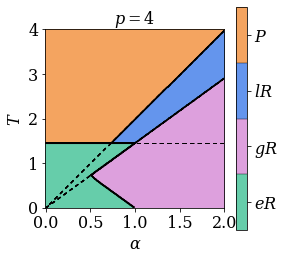}
    \includegraphics[width = 0.495\textwidth]{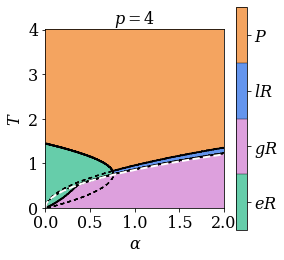}
    \caption{RS phase diagrams of inverse models with $p^* = 2$ and $p = 4$. The left plot is for $\alpha = \frac{M \left( p/2 + 1 \right)!}{N^{p/2}}$, and $\beta^* = 1 - \frac{1}{\sqrt{2}}$ such that $\eta = 1$ and the right plot is for $\alpha = \frac{M p!}{N^{p - 1}}$ and $\beta^* = \sqrt{\frac{2 \lambda}{N}}$ with $\lambda = \beta$. Accurate pattern retrieval is not possible in the Paramagnetic phase ($P$) or in the example Retrieval phase ($eR$), but it is possible in the local Retrieval phase ($lR$) and in the global Retrieval phase ($gR$). The ferromagnetic fixed point corresponding to accurate pattern retrieval is globally stable in the $gR$ phase, but locally stable in the $lR$ phase. The black dashed lines mark the metastable continuation of the $eR$, $lR$ and $gR$ phase boundaries through neighboring phases with a larger free entropy. The paramagnetic total entropy becomes negative below the white dashed line drawn on the right plot. However, the paramagnetic phase is no longer globally stable at that temperature.
    }
    \label{fig:inverse_phase_diagrams_p_s=/=p}
\end{figure}

We first consider a large noise scaling where $\beta^* \sim \mathcal{O} \left( N^{2/p - 1} \right)$ and $M \sim \mathcal{O} \left( N^{p - 1} \right)$, such that
$$\alpha = \frac{M p!}{N^{p-1}}\,, \quad \hbox{and} \quad  \lambda = \frac{\left[ \beta^* \right]^{p/2}}{(p/2)!} N^{p/2 - 1} \,,$$
are finite.
In this scaling, a $p \geq 3$ network requires $\mathcal{O} \left( N^{p - 2} \right)$ more training examples than a $p = 2$ network with finite load $\gamma = \frac{M}{N}$, but also has a higher tolerance to teacher noise. For instance, a student with $p = 4$ interactions is able to retrieve the pattern of a teacher with $T^* \sim \mathcal{O} \left( N^{1/2} \right)$ noise when it is shown enough examples $M \sim \mathcal{O} \left( N^3 \right)$ to be in the $gR$ phase (see Fig. \ref{fig:inverse_phase_diagrams_p_s=/=p}).

$\mathcal{O} \left( N^{1/2} \right)$ noise tolerance was also observed in the $p = 4$ direct model, where it is a consequence of the redundancy stemming from storing $\mathcal{O} \left( N \right)$ memories rather than the $\mathcal{O} \left( N^3 \right)$ needed to saturate the storage capacity \cite{agliari2020neural}.
\begin{figure}[t!]
    \centering
    \includegraphics[width = 0.495\textwidth]{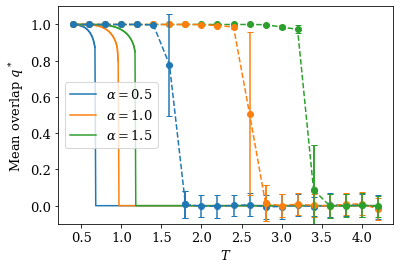}
    \caption{Monte Carlo simulations (dashed lines) and RS saddle-point solutions (full lines) of the inverse model in the large-noise scaling with $p^* = 2$ and $p = 4$. The teacher generates $M = \frac{\alpha N^{p-1}}{p!}$ examples $\sigma^a$ with $N = 256$ components each, and the simulation results are then averaged over $L = 100$ student patterns. The student patterns are all initialized to $\xi^*$.}
    \label{fig:high_T_s_monte_carlo_simulations}
\end{figure}
Our $p = 4$ inverse model exploits a different kind of redundancy by learning from $\mathcal{O} \left( N^3 \right)$ examples whereas $p = 2$ only needs $\mathcal{O} \left( N \right)$. In other terms, both storing extensively less memories than the maximum allowed amount and generating extensively more examples than the minimum required amount provide enough redundancy to recover a pattern muddled in an extensive amount of noise. In both cases, there is an $\mathcal{O} \left( N^2 \right)$ gap between the number of patterns used in the noise-tolerant and noise-susceptible regimes.
Going beyond $p = 4$, the inverse model has $\mathcal{O} \left( N^{1 - 2/p} \right)$ noise tolerance as a function of $p$. In particular, our theory predicts that the tolerance saturates at $T^* \sim \mathcal{O} \left( N \right)$ as $p \rightarrow \infty$, but at the cost of using an intractable number of examples. This behavior is different from the $\mathcal{O} \left( N^{1/2 - p/4} \right)$ tolerance of the direct $p$-body model in the noisy-learning regime studied in \cite{agliari2020tolerance}. In other terms, the dataset noise that we are facing is of a different nature than the learning noise of \cite{agliari2020tolerance}.
In any case, it is interesting that both the direct and inverse models are able to tolerate an extensive amount of noise. Overall, our results suggest that it could be advantageous to use a student network with a relatively large $p$ to learn from a large but noisy dataset when the
$p^*$ of the teacher generative model is unknown.

An unavoidable drawback of large teacher noise is that it always lead to uncorrelated examples, which makes accurate example retrieval impossible.
Instead, it is replaced by the inaccurate example retrieval phase where the student has finite overlap $m$
with a noisy example generated by the teacher but no overlap with the signal (see Fig. \ref{fig:inverse_phase_diagrams_p_s=/=p}). Depending on $T$ and $\alpha$, this phase can be either globally stable or locally stable.
For the sake of clarity, we plot only the globally stable phase on our phase diagram in Fig. (\ref{fig:inverse_phase_diagrams_p_s=/=p}). The locally stable phase is arguably less important to plot because it is identical to the locally stable ferromagnetic phase previously reported in the direct model when assuming replica symmetry (see \cite{albanese2022replica} and Fig. \ref{fig:direct_phase_diagrams}).

Given $m = 0$, the free entropy of the inverse model with $p \geq 3$, $p^* = 2$ and $\beta = \lambda$ is the same as on the Nishimori line (see Eq. \ref{eqn:saddle-point_p_s=p} and Appendix \ref{app:teacher-student_free_entropy}). As a direct consequence, the total entropy is positive outside of the $eR$ phase (see Fig. \ref{fig:inverse_phase_diagrams_p_s=/=p}).
Additionally, the $p^* = 2$, $p \geq 3$ phase diagrams with $\beta \neq \lambda$ are identical to the $p = p^*$ phase diagrams with $\beta \neq \beta^*$, which suggests that $\beta = \lambda$ is optimal for $p^* = 2$, $p \geq 3$ in the same sense as $\beta = \beta^*$ is optimal for $p = p^*$ (see Fig. \ref{fig:inverse_phase_diagrams_beta_s=/=beta}).
Monte Carlo simulations confirm that a student with $p \geq 3$ is able to retrieve the pattern of a teacher with $p = 2$ and $T^* \sim \mathcal{O} \left( N^{1/2} \right)$ (see Fig. \ref{fig:high_T_s_monte_carlo_simulations}). However, the $lR$ phase transition is at a higher $T$ in the simulations than on the $\beta = \lambda$ RS phase diagram (see Fig. \ref{fig:inverse_phase_diagrams_beta_s=/=beta}), which means that RSB is necessary to describe it accurately. One could check where replica symmetry holds by evaluating the stability of the RS saddle point throughout the phase diagram.

\subsubsection{Finite noise scaling}
We also consider a different scaling regime where $\beta^* \sim \mathcal{O} \left( 1 \right)$ and  $M \sim \mathcal{O} \left( N^{p/2} \right)$,  such that
$$\alpha = \frac{M (p/2 + 1)!}{N^{p/2}}\,,$$
is finite. In this finite-noise scaling, $p \geq 3$ requires $\mathcal{O} \left( N^{p/2 - 1} \right)$ more training examples than $p = 2$, which is a lot less than the first scaling. For instance, a student with $p = 4$ needs $\mathcal{O} \left( N^2 \right)$ examples to retrieve $\xi^*$. As before, the phase transitions are all first order, the overlap $q^*$ stays high throughout the $gR$ and $lR$ phase of $p = 4$ and $gR$ is a hard phase. The saddle-point equations (see Eqs. \ref{eqn:saddle-point_p_s=2})
are free from the pattern interference term $\sqrt{\alpha r} x$ present in their $p^* = p$ counterparts (see Eqs. \ref{eqn:saddle-point_p_s=p}) until $\beta^*$ becomes so small that is approaches $\mathcal{O} \left( N^{2/p - 1} \right)$. Therefore, contrary to $p^* = p = 2$, the network is never in the $SG$ phase. Practically, it means that $p \geq 3$ gives more freedom than $p = 2$ for tuning $\beta$ and $\alpha$. The only remaining restriction is that choosing $\alpha$ and $T$ too small puts the network into the inaccurate $eR$ phase resulting from the $k z$ term (see Fig. \ref{fig:inverse_phase_diagrams_p_s=/=p}). The saddle point equations can be derived without the RS ansatz because they do not involve $q$ and $r$. Consequently, we expect them to yield an exact solution.
Like on the Nishimori line, the total entropy of the paramagnetic phase is always positive, which is consistent with the solution being exact.

\subsection{Robustness against adversarial attacks}\label{sec:adv}
Inverse models with $p^* = 2$ and $p \geq 3$ offer an opportunity to study adversarial attacks in a simple setting because their phase diagrams have regions where the signal retrieval phases ($gR$ and $lR$) overlap with the inaccurate $eR$ phase. Recall that, in the $lR$ phase, a noisy student pattern $\xi$ either converges to $\xi^*$ or falls in the paramagnetic phase, depending on the amount of noise that $\xi$ contains initially. The quantity of noise needed to prevent pattern retrieval becomes smaller as one approaches the $lR$ to $P$ phase transition and the basin of attraction of $lR$ shrinks.
Similarly, in the region of inaccurate $eR$ where signal retrieval is metastable, patterns $\xi$ that are corrupted by replacing some of their entries $\xi_i$ by the components $\sigma^a_i$ of an example $\sigma^a$ may converge to $\sigma^a$ when enough entries are replaced. The fraction $\varepsilon$ of entries that need to be replaced becomes smaller as the basin of attraction of inaccurate $eR$ expands and overtakes that of signal retrieval.
In practice, an adversary can use this strategy to trick the student into converging to a pattern other than $\xi^*$.
This scenario is similar to an adversarial attack targeting the input of K \& H's dense Hopfield network model because the student pattern $\xi$ plays a similar role in the inverse model as the test data in K \& H's dense Hopfield networks (see Fig. \ref{fig:gardner-krotov_comparison_diagram}, Section \ref{sec:match_discussion} and Appendix \ref{app:hamiltonians}). In that analogy, the examples $\sigma$ are acting like the neural network weights rather than taking the role of the training data.

\begin{figure}[t!]
    \centering
    \includegraphics[width = 0.495\textwidth]{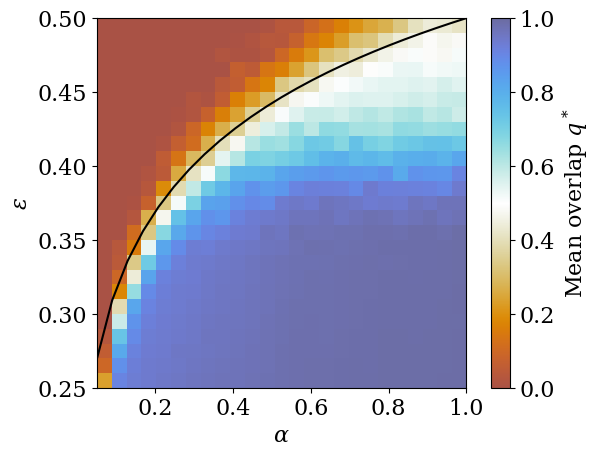}
    \includegraphics[width = 0.495\textwidth]{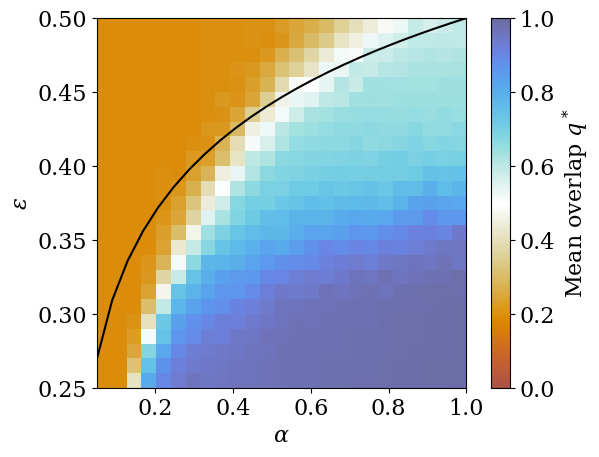}
    \caption{Monte Carlo simulations of the overlap $q^*$ as a function of $\alpha$ and adversarial attack size $\varepsilon$ in the inverse model with $p^* = 2$, $\beta^* = 1 - \frac{1}{\sqrt{2}}$, $p = 4$, $\beta = \infty$ and $N = 1024$. The simulation results are averaged over $L = 100$ student patterns. On the left plot, the inverse model is corrupted by an example $\sigma^a$ that has a small overlap with $\xi^*$ in absolute value. On the right plot, it is corrupted by the example that has the largest overlap with $\xi^*$.
        The black line $\varepsilon^* = \frac{\alpha^{1/3}}{\alpha^{1/3} + 1}$ is our analytical formula for the largest adversarial perturbation $\varepsilon$ such that the student retrieves $\xi^*$ rather than the example $\sigma^a$.}
    \label{fig:adv_phase_diagrams}
\end{figure}

We will now investigate what values of the perturbation size $\varepsilon$ are a threat by deriving a formula for the largest $\varepsilon$ such that the student converges to the signal at zero temperature. This largest $\varepsilon$ will be denoted $\varepsilon^*$, and we expect it to be a good measure of adversarial robustness. The saddle-point equations with $T = 0$ indicate that the student converges to one of the signal retrieval phases if and only if $k < \eta \alpha r^*$ (see Eqs. \ref{eqn:saddle-point_p_s=2}). Sampling the initial conditions of $\xi_i$ from $\left( 1 - \varepsilon \right) \delta \left( \xi_i-\xi^*_i \right) + \varepsilon \ \delta \left( \xi_i-\sigma^a_i \right)$ with $\varepsilon \in \left[ 0, 1 \right]$, we get
\begin{align*}
    r^*     & = p \left[ \frac{1}{N} \sum_{i = 1}^{\left( 1 - \varepsilon \right) N} \xi^*_i \xi^*_i + \frac{1}{N} \sum_{i = 1}^{\varepsilon N} \xi^*_i \sigma^a_i \right]^{p - 1}\,,       \\
    \quad k & = p \left[ \frac{1}{N} \sum_{i = 1}^{\left( 1 - \varepsilon \right) N} \xi^*_i \sigma^a_i + \frac{1}{N} \sum_{i = 1}^{\varepsilon N} \sigma^a_i \sigma^a_i \right]^{p - 1}\,.
\end{align*}
By the law of large numbers, $\frac{1}{\varepsilon N} \sum_{i = 1}^{\varepsilon N} \xi^*_i \sigma^a_i$ and $\frac{1}{\left( 1 - \varepsilon \right) N} \sum_{i = 1}^{\left( 1 - \varepsilon \right) N} \xi^*_i \sigma^a_i$ are both typically close to $m^* = \frac{1}{N} \sum_i^N \xi^*_i \sigma^a_i \approx 0$ as $N \rightarrow \infty$. If we take $\sigma^a$ to be a typical example, then $r^*$ and $k$ reduce to
\begin{align*}
    r^* & \approx p \left( 1 - \varepsilon \right)^{p - 1}\,, \\
    k   & \approx p \varepsilon^{p - 1}\,.
\end{align*}
Substituting these expressions back in $k < \eta \alpha r^*$ yields
\begin{align*}
    \varepsilon^{p - 1} & < \eta \alpha \left( 1 - \varepsilon \right)^{p - 1}   \,,                                                 \\
    \varepsilon         & < \frac{\left[ \eta \alpha \right]^{\frac{1}{p - 1}}}{\left[ \eta \alpha \right]^{\frac{1}{p - 1}} + 1}\,.
\end{align*}
In other terms, the inverse model with $p^* = 2$ and even $p \geq 3$ is resistant to adversarial attacks of size $\varepsilon^* = \frac{\left[ \eta \alpha \right]^{\frac{1}{p - 1}}}{\left[ \eta \alpha \right]^{\frac{1}{p - 1}} + 1}$ and smaller.
For $p = 4$, $\varepsilon^*$ is in good agreement with Monte Carlo simulations of the inverse model corrupted by a typical example (see Fig. \ref{fig:adv_phase_diagrams}).
This comparison is good evidence that our solution of the finite-noise scaling is indeed exact. Additionally, $\varepsilon^*$ is a decent approximation of empirical robustness even when the inverse model is corrupted by the example that has the largest overlap with $\xi^*$. A similar construction with the perturbation sampled uniformly at random gives $k \sim \mathcal{O} \left( N^{1/2 - p/2} \right) \approx 0$, so adversarial attacks are much more efficient at fooling the model than random noise. Just like adversarial attacks targeting more complicated neural networks \cite{biggio2013evasion, szegedy2013intriguing}, our example-based attack can be hard to detect at low $\varepsilon$ because a few adversarially perturbed entries $\xi_i$ do not look very different from a low amount of meaningless noise.
Moreover, $\varepsilon^*$ grows monotonically with $\alpha$, which is consistent with the common observation that larger neural networks
are also more adversarially robust \cite{madry2018towards, gowal2020uncovering, huang2021exploring, bubeck2021law, bubeck2021universal, puigcerver2022adversarial, ribeiro2023overparameterized}.
At first glance, this effect can be counter-intuitive because adversarial vulnerability looks like a form of overfitting \cite{goodfellow2015explaining}.
In our model, however, all examples work together to stabilize the $lR$ phase, and the best way to push the student into the $eR$ phase is to perturb it with a single example. Therefore, it is not surprising that increasing $\alpha$ makes the student more robust.
We recall that the examples $\sigma$ are a feature-based representation of $\xi^*$.
Interestingly, it means that the underlying mechanism of our example-based attack is conceptually similar to gradient-based attacks targeting many common types of neural networks \cite{goodfellow2015explaining}.
In fact, gradient-based attacks find features stored in neural network weights and add them to the data in order to fool the network \cite{goodfellow2015explaining, jetley2018friends, ilyas2019adversarial, tsipras2019robustness}. It would be interesting to investigate, both empirically and theoretically, if only a small number of weights are involved in constructing these adversarial attacks. If it is the case, it could explain why larger neural networks are often more robust.
In general, we expect this kind of one-example attack to be possible in any region of signal retrieval that overlaps with the inaccurate $eR$ phase.
Using $p \neq p^*$ may not be a necessary ingredient of adversarial vulnerability in more general models with other sources of mismatch, but in our case it ensures that the signal retrieval phases intersect the inaccurate $eR$ phase.
Conversely, the accurate $eR$ phase is by definition robust to adversarial attacks since retrieving an example $\sigma^a$ is the same as recovering $\xi^*$. This distinction clarifies why the dense Hopfield networks designed by K \& H are adversarially robust in the prototype phase despite being adversarially vulnerable in the feature phase. In fact, K \& H observed that adversarial attacks are unsuccessful in the prototype phase specifically because they retrieve stored examples that are semantically meaningful \cite{krotov2018dense}.
In summary, our model yields two main results concerning adversarial examples. First of all, it suggests a reason why large feature-based neural networks are more adversarially robust than smaller ones. Second of all, it clarifies why dense Hopfield networks are much more robust in the prototype phase than in the feature phase.

\section{Conclusion}
\label{sec:dense_HN_conclusion}

In this work, we derive the exact phase diagram of the $p$-dense networks in the teacher-student setting \cite{barra2017phase, barra2018phase, gardner1987multiconnected, alemanno2023hopfield}. On the Nishimori line, we find an example Retrieval phase ($eR$) and a global Retrieval phase ($gR$) reminiscent of the prototype and feature regimes observed empirically in dense Hopfield networks \cite{krotov2016dense}. We show that the phase transition towards $gR$ of the inverse model overlaps the Paramagnetic to Spin-Glass ($P$-$SG$) transition of the direct model, which allows us to locate the $P$-$SG$ transition much more precisely than before \cite{gardner1987multiconnected, albanese2022replica}. On the other hand, we discover that inverse models outside of the Nishimori line are able to resist an extensive amount of noise. In fact, a student with $p \geq 3$ is able to learn from a teacher with $p^* = 2$ even when the teacher's inverse temperature $\beta^*$ is as low as $\mathcal{O} \left( N^{2/p - 1} \right)$. Moreover, such a student is immune to pattern interference until $\beta^*$ reaches $\mathcal{O} \left( N^{2/p - 1} \right)$. In this setting, we derive a formula measuring the adversarial robustness of the student with $p \geq 3$ and $T = 0$. We then use this formula to describe how making a neural network larger can potentially increase its robustness to adversarial attacks constructed with only a few learned weights \cite{madry2018towards, gowal2020uncovering, huang2021exploring, bubeck2021law, bubeck2021universal, puigcerver2022adversarial, ribeiro2023overparameterized}. Our model also clarifies why the prototype phase of dense Hopfield networks is adversarially robust \cite{krotov2018dense}. We compare our key results against Monte Carlo simulations.

Dense networks with exponential interactions have been argued to be the $p \rightarrow \infty$ limit of the $p$-body models  \cite{demircigil2017model}. It would be interesting to see if they can achieve $\mathcal{O} \left( N \right)$ noise tolerance at the cost of an exponential number of training examples. More generally, studying exponential models in the teacher-student setting would be an interesting extension of this work and could be used to complement existing studies of the direct model \cite{demircigil2017model, lucibello2023exponential}.
A caveat of our model is that the teacher has only one pattern. In fact, we would need to use a teacher with at least two patterns to describe more completely the kind of adversarial attack aiming to misclassify data.
It should be possible to study this kind of teacher by using an approach similar to \cite{hou2019minimal}.
In particular, \cite{barra2017phase} and \cite{hou2019minimal} argue that the performance of restricted Boltzmann machines with a finite number $P$ of i.i.d. planted patterns is independent of $P$ in the teacher-student setting.
It would be interesting to investigate whether this characteristic also holds for $p$-body dense networks. On the practical side, we highlight the untapped benefits of using $p$-body models to either resist an extensive amount of noise in the feature phase or improve adversarial robustness in the prototype phase. Overall, we stress that further investigations of dense Hopfield networks could unlock their true potential.

\section*{Data availability}
The figures can be reproduced using the code available on \href{https://github.com/RobinTher/Dense_Associative_Network}{this public Github repository}.


\begin{subappendices}

\section{Gardner's Hamiltonian vs K \& H's Hamiltonian}
\label{app:hamiltonians}

Consider the generalized Hopfield Hamiltonian $H \left[ \sigma \mid \xi \right] = -\sum_{i_1 < ... < i_p = 1}^{N} J_{i_1 ... i_p} \sigma_{i_1} ... \sigma_{i_p}$ with $p$-body interactions $J_{i_1 ... i_p} = \frac{p!}{N^{p - 1}} \sum_{\mu=1}^{M} \xi_{i_1}^{\mu} ... \xi_{i_p}^{\mu}$ described by Gardner \cite{gardner1987multiconnected}, where $M$ indicates the number of patterns $\xi^{\mu}$ used to construct $J$, and $N$ denotes the number of components of each pattern $\xi^{\mu}$ and example $\sigma$. In this Section, we will omit $\xi$ in the argument of $H \left[ \sigma \mid \xi \right]$ and write $H \left[ \sigma \right]$ instead for notational simplicity. Unless indicated otherwise, we will assume a large number number of components $N \gg 1$ and patterns $M \sim \mathcal{O} \left( N^{p-1} \right)$. We will start by comparing it to the dense Hopfield network Hamiltonian $\mathcal{H} \left[ \sigma \right] = -\frac{1}{N^{p-1}} \sum_{\mu} \left( \sum_{i} \xi_{i}^{\mu} \sigma_{i} \right)^p$ studied by K \& H \cite{krotov2016dense}.

For that purpose, we rewrite $H$ in the form $H \left[ \sigma \right] = -\frac{1}{p!} \sum_{i_1 \neq ... \neq i_p} J_{i_1 ... i_p} \sigma_{i_1} ... \sigma_{i_p}$ by summing over all permutations of $\{ i_1 ... i_p \}$ in place of the restricted set $i_1 < ... < i_p$ and compensating for double counting with the prefactor $\frac{1}{p!}$. This manipulation leads to
\begin{align*}
    H \left[ \sigma \right] & = -\frac{1}{p!} \sum_{i_1 \neq ... \neq i_p} J_{i_1 ... i_p} \sigma_{i_1} ... \sigma_{i_p}                                        \\
                            & = -\frac{1}{N^{p-1}} \sum_{\mu} \sum_{i_1 \neq ... \neq i_p} \xi_{i_1}^{\mu} ... \xi_{i_p}^{\mu} \sigma_{i_1} ... \sigma_{i_p}\,.
\end{align*}
On the other hand, K \& H's Hamiltonian may be rewritten
\begin{align*}
    \mathcal{H} \left[ \sigma \right] & = -\frac{1}{N^{p-1}} \sum_{\mu} \left( \sum_{i} \xi_{i}^{\mu} \sigma_{i} \right)^p                                                                \\
                                      & = -\frac{1}{N^{p-1}} \sum_{\mu} \left( \sum_{i_1} \xi_{i_1}^{\mu} \sigma_{i_1} \right) ... \left( \sum_{i_p} \xi_{i_p}^{\mu} \sigma_{i_p} \right) \\
                                      & = -\frac{1}{N^{p-1}} \sum_{\mu} \sum_{i_1 ... i_p} \xi_{i_1}^{\mu} ... \xi_{i_p}^{\mu} \sigma_{i_1} ... \sigma_{i_p}\,,
\end{align*}
where the sum over $i_1 ... i_p$ includes both the set of indices $i_1 \neq ... \neq i_p$ found in $H \left[ \sigma \right]$ and other configurations where some indices are equal. For example, the configuration $i_1 \neq ... \neq i_{p-1} = i_p$ contains the fewest equal indices after $i_1 \neq ... \neq i_p$. In other words, $\mathcal{H} \left[ \sigma \right]$ can be expressed as an expansion around $H \left[ \sigma \right]$, and the two Hamiltonians are equivalent when the normalized residuals $\frac{\mathcal{H} \left[ \sigma \right] - H \left[ \sigma \right]}{N}$ vanish in the limit of large $N$. In this study, we encounter two cases which bring different results.
\begin{enumerate}[label = \textbf{\arabic*}]
    \item The Hamiltonians $\mathcal{H} \left[ \sigma \right]$ and $H \left[ \sigma \right]$ are dominated by a few closely packed configurations $\xi^{\mu}$ that have finite overlap $\frac{1}{N} \sum_i \xi^{\mu}_i \sigma_i \sim \mathcal{O} \left( 1 \right)$ with $\sigma$. We say that they are aligned with $\sigma$.
          \label{case:strongly-correlated}
    \item The Hamiltonians $\mathcal{H} \left[ \sigma \right]$ and $H \left[ \sigma \right]$ are dominated by many spread out configurations $\xi^{\mu}$ that have microscopic overlap $\frac{1}{N} \sum_i \xi^{\mu}_i \sigma_i \sim \mathcal{O} \left( N^{-1/2} \right)$ with $\sigma$. We say that they are misaligned with $\sigma$
          \label{case:weakly-correlated}
\end{enumerate}
We use the expansion of $\mathcal{H} \left[ \sigma \right]$ to discuss both the aligned case and the misaligned case. We start by writing the $i_1 \neq ... \neq i_p$ and $i_1 \neq ... \neq i_{p-1} = i_p$ terms explicitly, which leads to the form
\begin{align*}
    \mathcal{H} \left[ \sigma \right] & = -\frac{1}{N^{p-1}} \sum_{\mu} \sum_{i_1 \neq ... \neq i_p} \xi_{i_1}^{\mu} ... \xi_{i_p}^{\mu} \sigma_{i_1} ... \sigma_{i_p}                                                                                 \\
                                      & \quad- \frac{1}{2} \frac{p (p - 1)}{N^{p-1}} \sum_{\mu} \sum_{i_1 \neq ... \neq i_{p-1}} \xi_{i_1}^{\mu} ... \left( \xi_{i_{p-1}}^{\mu} \right)^2 \sigma_{i_1} ... \left( \sigma_{i_{p-1}} \right)^2 + ... \,,
\end{align*}
because there are $\binom{p}{2} = \frac{p (p - 1)}{2}$ ways for the indices $i_{p-1}$ and $i_p$ to be equal. This expression can be summarized by $\mathcal{H} \left[ \sigma \right] = H \left[ \sigma \right] + H' \left[ \sigma \right] + ...$, where $H' \left[ \sigma \right]$ simplifies to
\begin{align*}
    H' \left[ \sigma \right] & = -\frac{1}{2} \frac{p (p - 1)}{N^{p-1}} \sum_{\mu} \sum_{i_1 \neq ... \neq i_{p-1}} \xi_{i_1}^{\mu} ... \left( \xi_{i_{p-1}}^{\mu} \right)^2 \sigma_{i_1} ... \left( \sigma_{i_{p-1}} \right)^2 \\
                             & = -\frac{1}{2} \frac{p (p - 1)}{N^{p-2}} \sum_{\mu} \sum_{i_1 \neq ... \neq i_{p-2}} \xi_{i_1}^{\mu} ... \xi_{i_{p-2}}^{\mu} \sigma_{i_1} ... \sigma_{i_{p-2}}                                   \\
                             & = -\frac{1}{2} \frac{p!}{N^{p-2}} \sum_{\mu} \sum_{i_1 < ... < i_{p-2}} \xi_{i_1}^{\mu} ... \xi_{i_{p-2}}^{\mu} \sigma_{i_1} ... \sigma_{i_{p-2}}\,.
\end{align*}
In the aligned case, $H' \left[ \sigma \right]$ is $\mathcal{O} \left( 1 \right)$ in $N$ because the sum over $i_1 < ... < i_{p-2}$ is $\mathcal{O} \left( N^{p - 2} \right)$. The terms implied by the ellipsis are even smaller because their sums are resctricted by more equality constraints. Therefore, the residuals $\frac{\mathcal{H} \left[ \sigma \right] - H \left[ \sigma \right]}{N}$ vanish in the limit of large $N$, and the two Hamiltonians are equivalent. Conversely, we find that $\mathcal{H} \left[ \sigma \right]$ and $H \left[ \sigma \right]$ differ from each other in the misaligned case (see Appendix \ref{app:direct_cumulant} for more details).
Therefore, although the phases of $H \left[ \sigma \right]$ that we obtain in this study are qualitatively similar to the ones observed by K \& H \cite{krotov2016dense, krotov2018dense}, the phase diagram of $H \left[ \sigma \right]$ must be compared against a simulation of $H \left[ \sigma \right]$ rather than $\mathcal{H} \left[ \sigma \right]$ in order to test our theory quantitatively.

To understand how to sample $\sigma$ in both models, consider a Monte Carlo simulation used to find the statistical equilibrium of a spin ensemble $\sigma$ with Hamiltonian $G \left[ \sigma \right]$.
To be more specific, suppose $\sigma$ is updated to a new state $\sigma'$ with a randomly selected spin $\sigma_i$ flipped with acceptance probability $P_i = \frac{1}{1 + \exp \left[ \beta \left( G \left[ \sigma' \right] -  G \left[ \sigma \right]\right) \right]}$ for a large number of time-steps. This approach works well for $G \left[ \sigma \right] = \mathcal{H} \left[ \sigma \right]$. However, in the case of $H \left[ \sigma \right]$, we find that the simulation only converges when we use the local field $h_i = \frac{p!}{N^{p-1}} \sum_{\mu} \xi^{\mu}_i \sum_{i_1 < ... < i_{p-1}} \xi^{\mu}_{i_1} ... \xi^{\mu}_{i_{p-1}} \sigma_{i_1} ... \sigma_{i_{p-1}}$ mentioned by Gardner \cite{gardner1987multiconnected} to approximate $\frac{H \left[ \sigma' \right] -  H \left[ \sigma \right]}{2 \sigma_i}$ at large $N$. In other words, we iteratively flip randomly chosen spins $\sigma_i$ with acceptance probability $P_i = \frac{1}{1 + \exp \left( 2 \beta h_i \sigma_i \right)}$ for a large number of time steps. For arbitrary $p$, it is not obvious how to compute $h_i$ quickly as a sub-routine of the Monte Carlo simulation. However, we find that both $p = 3$ and $p = 4$ have closed-formed expressions that are easy to evaluate numerically in an efficient way. To be more precise,
\begin{itemize}
    \item $p = 3$ leads to $h_i = 3 \sum_{\mu} \xi^{\mu}_i \left[ \left( \frac{1}{N} \sum_j \xi^{\mu}_j \sigma_j \right)^2 - \frac{1}{N} \right]$,
    \item and $p = 4$ leads to $h_i = 4 \sum_{\mu} \xi^{\mu}_i \left( \frac{1}{N} \sum_j \xi^{\mu}_j \sigma_j \right) \left[ \left( \frac{1}{N} \sum_j \xi^{\mu}_j \sigma_j \right)^2 - \frac{3}{N} \right]$.
\end{itemize}
For this reason and also because the number $M \sim \mathcal{O} \left( N^{p-1} \right)$ of patterns $\xi^{\mu}$ used in a Monte Carlo simulations increases exponentially with $p$, we choose to simulate only $p = 3$ and $p = 4$.

The output of the neural network model that K \& H designed for classification of data is
\begin{equation*}
    c_j = \tanh \left[ \frac{1}{2} \beta \left( \mathcal{H} \left[ \sigma' \right] - \mathcal{H} \left[ \sigma \right] \right) \right] \approx \tanh \left[ \beta p \sum_{\mu} \xi^{\mu}_j \left( \frac{1}{N} \sum_i \xi^{\mu}_i \sigma_i \right)^{p-1} \right]
\end{equation*}
We omit the linear rectifier present in the original paper \cite{krotov2016dense} because the overlaps $\frac{1}{N} \sum_{i} \xi_{i}^{\mu} \sigma_{i}$ are almost always positive (see for example the Supplement of \cite{boukacem2024waddington}). The predicted class is then $j' = \argmax_j \left\{ c_i \right\}$. Using $1 - P_j = \frac{1}{1 + \exp \left[ \beta \left( \mathcal{H} \left[ \sigma' \right] - \mathcal{H} \left[ \sigma \right]\right) \right]}$ instead of $c_j$ does not change $j'$ because $1 - P_j$ and $c_j$ are related by $1 - P_j = \frac{1}{2} \left[ c_j + 1 \right]$. When we evaluate $P_i$ using $H$ instead of $\mathcal{H}$, this relation does not always hold exactly. Rather, it should be considered an approximation.

\section{Direct model cumulant expansions}
\label{app:direct_cumulant}
In the direct model, the average replicated partition function $\left\langle Z^L \right\rangle$ takes the form:
\begin{align*}
    \left\langle Z^L \right\rangle & = \left\langle \sum_{\sigma} \exp \left( -\beta \sum_{\gamma = 1}^{L} H \left[ \sigma^{\gamma} \mid \xi \right] \right) \right\rangle \,,
\end{align*}
with $\sigma = \begin{Bmatrix} \sigma^1 \ \dots \ \sigma^{L} \end{Bmatrix}$. Gardner simplifies it to
\begin{align}
    \label{eqn:direct_model_partition_function}
    \left\langle Z^L \right\rangle  \approx \Bigg\langle \sum_{\sigma} \exp \Bigg( \beta N \sum_{\gamma} \sum_{\mu \in \Gamma_{\gamma}} \left[ \frac{1}{N} \sum_i \xi^{\mu}_i \sigma^{\gamma}_i \right]^p
    + \beta \sum_{\gamma} \sum_{\mu \in \Bar{\Gamma}} \frac{p!}{N^{p-1}} \sum_{i_1 < ... < i_p} \xi^{\mu}_{i_1} ... \xi^{\mu}_{i_p} \sigma^{\gamma}_{i_1} ... \sigma^{\gamma}_{i_p} \Bigg) \Bigg\rangle\,,
\end{align}
where the sets $\Gamma_{\gamma}$ contain the patterns $\xi^{\mu}$ that have macroscopic overlap with $\sigma_{\gamma}$, and their complement $\Bar{\Gamma} = \cap_{\gamma} \Bar{\Gamma}_{\gamma}$ consists of the remaining patterns. Two approximations are used to obtain this expression:
\begin{itemize}
    \item $\sum_{\mu \in \Gamma_{\gamma}} \frac{p!}{N^{p-1}} \sum_{i_1 < ... < i_p} \xi^{\mu}_{i_1} ... \xi^{\mu}_{i_p} \sigma^{\gamma}_{i_1} ... \sigma^{\gamma}_{i_p} \approx N \sum_{\mu \in \Gamma_{\gamma}} \left[ \frac{1}{N} \sum_i \xi^{\mu}_i \sigma^{\gamma}_i \right]^p$ because this part of \linebreak$H \left[ \sigma^{\gamma} \mid \xi \right]$ is aligned with $\sigma$ (see Case \ref{case:strongly-correlated} of Appendix \ref{app:hamiltonians}).
    \item $\sum_{\mu \in \Bar{\Gamma}_{\gamma}} \frac{p!}{N^{p-1}} \sum_{i_1 < ... < i_p} \xi^{\mu}_{i_1} ... \xi^{\mu}_{i_p} \sigma^{\gamma}_{i_1} ... \sigma^{\gamma}_{i_p} \approx \sum_{\mu \in \Bar{\Gamma}} \frac{p!}{N^{p-1}} \sum_{i_1 < ... < i_p} \xi^{\mu}_{i_1} ... \xi^{\mu}_{i_p} \sigma^{\gamma}_{i_1} ... \sigma^{\gamma}_{i_p}$ since $\Bar{\Gamma}$ contains almost all of the elements in each $\Bar{\Gamma}_{\gamma}$ when $N$ is large.
\end{itemize}
Gardner evaluates the contribution of the $\mu \in \Bar{\Gamma}$ terms via a cumulant expansion, resulting in:
\begin{align*}
     & \log \left\langle \exp \left( \beta \sum_{\gamma} \frac{p!}{N^{p-1}} \sum_{i_1 < ... < i_p} \xi^{\mu}_{i_1} ... \xi^{\mu}_{i_p} \sigma^{\gamma}_{i_1} ... \sigma^{\gamma}_{i_p} \right) \right\rangle                                                                                                                                                                                                   \\
     & \quad\approx \beta \left\langle \sum_{\gamma} \frac{p!}{N^{p-1}} \sum_{i_1 < ... < i_p} \xi^{\mu}_{i_1} ... \xi^{\mu}_{i_p} \sigma^{\gamma}_{i_1} ... \sigma^{\gamma}_{i_p} \right\rangle + \frac{1}{2} \beta^2 \left\langle \left[ \sum_{\gamma} \frac{p!}{N^{p-1}} \sum_{i_1 < ... < i_p} \xi^{\mu}_{i_1} ... \xi^{\mu}_{i_p} \sigma^{\gamma}_{i_1} ... \sigma^{\gamma}_{i_p} \right]^2 \right\rangle \\
     & \quad\approx \frac{1}{2} \beta^2 \left\langle \left[ \sum_{\gamma} \frac{p!}{N^{p-1}} \sum_{i_1 < ... < i_p} \xi^{\mu}_{i_1} ... \xi^{\mu}_{i_p} \sigma^{\gamma}_{i_1} ... \sigma^{\gamma}_{i_p} \right] \left[ \sum_{\delta} \frac{p!}{N^{p-1}} \sum_{j_1 < ... < j_p} \xi^{\mu}_{j_1} ... \xi^{\mu}_{j_p} \sigma^{\delta}_{j_1} ... \sigma^{\delta}_{j_p} \right] \right\rangle \,,
\end{align*}
because the product of independent spins $\xi^{\mu}_{i_1} ... \xi^{\mu}_{i_p}$ averages to $0$. The sums are then regrouped to get
\begin{align*}
     & \log \left\langle \exp \left( \beta \sum_{\gamma} \frac{p!}{N^{p-1}} \sum_{i_1 < ... < i_p} \xi^{\mu}_{i_1} ... \xi^{\mu}_{i_p} \sigma^{\gamma}_{i_1} ... \sigma^{\gamma}_{i_p} \right) \right\rangle                                                                                                                                    \\
     & \quad= \frac{1}{2} \beta^2 \left[ \frac{p!}{N^{p-1}} \right]^2 \left\langle \sum_{\gamma} \sum_{\delta} \sum_{i_1 < ... < i_p} \sum_{j_1 < ... < j_p} \xi^{\mu}_{i_1} \xi^{\mu}_{j_1} ... \xi^{\mu}_{i_p} \xi^{\mu}_{j_p} \ \sigma^{\gamma}_{i_1} \sigma^{\delta}_{j_1} ... \sigma^{\gamma}_{i_p} \sigma^{\delta}_{j_p} \right\rangle\,.
\end{align*}
Consider $\xi^{\mu}_{i} \xi^{\mu}_{j}$ for an arbitrary pair of indices $i$ and $j$. There are two cases.
\begin{itemize}
    \item If $i = j$, then $\xi^{\mu}_{i} \xi^{\mu}_{j}$ is deterministic and equal to $1$.
    \item If $i \neq j$, then $\xi^{\mu}_{i} \xi^{\mu}_{j}$ can be either $+1$ and $-1$ with equal probabilities.
\end{itemize}
On the one hand, if $i_n = j_n$ for all $n$, then $\left\langle \xi^{\mu}_{i_1} \xi^{\mu}_{j_1} ... \xi^{\mu}_{i_p} \xi^{\mu}_{j_p} \right\rangle = 1$. On the other hand, if $i_n \neq j_n$ for some $n$, then $\left\langle \xi^{\mu}_{i_1} \xi^{\mu}_{j_1} ... \xi^{\mu}_{i_p} \xi^{\mu}_{j_p} \right\rangle = 0$ because $\xi^{\mu}_{i_1} \xi^{\mu}_{j_1} ... \xi^{\mu}_{i_p} \xi^{\mu}_{j_p}$ is still a product of independent random spins once the deterministic variables are removed. These two cases can be summarized by $\left\langle \xi^{\mu}_{i_1} \xi^{\mu}_{j_1} ... \xi^{\mu}_{i_p} \xi^{\mu}_{j_p} \right\rangle = \delta_{i_1 j_1} ... \delta_{i_p j_p}$, which then gives
\begin{align*}
     & \log \left\langle \exp \left( \beta \sum_{\gamma} \frac{p!}{N^{p-1}} \sum_{i_1 < ... < i_p} \xi^{\mu}_{i_1} ... \xi^{\mu}_{i_p} \sigma^{\gamma}_{i_1} ... \sigma^{\gamma}_{i_p} \right) \right\rangle                                                                        \\
     & \quad= \frac{1}{2} \beta^2 \left[ \frac{p!}{N^{p-1}} \right]^2 \sum_{\gamma} \sum_{\delta} \sum_{i_1 < ... < i_p} \sum_{j_1 < ... < j_p} \delta_{i_1 j_1} ... \delta_{i_p j_p} \ \sigma^{\gamma}_{i_1} \sigma^{\delta}_{j_1} ... \sigma^{\gamma}_{i_p} \sigma^{\delta}_{j_p} \\
     & \quad= \frac{1}{2} \beta^2 \left[ \frac{p!}{N^{p-1}} \right]^2 \sum_{\gamma} \sum_{\delta} \sum_{i_1 < ... < i_p} \sigma^{\gamma}_{i_1} \sigma^{\delta}_{i_1} ... \sigma^{\gamma}_{i_p} \sigma^{\delta}_{i_p}                                                                \\
     & \quad\approx \frac{1}{2} \beta^2 \frac{p!}{N^{p-1}} \frac{1}{N^{p-1}} \sum_{\gamma \delta} \left[ \sum_{i} \sigma^{\gamma}_{i} \sigma^{\delta}_{i} \right]^p                                                                                                                 \\
     & \quad= \beta^2 \frac{p!}{N^{p-1}} N \sum_{\gamma < \delta} \left[ \frac{1}{N} \sum_{i} \sigma^{\gamma}_{i} \sigma^{\delta}_{i} \right]^p + \frac{1}{2} \beta^2 \frac{p!}{N^{p-1}} L N \,.
\end{align*}
The order $n > 2$ terms are subdominant in $N$ and can be neglected when $p \geq 3$ \cite{gardner1987multiconnected}. The RS free entropy is then obtained through a standard approach to the replica method. Note that Gardner's Hamiltonian is misaligned with $\sigma$ when the free entropy is dominated by this cumulant expansion (see Case \ref{case:weakly-correlated} of Appendix \ref{app:hamiltonians}). In the case of K \& H's Hamiltonian, we must also take into account the correction $H' \left[ \sigma \right] = \frac{1}{2} \frac{p!}{N^{p-2}} \sum_{\gamma} \sum_{i_1 < ... < i_{p-2}} \xi^{\mu}_{i_1} ... \xi^{\mu}_{i_{p-2}} \sigma^{\gamma}_{i_1} ... \sigma^{\gamma}_{i_{p-2}}$ introduced in appendix \ref{app:hamiltonians} by imposing $i_{p-1} = i_p$. In fact, a cumulant expansion of this expression gives
\begin{align*}
     & \log \left\langle \exp \left( \beta p \sum_{\gamma} \frac{1}{2} \frac{p!}{N^{p-2}} \sum_{i_1 < ... < i_{p-2}} \xi^{\mu}_{i_1} ... \xi^{\mu}_{i_{p-2}} \sigma^{\gamma}_{i_1} ... \sigma^{\gamma}_{i_{p-2}} \right) \right\rangle \\
     & \quad\approx \frac{1}{4} \beta^2 \frac{p!}{N^{p-2}} \frac{p (p - 1)}{N^{p-2}} \sum_{\gamma < \delta} \left[ \sum_{i} \sigma^{\gamma}_{i} \sigma^{\delta}_{i} \right]^{p-2} + \frac{1}{8} \beta^2 \frac{p!}{N^{p-2}} L           \\
     & \quad= \frac{1}{4} p (p - 1) \beta^2 \frac{p!}{N^{p-1}} N \sum_{\gamma < \delta} \left[ \frac{1}{N} \sum_{i} \sigma^{\gamma}_{i} \sigma^{\delta}_{i} \right]^{p-2} + \frac{1}{8} \beta^2 \frac{p!}{N^{p-1}} L N\,,
\end{align*}
which contributes to the free energy on the same order in $N$ as Gardner's Hamiltonian. Therefore, K \& H's Hamiltonian is not equivalent to Gardner's Hamiltonian when the latter is misaligned with $\sigma$ (see Case \ref{case:weakly-correlated}). The index configurations with more equality constraints also contribute to the free entropy on the same order in $N$ because the factors of $N$ that are lost to equality constraints are restored when the sums get squared in the cumulant expansion.

$p = 2$ is the only positive integer such that Gardner's Hamiltonian and Krotov's Hamiltonian are equivalent \cite{amit1985storing, gardner1987multiconnected}. In the misaligned case with a single stored pattern $\xi^*$ (see Case \ref{case:weakly-correlated}), the free entropy of $p = 2$ simplifies to\
\begin{align*}
    \frac{\log \left( Z \right)}{N} & = \frac{1}{N} \log \left\langle \exp \left\{ \beta \frac{2}{N} \sum_{i_1 < i_2} \xi^*_{i_1} \xi^*_{i_2} \sigma^{\gamma}_{i_1} \sigma^{\gamma}_{i_2} \right\} \right\rangle + \log 2                                                \\
                                    & = \frac{1}{N} \log \left\langle \exp \left( -\beta \right) \int_{\mathbb{R}} dx \frac{1}{\sqrt{2 \pi}} \exp \left\{ -\frac{1}{2} x^2 + x \sqrt{\beta \frac{2}{N}} \sum_i \xi^*_i \sigma^{\gamma}_i \right\} \right\rangle + \log 2 \\
                                    & = \frac{1}{N} \log \left[ \int_{\mathbb{R}} dx \frac{1}{\sqrt{2 \pi}} \exp \left\{ -\frac{1}{2} x^2 \right\} \cosh^N \left( x \sqrt{\beta \frac{2}{N}} \right) \right] - \beta \frac{1}{N} + \log 2\,,
\end{align*}
by using the Hubbard-Stratonovich transformation. At large $N$, it approximates to:
\begin{align*}
    \frac{\log \left( Z \right)}{N} & \approx \frac{1}{N} \log \left[ \int dx \frac{1}{\sqrt{2 \pi}} \exp \left\{ -\frac{1}{2} x^2 \right\} \left( 1 + \beta \frac{1}{N} x^2 \right)^N \right] - \beta \frac{1}{N} + \log 2 \\
                                    & \approx \frac{1}{N} \log \left[ \int_{\mathbb{R}} dx \frac{1}{\sqrt{2 \pi}} \exp \left\{ -\frac{1}{2} x^2 \right\} \exp \left( \beta x^2 \right) \right] - \beta \frac{1}{N} + \log 2 \\
                                    & = \left( -\frac{1}{2} \log \left( 1 - 2 \beta \right) - \beta \right) \frac{1}{N} + \log 2\,,
\end{align*}
thanks to the well-known limit $\lim_{N \rightarrow \infty} \left( 1 + \frac{1}{N} z \right)^N = \exp \left( z \right)$. This free entropy is consistent with the one found in literature when $\alpha = \frac{1}{N}$ \cite{amit1985storing}.

\section{Teacher-student replicated partition function}
\label{app:teacher-student_partition}

Recall that the student samples its pattern from the posterior $P \left( \xi \mid \sigma \right) = \frac{P \left( \xi \right) \prod_a P \left( \sigma^a \mid \xi \right)}{P \left( \sigma \right)}$ (see Section \ref{sec:dense_HN_teacher-student}). Given $P \left( \xi \right)$ uniform, it can be rewritten as $P \left( \xi \mid \sigma \right) = \frac{\prod_a P \left( \sigma^a \mid \xi \right)}{\sum_{\xi} \prod_a P \left( \sigma^a \mid \xi \right)}$, where $P \left( \sigma^a \mid \xi \right)$ is the distribution of the direct model with a single pattern $\xi$. To simplify $P \left( \xi \mid \sigma \right)$ further, we need to manipulate the partition function $Z = \sum_{\sigma^a} \exp \left( -\beta H \left[ \sigma^a \mid \xi \right] \right)$ of $P \left( \sigma^a \mid \xi \right)$ (see Appendix \ref{app:hamiltonians} for the definition of $H \left[ \sigma \mid \xi \right]$). Under the gauge transformation $\sigma^a_i \rightarrow \xi_i \sigma^a_i$, we may write
\begin{align*}
    Z = \sum_{\sigma^a} \exp \left( \beta \frac{p!}{N^{p-1}} \sum_{i_1 < ... < i_p} \sigma^a_{i_1} ... \sigma^a_{i_p} \right)\,,
\end{align*}
without changing the configurations of $\sigma^a$ that we are summing over. Therefore, $Z$ does not depend on $\xi$, and we can factor it out of the sum $\sum_{\xi}$, which yields
\begin{align*}
    P \left( \xi \mid \sigma \right) & = \frac{\prod_a \frac{1}{Z} \exp \left( \beta \frac{p!}{N^{p-1}} \sum_{i_1 < ... < i_p} \xi_{i_1} ... \xi_{i_p} \sigma^{a}_{i_1} ... \sigma^{a}_{i_p} \right)}{\sum_{\xi} \prod_a \frac{1}{Z} \exp \left( \beta \frac{p!}{N^{p-1}} \sum_{i_1 < ... < i_p} \xi_{i_1} ... \xi_{i_p} \sigma^{a}_{i_1} ... \sigma^{a}_{i_p} \right)} \\
                                  & = \frac{\exp \left( \beta \frac{p!}{N^{p-1}} \sum_a \sum_{i_1 < ... < i_p} \xi_{i_1} ... \xi_{i_p} \sigma^{a}_{i_1} ... \sigma^{a}_{i_p} \right)}{\sum_{\xi} \exp \left( \beta \frac{p!}{N^{p-1}} \sum_a \sum_{i_1 < ... < i_p} \xi_{i_1} ... \xi_{i_p} \sigma^{a}_{i_1} ... \sigma^{a}_{i_p} \right)}\,.
\end{align*}
Therefore, we define the partition function of the inverse model to be $\mathcal{Z} = \sum_{\xi} \exp \left( -\beta H \left[ \xi \mid \sigma \right] \right)$ (again, see Appendix \ref{app:hamiltonians} for the definition of $H \left[ \xi \mid \sigma \right]$). The $L^{\text{th}}$ power of $\mathcal{Z}$ and its average then take the form
\begin{align*}
    \mathcal{Z}^L                            & = \sum_{\xi} \prod_b \exp \left( \beta \frac{p!}{N^{p-1}} \sum_a \sum_{i_1 < ... < i_p} \xi^b_{i_1} ... \xi^b_{i_p} \sigma^{a}_{i_1} ... \sigma^{a}_{i_p} \right) \,,                                  \\
    \left\langle \mathcal{Z}^L \right\rangle & = \sum_{\sigma} P \left( \sigma \right) \sum_{\xi} \exp \left( \beta \frac{p!}{N^{p-1}} \sum_{a b} \sum_{i_1 < ... < i_p} \xi^b_{i_1} ... \xi^b_{i_p} \sigma^{a}_{i_1} ... \sigma^{a}_{i_p} \right)\,,
\end{align*}
where $b \in \begin{Bmatrix} 1 \ \dots \ L \end{Bmatrix}$ label replicas in the set of patterns $\xi = \begin{Bmatrix} \xi^1 \ \dots \ \xi^L \end{Bmatrix}$ inferred by the student. Using the definition of conditional probability, we rewrite $P \left( \sigma \right)$ as
\begin{align*}
    P \left( \sigma \right) & = \sum_{\xi^*} P \left( \sigma  \mid  \xi^* \right) P \left( \xi^* \right)     \\
                            & = \frac{1}{2^N} \sum_{\xi^*} P \left( \sigma  \mid  \xi^* \right)              \\
                            & = \frac{1}{2^N} \sum_{\xi^*} \prod_a P \left( \sigma^a  \mid  \xi^* \right)\,,
\end{align*}
where $P \left( \sigma \mid \xi^* \right)$ has the same functional form as $P \left( \sigma \mid \xi^b \right)$, but has hyperparameters $p^*$ and $\beta^*$ in place of $p$ and $\beta$. As we did for $Z$, we factor the partition function $Z^*$ of $P \left( \sigma^a \mid \xi^* \right)$ out of the sum, which yields
\begin{align*}
    P \left( \sigma \right) & = \frac{1}{2^N} \frac{1}{\left[ Z^* \right]^M} \sum_{\xi^*} \prod_a \exp \left( \beta^* \frac{p^*!}{N^{p^*-1}} \sum_{i_1 < ... < i_{p^*}} \xi^*_{i_1} ... \xi^*_{i_{p^*}} \sigma^{a}_{i_1} ... \sigma^{a}_{i_{p^*}} \right) \\
                            & = \frac{1}{2^N} \frac{\mathcal{Z}^*}{\left[ Z^* \right]^M} = \frac{1}{2^{M N}} \frac{\mathcal{Z}^*}{\left[ 2^{N/M - N} Z^* \right]^M}\,,
\end{align*}
where $\mathcal{Z}^* = \sum_{\xi^*} \exp \left( -\beta^* H \left[ \xi^* \mid \sigma \right] \right)$ is the partition function of the inverse model with interaction order $p^*$. Using $\sum_{\sigma} P \left( \sigma \right) = 1$, we immediately deduce that $\left[ 2^{N/M - N} Z^* \right]^M = \left\langle \mathcal{Z}^* \right\rangle$. Plugging $P \left( \sigma \right) = \frac{1}{2^{M N}} \frac{\mathcal{Z}^*}{\left\langle \mathcal{Z}^* \right\rangle}$ back in $\left\langle \mathcal{Z}^L \right\rangle$ then gives
\begin{align*}
    \left\langle \mathcal{Z}^L \right\rangle & = \frac{1}{2^{M N}} \frac{1}{\left\langle \mathcal{Z}^* \right\rangle} \sum_{\xi^*} \sum_{\sigma} \exp \left( \beta^* \frac{p^*!}{N^{p^*-1}} \sum_a \sum_{i_1 < ... < i_{p^*}} \xi^*_{i_1} ... \xi^*_{i_{p^*}} \sigma^{a}_{i_1} ... \sigma^{a}_{i_{p^*}} \right) \\
                                             & \quad \sum_{\xi} \exp \left( \beta \frac{p!}{N^{p-1}} \sum_{a b} \sum_{i_1 < ... < i_p} \xi^b_{i_1} ... \xi^b_{i_p} \sigma^{a}_{i_1} ... \sigma^{a}_{i_p} \right)\,.
\end{align*}
We simplify this expression to:
\begin{align*}
    \left\langle \mathcal{Z}^L \right\rangle & = \frac{1}{2^{M N}} \frac{1}{\left\langle \mathcal{Z}^* \right\rangle} \sum_{\xi^*} \sum_{\sigma} \exp \Bigg( \beta^* \frac{p^*!}{N^{p^*-1}} \sum_{a \in \Gamma_*} \sum_{i_1 < ... < i_{p^*}} \xi^*_{i_1} ... \xi^*_{i_{p^*}} \sigma^{a}_{i_1} ... \sigma^{a}_{i_{p^*}} \\
                                             & \quad+ \beta^* \frac{p^*!}{N^{p^*-1}} \sum_{a \in \Bar{\Gamma}_*} \sum_{i_1 < ... < i_{p^*}} \xi^*_{i_1} ... \xi^*_{i_{p^*}} \sigma^{a}_{i_1} ... \sigma^{a}_{i_{p^*}} \Bigg)                                                                                           \\
                                             & \quad \sum_{\xi} \exp \Bigg( \beta \frac{p!}{N^{p-1}} \sum_{b} \sum_{a \in \Gamma_b} \sum_{i_1 < ... < i_p} \xi^b_{i_1} ... \xi^b_{i_p} \sigma^{a}_{i_1} ... \sigma^{a}_{i_p}                                                                                           \\
                                             & \quad+ \beta \frac{p!}{N^{p-1}} \sum_{b} \sum_{a \in \Bar{\Gamma}_b} \sum_{i_1 < ... < i_p} \xi^b_{i_1} ... \xi^b_{i_p} \sigma^{a}_{i_1} ... \sigma^{a}_{i_p} \Bigg)                                                                                                    \\
                                             & \approx \frac{1}{2^{M N}} \frac{1}{\left\langle \mathcal{Z}^* \right\rangle} \sum_{\xi^*} \sum_{\sigma} \exp \Bigg( \beta^* N \sum_{a \in \Gamma_*} \left[ \frac{1}{N} \sum_i \xi^*_i \sigma^a_i \right]^{p^*}                                                          \\
                                             & \quad+ \beta^* \frac{p^*!}{N^{p^*-1}} \sum_{a \in \Bar{\Gamma}} \sum_{i_1 < ... < i_{p^*}} \xi^*_{i_1} ... \xi^*_{i_{p^*}} \sigma^{a}_{i_1} ... \sigma^{a}_{i_{p^*}} \Bigg)                                                                                             \\
                                             & \quad \sum_{\xi} \exp \Bigg( \beta N \sum_{b} \sum_{a \in \Gamma_b} \left[ \frac{1}{N} \sum_i \xi^b_i \sigma^a_i \right]^p                                                                                                                                              \\
                                             & \quad+ \beta \frac{p!}{N^{p-1}} \sum_{b} \sum_{a \in \Bar{\Gamma}} \sum_{i_1 < ... < i_p} \xi^b_{i_1} ... \xi^b_{i_p} \sigma^{a}_{i_1} ... \sigma^{a}_{i_p} \Bigg)\,,
\end{align*}
where $\Gamma_b$ represents the set of inputs $\sigma^a$ which have macroscopic overlap with the pattern $\xi^b$, and $\Bar{\Gamma} = \left[ \cap_b \Bar{\Gamma}_b \right] \cap \Bar{\Gamma}_*$ contains almost all of the elements in each $\Bar{\Gamma}_b$ and $\Bar{\Gamma}_*$ for $N \rightarrow \infty$. The reasoning used to build the sets $\Gamma_*$, $\Gamma_b$ and $\Bar{\Gamma}$ is the same as outlined at the start of appendix \ref{app:direct_cumulant}.

\section{Teacher-student free entropy}
\label{app:teacher-student_free_entropy}

Assuming that the teacher is misaligned with $\sigma$ (see Case \ref{case:weakly-correlated} of Appendix \ref{app:hamiltonians}), the form of $\left\langle \mathcal{Z}^L \right\rangle$ obtained in appendix \ref{app:teacher-student_partition} simplifies to
\begin{align*}
    \left\langle \mathcal{Z}^L \right\rangle & \approx \frac{1}{2^{M N}} \frac{1}{\left\langle \mathcal{Z}^* \right\rangle} \sum_{\xi^* \xi} \sum_{\sigma} \exp \Bigg( \beta^* \frac{p^*!}{N^{p^*-1}} \sum_{a \in \Bar{\Gamma}} \sum_{i_1 < ... < i_{p^*}} \xi^*_{i_1} ... \xi^*_{i_{p^*}} \sigma^{a}_{i_1} ... \sigma^{a}_{i_{p^*}} \Bigg) \\
                                             & \quad \exp \Bigg( \beta N \sum_{b} \sum_{a \in \Gamma_b} \left[ \frac{1}{N} \sum_i \xi^b_i \sigma^a_i \right]^p + \beta \frac{p!}{N^{p-1}} \sum_{b} \sum_{a \in \Bar{\Gamma}} \sum_{i_1 < ... < i_p} \xi^b_{i_1} ... \xi^b_{i_p} \sigma^{a}_{i_1} ... \sigma^{a}_{i_p} \Bigg)\,.
\end{align*}
In order to evaluate $\left\langle \mathcal{Z}^* \right\rangle = \left[ 2^{N/M - N} Z^* \right]^M$, we recall that the teacher is a special case of the direct model with a single memory (see Section \ref{sec:dense_HN_teacher-student}). Since the teacher is in the misaligned case, its free entropy is
\begin{align*}
    \frac{\log \left( Z^* \right)}{N}
     & = \begin{cases}
             \left( -\frac{1}{2} \log \left( 1 - 2 \beta^* \right) - \beta^* \right) \frac{1}{N} + \log 2       \,,                         & \quad p^* = 2\,,  \\
             \frac{1}{2} \left[ \beta^* \right]^2 \frac{p^*!}{N^{p^*-1}} + \log 2 + \mathcal{O} \left( \frac{1}{N^{3p^*/2 - 2)}} \right)\,, & \quad p^* \geq 3,
         \end{cases}
\end{align*}
as derived in Appendix \ref{app:direct_cumulant}. Given $\alpha^* = \frac{M p^*!}{N^{p^* - 1}}$, we use it to simplify $\frac{\log \left\langle \mathcal{Z}^* \right\rangle}{N}$ to
\begin{align*}
    \frac{\log \left\langle \mathcal{Z}^* \right\rangle}{N} & = \frac{M \log \left[ 2^{N/M - N} Z^* \right]}{N}                                                                                                               \\
                                                            & = \begin{cases}
                                                                    \frac{1}{2} \left( -\frac{1}{2} \log \left( 1 - 2 \beta^* \right) - \beta^* \right) \alpha^* + \log 2   \,,     & \quad p^* = 2\,,    \\
                                                                    \frac{1}{2} \left[ \beta^* \right]^2 \alpha^* + \log 2 + \mathcal{O} \left( \frac{1}{N^{p^*/2 - 1}} \right) \,, & \quad p^* \geq 3\,,
                                                                \end{cases}
\end{align*}
which is the paramagnetic free entropy of a $p^*$-body Hopfield network \cite{amit1985storing, gardner1987multiconnected}.
Coming back to $\left\langle \mathcal{Z}^L \right\rangle$, we fix order parameters $q^{* b}$, $q^{b c}$ and $m^b_a$ using the delta functions $\delta \left( N q^{* b} - \sum_i \xi^*_i \xi^b_i \right)$, $\delta \left( N q^{b c} - \sum_i \xi^b_i \xi^c_i \right)$ and $\delta \left( N m^b_a - \sum_i \xi^b_i \sigma^a_i \right)$, which results in
\begin{align*}
    \left\langle \mathcal{Z}^L \right\rangle & = \frac{1}{2^{M N}} \frac{1}{\left\langle \mathcal{Z}^* \right\rangle} \sum_{\xi^* \xi} \sum_{\sigma} \int_{\mathbb{R}} \prod_b dq^{* b} \prod_{b < c} dq^{b c} \prod_b \prod_{a \in \Gamma_b} d m^b_a \\
                                             & \quad \delta \left( N q^{* b} - \sum_i \xi^*_i \xi^b_i \right) \delta \left( N q^{b c} - \sum_i \xi^b_i \xi^c_i \right) \delta \left( N m^b_a - \sum_i \xi^b_i \sigma^a_i \right)                      \\
                                             & \quad \exp \Bigg( \beta N \sum_{b} \sum_{a \in \Gamma_b} \left[ \frac{1}{N} \sum_i \xi^b_i \sigma^a_i \right]^p                                                                                        \\
                                             & \quad+ \beta^* \frac{p^*!}{N^{p^*-1}} \sum_{a \in \Bar{\Gamma}} \sum_{i_1 < ... < i_{p^*}} \xi^*_{i_1} ... \xi^*_{i_{p^*}} \sigma^{a}_{i_1} ... \sigma^{a}_{i_{p^*}}                                   \\
                                             & \quad+ \beta \frac{p!}{N^{p-1}} \sum_{b} \sum_{a \in \Bar{\Gamma}} \sum_{i_1 < ... < i_p} \xi^b_{i_1} ... \xi^b_{i_p} \sigma^{a}_{i_1} ... \sigma^{a}_{i_p} \Bigg).
\end{align*}
In Fourier space, this expression takes the form
\begin{align*}
    \left\langle \mathcal{Z}^L \right\rangle & = \frac{1}{\left\langle \mathcal{Z}^* \right\rangle} \sum_{\xi^* \xi} \Bigg\langle \int \prod_b dq^{* b} dr^{* b} \prod_{b < c} dq^{b c} dr^{b c} \prod_b \prod_{a \in \Gamma_b} d m^b_a d k^b_a               \\
                                             & \quad \exp \left\{ \beta^* \beta \alpha \sum_b \left( \sum_i \xi^*_i \xi^b_i - N q^{* b} \right) r^{* b} + \beta^2 \alpha \sum_{b < c} \left( \sum_i \xi^b_i \xi^c_i - N q^{b c} \right) r^{b c} \right\}      \\
                                             & \quad \exp \Bigg\{ \beta \sum_b \sum_{a \in \Gamma_b} \left( \sum_i \xi^b_i \sigma^a_i - N m^b_a \right) k^b_a + \beta N \sum_{b} \sum_{a \in \Gamma_b} \left[ \frac{1}{N} \sum_i \xi^b_i \sigma^a_i \right]^p \\
                                             & \quad+ \beta^* \frac{p^*!}{N^{p^*-1}} \sum_{a \in \Bar{\Gamma}} \sum_{i_1 < ... < i_{p^*}} \xi^*_{i_1} ... \xi^*_{i_{p^*}} \sigma^{a}_{i_1} ... \sigma^{a}_{i_{p^*}}                                           \\
                                             & \quad+ \beta \frac{p!}{N^{p-1}} \sum_{b} \sum_{a \in \Bar{\Gamma}} \sum_{i_1 < ... < i_p} \xi^b_{i_1} ... \xi^b_{i_p} \sigma^{a}_{i_1} ... \sigma^{a}_{i_p} \Bigg\} \Bigg\rangle_{\sigma}\,,
\end{align*}
where the sum over $\sigma$ with a pre-factor of $\frac{1}{2^{M N}}$ was replaced by the uniform average $\langle \rangle_{\sigma}$. Following the same reasoning as in appendix \ref{app:direct_cumulant}, a second order cumulant expansion of the last two terms for any $a \in \Bar{\Gamma}$ yields
\begin{align*}
     & \log \Bigg\langle \exp \Bigg\{ \beta^* \frac{p^*!}{N^{p^*-1}} \sum_{i_1 < ... < i_{p^*}} \xi^*_{i_1} ... \xi^*_{i_{p^*}} \sigma^{a}_{i_1} ... \sigma^{a}_{i_{p^*}} + \beta \frac{p!}{N^{p-1}} \sum_{b} \sum_{i_1 < ... < i_p} \xi^b_{i_1} ... \xi^b_{i_p} \sigma^{a}_{i_1} ... \sigma^{a}_{i_p} \Bigg\} \Bigg\rangle \\
     & \quad\approx \frac{1}{2} \beta^2 \left[ \frac{p!}{N^{p-1}} \right]^2 \sum_{b \neq c} \sum_{i_1 < ... < i_p} \sum_{j_1 < ... < j_p} \xi^b_{i_1} \xi^c_{j_1} ... \xi^b_{i_p} \xi^c_{j_p} \left\langle \sigma^a_{i_1} \sigma^a_{j_1} ... \sigma^a_{i_p} \sigma^a_{j_p} \right\rangle                                    \\
     & \qquad+ \beta^* \beta \frac{p^*!}{N^{p^*-1}} \frac{p!}{N^{p-1}} \sum_b \sum_{i_1 < ... < i_{p^*}} \sum_{j_1 < ... < j_p} \left\langle \xi^*_{i_1} \sigma^a_{i_1} ... \xi^*_{i_{p^*}} \sigma^a_{i_{p^*}} \xi^b_{j_1} \sigma^a_{j_1} ... \xi^b_{j_p} \sigma^a_{j_p} \right\rangle                                      \\
     & \qquad+ \frac{1}{2} \beta^2 \frac{p!}{N^{p-1}} L N + \frac{1}{2} \left[ \beta^* \right]^2 \frac{p^*!}{N^{p^*-1}} N\,.
\end{align*}
When $p^* = p$, it reduces to
\begin{align*}
     & \log \Bigg\langle \exp \Bigg\{ \beta^* \frac{p^*!}{N^{p^*-1}} \sum_{i_1 < ... < i_{p^*}} \xi^*_{i_1} ... \xi^*_{i_{p^*}} \sigma^{a}_{i_1} ... \sigma^{a}_{i_{p^*}} + \beta \frac{p!}{N^{p-1}} \sum_{b} \sum_{i_1 < ... < i_p} \xi^b_{i_1} ... \xi^b_{i_p} \sigma^{a}_{i_1} ... \sigma^{a}_{i_p} \Bigg\} \Bigg\rangle \\
     & \quad= \beta^2 \frac{p!}{N^{p-1}} N \sum_{b < c} \left[ \frac{1}{N} \sum_i \xi^b_i \xi^c_i \right]^p + \beta^* \beta \frac{p!}{N^{p-1}} N \sum_{b} \left[ \frac{1}{N} \sum_i \xi^*_i \xi^b_i \right]^p                                                                                                               \\
     & \qquad+ \frac{1}{2} \beta^2 \frac{p!}{N^{p-1}} L N + \frac{1}{2} \left[ \beta^* \right]^2 \frac{p!}{N^{p-1}} N\,,
\end{align*}
because $\left\langle \sigma^a_{i_n} \sigma^a_{j_n} \right\rangle = \delta_{i_n j_n}$ (see Appendix \ref{app:direct_cumulant} for more details). On the contrary, the second order expectation $\left\langle \xi^*_{i_1} \sigma^a_{i_1} ... \xi^*_{i_{p^*}} \sigma^a_{i_{p^*}} \xi^b_{j_1} \sigma^a_{j_1} ... \xi^b_{j_p} \sigma^a_{j_p} \right\rangle$ vanishes when $p^* \neq p$. In fact, spins come in pairs $\left\langle \sigma^a_{i_n} \sigma^a_{j_n} \right\rangle = \delta_{i_n j_n}$ only up to $n \leq \min \left\{ p^*, p \right\}$, and the remaining single-spin averages $\left\langle \sigma^a_{i_n} \right\rangle = 0$ make the second order expectation vanish.

We need to go beyond second order to treat $p^* \neq p$. We will focus on $p^* = 2$ and $p \geq 3$ to investigate the consequences of using a $p$-body model to learn examples generated by the original $2$-body Hopfield model. For simplicity, we take $p$ even so that the spins of both terms can be grouped in pairs at order $\frac{p}{2} + 1$, when the teacher term $\beta^* \frac{2}{N} \sum_{i_1 < i_2} \xi^*_{i_1} \xi^*_{i_2} \sigma^{a}_{i_1} \sigma^{a}_{i_2}$ is raised to the power of $\frac{p}{2}$ and the student term is raised to the power of $1$. This restriction will simplify some of the incoming calculations. To leading order in $N$, the cumulant generating function reduces to
\begin{align*}
     & \log \Bigg\langle \exp \Bigg\{ \beta^* \frac{2}{N} \sum_{i_1 < i_2} \xi^*_{i_1} \xi^*_{i_2} \sigma^{a}_{i_1} \sigma^{a}_{i_2} + \beta \frac{p!}{N^{p-1}} \sum_{b} \sum_{i_1 < ... < i_p} \xi^b_{i_1} ... \xi^b_{i_p} \sigma^{a}_{i_1} ... \sigma^{a}_{i_p} \Bigg\} \Bigg\rangle              \\
     & \quad \approx \log \Bigg[ \left\langle \exp \left\{ \beta^* \frac{2}{N} \sum_{i_1 < i_2} \xi^*_{i_1} \xi^*_{i_2} \sigma^{a}_{i_1} \sigma^{a}_{i_2} \right\} \right\rangle                                                                                                                    \\
     & \qquad\left\langle \exp \left\{ \beta \frac{p!}{N^{p-1}} \sum_{b} \sum_{i_1 < ... < i_p} \xi^b_{i_1} ... \xi^b_{i_p} \sigma^{a}_{i_1} ... \sigma^{a}_{i_p} \right\} \right\rangle                                                                                                            \\
     & \qquad+ \left\langle \beta \frac{p!}{N^{p-1}} \sum_{b} \sum_{j_1 < ... < j_p} \xi^b_{j_1} ... \xi^b_{j_p} \sigma^{a}_{j_1} ... \sigma^{a}_{j_p} \exp \left\{ \beta^* \frac{2}{N} \sum_{i_1 < i_2} \xi^*_{i_1} \xi^*_{i_2} \sigma^{a}_{i_1} \sigma^{a}_{i_2} \right\} \right\rangle \Bigg]\,,
\end{align*}
where the last term encompasses the teacher-student coupling that allows retrieval to take place. The teacher term
\begin{align*}
    \log \left\langle \exp \left\{ \beta^* \frac{2}{N} \sum_{i_1 < i_2} \xi^*_{i_1} \xi^*_{i_2} \sigma^{a}_{i_1} \sigma^{a}_{i_2} \right\} \right\rangle \approx -\frac{1}{2} \log \left( 1 - 2 \beta^* \right) - \beta^* \,,
\end{align*}
and the student term
\begin{align*}
     & \log \left\langle \exp \left\{ \beta \frac{p!}{N^{p-1}} \sum_{b} \sum_{i_1 < ... < i_p} \xi^b_{i_1} ... \xi^b_{i_p} \sigma^{a}_{i_1} ... \sigma^{a}_{i_p} \right\} \right\rangle \\
     & \quad \approx \beta^2 \frac{p!}{N^{p-1}} N \sum_{b < c} \left[ \frac{1}{N} \sum_i \xi^b_i \xi^c_i \right]^p + \frac{1}{2} \beta^2 \frac{p!}{N^{p-1}} L N \,,
\end{align*}
are both known from Appendix \ref{app:direct_cumulant}. Later on, we will use $\log \left( z^* \right)$ and $z^*$ as shorthands for $-\frac{1}{2} \log \left( 1 - 2 \beta^* \right) - \beta^*$ and $\exp \left( -\frac{1}{2} \log \left( 1 - 2 \beta^* \right) - \beta^* \right)$, respectively. The coupling between the teacher and the student can be rewritten as
\begin{align*}
     & \left\langle \beta \frac{p!}{N^{p-1}} \sum_{b} \sum_{j_1 < ... < j_p} \xi^b_{j_1} ... \xi^b_{j_p} \sigma^{a}_{j_1} ... \sigma^{a}_{j_p} \exp \left\{ \beta^* \frac{2}{N} \sum_{i_1 < i_2} \xi^*_{i_1} \xi^*_{i_2} \sigma^{a}_{i_1} \sigma^{a}_{i_2} \right\} \right\rangle                                                                     \\
     & \quad = \left\langle \beta \frac{p!}{N^{p-1}} \sum_{b} \sum_{j_1 < ... < j_p} \xi^*_{j_1} ... \xi^*_{j_p} \xi^b_{j_1} ... \xi^b_{j_p} \ \xi^*_{j_1} ... \xi^*_{j_p} \sigma^{a}_{j_1} ... \sigma^{a}_{j_p} \exp \left\{ \beta^* \frac{2}{N} \sum_{i_1 < i_2} \xi^*_{i_1} \xi^*_{i_2} \sigma^{a}_{i_1} \sigma^{a}_{i_2} \right\} \right\rangle   \\
     & \quad = \beta \frac{p!}{N^{p-1}} \sum_{b} \sum_{j_1 < ... < j_p} \xi^*_{j_1} ... \xi^*_{j_p} \xi^b_{j_1} ... \xi^b_{j_p} \left\langle \xi^*_{j_1} ... \xi^*_{j_p} \sigma^{a}_{j_1} ... \sigma^{a}_{j_p} \exp \left\{ \beta^* \frac{2}{N} \sum_{i_1 < i_2} \xi^*_{i_1} \xi^*_{i_2} \sigma^{a}_{i_1} \sigma^{a}_{i_2} \right\} \right\rangle \,,
\end{align*}
because $\left[ \xi^*_{j_n} \right]^2 = 1$ for every index $j_n$. All interacting spin tuples of the form $\xi^*_{j_1} ... \xi^*_{j_p} \sigma^a_{j_1} ... \sigma^a_{i_p}$ are statistically equivalent as long as $j_1 < ... < j_p$, so the teacher-student coupling simplifies to
\begin{align*}
     & \left\langle \beta \frac{p!}{N^{p-1}} \sum_{b} \sum_{j_1 < ... < j_p} \xi^b_{j_1} ... \xi^b_{j_p} \sigma^{a}_{j_1} ... \sigma^{a}_{j_p} \exp \left\{ \beta^* \frac{2}{N} \sum_{i_1 < i_2} \xi^*_{i_1} \xi^*_{i_2} \sigma^{a}_{i_1} \sigma^{a}_{i_2} \right\} \right\rangle \\
     & \quad= \beta \frac{p!}{N^{p-1}} \sum_{b} \sum_{i_1 < ... < i_p} \xi^*_{i_1} ... \xi^*_{i_p} \xi^b_{i_1} ... \xi^b_{i_p}                                                                                                                                                    \\
     & \qquad \left\langle \frac{p!}{N^p} \sum_{j_1 < ... < j_p} \xi^*_{j_1} ... \xi^*_{j_p} \sigma^{a}_{j_1} ... \sigma^{a}_{i_p} \exp \left\{ \beta^* \frac{2}{N} \sum_{i_1 < i_2} \xi^*_{i_1} \xi^*_{i_2} \sigma^{a}_{i_1} \sigma^{a}_{i_2} \right\} \right\rangle             \\
     & \quad = V \left( \beta^*, p \right) \beta \frac{p!}{N^{p-1}} \sum_{b} \sum_{i_1 < ... < i_p} \xi^*_{i_1} ... \xi^*_{i_p} \xi^b_{i_1} ... \xi^b_{i_p} \,,
\end{align*}
where $V \left( \beta^*, p \right) = \left\langle \frac{p!}{N^p} \sum_{j_1 < ... < j_p} \xi^*_{j_1} ... \xi^*_{j_p} \sigma^{a}_{j_1} ... \sigma^{a}_{i_p} \exp \left( \beta^* \frac{2}{N} \sum_{i_1 < i_2} \xi^*_{i_1} \xi^*_{i_2} \sigma^{a}_{i_1} \sigma^{a}_{i_2} \right) \right\rangle$ does not depend on the microscopic details of the system. In fact, it can be expressed as a combination of the moments of $z^*$, which can all be derived from $\log \left( z^* \right)$. To leading order in $N$, the cumulant generating function expands to
\begin{align*}
     & \log \Bigg\langle \exp \Bigg\{ \beta^* \frac{2}{N} \sum_{i_1 < i_2} \xi^*_{i_1} \xi^*_{i_2} \sigma^{a}_{i_1} \sigma^{a}_{i_2} + \beta \frac{p!}{N^{p-1}} \sum_{b} \sum_{i_1 < ... < i_p} \xi^b_{i_1} ... \xi^b_{i_p} \sigma^{a}_{i_1} ... \sigma^{a}_{i_p} \Bigg\} \Bigg\rangle \\
     & \quad\approx -\frac{1}{2} \log \left( 1 - 2 \beta^* \right) - \beta^* + \beta^2 \frac{p!}{N^{p-1}} N \sum_{b < c} \left[ \frac{1}{N} \sum_i \xi^b_i \xi^c_i \right]^p + \frac{1}{2} \beta^2 \frac{p!}{N^{p-1}} L N                                                              \\
     & \qquad+ \left[ 1 - 2 \beta^* \right]^{1/2} \exp \left( \beta^* \right) V \left( \beta^*, p \right) \beta N \sum_b \left[ \frac{1}{N} \sum_i \xi^*_i \xi^b_i \right]^p\,.
\end{align*}
At this stage, we only need to find $V \left( \beta^*, p \right)$ in order to solve the system. We focus on two different scalings of $M$ and $\beta^*$ that make the teacher-student coupling leading order in $N$:
\begin{enumerate}[label = \textbf{\arabic*}]
    \item $M \sim \mathcal{O} \left( N^{p - 1} \right)$ and $\beta^* \sim \mathcal{O} \left( N^{2/p - 1} \right)$ will be called the large-noise scaling.
    \item $M \sim \mathcal{O} \left( N^{p/2} \right)$ and $\beta^* \sim \mathcal{O} \left( 1 \right)$ will be called the finite-noise scaling.
\end{enumerate}
The student term vanishes in the first scenario but is leading order in the second one.
The case of the teacher-student coupling is more subtle. When $\beta^*$ is small, we may keep only the first non-vanishing order of the exponential function present in the definition of $V \left( \beta^*, p \right)$.
Since $p$ is even, it leads to
\begin{align*}
    \label{eqn:large_noise_cumulant_expansion}
    V \left( \beta^*, p \right) & \approx \frac{1}{(p/2)!} \left\langle \frac{p!}{N^p} \sum_{j_1 < ... < j_p} \xi^*_{j_1} ... \xi^*_{j_p} \sigma^{a}_{j_1} ... \sigma^{a}_{j_p} \left( \beta^* \frac{2}{N} \sum_{i_1 < i_2} \xi^*_{i_1} \xi^*_{i_2} \sigma^{a}_{i_1} \sigma^{a}_{i_2} \right)^{p/2} \right\rangle \numberthis \\
                                & = \frac{\left[ \beta^* \right]^{p/2}}{(p/2)!} \frac{2^{p/2}}{N^{p/2}} \frac{p!}{2^{p/2}}                                                                                                                                                                                                    \\
                                & = \frac{\left[ \beta^* \right]^{p/2}}{(p/2)!} \frac{p!}{N^{p/2}} \,,
\end{align*}
because there are $\prod^{p/2}_{n = 1} \binom{2 n}{2} = \frac{p!}{2^{p/2}}$ spin pairings with nonzero expectation that satisfy the inequality constraints. In the large-noise scaling, we set
\begin{align*}
    \lambda & = \frac{\left[ \beta^* \right]^{p/2}}{(p/2)!} N^{p/2 - 1} \sim \mathcal{O} \left( 1 \right) \,,
\end{align*}
to get the asymptotically exact expression $V \left( \left[ (p/2)! \right]^{2/p} N^{1 - 2/p}, p \right) = \lambda \frac{p!}{N^{p - 1}}$. In the finite-noise scaling, this expansion is only an order of magnitude approximation. However, it still indicates that $V \left( \beta^*, p \right)$ is $\mathcal{O} \left( N^{-p/2} \right)$ when $\beta^*$ is $\mathcal{O} \left( 1 \right)$ in $N$. In other words, it shows that there is an $\mathcal{O} \left( 1 \right)$ parameter $\eta$ such that $V \left( \beta^* \left( \eta, p \right), p \right) = \eta \frac{\left( p/2 + 1 \right)!}{N^{p/2}}$. We will now use the cumulants $\frac{\partial \log \left( z^* \right)}{\partial \beta^*}$ and $\frac{\partial \log \left( z^* \right)}{\partial \beta^{* 2}}$ of $z^*$ to derive the value of $\eta$ corresponding to $p = 4$. First of all, note that $\frac{4!}{N^4} \sum_{j_1 < ... < j_4} \xi^*_{j_1} ... \xi^*_{j_4} \sigma^{a}_{j_1} ... \sigma^{a}_{j_4}$ can be expressed as:
\begin{align*}
     & \frac{24}{N^4} \sum_{j_1 < j_2 < j_3 < j_4} \xi^*_{j_1} \xi^*_{j_2} \xi^*_{j_3} \xi^*_{j_4} \sigma^a_{j_1} \sigma^a_{j_2} \sigma^a_{j_3} \sigma^a_{j_4}                                                                                                                                                                         \\
     & \quad= \frac{1}{N^4} \sum_{j_1 \neq j_2 \neq j_3 \neq j_4} \xi^*_{j_1} \xi^*_{j_2} \xi^*_{j_3} \xi^*_{j_4} \sigma^a_{j_1} \sigma^a_{j_2} \sigma^a_{j_3} \sigma^a_{j_4}                                                                                                                                                          \\
     & \quad= \frac{1}{N^4} \left[ \sum_{j_1 \neq j_2} \xi^*_{j_1} \xi^*_{j_2} \sigma^a_{j_1} \sigma^a_{j_2} \right] \left[ \sum_{j_3 \neq j_4} \xi^*_{j_3} \xi^*_{j_4} \sigma^a_{j_3} \sigma^a_{j_4} \right] - \frac{4}{N^3} \left[ \sum_{j_1 \neq j_2} \xi^*_{j_1} \xi^*_{j_2} \sigma^a_{j_1} \sigma^a_{j_2} \right] - \frac{2}{N^2} \\
     & \quad= \frac{1}{N^2} \left[ \frac{2}{N} \sum_{j_1 < j_2} \xi^*_{j_1} \xi^*_{j_2} \sigma^a_{j_1} \sigma^a_{j_2} \right]^2 - \frac{4}{N^2} \left[ \frac{2}{N} \sum_{j_1 < j_2} \xi^*_{j_1} \xi^*_{j_2} \sigma^a_{j_1} \sigma^a_{j_2} \right] - \frac{2}{N^2} \,,
\end{align*}
by subtracting the diagonals where pairs of indices are equal. Therefore, $\frac{1}{z^*} V \left( \beta^*, p \right)$ reduces to
\begin{align*}
    \frac{1}{z^*} V \left( \beta^*, p \right) & = \left\langle \frac{24}{N^4} \sum_{j_1 < j_2 < j_3 < j_4} \xi^*_{j_1} \xi^*_{j_2} \xi^*_{j_3} \xi^*_{j_4} \sigma^a_{i_1} \sigma^a_{i_2} \sigma^a_{i_3} \sigma^a_{i_4} \exp \left\{ \beta^* \frac{2}{N} \sum_{i_1 < i_2} \xi^*_{i_1} \xi^*_{i_2} \sigma^a_{i_1} \sigma^a_{i_2} \right\} \right\rangle \\
                                              & = \frac{1}{z^*} \frac{1}{N^2} \Bigg[ \left\langle \left[ \frac{2}{N} \sum_{j_1 < j_2} \xi^*_{j_1} \xi^*_{j_2} \sigma^a_{i_1} \sigma^a_{i_2} \right]^2 \exp \left\{ \beta^* \frac{2}{N} \sum_{i_1 < i_2} \xi^*_{i_1} \xi^*_{i_2} \sigma^a_{i_1} \sigma^a_{i_2} \right\} \right\rangle                  \\
                                              & \quad- 4 \left\langle \left[ \frac{2}{N} \sum_{j_1 < j_2} \xi^*_{j_1} \xi^*_{j_2} \sigma^a_{i_1} \sigma^a_{i_2} \right] \exp \left\{ \beta^* \frac{2}{N} \sum_{i_1 < i_2} \xi^*_{i_1} \xi^*_{i_2} \sigma^a_{i_1} \sigma^a_{i_2} \right\} \right\rangle                                                \\
                                              & \quad- 2 \left\langle \exp \left\{ \beta^* \frac{2}{N} \sum_{i_1 < i_2} \xi^*_{i_1} \xi^*_{i_2} \sigma^a_{i_1} \sigma^a_{i_2} \right\} \right\rangle \Bigg]                                                                                                                                           \\
                                              & = \frac{1}{N^2} \left[ \frac{\partial \log \left( z^* \right)}{\partial \beta^{* 2}} + \left[ \frac{\partial \log \left( z^* \right)}{\partial \beta^*} \right]^2 - 4 \frac{\partial \log \left( z^* \right)}{\partial \beta^*} - 2 \right]\,.
\end{align*}
The cumulants evaluate to
\begin{align*}
    \frac{\partial \log \left( z^* \right)}{\partial \beta^*}     & = \frac{\partial}{\partial \beta^*} \left[ -\frac{1}{2} \log \left( 1 - 2 \beta^* \right) - \beta^* \right] = \frac{2 \beta^*}{1 - 2 \beta^*}   \,,          \\
    \frac{\partial \log \left( z^* \right)}{\partial \beta^{* 2}} & = \frac{\partial}{\partial \beta^{*2}} \left[ -\frac{1}{2} \log \left( 1 - 2 \beta^* \right) - \beta^* \right] = \frac{2}{\left( 1 - 2 \beta^* \right)^2}\,,
\end{align*}
so we obtain
\begin{align*}
    \frac{1}{z^*} V \left( \beta^*, p \right) & = \frac{1}{N^2} \left[ \frac{2}{\left( 1 - 2 \beta^* \right)^2} + \frac{4 \left[ \beta^* \right]^2}{\left( 1 - 2 \beta^* \right)^2} - \frac{8 \beta^*}{1 - 2 \beta^*} - 2 \right] \\
                                              & = \frac{6}{N^2} \frac{2 \left[ \beta^* \right]^2}{\left( 1 - 2 \beta^* \right)^2}\,.
\end{align*}
In other terms, we find $\eta = \frac{2 \left[ \beta^* \right]^2}{\left( 1 - 2 \beta^* \right)^2}$ when $p = 4$. In summary, depending on the scaling, the teacher student coupling either simplifies to
\begin{enumerate}[label = \textbf{\arabic*}]
    \item $\beta \lambda \alpha \frac{N}{M} \sum_b \left[ \frac{1}{N} \sum_i \xi^*_i \xi^b_i \right]^p$ where $\alpha = \frac{M p!}{N^{p-1}}$ and $\lambda = \frac{\left[ \beta^* \right]^{p/2}}{(p/2)!} N^{p/2 - 1}$ are finite,
    \item or $\beta \eta \alpha \frac{N}{M} \sum_b \left[ \frac{1}{N} \sum_i \xi^*_i \xi^b_i \right]^p$ where $\alpha = \frac{M \left( p/2 + 1 \right)!}{N^{p/2}}$ and $\eta$ are finite.
\end{enumerate}
In either case, the result is similar to $p^* = p$ except for its pre-factor. We describe the rest of the derivation only for $p^* = p$ because the $p^* = 2$ and $p \geq 3$ calculations are almost identical. Putting the result of the $p^* = p$ cumulant expansion back in $\left\langle \mathcal{Z}^L \right\rangle$, we get:
\begin{align*}
    \left\langle \mathcal{Z}^L \right\rangle & \approx \frac{1}{\left\langle \mathcal{Z}^* \right\rangle} \sum_{\xi^* \xi} \Bigg\langle \int \prod_b dq^{* b} dr^{* b} \prod_{b < c} dq^{b c} dr^{b c} \prod_b \prod_{a \in \Gamma_b} d m^b_a \ d k^b_a                                       \\
                                             & \quad \exp \left\{ \beta^* \beta \alpha \sum_b \left( \sum_i \xi^*_i \xi^b_i - N q^{* b} \right) r^{* b} + \beta^2 \alpha \sum_{b < c} \left( \sum_i \xi^b_i \xi^c_i - N q^{b c} \right) r^{b c} \right\}                                      \\
                                             & \quad \exp \left\{ \beta \sum_{b} \sum_{a \in \Gamma_b} \left( \sum_i \xi^b_i \sigma^a_i - N m^b_a \right) k^b_a + \beta N \sum_b \sum_{a \in \Gamma_b} \left[ m^b_a \right]^p \right\}                                                        \\
                                             & \quad \exp \left\{ \beta^* \beta \alpha N \sum_{b} \left[ q^{* b} \right]^p + \beta^2 \alpha N \sum_{b < c} \left[ q^{b c} \right]^p + \frac{1}{2} \beta^2 \alpha L N + \frac{1}{2} \left[ \beta^* \right]^2 \alpha N \right\} \Bigg\rangle\,,
\end{align*}
where $\alpha = \frac{M p!}{N^{p-1}}$. The saddle point of $\left\langle \mathcal{Z}^L \right\rangle$ then evaluates to
\begin{align*}
    \frac{\log \left\langle \mathcal{Z}^L \right\rangle}{N} & \approx \Extr_{m,k,q,r,q^*,r^*} \Bigg[ \beta^* \beta \alpha \sum_b \left[ q^{* b} \right]^p + \beta^2 \alpha \sum_{b < c} \left[ q^{b c} \right]^p + \beta \sum_b \sum_{a \in \Gamma_b} \left[ m^b_a \right]^p \\
                                                            & \quad- \beta^* \beta \alpha \sum_b r^{* b} q^{* b} - \beta^2 \alpha \sum_{b < c} r^{b c} q^{b c} - \beta \sum_b \sum_{a \in \Gamma_b} m^b_a k^b_a                                                              \\
                                                            & \quad+ \frac{1}{2} \beta^2 \alpha L + \frac{1}{2} \left[ \beta^* \right]^2 \alpha - \frac{\log \left\langle \mathcal{Z} \right\rangle}{N} + \log 2                                                             \\
                                                            & \quad+ \frac{1}{N} \log \Bigg\langle \sum_{\xi} \exp \Bigg\{ \beta \sum_b \sum_{a \in \Gamma_b} k^b_a \sum_i \xi^b_i \sigma^a_i                                                                                \\
                                                            & \quad+ \beta^* \beta \alpha \sum_b r^{* b} \sum_i \xi^*_i \xi^b_i + \beta^2 \alpha \sum_{b < c} r^{b c} \sum_i \xi^b_i \xi^c_i \Bigg\} \Bigg\rangle_{\xi^* \sigma} \Bigg]                                      \\
                                                            & = \Extr \Bigg[ \beta^* \beta \alpha \sum_b \left[ q^{* b} \right]^p + \beta^2 \alpha \sum_{b < c} \left[ q^{b c} \right]^p + \beta \sum_b \sum_{a \in \Gamma_b} \left[ m^b_a \right]^p                         \\
                                                            & \quad- \beta^* \beta \alpha \sum_b r^{* b} q^{* b} - \beta^2 \alpha \sum_{b < c} r^{b c} q^{b c} - \beta \sum_b \sum_{a \in \Gamma_b} m^b_a k^b_a                                                              \\
                                                            & \quad+ \frac{1}{2} \beta^2 \alpha L + \frac{1}{2} \left[ \beta^* \right]^2 \alpha - \frac{\log \left\langle \mathcal{Z} \right\rangle}{N} + \log 2                                                             \\
                                                            & \quad+ \frac{1}{N} \sum_i \log \Bigg\langle \sum_{\xi_i} \exp \Bigg\{ \beta \sum_b \sum_{a \in \Gamma_b} k^b_a \xi^b_i \sigma^a_i                                                                              \\
                                                            & \quad+ \beta^* \beta \alpha \sum_b r^{* b} \xi^*_i \xi^b_i + \beta^2 \alpha \sum_{b < c} r^{b c} \xi^b_i \xi^c_i \Bigg\} \Bigg\rangle_{\xi^*_i \sigma_i} \Bigg]\,,
\end{align*}
where the average over $\xi^*$ and $\sigma$ is uniform.
We use $\frac{\log \left\langle \mathcal{Z}^* \right\rangle}{N} = \frac{1}{2} \left[ \beta^* \right]^2 \alpha + \log 2$ to simplify $\frac{\log \left\langle \mathcal{Z}^L \right\rangle}{N}$ to
\begin{align*}
    \frac{\log \left\langle \mathcal{Z}^L \right\rangle}{N} & \approx \Extr \Bigg[ \beta^* \beta \alpha \sum_b \left[ q^{* b} \right]^p + \beta^2 \alpha \sum_{b < c} \left[ q^{b c} \right]^p + \beta \sum_b \sum_{a \in \Gamma_b} \left[ m^b_a \right]^p \\
                                                            & \quad- \beta^* \beta \alpha \sum_b r^{* b} q^{* b} - \beta^2 \alpha \sum_{b < c} r^{b c} q^{b c} - \beta \sum_b \sum_{a \in \Gamma_b} m^b_a k^b_a                                            \\
                                                            & \quad+ \frac{1}{2} \beta^2 \alpha L + \frac{1}{N} \sum_i \log \Bigg\langle \sum_{\xi_i} \exp \Bigg\{ \beta \sum_b \sum_{a \in \Gamma_b} k^b_a \xi^b_i \sigma^a_i                             \\
                                                            & \quad+ \beta^* \beta \alpha \sum_b r^{* b} \xi^*_i \xi^b_i + \beta^2 \alpha \sum_{b < c} r^{b c} \xi^b_i \xi^c_i \Bigg\} \Bigg\rangle_{\xi^*_i \sigma_i} \Bigg]\,.
\end{align*}
Assuming each $\xi^b$ has macroscopic overlap with at most one pattern $\sigma^a$ and using the replica-symmetric ansatz $q^{* b} = q^*$, $q^{b c} = q$, $r^{* b} = r^*$, $r^{b c} = r$, $m^b_a = m$, $k^b_a = k$, the free entropy approximates to
\begin{align*}
    f & = \lim_{N \rightarrow \infty, L \rightarrow 0} \left( \frac{\partial}{\partial L} \left[ \frac{1}{N} \log \left\langle Z^L \right\rangle \right] \right)                                       \\
      & \approx \Extr \Bigg[ \beta^* \beta \alpha \left[ q^{*} \right]^p - \frac{1}{2} \beta^2 \alpha q^p + \beta m^p - \beta^* \beta \alpha r^{*} q^{*} + \frac{1}{2} \beta^2 \alpha r q              \\
      & \quad- \beta m k + \frac{1}{2} \beta^2 \alpha + \lim_{L \rightarrow 0} \Bigg( \frac{\partial}{\partial L} \Bigg[ \log \Bigg\langle \sum_{\xi_i} \exp \Bigg\{ \beta k \sum_b \xi^b_i \sigma^a_i \\
      & \quad+ \beta^* \beta \alpha r^* \sum_b \xi^*_i \xi^b_i + \beta^2 \alpha r \sum_{b < c} \xi^b_i \xi^c_i \Bigg\} \Bigg\rangle \Bigg] \Bigg) \Bigg]\,.
\end{align*}
Furthermore, the Hubbard-Stratonovich transformation gives
\begin{align*}
    \exp \left\{ \beta^2 \alpha r \sum_{b < c} \xi^b_i \xi^c_i \right\} \propto \exp \left\{ -\frac{1}{2} \beta^2 \alpha r L \right\} \int_{\mathbb{R}} dx \exp \left\{ -\frac{1}{2} x^2 + x \beta \sqrt{\alpha r} \sum_b \xi^b_i \right\}\,.
\end{align*}
We can then use the factorization
\begin{align*}
     & \sum_{\xi_i} \exp \Bigg\{ \beta k \sum_b \xi^b_i \sigma^a_i + \beta^* \beta \alpha r^* \sum_b \xi^*_i \xi^b_i +  x \beta \sqrt{\alpha r} \sum_b \xi^b_i \Bigg\} \\
     & \quad= \prod_b \sum_{\xi^b_i} \exp \left\{ \beta k \xi^b_i \sigma^a_i + \beta^* \beta \alpha r^* \xi^*_i \xi^b_i + x \beta \sqrt{\alpha r} \xi^b_i \right\}     \\
     & \quad= \prod_b \left[ 2 \cosh \left( \beta k \sigma^a_i + \beta^* \beta \alpha r^* \xi^*_i + x \beta \sqrt{\alpha r} \right) \right]                            \\
     & \quad= 2^L \cosh^L \left( \beta k \sigma^a_i + \beta^* \beta \alpha r^* \xi^*_i + x \beta \sqrt{\alpha r} \right)\,,
\end{align*}
in order to simplify the free energy to
\begin{align*}
    f & = \Extr \Bigg\{ \beta^* \beta \alpha \left[ q^{*} \right]^p - \frac{1}{2} \beta^2 \alpha q^p + \beta m^p - \beta^* \beta \alpha r^{*} q^{*} + \frac{1}{2} \beta^2 \alpha r q - \beta m k                                       \\
      & \quad- \frac{1}{2} \beta^2 \alpha r + \frac{1}{2} \beta^2 \alpha + \lim_{L \rightarrow 0} \Bigg( \frac{\partial}{\partial L} \Bigg[ \log \Bigg\langle \sum_{\xi_i} \int_{\mathbb{R}} dx \exp \left\{ -\frac{1}{x} x^2 \right\} \\
      & \qquad\exp \Bigg\{ \beta k \sum_b \xi^b_i \sigma^a_i + \beta^* \beta \alpha r^* \sum_b \xi^*_i \xi^b_i + x \beta \sqrt{\alpha r} \sum_b \xi^b_i \Bigg\} \Bigg\rangle \Bigg] \Bigg) \Bigg\}                                     \\
      & = \Extr \Bigg\{ \beta^* \beta \alpha \left[ q^{*} \right]^p - \frac{1}{2} \beta^2 \alpha q^p + \beta m^p - \beta^* \beta \alpha r^{*} q^{*} + \frac{1}{2} \beta^2 \alpha r q - \beta m k                                       \\
      & \quad- \frac{1}{2} \beta^2 \alpha r + \frac{1}{2} \beta^2 \alpha + \log 2 + \lim_{L \rightarrow 0} \Bigg( \frac{\partial}{\partial L} \Bigg[ \log \Bigg\langle \int_{\mathbb{R}} dx \exp \left\{ -\frac{1}{x} x^2 \right\}     \\
      & \qquad\cosh^L \left( \beta k \sigma^a_i + \beta^* \beta \alpha r^* \xi^*_i + x \beta \sqrt{\alpha r} \right) \Bigg\rangle \Bigg] \Bigg) \Bigg\}\,.
\end{align*}
After differentiating and taking the limit, we get
\begin{align*}
    f & = \Extr_{m,k,q,r,q^*,r^*} \Bigg\{ \beta^* \beta \alpha \left[ q^{*} \right]^p - \frac{1}{2} \beta^2 \alpha q^p + \beta m^p - \beta^* \beta \alpha r^{*} q^{*}                                                               \\
      & \quad+ \frac{1}{2} \beta^2 \alpha r q - \frac{1}{2} \beta^2 \alpha r - \beta m k + \frac{1}{2} \beta^2 \alpha + \log 2                                                                                                      \\
      & \quad+ \int dx \frac{1}{\sqrt{2\pi}} \exp \left\{ -\frac{1}{2} x^2 \right\} \Big\langle \log \left[ \cosh \left( \beta \left[ \sqrt{\alpha r} x + \beta^* \alpha r^* + k z \right] \right) \right] \Big\rangle_z \Bigg\}\,.
\end{align*}
In the case of $p^* = 2$ and $p \geq 3$ with finite $\alpha = \frac{M p!}{N^{p-1}}$ and $\lambda = \frac{\left[ \beta^* \right]^{p/2}}{(p/2)!} N^{p/2 - 1}$, the free energy has the same form but with $\beta^*$ replaced by $\lambda$. On the other other hand, the free energy with finite $\alpha = \frac{M (p/2 + 1)!}{N^{p/2}}$ and $\eta$ evaluates to:
\begin{align*}
    f = \Extr_{m,k,q^*,r^*} \Bigg\{ \beta \eta \alpha \left[ q^{*} \right]^p - \beta m^p - \beta \eta \alpha r^{*} q^{*} - \beta m k + \log 2 + \Big\langle \log \left[ \cosh \left( \beta \left[ \eta \alpha r^* + k z \right] \right) \right] \Big\rangle_z \Bigg\}\,.
\end{align*}

\section{Direct model RSB ansatz}
\label{app:rsb_ansatz}

Recall that the average replicated partition function of the direct model (see Eq. \ref{eqn:direct_model_partition_function}) takes the form
\begin{equation}
    \fontsize{10pt}{12}\selectfont
    \left\langle Z^L \right\rangle  \approx \Bigg\langle \sum_{\sigma} \exp \Bigg( \beta N \sum_{\gamma} \sum_{\mu \in \Gamma_{\gamma}} \left[ \frac{1}{N} \sum_i \xi^{\mu}_i \sigma^{\gamma}_i \right]^p + \beta \sum_{\gamma} \sum_{\mu \in \Bar{\Gamma}} \frac{p!}{N^{p-1}} \sum_{i_1 < ... < i_p} \xi^{\mu}_{i_1} ... \xi^{\mu}_{i_p} \sigma^{\gamma}_{i_1} ... \sigma^{\gamma}_{i_p} \Bigg) \Bigg\rangle\,.
\end{equation}
Introducing a new replica $\sigma^0$, we rewrite it as
\begin{equation}
    \fontsize{10pt}{12}\selectfont
    \nonumber \left\langle Z^L \right\rangle  = \Bigg\langle \sum_{\sigma} \exp \Bigg( \beta N \sum_{\gamma} \sum_{\mu \in \Gamma_{\gamma}} \left[ \frac{1}{N} \sum_i \xi^{\mu}_i \sigma^{\gamma}_i \right]^p + \beta \sum_{\gamma} \sum_{\mu \in \Bar{\Gamma}} \frac{p!}{N^{p-1}} \sum_{i_1 < ... < i_p} \xi^{\mu}_{i_1} ... \xi^{\mu}_{i_p} \sigma^{\gamma}_{i_1} ... \sigma^{\gamma}_{i_p} \Bigg) \frac{\left\langle Z \right\rangle}{\left\langle Z \right\rangle} \Bigg\rangle\,,
\end{equation}
where $Z = \sum_{\sigma_0} \exp \left( \beta \sum_{\mu \in \Bar{\Gamma}} \frac{p!}{N^{p-1}} \sum_{i_1 < ... < i_p} \xi^{\mu}_{i_1} ... \xi^{\mu}_{i_p} \sigma^{0}_{i_1} ... \sigma^{0}_{i_p} \right)$. Recall that, in the paramagnetic phase, we have (see \cite{gardner1987multiconnected} and also Appendix \ref{app:direct_cumulant})
\begin{align*}
    \left\langle Z \right\rangle  = \exp \left( \frac{1}{2} \beta^2 \alpha + \log 2 + \mathcal{O} \left( \frac{1}{N^{p/2 - 2}} \right) \right)  = Z \exp \left( \mathcal{O} \left( \frac{1}{N^{p/2 - 2}} \right) \right),
\end{align*}
so $\left\langle Z^L \right\rangle$ can be expressed as
\begin{align*}
    \left\langle Z^L \right\rangle & = \frac{1}{\left\langle Z \right\rangle} \Bigg\langle \sum_{\sigma} \exp \Bigg( \beta N \sum_{\gamma} \sum_{\mu \in \Gamma_{\gamma}} \left[ \frac{1}{N} \sum_i \xi^{\mu}_i \sigma^{\gamma}_i \right]^p  + \beta \sum_{\gamma} \sum_{\mu \in \Bar{\Gamma}} \frac{p!}{N^{p-1}} \sum_{i_1 < ... < i_p} \xi^{\mu}_{i_1} ... \xi^{\mu}_{i_p} \sigma^{\gamma}_{i_1} ... \sigma^{\gamma}_{i_p} \Bigg) \\
                                   & \quad \sum_{\sigma_0} \exp \left( \beta \sum_{\mu \in \Bar{\Gamma}} \frac{p!}{N^{p-1}} \sum_{i_1 < ... < i_p} \xi^{\mu}_{i_1} ... \xi^{\mu}_{i_p} \sigma^{0}_{i_1} ... \sigma^{0}_{i_p} + \mathcal{O} \left( \frac{1}{N^{p/2 - 2}} \right) \right) \Bigg\rangle\,.
\end{align*}
The $\mathcal{O} \left( \frac{1}{N^{p/2 - 2}} \right)$ corrections vanish to leading order in $N$ when we calculate the free entropy.

\clearpage
\section{Monte Carlo simulations for various system sizes}

\label{app:monte_carlo_simulations}

\begin{figure}[ht!]
    \centering
    \includegraphics[width = 0.325\textwidth]{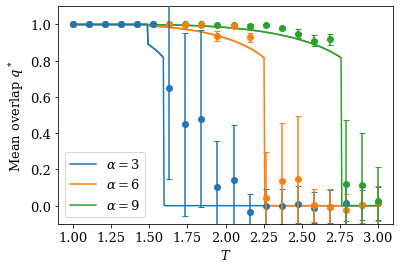}
    \includegraphics[width = 0.325\textwidth]{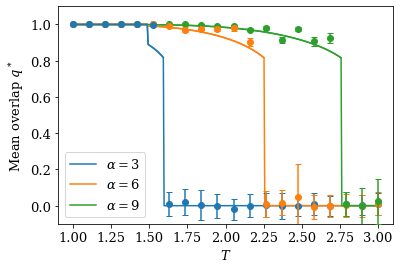}
    \includegraphics[width = 0.325\textwidth]{Dense_Hopfield_network/monte_carlo_global_N=512.png}
    \caption{Monte Carlo simulations of the $p = 3$ inverse model compared against saddle-point solutions for different values of $N$. The $lR$ phase is not included in these plots. The left plot has $N = 128$, the center plot has $N = 256$, and the right plot has $N = 512$. The dots are simulation data at a few values of $\alpha$, and the lines are slices of the saddle-point solutions at the same $\alpha$. There are $M = \frac{\alpha N^{p-1}}{p^!}$ examples $\sigma^a$, and simulation results are averaged over $L = 100$ student patterns. The simulation data is sometimes systematically shifted up with respect to the saddle-point solution, but the size of the difference tends to decrease with $N$. The shift is the most visible when $\alpha = 6$ and right after the fall from $eR$ to $gR$ when $\alpha = 3$. As expected, the fluctuations of the paramagnetic phase also decrease with $N$.}
    \label{fig:finite_size_effects}
\end{figure}

\end{subappendices}

\chapter{Modeling structured data learning with Restricted Boltzmann machines in
the teacher–student setting}
\label{chap:RBM_paper}
\begin{table}[h!]
    \centering
    \begin{tabular}{cc}
        Based on the article \cite{theriault2025modeling}, & doi: \href{https://doi.org/10.1016/j.neunet.2025.107542}{10.1016/j.neunet.2025.107542} \\ available under the CC BY 4.0 license \includegraphics[width=0.1\linewidth]{cc_by_licence_badge.png} & \href{http://crossmark.crossref.org/dialog/?doi=10.1016/j.neunet.2025.107542&domain=pdf}{\includegraphics[width=0.2\linewidth]{crossmark_badge.png}}
    \end{tabular}
\end{table}
\section{Introduction}
\label{sec:RBM_introduction}
Restricted Boltzmann Machines (RBMs) \cite{ackley1985learning, smolensky1986information, freund1991unsupervised, hinton2002training} are empirically known to fit complicated data and then sample new instances that are faithful to the underlying distribution \cite{hinton2002training, leroux2011learning, tubiana2017emergence}. For example, \cite{Salakhutdinov2007restricted} trained RBMs to predict the interests of Netflix users by fitting a database of movie ratings, \cite{kiviken2012multiple} used them to generate realistic textures and \cite{srivastava2013modelling} employed them for topic modelling. There are even universal approximation theorems stating that an RBM with an arbitrary number of hidden units can approximate any distribution with binary support arbitrarily well \cite{leroux2008representational, montufar2011refinements}. The current theoretical understanding of RBMs is largely based on statistical mechanics.
For example, \cite{decelle2017spectral, decelle2018thermodynamics} used statistical mechanics to model RBM training on real data. The statistical mechanics community also studied RBMs trained on simpler, synthetic datasets \cite{decelle2017spectral, decelle2018thermodynamics}. In particular, many works investigated the \textit{teacher-student setting} where a \textit{student} RBM is trained with data produced by a \textit{teacher} RBM \cite{huang2016unsupervised, huang2017statistical, barra2017phase, barra2018phase, huang2018role, hou2019minimal, decelle2021inverse, manzan2025effect}. Such studies are crucial to isolate individual characteristics of structured datasets and Neural Network (NN) design choices in a controlled environment and explain their effects on NN training. For example, \cite{barra2017phase, barra2018phase, manzan2025effect} investigated the effects of the prior chosen for the data on RBM learning. In this paper, we study the effects of data correlations and number of hidden units on RBMs in the teacher-student setting. In particular, we show that RBMs with a few hidden units in the teacher-student setting can serve as a toy model of the lottery ticket hypothesis \cite{frankle2018lottery, zhou2019deconstructing, ramanujan2020hidden, eran2020proving, frankle2020linear}.

RBMs have a visible layer $\sigma = \left\{ \sigma_i \right\}_{i = 1}^N$, a hidden layer $\tau = \left\{ \tau_{\mu} \right\}_{\mu = 1}^P$ and a set of internal connections $\xi = \left\{ \xi^{\mu}_i \right\}_{1 \leq i \leq N}^{1 \leq \mu \leq P}$, which are commonly referred to as weights. The visible layer is a set of concrete features found directly in the data, the hidden layer is an internal representation of the data in terms of abstract concepts, and the weights represent how the input features and abstract concepts are correlated with one another. To give a caricatural example, the presence of pointy ears as an input feature could be correlated with the abstract concept of a cat in the weights of a particular RBM. Given $\xi$, the visible and hidden layer follow the joint distribution
\begin{align}\label{eq:Gibbs}
    \Prob_\beta \left( \sigma, \tau \mid \xi \right) &= Z_\beta \left( \xi \right)^{-1} \Prob \left( \sigma \right) \Prob \left( \tau \right) \exp \left(-\beta H \left[ \sigma, \tau ; \xi \right] \right),
\end{align}
where the Hamiltonian $H \left[ \sigma, \tau ; \xi \right] = -\frac{1}{\sqrt{N}} \sum_{\mu = 1}^P \tau_{\mu} \sum_{i = 1}^N \xi^{\mu}_i \sigma_i$ weighs the cost of every RBM configuration, $\Prob \left( \sigma \right)$ and $\Prob \left( \tau \right)$ are priors on the visible and hidden layers, respectively, and $Z_\beta \left( \xi \right) = \mathbb{E}_{\sigma, \tau} \left[ \exp \left( -\beta H \left[ \sigma, \tau ; \xi \right] \right) \right]$ is the partition function normalizing the distribution, with $\mathbb{E}_{\sigma, \tau}$ the joint expectation over the visible and hidden unit priors. Intuitively, the priors are the default distributions of $\sigma$ and $\tau$ when the Hamiltonian does not contribute to $\Prob_\beta \left( \sigma, \tau \mid \xi \right)$, i.e. when $\beta$ is zero. Gibbs ditributions of the form (\ref{eq:Gibbs}) have been deeply investigated in the mathematical physics community for their link with the Parisi theory of spin glasses \cite{barra2015multi,panchenko2015free, barra2012glassy, genovese2020legendre, genovese2017overlap, genovese2016non}.
An RBM with a given $\xi$ can generate data $\sigma$ by sampling the marginal distribution
\begin{equation}
    \label{eq:RBM_direct_distribution}
    \Prob_\beta \left( \sigma \mid \xi \right) = Z_\beta \left( \xi \right)^{-1} \Prob \left( \sigma \right) \psi_\beta \left( \sigma ; \xi \right)= Z_\beta \left( \xi \right)^{-1} \Prob(\sigma) \prod_{\mu=1}^P \phi_\beta(\xi^\mu\cdot \sigma),
\end{equation}
where $\phi_\beta \left(\xi^\mu\cdot \sigma \right) = \mathbb{E}_{\tau_\mu} \left[ \exp \left( \frac{\beta}{\sqrt{N}}  \tau_{\mu} \sum_{i = 1}^N \xi^{\mu}_i \sigma_i \right) \right]$,
$\psi_\beta \left( \sigma ; \xi \right)$ factorizes as $\psi_\beta \left( \sigma ; \xi \right) = \prod_{\mu=1}^P \phi_\beta(\xi^\mu\cdot \sigma)$ and $Z_\beta \left( \xi \right) = \mathbb{E}_{\sigma} \left[ \psi_\beta \left( \sigma ; \xi \right) \right]$.  Marginal Gibbs distributions of the form (\ref{eq:RBM_direct_distribution}) are also known as generalized Hopfield networks \cite{barra2017phase, barra2018phase, hopfield1982neural, sollich2014extensive, agliari2018non, agliari2015retrieval, agliari2015hierarchical}.
Conversely, following Bayes' theorem, the weights of an RBM can be trained on a dataset $\boldsymbol{\sigma} = \left\{ \sigma^a_i \right\}_{1 \leq i \leq N}^{1 \leq a \leq M}$ of $M$ examples by sampling the posterior distribution
\begin{equation}
    \label{eq:inverse_distribution}
    \Prob_{\beta} \left( \xi | \boldsymbol{\sigma} \right)
    = \mathcal{Z}_{\beta} \left( \boldsymbol{\sigma} \right)^{-1} \Prob \left( \xi \right) \prod_{a = 1}^M \Prob_\beta \left( \sigma^a | \xi \right),
\end{equation}
where $\Prob \left( \xi \right)$ is a prior on the weights and $\mathcal{Z}_{\beta} \left( \boldsymbol{\sigma} \right) = \mathbb{E}_\xi \left[ \prod_{a = 1}^M \Prob_\beta \left( \sigma^a | \xi \right) \right]$ is the posterior partition function normalizing the distribution. In the teacher-student setting, the data $\boldsymbol{\sigma}$ used to train the RBM is produced by another RBM \cite{huang2016unsupervised, huang2017statistical, barra2017phase, barra2018phase, huang2018role, hou2019minimal, decelle2021inverse, manzan2025effect}. In other words, a \textit{student} RBM is trained using a dataset supplied by a \textit{teacher} RBM. The student's ability to fit the teacher's data can then be evaluated in terms of the so-called overlaps $Q(\xi^{*\mu},\xi^\nu)=\frac{1}{N} \sum_{i = 1}^N \xi^{* \mu}_i \xi^{\nu}_i$ between the teacher's weights $\xi^* = \left\{ \xi^{* \mu}_i \right\}_{1 \leq i \leq N}^{1 \leq \mu \leq P^*}$ and the student's weights $\xi = \left\{ \xi^{\nu}_i \right\}_{1 \leq i \leq N}^{1 \leq \mu \leq P}$, whose rows $\xi^{* \mu} = \left\{ \xi^{* \mu}_i \right\}_{i = 1}^N$ and $\xi^{\mu} = \left\{ \xi^{\nu}_i \right\}_{i = 1}^N$ are also called patterns. As such, we refer to the expected value of $Q(\xi^{*\mu},\xi^\nu)$ as the student's \textit{performance}.
This framework was used to predict the critical amount of data needed to train multiple variants of RBMs \cite{huang2016unsupervised, huang2017statistical, barra2017phase, barra2018phase, huang2018role, hou2019minimal, manzan2025effect}.

In \cite{barra2017phase}, it was observed empirically and conjectured that the performance of a student RBM is independent of the number of hidden units when the teacher patterns are uncorrelated. It was later shown analytically that a student RBM taught by a teacher with uncorrelated
patterns achieves the same performance whether the teacher and the student have $1$ or $2$ hidden units each \cite{hou2019minimal}. In a nutshell, this simplification occurs because an RBM with $2$ uncorrelated hidden units effectively factorizes into $2$ RBMs with $1$ hidden unit each. The conjecture of \cite{barra2017phase} was explicitly rephrased in terms of this factorization property in \cite{hou2019minimal}\footnote{See the end of Appendix C of the cited paper.}.
Beyond this idealized scenario, it has long been known that machine learning models benefit from the correlations found in structured data to build an efficient internal representation \cite{gardner1988space}. This phenomenon has received considerable attention in recent theoretical studies \cite{decelle2017spectral, decelle2018thermodynamics, dobriban2018high, goldt2020modelling, wu2020optimal, ghorbani2021neural, liao2021random, chen2021multiple, nakkiran2021optimal, richards2021asymptotics, ascoli2021interplay, loureiro2021learning, refinetti2021classifying, ichikawa20222statistical}, notably for RBMs with $2$ hidden units in the teacher-student setting \cite{hou2019minimal}. Extending the latter study to an arbitrary number of hidden units remains an intriguing open problem. In fact, although increasing the number of hidden units in the teacher model may bring us closer to the complexity of real data—complexity that RBMs are known to capture due to universal approximation theorems \cite{leroux2008representational, montufar2011refinements}—there is still no theoretical framework that fully explains the learning performance of RBMs for an arbitrary number of hidden units.

In this work, we evaluate the student's learning performance in the teacher-student setting where both the teacher and the student have an arbitrary finite number of hidden units and the teacher patterns are allowed to be correlated with one another. In particular, we evaluate the critical data load $\alpha_{\text{crit}} = \frac{M}{N}$ above which learning becomes possible. In Section \ref{sec:RBM_teacher-student}, we introduce the teacher-student setting and the replica method used in our calculations. In Section \ref{sec:core_results}, we present the so-called \textit{saddle-point} equations governing the performance and the critical load that we obtain from it. In Section \ref{sec:no_correlations}, we discuss the case where there are no correlations between the teacher patterns.
This Section is divided into three Subsections. In Subsection \ref{sec:independence} and \ref{sec:independence_g}, we show that the student's performance is independent of the number of hidden units when the teacher patterns are uncorrelated. In Subsection \ref{sec:lottery}, we argue that our teacher-student setting with uncorrelated teacher patterns can serve as a toy model of the lottery ticket hypothesis. Next, in Section \ref{sec:uniform_correlations}, we discuss the effects of uniform teacher pattern correlations on the student's performance. In particular, in Subsection \ref{sec:concentration}, we discuss the different learning phases of the teacher-student problem as a function of the correlations and the number of hidden units. Finally, in Section \ref{sec:random_correlations}, we discuss random correlations and compare their effect on the performance to that of uniform correlations. Throughout the paper, we compare key results against Monte Carlo simulations. The code and hyperparameter values of the training algorithms used to make the figures are available at the following public Github repository \cite{theriault2025modellingsoftware}.

\section{Model}
\label{sec:RBM_teacher-student}
In the teacher-student setting, a student RBM with marginal likelihood $\Prob_\beta \left( \sigma | \xi \right)$ (Eq. \ref{eq:RBM_direct_distribution}) is trained using a dataset $\boldsymbol{\sigma} = \left\{ \sigma^a_i \right\}_{1 \leq i \leq N}^{1 \leq a \leq M}$ of $M$ examples $\sigma^a = \left\{ \sigma^a_i \right\}_{i = 1}^{N}$ of dimension $N$, generated by a teacher RBM with a prescribed marginal likelihood $\Prob_{\beta^*} \left( \sigma | \xi^* \right)$. We call $\alpha=\frac{M}{N}$ the ratio between the size of the training set and the input dimension. In this scenario, the student knows that the correct model for the data is an RBM. However, it does not necessarily know the number of hidden units $P^*$ and the inverse temperature $\beta^*$ used by the teacher. Therefore, unless explicitly stated otherwise, we will assume that the number of hidden units $P$ and the inverse inference temperature $\beta$ of the student are not necessarily the same as the teacher's. For convenience, we will frequently state our results in terms of the temperatures $T^* = 1/\beta^*$ and $T = 1/\beta$ rather than in terms of $\beta^*$ and $\beta$.

For simplicity, we assume that the visible and hidden units of both the student and the teacher take values in $\left\{ -1, +1 \right\}$ with a uniform prior, i.e. they are binary random variables with no prior bias towards $-1$ or $+1$.
We impose structure in the data by taking the teacher patterns to be random variables with a fixed covariance matrix $\mathcal{Q}$. To be more precise, we assume that the columns $\xi^*_i = \left\{ \xi^{* \mu}_i \right\}_{\mu = 1}^{P^*}$ of $\xi^*$ are i.i.d. random variables, with mean $0$ and a well-defined $P^* \times P^*$ covariance matrix $\mathcal{Q}$. Uncorrelated teacher patterns are obtained by setting $\mathcal{Q} = \mathbf{I}$.

As previously mentioned, the student learns its patterns $\xi$ by sampling them from the posterior distribution
\begin{eqnarray} \label{eq:RBM_posterior}
    \Prob_{\beta} \left( \xi | \boldsymbol{\sigma} \right) &=& \mathcal{Z}_{ \beta} \left( \boldsymbol{\sigma} \right)^{-1} \Prob \left( \xi \right) \prod_{a = 1}^M \Prob_\beta \left( \sigma^a | \xi \right) =
    \mathcal{Z}_{ \beta} \left( \boldsymbol{\sigma} \right)^{-1} \Prob \left( \xi \right)
    \prod_{a = 1}^M \left[ Z_{\beta} \left( \xi \right)^{-1} \psi_\beta \left( \sigma^a ; \xi \right) \right]\nonumber \\
    &=& \mathcal{Z}_{ \beta} \left( \boldsymbol{\sigma} \right)^{-1}\Prob \left( \xi \right)
    \prod_{a = 1}^M \left[ Z_{\beta} \left( \xi \right)^{-1} \Prob(\sigma^a)\prod_{\mu=1}^P\phi_\beta \left( \sigma^a \cdot \xi^\mu \right) \right]\nonumber \\
    &=& \mathcal{Z}_{ \beta} \left( \boldsymbol{\sigma} \right)^{-1}\Prob \left( \sigma \right)Z_{\beta} \left( \xi \right)^{-M}\prod_{\mu=1}^P \psi_\beta(\xi^\mu;\sigma).
\end{eqnarray}
Compared with Eq. (\ref{eq:RBM_direct_distribution}), the posterior distribution is composed of $P$ generalized Hopfield distributions, one for each hidden unit. The data now plays the role of dual patterns, which interact only through the term $Z_{\beta} \left( \xi \right)^{-M}$. The latter is therefore responsible for encoding mutual correlation between student patterns, otherwise independent \cite{barra2017phase,decelle2021inverse,alemanno2023hopfield}. The student has no access to the structure of the teacher patterns, so we assume an identity covariance matrix for the student pattern prior $\Prob (\xi)$.

In general, the partition functions $Z_{\beta} \left( \xi \right)$ and $\mathcal{Z}_{ \beta} \left( \boldsymbol{\sigma} \right)$ are intractable, which makes the properties of Eq. (\ref{eq:RBM_posterior}) difficult to study. However, our assumptions about the priors on the visible units, hidden units and patterns make analytical computations possible in the limit of $N, M \rightarrow \infty$.

The overlaps $Q(\xi^{*\mu},\xi^\nu)=\frac{1}{N} \sum_{i = 1}^N \xi^{* \mu}_i \xi^{\nu}_i$ are a good measure of the student's learning performance because they quantify how close the student patterns are to the teacher patterns. We exploit techniques from statistical mechanics to compute the expected value $\mathbb{E}_{\xi^*,\boldsymbol{\sigma},\xi} \left[Q(\xi^{*\mu},\xi^\nu)\right]$ of the overlaps with respect to the distribution of the teacher patterns, the generated dataset and the inferred student patterns. Specifically, in the limit of large dataset and data dimension $M,N\to \infty$, we obtain the expected value of the overlaps as a byproduct of the limiting quenched free entropy \begin{equation}
\label{eq:def_free_entropy}
f (\alpha, \beta^*, \beta, P^*, P, \mathcal{Q}) = \lim_{M,N\to\infty} \frac{1}{N} \mathbb{E}_{\xi^*,\boldsymbol{\sigma}} \log \left[ \mathcal{Z}_{ \beta}(\boldsymbol{\sigma}) \right],
\end{equation}
where $\mathbb{E}_{\xi^*,\boldsymbol{\sigma}}$ is the expected value w.r.t. the joint distribution of the teacher patterns and generated dataset. In fact, we show in Section \ref{sec:core_results} that Eq. (\ref{eq:def_free_entropy}) can be expressed as the result of a variational principle w.r.t. a set of order parameters, whose optimum gives the expected value of the overlaps. We then use this result to investigate the effects of $P$ and $\beta$ on the student's ability to learn a dataset characterized by $P^*, \beta^*$ and $\mathcal{Q}$, as well as the impact of these hyperparameters on the critical threshold $\alpha_{\text{crit}}$ beyond which learning becomes possible. We focus our quantitative discussion on Gaussian and binary priors $\Prob (\xi), \Prob (\xi^*)$ (see \ref{sec:binary_random}), but our results can be generalized to other pattern distributions that meet the other assumptions stated earlier in this Section. The Gaussian case is particularly interesting because it is closely related to RBMs used in practical applications \cite{kiviken2012multiple, Salakhutdinov2007restricted, srivastava2013modelling}, which usually have continuous weights.

\section{Results and discussion}
\label{sec:results}

\subsection{Free entropy and saddle point equations}
\label{sec:core_results}

The free entropy $f$ can be computed by exploiting a well-established statistical mechanics technique called the replica trick, which is based on the identity
\begin{align*}
   f = \lim_{N\to\infty}\frac{1}{N} \mathbb{E}\log \left[ \mathcal{Z} \right] &= \lim_{N\to\infty, L \rightarrow 0} \left( \frac{1}{LN} \log \mathbb{E} \left[\mathcal{Z}^L  \right] \right).
\end{align*}
Calculations are shown in the Appendices. In \ref{app:RBM_partition_function}, we calculate the average replicated partition function $\mathbb{E}_{\xi^*,\boldsymbol{\sigma}}\left[ \mathcal{Z}^L \right]$ in the limit of $N \rightarrow \infty$. In \ref{app:free_entropy}, we use $\mathbb{E}_{\xi^*,\boldsymbol{\sigma}}\left[ \mathcal{Z}^L \right]$ to evaluate the quenched free entropy $f$ under the so-called Replica-Symmetric (RS) approximation. We find that the RS free entropy can be expressed in terms of the variational principle
\begin{align*}
    \label{eq:RBM_free_entropy}
    f &= \Extr_{m, \hat{m}, q, \hat{q}, s, \hat{s}} \Bigg\{ -\sum_{\mu = 1}^{P^*} \sum_{\nu = 1}^{P} m^{\mu \nu} \hat{m}^{\mu \nu} - \frac{1}{2} \sum_{\mu \neq \nu}^{P} s^{\mu \nu} \hat{s}^{\mu \nu} + \frac{1}{2} \sum_{\mu, \nu = 1}^P q^{\mu \nu} \hat{q}^{\mu \nu} \\
    &\quad+ \mathbb{E}_{\xi^*} \mathbb{E}_{z} \log \left[ \mathcal{Z} \left( \mathcal{L}^C \right) \right] + \alpha \left\langle \mathbb{E}_z \log \left[ \mathcal{Z} \left( \mathcal{L}^O \right) \right]\right\rangle_{\mathcal{M}_*} - \alpha \log \left[ \mathcal{Z} \left( \mathcal{M} \right) \right] \Bigg\}, \numberthis
\end{align*}
where $z = \left[ z_{\mu \nu} \right]_{\mu,\nu = 1}^{P}$ is a set of i.i.d. standard Gaussian random variables, and where the effective energies $\mathcal{M}_*$, $\mathcal{M}$, $\mathcal{L}^O$ and $\mathcal{L}^C$ are defined as
\begin{eqnarray}
    \mathcal{M}_* \left( \tau_* \right) &=& \frac{1}{2} \left[ \beta^* \right]^2 \sum_{\mu, \nu = 1}^{P^*} \mathcal{Q}_{\mu \nu} \tau^*_{\mu} \tau^*_{\nu} \label{hamiltonian_M_s} \\
    \mathcal{M} \left( \tau \right) &=& \frac{1}{2} \beta^2 \sum_{\mu, \nu = 1}^{P} s^{\mu \nu} \tau_{\mu} \tau_{\nu} \label{hamiltonian_M} \\
    \mathcal{L}^O \left( \tau; \tau^*, z \right) &=& \mathcal{L}_{\beta^*, \beta} \left( \tau, \tau^*, z ; m, s, q \right)\label{hamiltonian_L_O} \\
    \mathcal{L}^C \left( \xi; \xi^*, z \right) &=& \mathcal{L}_{1, 1} \left( \xi, \xi^*, z ; \hat{m}, \hat{s}, \hat{q} \right) \label{hamiltonian_L_C} \\
    \text{with} \quad \mathcal{L}_{\lambda_1, \lambda_2} \left( \xi, \xi^*, z ; m, s, q \right) &=& \frac{1}{2} \left[ \lambda_2 \right]^2 \sum_{\mu, \nu = 1}^P \left( s^{\mu \nu} - q^{\mu \nu} \right) \xi^{\mu} \xi^{\nu} \label{eq:sym_L} \\
    &\quad&+ \lambda_1 \lambda_2 \sum_{\mu = 1}^{P^*} \sum_{\nu = 1}^{P} m^{\mu \nu} \xi^{* \mu} \xi^{\nu} \\
    &\quad&+ \lambda_2 \sum_{\mu, \nu = 1}^P A_{\mu \nu} \left( q \right) z_{\mu \nu} \frac{\xi^{\mu} + \xi^{\nu}}{2} \nonumber \\
    \text{and} \quad A_{\mu \nu} \left( q \right) &=& \sqrt{2 q^{\mu \nu} - \delta_{\mu \nu} \sum_{\eta = 1}^P q^{\mu \eta}} \label{eq:diagonal_dominance}.
\end{eqnarray}
In particular, we use the notation $\mathcal{Z}(f) = \mathbb{E}_x \left[ \exp \left\{ f \left( x ; \cdot \right) \right\} \right]$ for the partition function and $\left\langle g \right\rangle_f=\mathbb{E}_x \left\{ g(x) \mathcal{P} \left[ f \right] \left( x ; \cdot \right) \right\}$ for the expectation value of an observable $g$ w.r.t. the Gibbs distribution $\mathcal{P} \left[ f \right] \left( x ; \cdot \right) = \mathcal{Z}(f)^{-1} \exp \left[ f \left( x ; \cdot \right) \right]$ of an effective energy $f$, where $\mathbb{E}_x$ is the expectation over the prior of $x$. As explained in \ref{sec:effective_hamiltonian}, $q^{\mu \nu}$ is assumed symmetric in Eq. (\ref{eq:sym_L}).
The solution of this variational principle (see \ref{app:RBM_saddle-point} for a detailed derivation) must obey the saddle-point equations
\begin{align*}
    \label{eq:RBM_saddle-point}
    m^{\mu \nu} &= \mathbb{E}_{\xi^*} \mathbb{E}_{z} \left[ \xi^{* \mu} \left\langle \xi^{\nu} \right\rangle_{\mathcal{L}^C} \right] \\
    s^{\mu \nu} &= \mathbb{E}_{\xi^*} \mathbb{E}_{z} \left[ \left\langle \xi^{\mu} \xi^{\nu} \right\rangle_{\mathcal{L}^C} \right] \\
    q^{\mu \nu} &= \mathbb{E}_{\xi^*} \mathbb{E}_{z} \left[ \left\langle \xi^{\mu} \right\rangle_{\mathcal{L}^C} \left\langle \xi^{\nu} \right\rangle_{\mathcal{L}^C} \right],\\
    \hat{m}^{\mu \nu} &= \beta^* \beta \alpha \ \left\langle \mathbb{E}_{z} \left[ \tau^*_{ \mu} \left\langle \tau_{\nu} \right\rangle_{\mathcal{L}^O} \right] \right\rangle_{\mathcal{M}_*} \numberthis \\
    \hat{s}^{\mu \nu} &= \beta^2 \alpha \left( \left\langle \mathbb{E}_{z} \left[ \left\langle \tau_{\mu} \tau_{\nu} \right\rangle_{\mathcal{L}^O} \right]\right\rangle_{\mathcal{M}_*} - \left\langle \tau_{\mu} \tau_{\nu} \right\rangle_{\mathcal{M}} \right) \\
    \hat{q}^{\mu \nu} &= \beta^2 \alpha \left\langle \mathbb{E}_{z} \left[ \left\langle \tau_{\mu} \right\rangle_{\mathcal{L}^O} \left\langle \tau_{\nu} \right\rangle_{\mathcal{L}^O} \right] \right\rangle_{\mathcal{M}_*}.
\end{align*}
The optimal order parameters $m\in \mathbb{R}^{P^*\times P}$, $s\in \mathbb{R}^{P\times P}$ and $q\in\mathbb{R}^{P\times P}$ solving Eqs. (\ref{eq:RBM_saddle-point}) are to be interpreted as expected overlaps. First of all, $m^{\mu \nu}$, which is commonly known as the Mattis magnetization, is the limiting expected overlap between the teacher pattern $\xi^{* \mu}$ and the student pattern $\xi^{\nu}$, i.e. 
\begin{equation}
m^{\mu \nu} = \lim_{N,M \rightarrow \infty} \mathbb{E}_{\xi^*,\boldsymbol{\sigma},\xi} \left[Q(\xi^{*\mu},\xi^\nu)\right]\ \ \ \ \mu=1,\ldots, P^* \  \nu=1,\ldots, P,
\end{equation}
where $\mathbb{E}_{\xi^*,\boldsymbol{\sigma},\xi}$ is the expectation w.r.t. the joint distribution $\Prob(\xi^*)\Prob_{\beta^*}(\boldsymbol{\sigma}|\xi^*)\Prob_\beta(\xi|\boldsymbol{\sigma})$ of teacher patterns, dataset and student patterns.
Second of all, $s^{\mu \nu}$
with $\mu \neq \nu$
is the limiting expected overlap between any two student patterns $\xi^{\mu}$ and $\xi^{\nu}$ from the same sample $\xi \sim \Prob_\beta \left( \xi | \boldsymbol{\sigma} \right)$, i.e. 
\begin{equation}
s^{\mu \nu} = \lim_{N,M \rightarrow \infty} \mathbb{E}_{\xi^*,\boldsymbol{\sigma},\xi} \left[Q(\xi^\mu,\xi^\nu)\right]\ \ \ \mu,\nu=1,\ldots,P.
\end{equation}
Finally, $q^{\mu \nu}$ is the limiting expected overlap between any two student patterns $\xi^{1 \mu}$ and $\xi^{2 \nu}$ from two independent posterior samples $\xi^1$ and $\xi^2$, i.e. 
\begin{align*}
q^{\mu \nu} &= \lim_{N,M \rightarrow \infty} \mathbb{E}_{\xi^*,\boldsymbol{\sigma}, \xi^1\times\xi^2} \left[Q(\xi^{1 \mu},\xi^{2 \nu})\right] \\
&= \lim_{N,M \rightarrow \infty} \mathbb{E}_{\xi^*,\boldsymbol{\sigma}} \left[Q(\mathbb{E}_{\xi|\boldsymbol{\sigma}}\left[\xi^{\mu}\right],\mathbb{E}_{\xi|\boldsymbol{\sigma}}\left[\xi^{\nu}\right])\right]\ \ \ \mu,\nu=1,\ldots,P, \numberthis
\end{align*}
where $\mathbb{E}_{\xi|\boldsymbol{\sigma}}$ indicates the expectation w.r.t. the posterior distribution (Eq. \ref{eq:RBM_posterior}).
While $s$ measures the effective correlation between student patterns, the so-called spin-glass order parameter $q$ quantifies the tendency of the student to stay frozen in specific configurations rather than sampling all possible patterns uniformly. For simplicity's sake, we will usually call $m$ the magnetization and $q$ the Spin-Glass (SG) overlap. As in the Introduction, we will also occasionally refer to $m$ as the student's performance.

\begin{figure}
    \centering
    \includegraphics[width = \textwidth]{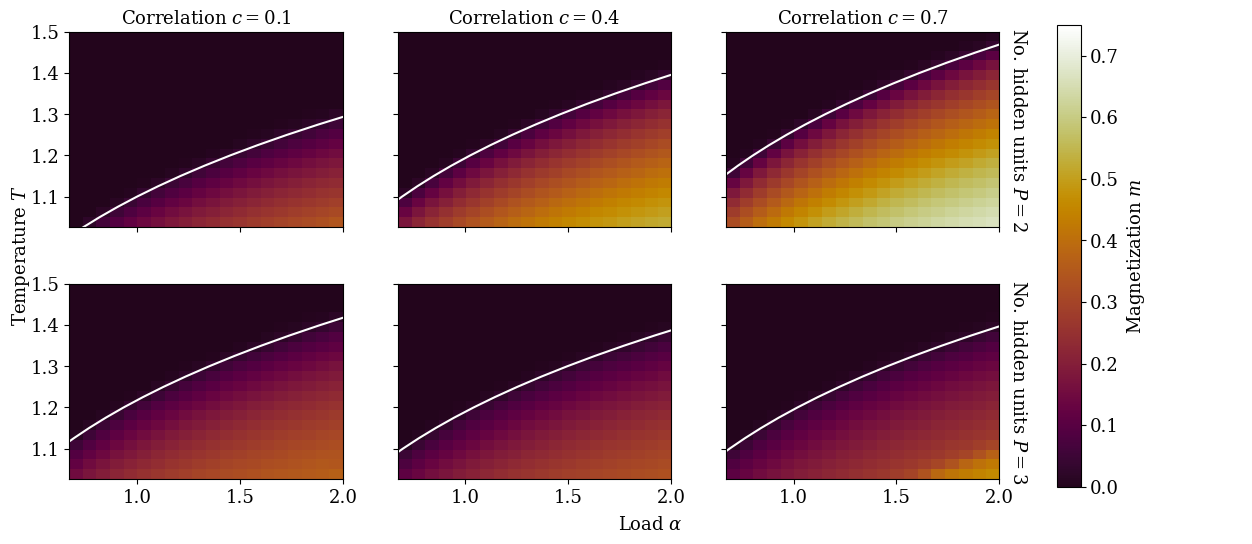}
    \caption{Mattis magnetization $m$ for $\beta = \beta^*$ and $P = P^*$ as a function of the number of hidden units $P$, the correlation $c$, the temperature $T$ and the data load $\alpha$. $m$ is obtained by solving Eqs. (\ref{eq:RBM_saddle-point}) for binary student patterns with a uniform prior and binary teacher patterns with covariance $\mathcal{Q}_{\mu \nu} \sim \mathcal{W} \left( c, P \right)$, where $c \in \left[ 0, 1 \right)$ (see \ref{sec:projected_wishart}). The top and bottom rows feature $P = 2$ and $P = 3$, respectively. The white lines mark the phase transition described by Eq. (\ref{eq:critical_load}) with $\lambda^{\mathcal{S}}_{\text{max}}$ given by Eq. (\ref{eq:max_eigval}).}
    \label{fig:random_phase_diagram}
\end{figure}

The RS saddle point equations (Eqs. \ref{eq:RBM_saddle-point}) can be solved by numerical iteration for any values of the hyperparameters $\beta^*$, $\beta$, $\alpha$, $P^*$ and $P$ (see \ref{app:numerical_methods}). We expect the RS solution to be exact when the student is fully informed about the teacher's prior and hyperparameters and matches them with its own, i.e. $\beta = \beta^*$, $P = P^*$ and $\Prob \left( \xi \right) = \Prob \left( \xi^* \right)$. This regime of complete information is commonly referred to as the Nishimori line \cite{nishimori1980exact, iba1999nishimori, nishimori2001statistical, contucci2009spin}. When $\beta = \beta^*$, we find two different phases (see Figs. \ref{fig:BM_uncorrelated_overlaps_beta=1.2}, \ref{fig:BM_unstructured_overlaps_beta=1.2}, \ref{fig:RBM_phase_diagram} and \ref{fig:random_phase_diagram}) : 
\begin{itemize}
\item the \textit{Paramagnetic}  (P) phase, where the order parameters $m$, $s$ and $q$ all vanish;
\item the \textit{ Ferromagnetic} (F) phase, where they are all larger than zero.  
\end{itemize}
Intuitively, the student RBM can partially learn the teacher patterns in the F phase but becomes unable to do so in the P phase. The corresponding P-F phase transition is thus also the onset of learning. When $\beta \neq \beta^*$, we also find 
\begin{itemize}
\item the \textit{Spin-Glass} (SG) phase, where $m= 0$ but $q>0$.
\end{itemize}
In this phase, the student converges to spurious low-energy states ($q > 0$) unrelated to the teacher patterns ($m = 0$) (see Figs. \ref{fig:m_diagram_fixed_T_s} and \ref{fig:q_diagram_fixed_T_s}). Looking at Figs. (\ref{fig:BM_uncorrelated_overlaps_beta=1.2}), (\ref{fig:BM_unstructured_overlaps_beta=1.2}), (\ref{fig:RBM_phase_diagram}) and (\ref{fig:random_phase_diagram}), the P-F phase transition appears to be second order.
In general, such second-order phase transitions coincide with the onset of instability of the paramagnetic solution in the saddle-point equations (Eqs. \ref{eq:RBM_saddle-point}). The instability condition (see \ref{app:critical_load}) reads as
\begin{align}
    \label{eq:critical_load}
    \alpha \geq \alpha_{\text{crit}} = \frac{1}{\left[ \beta^* \beta \right]^2 \lambda^{\mathcal{S}}_{\text{max}}},
\end{align}
where $\lambda^{\mathcal{S}}_{\text{max}}$ is the largest eigenvalue of the matrix $\mathcal{S} = \mathcal{Q} \mathcal{R}$, with the covariance matrix $\mathcal{R}$ defined as $\mathcal{R}_{\mu\nu}=\left\langle \tau^{*\mu}\tau^{*\nu}\right\rangle_{\mathcal{M}_*}$. As expected, the $\alpha_{\text{crit}}$ of Eq. (\ref{eq:critical_load}) coincides with the P-F phase transition of Figs. (\ref{fig:RBM_phase_diagram}) and (\ref{fig:random_phase_diagram}). Interestingly, $\alpha_{\text{crit}}$ does not depend on the number of hidden units $P$ of the student nor on its prior $\Prob \left( \xi \right)$. Outside the Nishimori line, i.e. when $(\beta,P,P(\xi)) \neq (\beta^*,P^*,P(\xi^*))$, the RS approximation is not expected to be exact. Therefore, in principle, one would need to calculate the so-called Replica Symmetry Breaking (RSB) corrections of $\alpha_{\text{crit}}$. However, we find that Eq. (\ref{eq:critical_load}) is consistent with Monte Carlo simulations even when $P\neq P^*$ (see Fig. \ref{fig:BM_simulated_magnetization_P=1_P_t=2}) and $\Prob(\xi)\neq \Prob(\xi^*)$ (see Fig. \ref{fig:magnetization_and_overlap_P=3_P_t=3}). As such, the RS approximation of $\alpha_{\text{crit}}$ seems robust to the priors and number of hidden units in practice.

\subsection{Learning uncorrelated patterns}
\label{sec:no_correlations}
In this Section, including Subsections \ref{sec:independence}, \ref{sec:independence_g} and \ref{sec:lottery}, we take $\mathcal{Q} = \mathbf{I}$, i.e. there are no correlations between the teacher patterns. In that regime, we find that the teacher-student setting exhibits an effective factorization property on the student patterns, which makes the behavior of the student RBM explainable in terms of that with a single hidden unit.
A first indication of this result is given by the critical load $\alpha_{\text{crit}}$ (see Eq. \ref{eq:critical_load}). As we explained in the previous Section, no matter $\mathcal{Q}$, Eq. (\ref{eq:critical_load}) does not depend on the number $P$ of student patterns. When $\mathcal{Q} = \mathbf{I}$, and thus $\mathcal{S} = \mathbf{I}$ (see Eqs. \ref{eq:critical_load}), we find that
\begin{align}
\label{eq:critical_load_without_correlations}
    \alpha_{\text{crit}} = \frac{1}{\left[ \beta^* \beta \right]^2},
\end{align}
which does not depend on the number $P^*$ of teacher patterns either. Eq. (\ref{eq:critical_load_without_correlations}) generalizes to arbitrary finite $P^*$ and $P$ the critical load $\alpha_{\text{crit}} = \beta^{-4}$ previously found for $P^* = P = 1$ hidden units \cite{huang2016unsupervised, huang2017statistical} and $P^* = P = 2$ uncorrelated hidden units \cite{hou2019minimal} when $\beta^* = \beta$.
As we show in the next Subsections, the previous universality result also holds for the order parameters, which we recall are a good measure of the student's learning performance.

\subsubsection{Independence of the number of hidden units: binary patterns}
\label{sec:independence}
\begin{figure}
    \centering
    \includegraphics[width=0.495\linewidth]{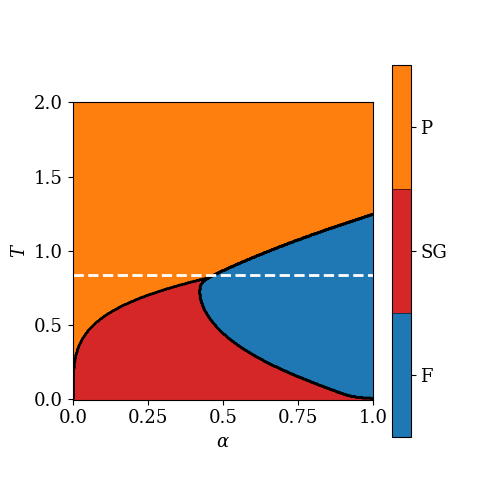}
    \includegraphics[width=0.495\linewidth]{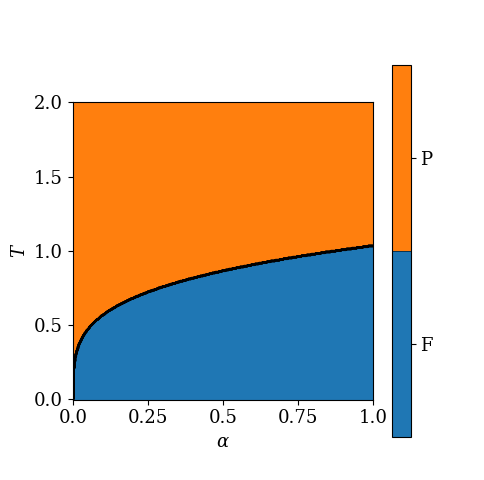}
    \caption{RS phase diagrams of the teacher-student setting with $P = P^* = 1$ obtained by solving Eqs. (\ref{eq:reduced_saddle-point_P1}). The left diagrams has $\beta^* = 1.2$, and the right one, $\beta^* = \beta = 1/T$. The student is unable to learn the teacher patterns in the Paramagnetic phase (P) and in the Spin-Glass phase (SG), but it is able to do so in the Ferromagnetic phase (F). The white dashed line in the left plot is the Nishimori line $\beta^* = \beta$. The magnetization $m$ and SG overlap $q$ solving Eqs. (\ref{eq:reduced_saddle-point_P1}) along this line are plotted on Fig. (\ref{fig:BM_uncorrelated_overlaps_beta=1.2}).}
    \label{fig:RBM_phase_diagram_P1}
\end{figure}

In the case of binary patterns, Eqs. (\ref{eq:RBM_saddle-point}) with $P=P^*=1$ simplify to 
 \begin{align*}
    \label{eq:reduced_saddle-point_P1}
    m &= \mathbb{E}_{z} \left[ \tanh \left( \hat{m} + \sqrt{\hat{q}} z \right) \right] \\
    q &= \mathbb{E}_{z} \left[ \tanh^2 \left( \hat{m} + \sqrt{\hat{q}} z \right) \right] \numberthis \\
    \hat{m} &= \beta^* \beta \alpha \ \mathbb{E}_{z} \left[ \tanh \left( \beta^* \beta m + \beta \sqrt{q} z \right) \right] \\
    \hat{q} &= \beta^2 \alpha \ \mathbb{E}_{z} \left[ \tanh^2 \left( \beta^2 m + \beta \sqrt{q} z \right) \right],
\end{align*}
whose solution is summarized in the phase diagrams of Fig. (\ref{fig:RBM_phase_diagram_P1}). To be more precise, the left and right panels of Fig. (\ref{fig:RBM_phase_diagram_P1}) show the fixed-$\beta^*$ regime and the Nishimori regime $\beta=\beta^*$, respectively. For clarity's sake, we will discuss Fig. (\ref{fig:RBM_phase_diagram_P1}) and the solution of Eq. (\ref{eq:reduced_saddle-point_P1}) in terms of $T^* = 1/\beta^*$ and $T = 1/\beta$ for the remainder of this paragraph. We recall that the student can learn the teacher patterns in the Ferromagnetic (F) phase ($m \neq 0$), but not in the Paramagnetic (P) phase ($m = q = 0$) nor in the Spin-Glass (SG) phase ($m = 0$ and $q > 0$).
For $T < T^*$, we find the SG phase between the P phase and the F phase. The corresponding P-SG transition line $\alpha_{\text{P-SG}} = T^4$ extends until the Nishimori line $T = T^*$, where it meets the F phase. Therefore, the Nishimori line crosses the triple point of the P, SG and F phases, as expected from spin glass theory \cite{nishimori1980exact, nishimori2001statistical}.
On the Nishimori line, it is straightforward to verify that $m = q$, which is also expected from spin glass theory \cite{nishimori1980exact, nishimori2001statistical}. In particular, $m$ and $q$ simultaneously become nonzero above the critical load $\alpha_{\text{crit}} = \left[ T^* T \right]^2 = T^4$ of the P-F phase transition on the Nishimori line, which prevents the SG phase from forming (see Fig. \ref{fig:RBM_phase_diagram_P1}, right panel).
As in some related inference problems in the teacher-student setting, $\alpha_{\text{P-SG}} = T^4$ is identical to the $\alpha_{\text{crit}}$ found on the Nishimori line \cite{manzan2025effect, alemanno2023hopfield, theriault2024dense}. For $T > T^*$, the SG phase does not exist either, and the P phase transitions directly to the F phase at $\alpha_{\text{crit}} = \left[ T^* T \right]^2$.

Looking at the SG-F transition, we see that decreasing the inference temperature $T$ too much can make it harder for the student to learn the teacher patterns.
In particular, a student with enough data to learn the teacher patterns at a given inference temperature $T_1$ may fail to do so at a lower inference temperature $T_2 < T_1$. The RS approximation is probably not completely accurate in this regime, so one would need to calculate RSB corrections to study this behavior quantitatively.

\begin{figure}
    \centering
    \includegraphics[width = \textwidth]{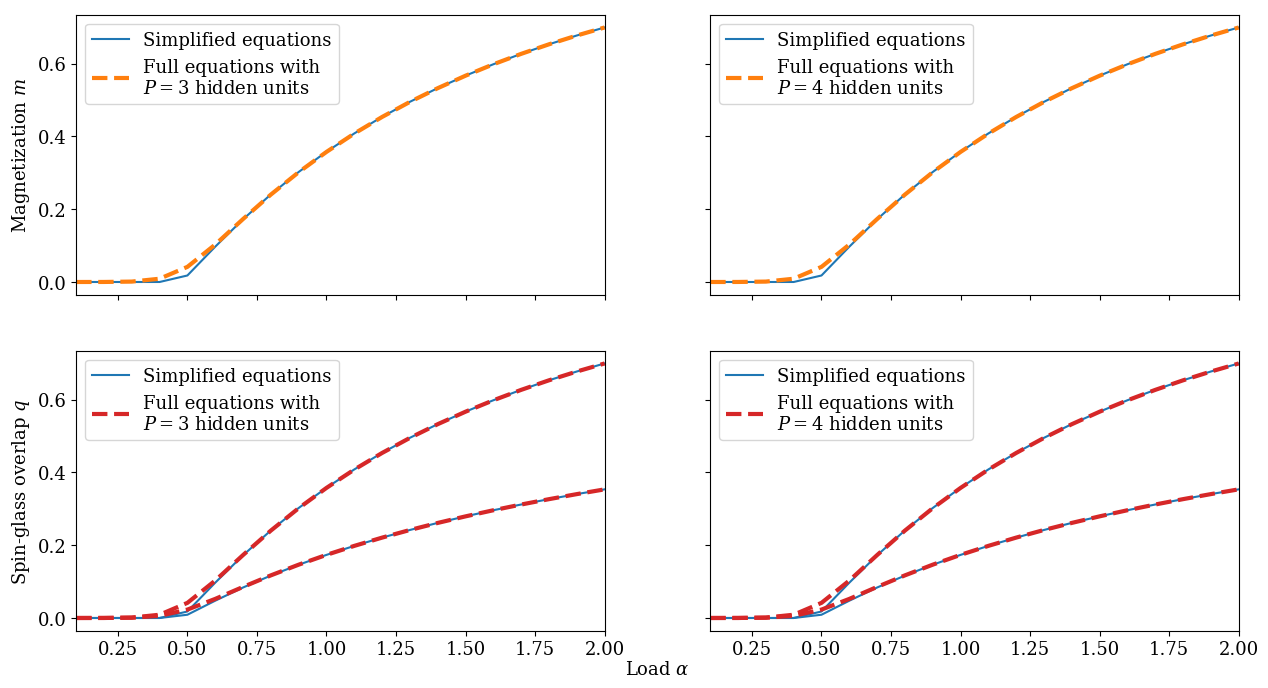}
    \caption{Permutation Symmetry Breaking (PSB) solution of Eqs. (\ref{eq:RBM_saddle-point}) for binary student patterns with a uniform prior and binary teacher patterns with covariance $\mathcal{Q} = \mathbf{I}$, in red and orange, compared against the solution of Eqs. (\ref{eq:reduced_saddle-point_P1}, \ref{eq:reduced_saddle-point}), in blue. We plot the Mattis magnetization $m$ in the top row, and the Spin-Glass (SG) overlap $q$ in the bottom row. The magnetization plots and the top lines of the SG overlap plots show that the student patterns that converge to teacher patterns have the same $m$ and $q$ as the solution of Eqs. (\ref{eq:reduced_saddle-point_P1}), and thus also satisfy $m = q$. Conversely, the bottom lines of the SG overlap plots show that the student patterns that do not converge to a teacher pattern have the same SG overlap $g$ as the solution of Eqs. (\ref{eq:reduced_saddle-point}). We use $P = 3$ and $P^* = 2$ in the left column and $P = 4$ and $P^* = 3$ in the right column. All plots have $\beta^* = \beta = 1.2$.}
    \label{fig:BM_uncorrelated_overlaps_beta=1.2}
\end{figure}

When $P, P^*$ are arbitrary finite numbers and $\mathcal{Q} = \mathbf{I}$, we search for solutions of Eqs. (\ref{eq:RBM_saddle-point}) by making the ansatz
\begin{equation}
\label{eq:no_correlations_ansatz}
\begin{aligned}
    &m^{\mu \nu} = \delta_{\mu \nu} m, &\quad &\hat{m}^{\mu \nu} = \delta_{\mu \nu} \hat{m}, \\
    &s^{\mu \nu} = \delta_{\mu \nu}, &\quad &\hat{s}^{\mu \nu} = 0, \\
    &q^{\mu \nu}
    = \begin{cases}
        \delta_{\mu \nu} q &\quad \mu, \nu \leq P^* \\
        \delta_{\mu \nu} g &\quad \text{otherwise},
    \end{cases} &\quad &\hat{q}^{\mu \nu}
    = \begin{cases}
        \delta_{\mu \nu} \hat{q} &\quad \mu, \nu \leq P^* \\
        \delta_{\mu \nu} \hat{g} &\quad \text{otherwise},
    \end{cases}
\end{aligned}
\end{equation}
which describes the situation where the student learns the teacher patterns one-to-one. Following the nomenclature introduced in \cite{hou2019minimal}, we will call Eq. (\ref{eq:no_correlations_ansatz}) the Permutation Symmetry Breaking (PSB) ansatz. We have assumed without loss of generality that the $\min \left\{ P, P^* \right\}$ nonzero magnetizations are on the main diagonal, as any hidden unit permutation would give an equivalent solution. According to this ansatz, the first $\min \left\{ P, P^* \right\}$ student patterns converge one-to-one to the first $\min \left\{ P, P^* \right\}$ teacher patterns with magnetization $m$ and SG overlap $q$. When $P>P^*$, the remaining $P - P^*$ student patterns (i.e. $P\geq\nu>P^*$) are aligned in spurious directions (i.e. $m^{\mu\nu}=0\  \forall \mu=1,...,P^*$) with a different SG overlap $g \neq q$. Conversely, when $P<P^*$, $P^*-P$ of the teacher patterns are not learned for lack of student hidden units. The latter case is described in terms of only $m$ and $q$, and so is $P = P^*$.
Under the PSB ansatz, Eqs. (\ref{eq:RBM_saddle-point}) decouple into $P$ independent systems of equations for the $P$ hidden units $\nu = 1, ..., P$ (see \ref{app:no_correlations}). In other words, the student factorizes into $P$ students with one hidden unit each. The systems of equations for the first $\min \left\{ P, P^* \right\}$ patterns are all identical to each other and equivalent to Eqs. (\ref{eq:reduced_saddle-point_P1}). When $P>P^*$, the $P - P^*$ remaining systems of equations all take the form
\begin{align} 
    \label{eq:reduced_saddle-point}
    g &= \mathbb{E}_{z} \left[ \tanh^2 \left( \sqrt{\hat{g}} z \right) \right]\\
    \hat{g} &= \beta^2 \alpha \ \mathbb{E}_{z} \left[ \tanh^2 \left( \beta \sqrt{g} z \right) \right].\nonumber
\end{align}
In summary, Eqs. (\ref{eq:RBM_saddle-point}) with binary patterns and $\mathcal{Q} = \mathbf{I}$ reduce to Eqs. (\ref{eq:reduced_saddle-point_P1}, \ref{eq:reduced_saddle-point}) under the PSB ansatz. In particular, the solution of Eqs. (\ref{eq:reduced_saddle-point_P1}, \ref{eq:reduced_saddle-point}), which we will call the PSB solution, has $m = q$ even in the region outside the Nishimori line where $\beta = \beta^*$ and $P \neq P^*$. In Fig. (\ref{fig:BM_uncorrelated_overlaps_beta=1.2}), we verify that the PSB solution can be found by iterating the original saddle-point equations (Eqs. \ref{eq:RBM_saddle-point}) with binary student patterns (see \ref{sec:binary_random}), $\mathcal{Q} = \mathbf{I}$ and near-diagonal initial conditions $m^{\mu \nu} = \delta_{\mu \nu} m_0 + \left( 1 - \delta_{\mu \nu} \right) \varepsilon$, where $\varepsilon \ll m_0$. In other words, we verify that the solution of Eqs. (\ref{eq:reduced_saddle-point_P1}, \ref{eq:reduced_saddle-point}) is stable for the full saddle-point iteration. We plot only the case of $P \geq P^*$ because $P < P^*$ yields similar results.

\begin{figure}
    \centering
    \includegraphics[width = \textwidth]{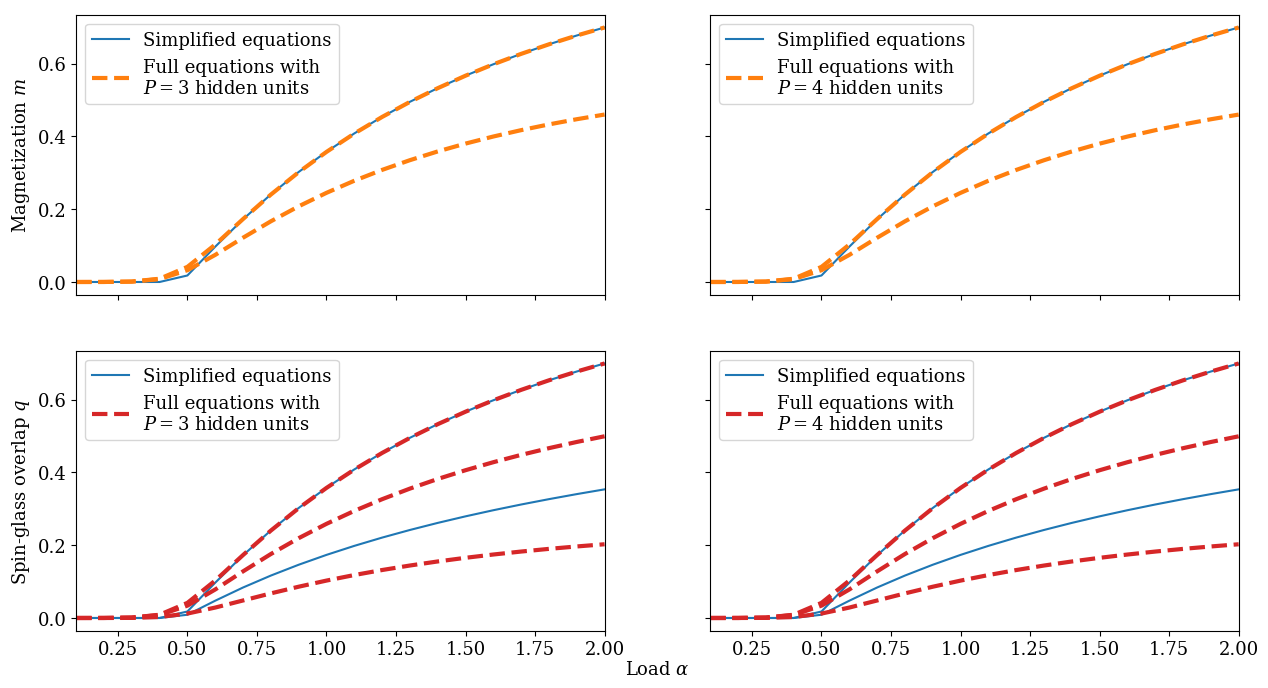}
    \caption{Partial Permutation Symmetry Breaking (partial PSB) solutions of Eqs. (\ref{eq:RBM_saddle-point}) for binary student patterns with a uniform prior and binary teacher patterns with covariance $\mathcal{Q} = \mathbf{I}$, in red and orange, compared against the solution of Eqs. (\ref{eq:reduced_saddle-point_P1}, \ref{eq:reduced_saddle-point}), in blue. We plot the Mattis magnetization $m$ in the top row, and the SG overlap $q$ in the bottom row. The top lines of the plots show that the student patterns $\xi^{\mu}_{\text{PSB}}$ that converge to teacher patterns one-to-one have the same $m$ and $q$ as the solution of Eqs. (\ref{eq:reduced_saddle-point_P1}), and thus also satisfy $m = q$. Conversely, the other lines show that the student patterns $\xi^{\mu}_{\text{PS}}$ that converge to a common teacher pattern have a smaller $m$ and a different $q$. To be more precise, the central and bottom branches of $q$ are the spin-glass order parameters corresponding to
    $Q(\xi^{1 \mu}_{PS},\xi^{2 \mu}_{PS})$ and $Q(\xi^{1 \mu}_{PS},\xi^{2 \nu}_{PS})$ with $\mu \neq \nu$, respectively (see Section \ref{sec:RBM_teacher-student}). They are both different from the $g$ of Eq. (\ref{eq:reduced_saddle-point}). The Mattis magnetization and SG overlaps omitted from this Figure all vanish. We use $P = 3$ and $P^* = 2$ in the left column and $P = 4$ and $P^* = 3$ in the right column. All plots have $\beta^* = \beta = 1.2$.}
    \label{fig:BM_unstructured_overlaps_beta=1.2}
\end{figure}

\begin{figure}
    \centering
    \includegraphics[width = \textwidth]{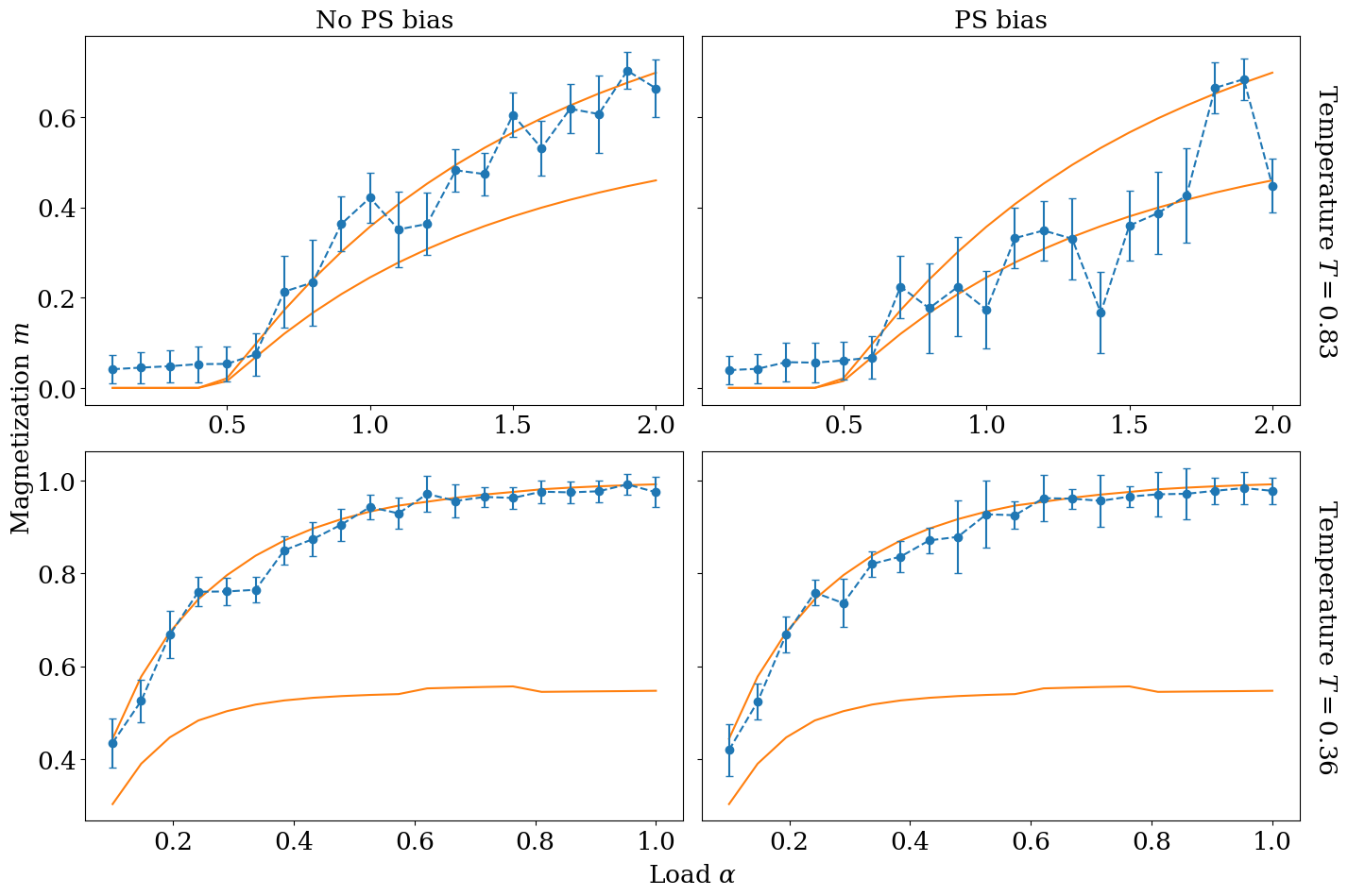}
    \caption{Solutions of Eqs. (\ref{eq:RBM_saddle-point}) shown in Fig. (\ref{fig:BM_unstructured_overlaps_beta=1.2}), in orange, compared against $N = 512$ dimensional Monte Carlo simulations, in blue, of the teacher-student problem where the student has $P = 2$ binary patterns with a uniform prior and the teacher has a single binary pattern. The blue dots and error bars represent the means and standard deviations, respectively, of a single diagonal coefficient of $m$ during the simulations. The inverse temperature is set to $\beta^* = \beta = 1.2 \approx 1/0.83$ and $\beta^* = \beta = 2.8 \approx 1/0.36$ in the top and bottom rows, respectively.
    In the left column, the student with $T^* = T = 0.83$ is biased towards the PSB solution by near-diagonal initial conditions, while the student with $T^* = T = 0.36$ has random initial conditions.
    The students of the right column have off-diagonal initial conditions.
    The top left and top right students are also biased by a small external field pointing towards the PSB and partial PSB solutions, respectively. The bottom students are not biased by an external field.
    }
    \label{fig:BM_simulated_magnetization_P=1_P_t=2}
\end{figure}

\begin{figure}[h!]
    \centering
    \includegraphics[width=0.75\linewidth]{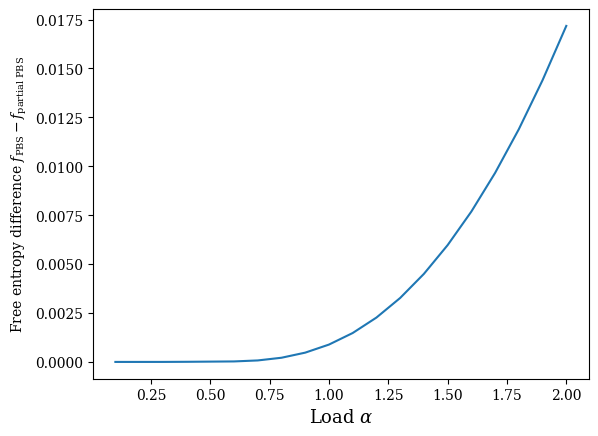}
    \caption{Free entropy difference of the so-called PSB and partial PSB solutions of Eqs. (\ref{eq:RBM_saddle-point}) shown in Figs. (\ref{fig:BM_uncorrelated_overlaps_beta=1.2}) and (\ref{fig:BM_unstructured_overlaps_beta=1.2}). This plot was made using $P^* = 2$ and $P = 3$, but the free entropy of $P^* = 3$ and $P = 4$ looks identical.}
    \label{fig:free_entropy_difference}
\end{figure}

The original saddle-point equations (Eq. \ref{eq:RBM_saddle-point}) with binary patterns, $\mathcal{Q} = \mathbf{I}$ and $P > P^*$ also have other solutions that cannot be expressed as cleanly by a set of simplified saddle-point equations (see Fig. \ref{fig:BM_unstructured_overlaps_beta=1.2}).
These solutions can be found by initializing $m^{ \mu\nu}$ with $m_0 \gg \varepsilon$ on the diagonal and at least one off-diagonal entry close to $m_0$, a process we will refer to as off-diagonal initialization. They correspond to a distributed representation where some of the student patterns learn the same teacher pattern, so we dub these solutions partial PSB. Throughout this paper and its figures, we focus on the case where at most two student patterns learn the same teacher pattern, but larger numbers are also possible. Within a partial PSB solution, some student patterns may still learn teacher patterns one-to-one. These student patterns $\xi^{\mu}_{\text{PSB}}$ have the same $m$ and $q$ as Eq. (\ref{eq:reduced_saddle-point_P1}), and in particular satisfy $m = q$ (see Fig. \ref{fig:BM_unstructured_overlaps_beta=1.2}). Comparison of the RS free entropy (Eq. \ref{eq:RBM_free_entropy}) of the PSB and partial PSB solutions of Eqs. (\ref{eq:RBM_saddle-point}) suggests that the former is always favored in the limit $N,M\to\infty$. The free entropy difference between them decreases with increasing $T = T^*$ and vanishes at the onset of the paramagnetic phase (see Fig. \ref{fig:free_entropy_difference}).
These results are confirmed by numerical simulations. In low $T = T^*$ simulations, students with binary patterns always converge to the PSB solution, even when we use random initial conditions (see Figs. \ref{fig:BM_simulated_magnetization_P=1_P_t=2}), which is solid evidence that PSB is favored at low $T = T^*$. When $T = T^*$ is relatively close to the paramagnetic phase, the simulations have relatively large error bars and repeatedly jump between the different solutions rather than converging to a single mode. This kind of instability can be attributed to finite-size fluctuations.


\begin{figure}
    \centering
    \includegraphics[width = 0.495\textwidth]{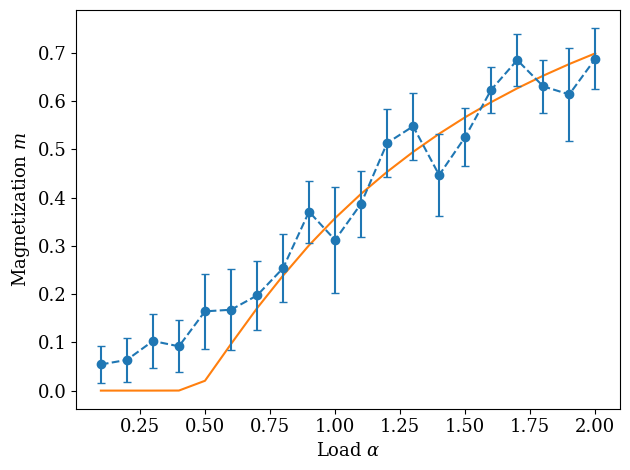}
    \caption{The magnetization $m$ solving Eqs. (\ref{eq:reduced_saddle-point}), in orange, compared against $N = 512$ dimensional Monte Carlo simulations, in blue, of the teacher-student problem where the student has $P = 2$ binary patterns with a uniform prior and the teacher has $P^* = P = 2$ binary patterns with covariance $\mathcal{Q} = \mathbf{I}$. The blue dots and error bars represent the means and standard deviations, respectively, of the diagonal of the magnetization $m$ during the simulations. The inverse temperature is set to $\beta^* = \beta = 1.2$, and the simulations have a small external field biasing the student towards the PSB solution.}
    \label{fig:BM_simulated_magnetization}
\end{figure}

The PSB solution is independent of $P^*$ and $P$, which means that, in terms of the student's performance $m$, learning $P^*$ i.i.d. patterns stored in a single teacher RBM is as difficult as learning $P^*$ i.i.d. patterns from $P^*$ separate RBMs with one hidden unit each. Such independence of the student's performance from the number of hidden units was first conjectured in \cite{barra2017phase} based on empirical observations.
The performance on the Nishimori line was then shown to be the same for $P = P^* = 2$ \cite{hou2019minimal} as for $P = P^* = 1$ \cite{huang2016unsupervised, huang2017statistical}. Our PSB solution extends this result to $\beta = \beta^*$ and any finite $P, P^*$. On the Nishimori line, i.e. when $\beta = \beta^*$ and $P = P^*$, the teacher-student setting is replica symmetric, so we expect the PSB solution to be exact, which is confirmed by simulations (see Fig. \ref{fig:BM_simulated_magnetization}).
When $\beta = \beta^*$ and $P \neq P^*$, replica symmetry is not guaranteed, but the PSB solution is still in good agreement with simulations (see Fig. \ref{fig:BM_simulated_magnetization_P=1_P_t=2}). Finally, we expect RSB corrections when $\beta \neq \beta^*$.
Interestingly, we observe a weaker form of independence from the number of hidden units in the suboptimal partial PSB solutions, in the sense that the partial PSB solutions that we find at given $P^*$ and $P$ are also present at larger $P^*$ and $P$. These results seem related to the embedding principle stating that a neural network contains all minima of narrower neural networks with the same architecture \cite{zhang2021embedding}.

\subsubsection{Independence of the number of hidden units: Gaussian patterns}
\label{sec:independence_g}
\begin{figure}
    \centering
    \includegraphics[width = 0.495\textwidth]{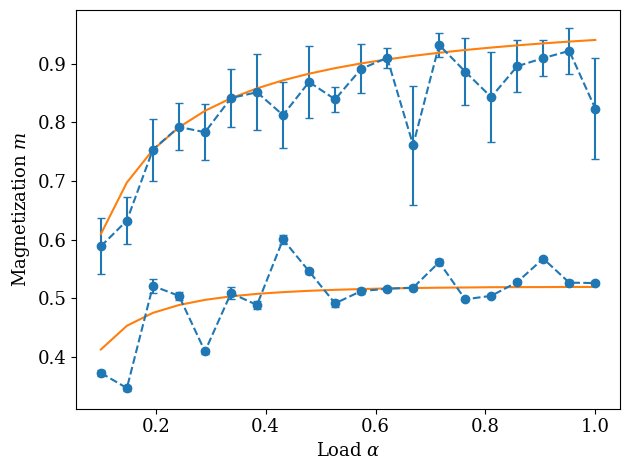}
    \caption{Solutions of Eqs. (\ref{eq:RBM_saddle-point}) for real-valued student patterns with a standard Gaussian prior and teacher pattern covariance $\mathcal{Q} = \mathbf{I}$, in orange,
    compared against $N = 512$ dimensional Monte Carlo simulations, in blue, of the teacher-student problem where the student has $P = 2$ real-valued patterns with a standard Gaussian prior and the teacher has a single Gaussian pattern. The top orange line is the magnetization $m$ of student patterns that converge to teacher patterns one-to-one, so it is also the $m$ solving Eqs. (\ref{eq:reduced_normal_saddle-point}). On the other hand, the bottom orange line is the $m$ of student patterns that converge to a common teacher pattern in the partial PSB solution. The blue dots and error bars represent the means and standard deviations, respectively, of the largest entry of the magnetization $m$ during the simulations. The top and bottom blue lines result from simulations with i.i.d. standard Gaussian initial conditions and off-diagonal initial conditions, respectively. The learning rate was reduced more quickly for the bottom line than for the top line. The inverse temperature is set to $\beta^* = \beta = 4$.}
    \label{fig:normal_BM_simulated_magnetization_P=1_P_t=2}
\end{figure}

\begin{figure}[h!]
    \centering
    \includegraphics[width = \textwidth]{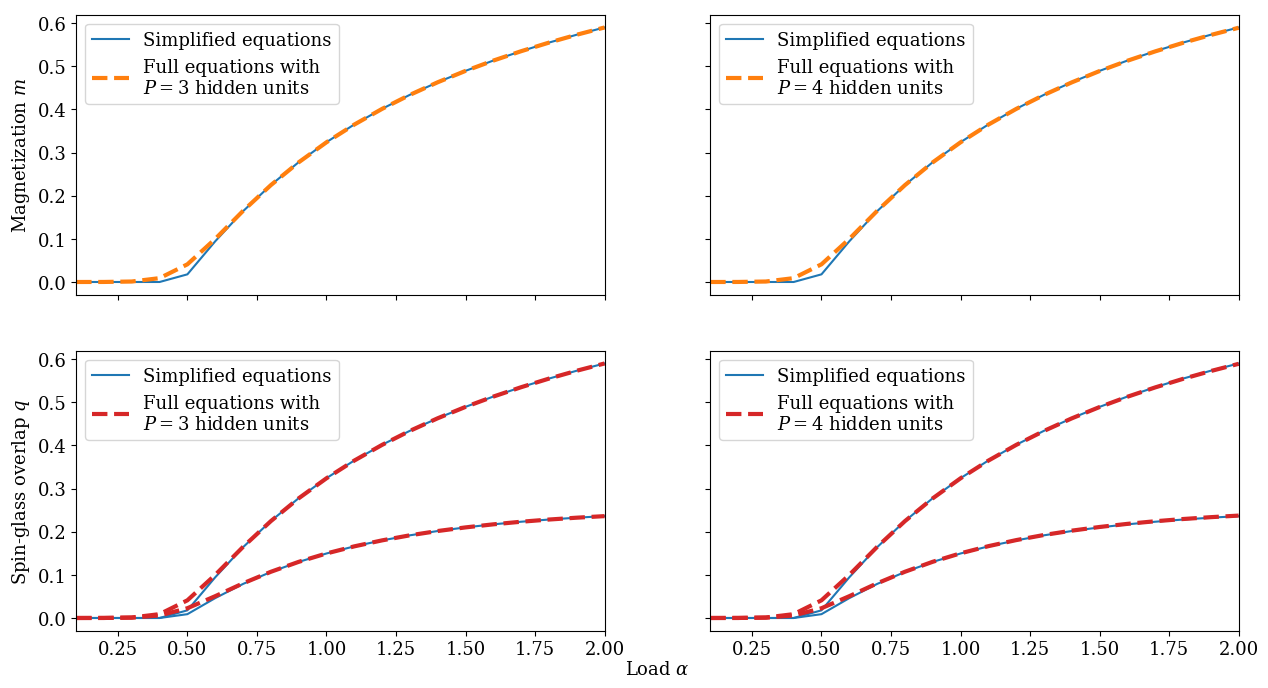}
    \caption{Permutation Symmetry Breaking (PSB) solution of Eqs. (\ref{eq:RBM_saddle-point}) for real-valued student patterns with a standard Gaussian prior and teacher pattern covariance $\mathcal{Q} = \mathbf{I}$, in red and orange, compared against the solution of Eqs. (\ref{eq:reduced_normal_saddle-point}), in blue. We plot the Mattis magnetization $m$ in the top row, and the SG overlap $q$ in the bottom row. The magnetization plots and the top lines of the SG overlap plots show that the student patterns that converge to teacher patterns have the same $m$ and $q$ as the solution of Eqs. (\ref{eq:reduced_normal_saddle-point}), and thus also satisfy $m = q$. Conversely, the bottom lines of the SG overlap plots student patterns that do not converge to a teacher pattern have a spin-glass overlap of $g$ as in Eqs. (\ref{eq:reduced_normal_saddle-point}). The top branch of $q$ is  We use $P = 3$ and $P^* = 2$ in the left column and $P = 4$ and $P^* = 3$ in the right column. All plots have $\beta^* = \beta = 1.2$.}
    \label{fig:normal_BM_uncorrelated_overlaps_beta=1.2}
\end{figure}

\begin{figure}
    \centering
    \includegraphics[width = \textwidth]{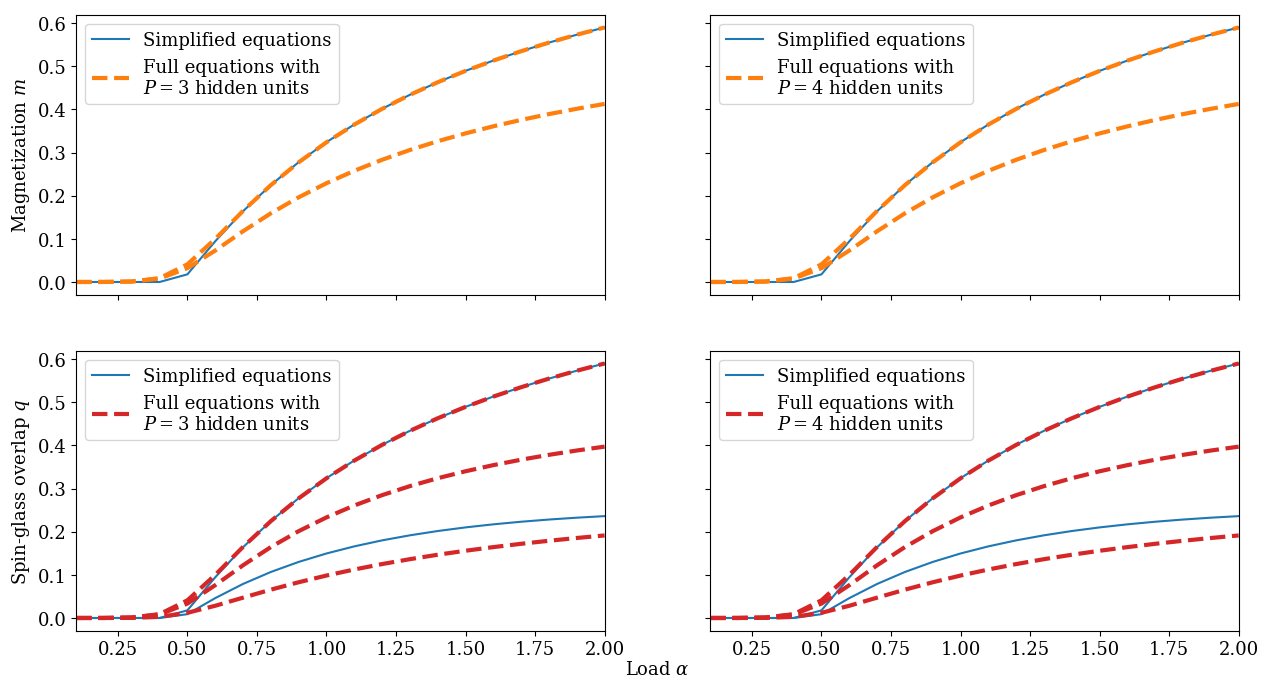}
    \caption{Partial Permutation Symmetry Breaking (partial PSB) solutions of Eqs. (\ref{eq:RBM_saddle-point}) for real-valued student patterns with a standard Gaussian prior and teacher pattern covariance $\mathcal{Q} = \mathbf{I}$, in red and orange, compared against the solution of Eqs. (\ref{eq:reduced_normal_saddle-point}), in blue. We plot the Mattis magnetization $m$ in the top row, and the SG overlap $g$ in the bottom row. The top lines of the plots show that the student patterns $\xi^{\mu}_{\text{PSB}}$ that converge to teacher patterns one-to-one have the same $m$ and $q$ as the solution of Eqs. (\ref{eq:reduced_normal_saddle-point}), and thus also satisfy $m = q$. Conversely, the other lines show that the student patterns $\xi^{\mu}_{\text{PS}}$ that converge to a common teacher pattern have a smaller $m$ and a different $q$. To be more precise, the central and bottom branches of $q$ are the spin-glass order parameters corresponding to $Q(\xi^{1 \mu}_{PS},\xi^{2 \mu}_{PS})$ and $Q(\xi^{1 \mu}_{PS},\xi^{2 \nu}_{PS})$ with $\mu \neq \nu$, respectively (see Section \ref{sec:RBM_teacher-student}). They are both different from the $g$ of Eq. (\ref{eq:reduced_normal_saddle-point}). The Mattis magnetization and SG overlaps omitted from this Figure all vanish. We use $P = 3$ and $P^* = 2$ in the left column and $P = 4$ and $P^* = 3$ in the right column. All plots have $\beta^* = \beta = 1.2$.}
    \label{fig:normal_BM_unstructured_overlaps_beta=1.2}
\end{figure}

When the student patterns $\xi$ are real-valued variables with an i.i.d. standard Gaussian prior, Eqs. (\ref{eq:RBM_saddle-point}) under the PSB ansatz (Eqs. \ref{eq:no_correlations_ansatz}) simplify to (see \ref{app:no_correlations})
\begin{align*}
    \label{eq:reduced_normal_saddle-point}
    m &= \frac{\hat{m}}{1 + \hat{q}} \\
    q &= \frac{\hat{m}^2}{\left( 1 + \hat{q} \right)^2} + \frac{\hat{q}}{\left( 1 + \hat{q} \right)^2} \\
    g &= \frac{\hat{g}}{\left( 1 + \hat{g} \right)^2} \numberthis \\
    \hat{m} &= \beta^* \beta \alpha \ \mathbb{E}_{z} \left[ \tanh \left( \beta^* \beta m + \beta \sqrt{q} z \right) \right] \\
    \hat{q} &= \beta^2 \alpha \ \mathbb{E}_{z} \left[ \tanh^2 \left( \beta^2 m + \beta \sqrt{q} z \right) \right] \\
    \hat{g} &= \beta^2 \alpha \ \mathbb{E}_{z} \left[ \tanh^2 \left( \beta \sqrt{g} z \right) \right].
\end{align*}
This regime is interesting because RBMs used in practical applications usually have continous weights \cite{kiviken2012multiple, Salakhutdinov2007restricted, srivastava2013modelling}.
Similarly as before, we compare Eqs. (\ref{eq:reduced_normal_saddle-point}) against the PSB and partial PSB solutions of Eqs. (\ref{eq:RBM_saddle-point}) for real-valued student patterns with an i.i.d. standard Gaussian prior and teacher pattern covariance $\mathcal{Q} = \mathbf{I}$.
The resulting plots are qualitatively similar to Figs. (\ref{fig:BM_uncorrelated_overlaps_beta=1.2}) and (\ref{fig:BM_unstructured_overlaps_beta=1.2}), so we show them in \ref{app:supplementary_figures} rather than in the main text (see Figs. \ref{fig:normal_BM_uncorrelated_overlaps_beta=1.2} and \ref{fig:normal_BM_unstructured_overlaps_beta=1.2}). As before, we observe that the solutions of the Gaussian equations are independent of $P^*$ and $P$. One important difference between the student model with binary patterns and the student model with real-valued patterns is that they are simulated differently. In fact, binary patterns are obtained using the standard random walk Metropolis-Hastings algorithm, while real-valued patterns are learned via underdamped stochastic Langevin dynamics \cite{welling2011bayesian, zhang2023improved} of the RBM marginal likelihood (Eq. \ref{eq:RBM_direct_distribution}) gradient. In a nutshell, the latter algorithm is a variant of stochastic gradient ascent where noise is added to the momentum vector to sample the RBM posterior (Eq. \ref{eq:inverse_distribution}) rather than to find one of its modes. We use contrastive divergence, i.e. alternate Gibbs sampling of the visible and hidden units, to evaluate the marginal likelihood \cite{hinton2002training, hinton2012practical}.
At low temperature, simulations of the student model with real-valued patterns usually converge to the solution of Eqs. (\ref{eq:reduced_normal_saddle-point}) (see Fig. \ref{fig:normal_BM_simulated_magnetization_P=1_P_t=2}). They can also sometimes stay stuck near the partial PSB solution without converging to it when the learning rate is reduced too aggressively during training. This behavior is shown in Fig. (\ref{fig:normal_BM_simulated_magnetization_P=1_P_t=2}) by the simulation data points that are scattered around the partial PSB solution but do not agree with it within their error bars. In fact, although simulation error is useful for measuring the convergence of the Langevin learning algorithm, its sensitivity to the learning rate schedule makes it poorly representative of the equilibrium distribution of the magnetization $m$.

\subsubsection{A simple model of the lottery ticket hypothesis}
\label{sec:lottery}
Many works studying feedforward neural networks found a training regime where some of the hidden units learn the underlying data distribution while the others take a back seat \cite{chizat2018global, ma2020quenching, luo2021phase, zhang2021embedding}. Interestingly, this behavior is similar to our PSB solution with $P > P^*$, where only a subset of the student patterns learn the teacher patterns (see Sections \ref{sec:independence} and \ref{sec:independence_g}). As proposed in \cite{zhang2021embedding}, we investigate the relationship between this kind of training regime and the lottery ticket hypothesis \cite{frankle2018lottery}.

\begin{figure}
    \centering
    \includegraphics[width = \textwidth]{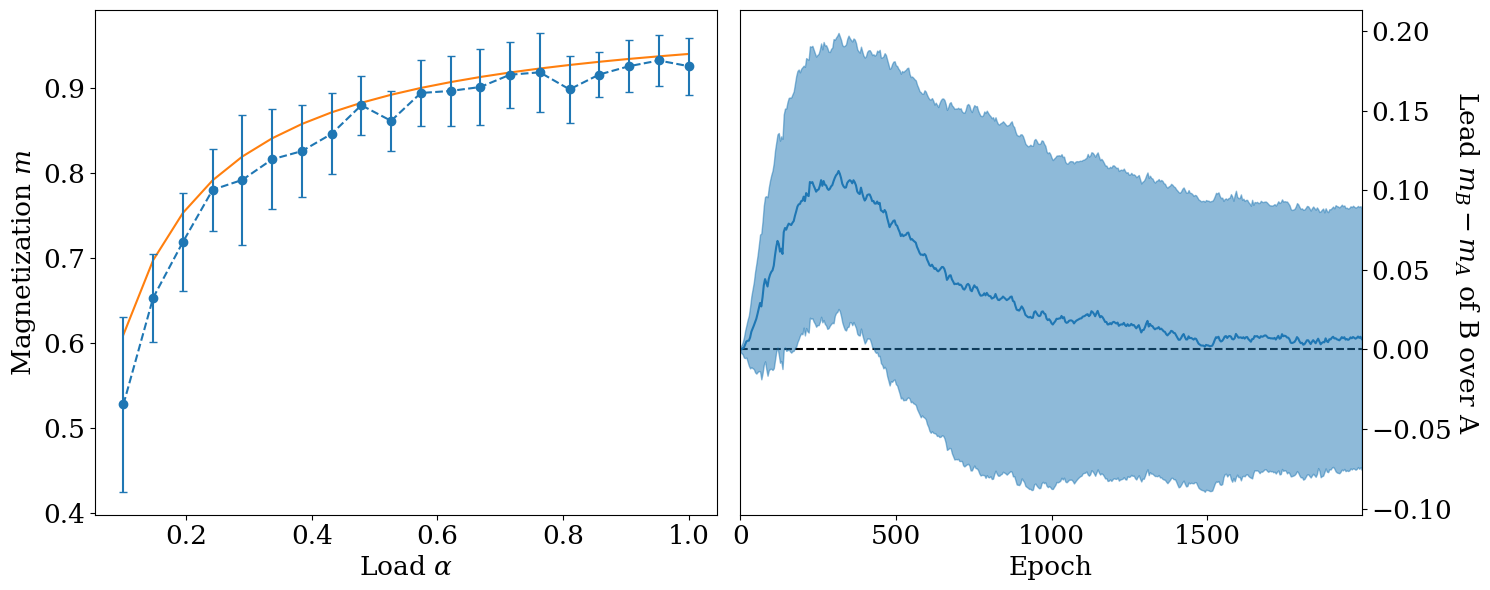}
    \caption{Results of the lottery ticket experiment described in Section \ref{sec:lottery}. In the left panel, $N = 512$ dimensional Monte Carlo simulations of student B, in blue, are compared against the solution of Eqs. (\ref{eq:reduced_normal_saddle-point}), in orange. The blue dots and error bars represent the means and standard deviations, respectively, of the diagonal of the magnetization $m$ during the simulations. The right panel shows the difference $m_B - m_A$ of the magnetizations of A and B as a function of the simulation epochs. The solid blue line and the shaded region represent the median of $m_A - m_B$ over $\alpha \in \left[ 0, 1 \right]$ and the corresponding mean absolute deviation around the median, respectively. $m_A - m_B$ goes to zero when the number of elapsed epochs is large, so student A converges to the solution of Eqs. (\ref{eq:reduced_normal_saddle-point}) like student B. The inverse temperature is set to $\beta^* = \beta = 4$.
    }
    \label{fig:winning_ticket_lead}
\end{figure}

The lottery ticket hypothesis states that a generic randomly initialized overparameterized neural network contains subnetworks that fit data with similar accuracy as the entire trained network when they are extracted from it and trained independently \cite{frankle2018lottery, zhou2019deconstructing, ramanujan2020hidden, eran2020proving, frankle2020linear}. These special subnetworks, which are commonly referred to as winning lottery tickets, are thought to have fortuitous initial conditions that facilitate training \cite{frankle2018lottery}.
The PSB solution studied in Sections \ref{sec:independence} and \ref{sec:independence_g} makes it clear that any student network of size $P > P^*$ contains subnetworks of size $\Tilde{P} \in \left\{ P^*, ..., P - 1 \right\}$ that learn the teacher patterns to the same extent, and thus fit the data at least as well.
It is not obvious, however, whether any of these subnetworks can have lucky initial conditions such that they train more easily than with i.i.d. random initial conditions. As such, we apply to our teacher-student setting a variant of the magnitude pruning algorithm traditionally used to identify winning tickets \cite{frankle2018lottery, frankle2020linear} and check if it finds a subnetwork that converges especially quickly. Consider one teacher and three students with real-valued patterns, hereby referred to as \textit{the teacher}, \textit{student 0}, \textit{student A} and \textit{student B}, respectively, where A is a control network and B is a candidate winning ticket obtained from 0. We perform the following numerical experiment:
\begin{itemize}
    \item Initialize student 0 with $P = 8$ i.i.d. Gaussian patterns and the teacher with $P^* = 4$ i.i.d. Gaussian patterns. Save the initial value of the patterns of student $0$ as $\xi_0$.
    \item Train student 0 on data generated by the teacher. Compute the Euclidean norms of the learned patterns.
    \item Initialize student A with $P = 4$ i.i.d. Gaussian patterns and student B with the $P = 4$ patterns of $\xi_0$ that evolved to have the largest Euclidean norm when trained.
    \item Train student A and student B on data generated by the teacher. Record their respective magnetizations $m_A$ and $m_B$ as a function of the training epochs in order to compare their convergence speeds.
\end{itemize}
This procedure is slightly different from the usual magnitude pruning algorithm in that it prunes the patterns with the smallest norms rather than the entries of the weight matrix with the smallest absolute values \cite{frankle2018lottery, frankle2020linear}. As in Section \ref{sec:independence_g}, we train students 0, A and B with underdamped stochastic Langevin dynamics \cite{welling2011bayesian, zhang2023improved} of the RBM marginal likelihood (Eq. \ref{eq:RBM_direct_distribution}).
Our results are shown in Fig. (\ref{fig:winning_ticket_lead}). In the left panel, we verify that student B converges to the PSB solution (i.e. the solution of Eqs. \ref{eq:reduced_normal_saddle-point}) in the range of $\alpha$ that we are studying. As in Section \ref{sec:independence_g}, the error in $m$ is useful for measuring the convergence of the learning algorithm, but poorly representative of the equilibrium distribution.
There seems to be a small systematic shift in the simulations at low $\alpha$, which could be due to finite-size effects or order $\frac{P}{M}$ corrections \cite{barra2017phase}. In the right panel, we plot the median of $m_B - m_A$ over $\alpha$ as a function of the epochs. B initially converges more quickly than A, and the lead of B eventually shrinks to zero because of the ergodicity of the training algorithm. In other words, magnitude pruning is able to identify a winning ticket that converges more quickly than networks with i.i.d. random initial conditions.
The patterns of student 0 that converge to the teacher patterns the fastest have a small initial magnetization (see Fig. \ref{fig:winning_ticket_lead}, epoch 0). Therefore, the shape of the loss function at the initial conditions and the basin of attraction in which they are located may be more important than the initial conditions themselves, as suggested by \cite{frankle2018lottery}.

\subsection{Learning uniformly correlated  patterns}
\label{sec:uniform_correlations}
\begin{figure}
    \centering
    \includegraphics[width = \textwidth]{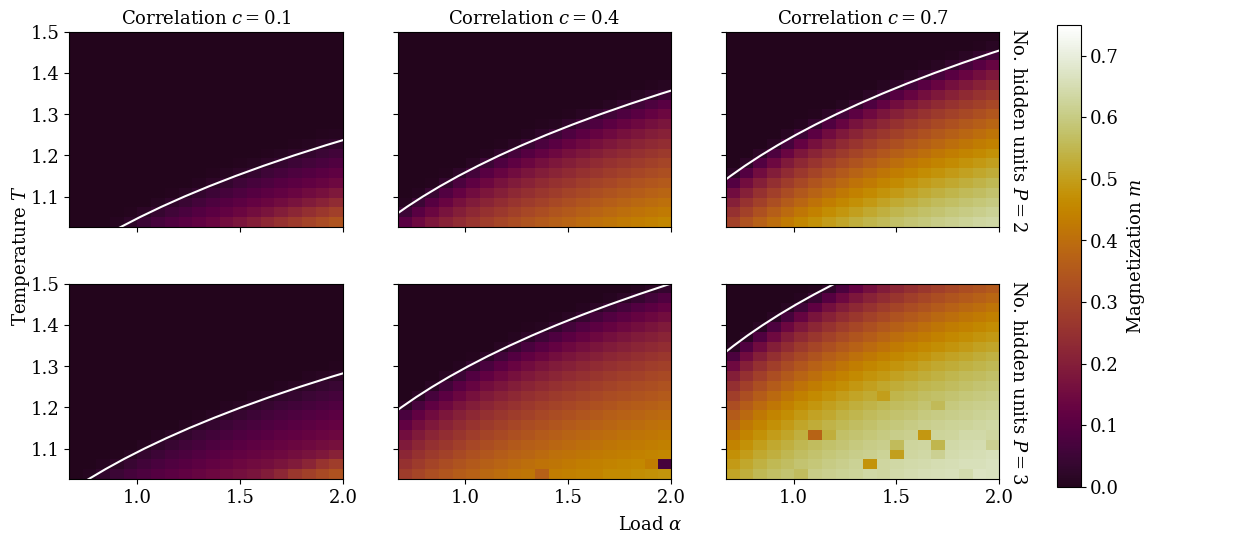}
    \caption{Mattis magnetization $m$ for $\beta = \beta^*$ and $P = P^*$ as a function of the number of hidden units $P$, the correlation $c$, the temperature $T$ and the data load $\alpha$. $m$ is obtained by solving Eqs. (\ref{eq:RBM_saddle-point}) numerically for binary student patterns with a uniform prior and binary teacher patterns with covariance $\mathcal{Q}_{\mu \nu} = \delta_{\mu \nu} + \left( 1 - \delta_{\mu \nu} \right) c$, where $c \in \left[ 0, 1 \right)$ (see \ref{app:uniform_correlations}). The top and bottom rows feature $P = 2$ and $P = 3$, respectively. The white lines mark the phase transition described by Eq. (\ref{eq:critical_load}) with $\lambda^{\mathcal{S}}_{\text{max}}$ given by Eq. (\ref{eq:max_eigval}).
    The speckles in the plots with $P = 3$, $c = 0.4$ and $P = 3$, $c = 0.7$ are due to the occasional numerical instability of the saddle-point equations (see \ref{app:numerical_methods}).}
    \label{fig:RBM_phase_diagram}
\end{figure}

\begin{figure}
    \centering
    \includegraphics[width = \textwidth]{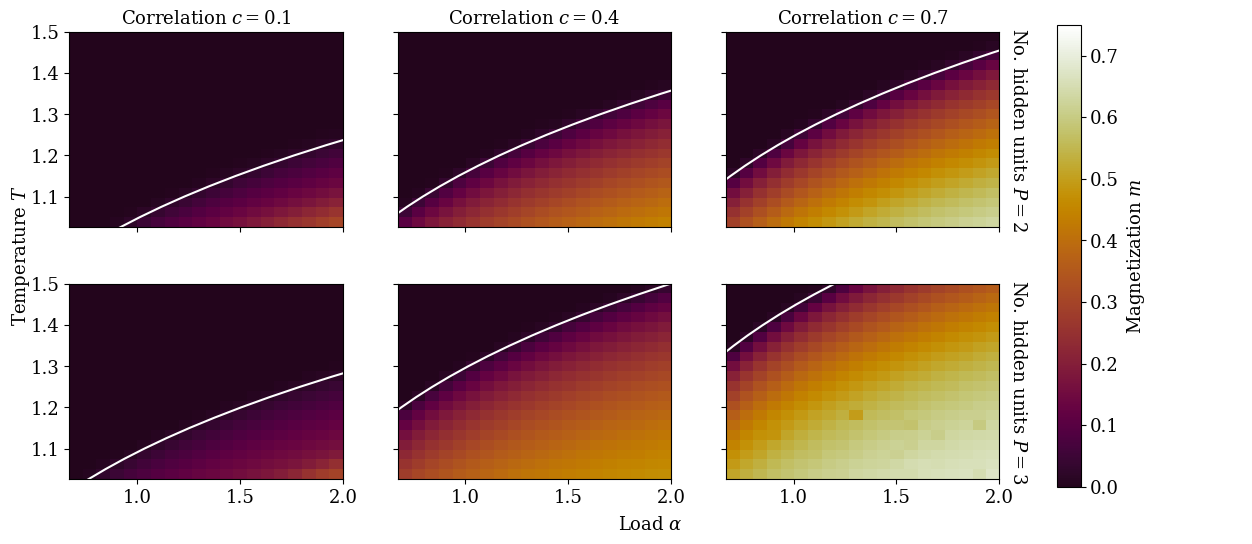}
    \caption{Mattis magnetization $m$ for $\beta = \beta^*$ and $P = P^*$ as a function of the number of hidden units $P$, the correlation $c$, the temperature $T$ and the data load $\alpha$. $m$ is obtained by solving Eqs. (\ref{eq:RBM_saddle-point}) numerically for real-valued student patterns with a standard Gaussian prior and teacher pattern covariance $\mathcal{Q}_{\mu \nu} = \delta_{\mu \nu} + \left( 1 - \delta_{\mu \nu} \right) c$, where $c \in \left[ 0, 1 \right)$ (see \ref{app:uniform_correlations}). The top and bottom rows feature $P = 2$ and $P = 3$, respectively. The white lines mark the phase transition of Eq. (\ref{eq:critical_load}) with $\lambda^{\mathcal{S}}_{\text{max}}$ given by Eq. (\ref{eq:max_eigval}).
    The speckles in the plots with $P = 3$, $c = 0.4$ and $P = 3$, $c = 0.7$ are due to the saddle-point iteration converging to a different solution of Eqs. (\ref{eq:normal_saddle-point}) than at neighboring $\alpha$ and $T$.
    }
    \label{fig:normal_phase_diagram}
\end{figure}

In this Section, including Subsection \ref{sec:concentration}, we introduce uniform correlations in the teacher patterns by fixing the covariance matrix $\mathcal{Q}$ of $\xi^*$ to $\mathcal{Q}_{\mu \nu} = \delta_{\mu \nu} + \left( 1 - \delta_{\mu \nu} \right) c$, where $c \in \left[ 0, 1 \right)$ controls the correlation strength.
In the presence of uniform correlation, the Hamiltonian $\mathcal{M}_*$ (Eq. \ref{hamiltonian_M_s}) is that of the Curie-Weiss model with coupling constant $\frac{1}{2} \left[ \beta^* \right]^2 c$. Therefore, its correlation matrix $\mathcal{R}$ has the same form as $\mathcal{Q}$ but with a different off-diagonal element $d$.
In \ref{app:uniform_correlations}, we show that the maximum eigenvalue $\lambda^{\mathcal{S}}_{\text{max}}$ of Eq. (\ref{eq:critical_load}) is
\begin{align}
    \label{eq:max_eigval}
    \lambda^{\mathcal{S}}_{\text{max}} = \left( P^* - 1 \right)^2 c d + \left( P^* - 1 \right) \left( c + d \right) + 1.
\end{align}
As expected, the corresponding critical load is also the onset of nonzero Mattis magnetization in the regime of $\beta=\beta^*$ where only the paramagnetic and ferromagnetic phases exist, which we show explicitly in Fig. (\ref{fig:RBM_phase_diagram}) for binary patterns. We show in \ref{app:supplementary_figures} that the same holds for real-valued student patterns with a standard Gaussian prior (see Fig. \ref{fig:normal_phase_diagram}).
Eq. (\ref{eq:max_eigval}) extends to arbitrary finite $P^*$ the critical load obtained for $P^* = 2$ in \cite{hou2019minimal}. In fact, when $P^* = 2$, we find $d = \tanh \left( \left[ \beta^* \right]^2 c \right)$ and Eq. (\ref{eq:critical_load}) reduces to the critical load of \cite{hou2019minimal}.
\begin{figure}
    \centering
    \includegraphics[width = \textwidth]{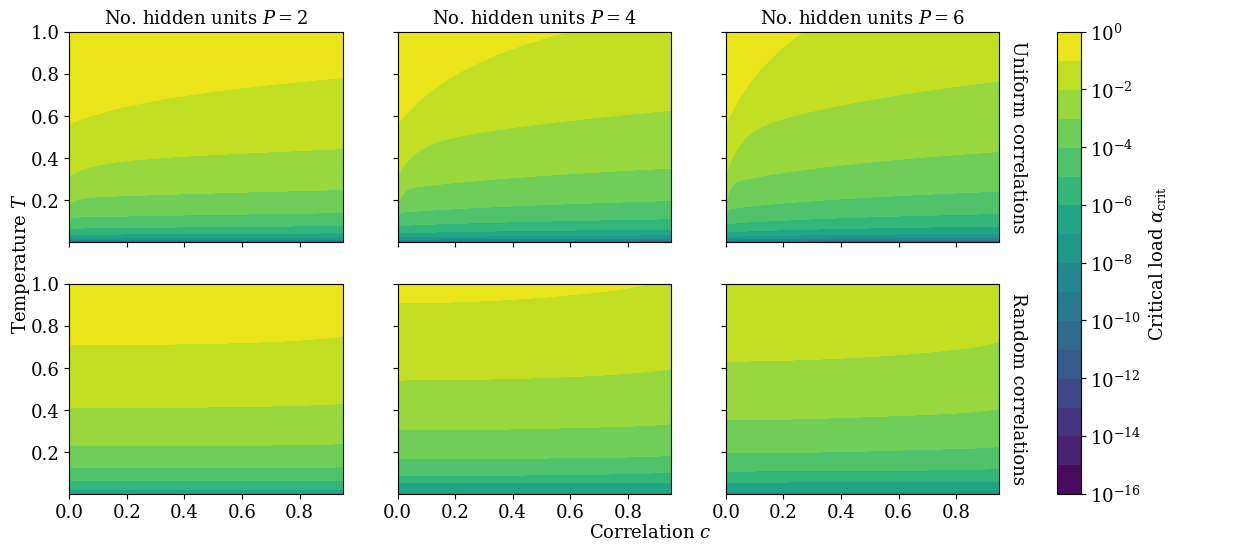}
    \caption{Critical load $\alpha_{\text{crit}}$ for $\beta = \beta^*$ and $P = P^*$ as a function of the number of hidden units $P$, the temperature $T$ and the correlation $c$. $\alpha_{\text{crit}}$ is obtained from Eq. (\ref{eq:critical_load}). The top row has $\mathcal{Q}_{\mu \nu} = \delta_{\mu \nu} + \left( 1 - \delta_{\mu \nu} \right) c$, so the max eigenvalue $\lambda^{\mathcal{S}}_{\text{max}}$ is that of Eq. (\ref{eq:max_eigval}). The bottom row is the arithmetic mean $\overline{\alpha_{\text{crit}}}$ over correlation matrices $\mathcal{Q}$ sampled from the projected Wishart distribution $\mathcal{W} \left( c, P \right)$ defined in \ref{sec:projected_wishart}.}
    \label{fig:alpha_crit}
\end{figure}
\begin{figure}
    \centering
    \includegraphics[width = \textwidth]{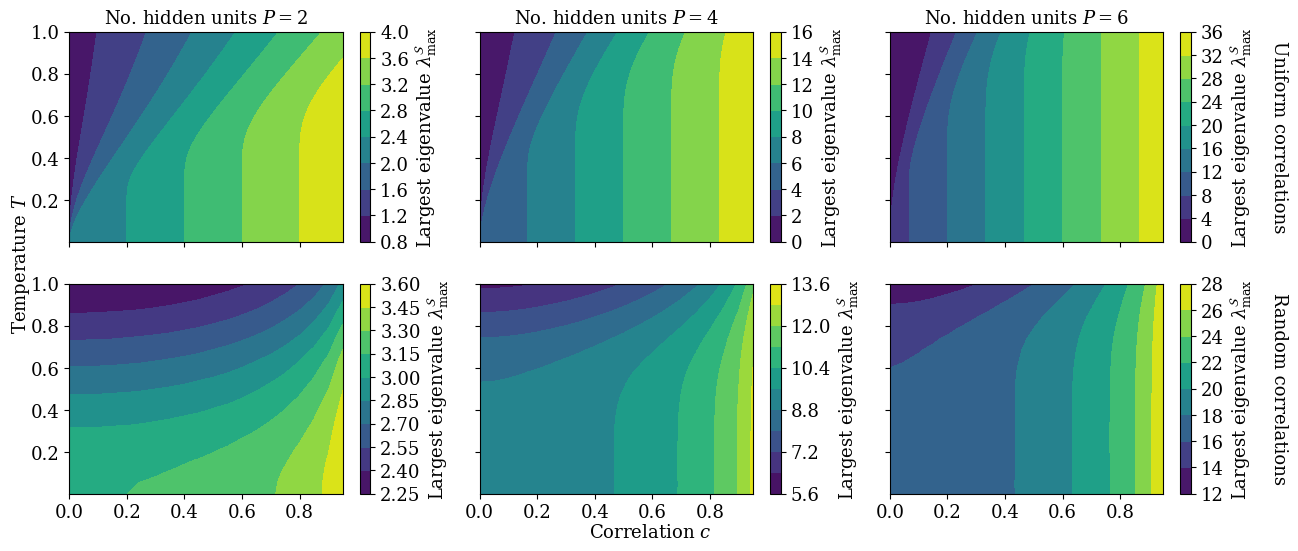}
    \caption{Largest eigenvalue $\lambda^{\mathcal{S}}_{\text{max}}$ of $\mathcal{S} = \mathcal{Q} \mathcal{R}$ (see \ref{app:critical_load}) for $\beta = \beta^*$ and $P = P^*$ as a function of the number of hidden units $P$, the temperature $T$ and the correlation $c$. The top row has $\mathcal{Q}_{\mu \nu} = \delta_{\mu \nu} + \left( 1 - \delta_{\mu \nu} \right) c$, so the max eigenvalue $\lambda^{\mathcal{S}}_{\text{max}}$ is that of Eq. (\ref{eq:max_eigval}). The bottom row is the harmonic mean $\left[ \overline{1 / \lambda^{\mathcal{S}}_{\text{max}}} \right]^{-1}$ over correlation matrices $\mathcal{Q}$ sampled from the projected Wishart distribution $\mathcal{W} \left( c, P \right)$ defined in \ref{sec:projected_wishart}.}
    \label{fig:eigval_range}
\end{figure}%
Plotting Eq. (\ref{eq:max_eigval}) (see Fig. \ref{fig:eigval_range}) and the corresponding $\alpha_{\text{crit}}$ (Fig. \ref{fig:alpha_crit}) as a function of $T$, $c$ and $P^*$, we see that $\alpha_{\text{crit}}$ decreases with $P^*$ and $c$.
For $c \gg \left[ T^* \right]^2$, the spins of Curie-Weiss Hamiltonian $\mathcal{M}_*$ all align, so their correlation is $d = 1$, and we obtain
\begin{align*}
    \lambda^{\mathcal{S}}_{\text{max}} = \left( \left( P^* - 1 \right) c + 1 \right) P^*.
\end{align*}
In this limit, $\alpha_{\text{crit}}$ is roughly inversely proportional to both $c$ and $\left[ P^* \right]^2$.
Conversely, for $c \ll \left[ T^* \right]^2$, the correlation of $\mathcal{M}_*$ is $d = 0$, and we obtain
\begin{align}
    \lambda^{\mathcal{S}}_{\text{max}} = \left( P^* - 1 \right) c + 1.
\end{align}
In this limit, $\alpha_{\text{crit}}$ decreases less quickly by a factor of $P^*$ than for $c \gg \left[ T^* \right]^2$. However, relatively small correlations can still significantly decrease $\alpha_{\text{crit}}$ when $P^*$ is large. In other words, the critical load benefits from structured data with many correlated underlying abstract concepts even when the correlation between the different concepts is rather small. Overall, these results shed light upon the way $\alpha_{\text{crit}}$ depends on $P^*$, which was previously unclear given that previous work \cite{hou2019minimal} focused on $P = P^* = 2$.

Neural networks are often regularized by feature decorrelation techniques that reduce or eliminate correlations in their inputs and hidden layers \cite{lecun1998efficient, ioffe2015batch, huangi2018decorrelated, zhang2021stochastic}.
Decorrelation is known to increase training speed \cite{ioffe2015batch, huangi2018decorrelated, zhang2021stochastic, wiesler2011convergence}. However, our work suggests that it can also increase the critical load as a drawback, which may hinder performance when the data is noisy (i.e. $\beta^*$ finite). \cite{galloway2019batch, fagbohungbe2022effect} observed that batch normalization \cite{ioffe2015batch}, which was motivated by decorrelation \cite{huangi2018decorrelated, lecun1998gradient}, makes neural networks less robust to noise. This effect could potentially be related to our findings.
Arguably, the main caveat to our analysis is that we assumed that $P^*$ was finite when deriving Eq. (\ref{eq:max_eigval}), so the critical load is probably different when $P^*$ is $\mathcal{O} \left( N \right)$ \cite{barra2017phase}.
\begin{figure}
    \centering
    \includegraphics[width = \textwidth]{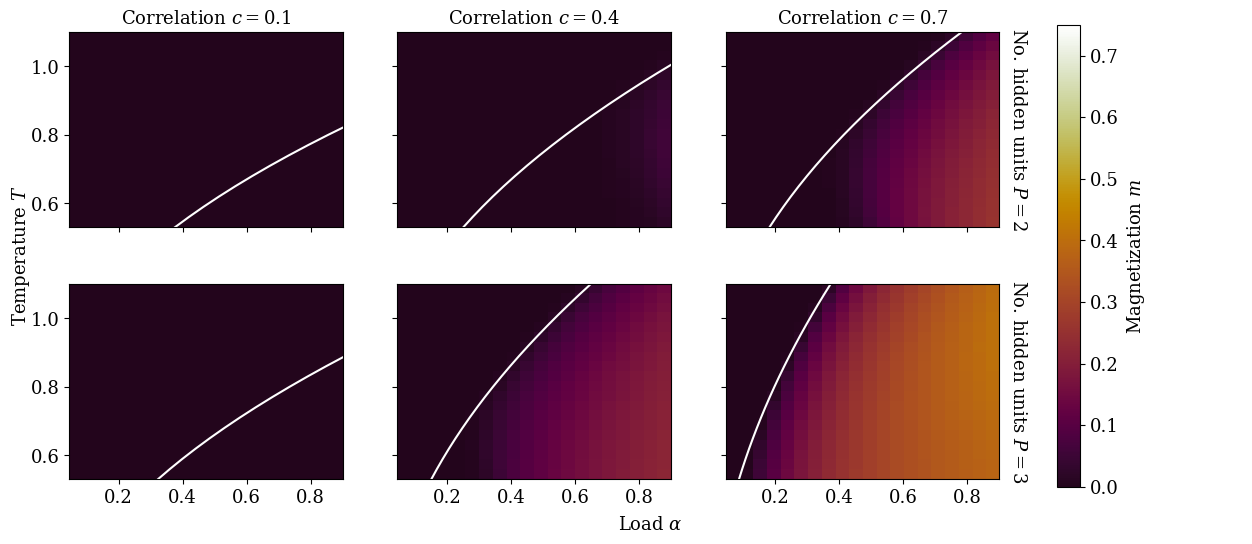}
    \caption{Mattis magnetization $m$ for $\beta^* = 0.8$ and $P = P^*$ as a function of the number of hidden units $P$, the correlation $c$, the temperature $T$ and the data load $\alpha$. $m$ is obtained by solving Eqs. (\ref{eq:RBM_saddle-point}) numerically for binary student patterns with a uniform prior and binary teacher patterns with covariance $\mathcal{Q}_{\mu \nu} = \delta_{\mu \nu} + \left( 1 - \delta_{\mu \nu} \right) c$, where $c \in \left[ 0, 1 \right)$ (see \ref{app:uniform_correlations}). The top and bottom rows feature $P = 2$ and $P = 3$, respectively. The white lines mark the critical load of Eq. (\ref{eq:critical_load}) with $\beta^* = 0.8$ and $\lambda^{\mathcal{S}}_{\text{max}}$ given by Eq. (\ref{eq:max_eigval}).}
    \label{fig:m_diagram_fixed_T_s}
\end{figure}
\begin{figure}
    \centering
    \includegraphics[width = \textwidth]{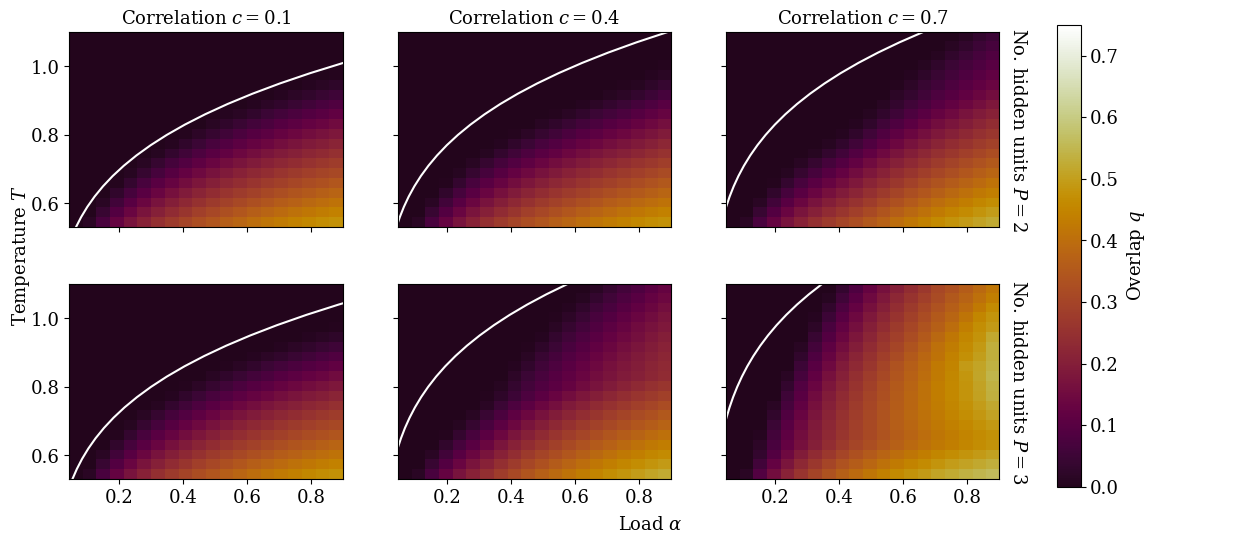}
    \caption{SG overlap $q$ for $\beta^* = 0.8$ and $P = P^*$ as a function of the number of hidden units $P$, the correlation $c$, the temperature $T$ and the data load $\alpha$. $q$ is obtained by solving Eqs. (\ref{eq:RBM_saddle-point}) numerically for binary student patterns with a uniform prior and binary teacher patterns with covariance $\mathcal{Q}_{\mu \nu} = \delta_{\mu \nu} + \left( 1 - \delta_{\mu \nu} \right) c$, where $c \in \left[ 0, 1 \right)$ (see \ref{app:uniform_correlations}). The top and bottom rows feature $P = 2$ and $P = 3$, respectively. The white lines mark the critical load of Eq. (\ref{eq:critical_load}) with $\beta = \beta^*$ and $\lambda^{\mathcal{S}}_{\text{max}}$ given by Eq. (\ref{eq:max_eigval}).}
    \label{fig:q_diagram_fixed_T_s}
\end{figure}

Figs. (\ref{fig:m_diagram_fixed_T_s}) and (\ref{fig:q_diagram_fixed_T_s}) display the magnetization $m$ and the SG overlap $q$, respectively, found by solving Eqs. (\ref{eq:RBM_saddle-point}) at fixed $\beta^*$. As usual, we find the paramagnetic (P) phase ($m = q = 0$) at small $\alpha$ and low $T$. At low $c$ and $P$, the spin glass (SG) phase ($m = 0$ and $q > 0$) occupies the medium $\alpha$ and low $T$ region of the $\alpha, T$ plane. It transitions to the ferromagnetic (F) phase ($m \neq 0$) as $\alpha$, $c$ and $P$ increase.
At high $c$, $P$ and $\alpha$, the SG overlap $q$ seems to be nonmonotonic in $T$. As expected, the critical load of Eq. (\ref{eq:critical_load}) with $\lambda^{\mathcal{S}}_{\text{max}}$ given by Eq. (\ref{eq:max_eigval}) follows the onset of nonzero magnetization corresponding to the P-F phase transition, but not the SG-F phase transition.
The critical load of $\beta = \beta^*$ approximately follows the P-SG phase transition when $c$ is small, which is consistent with Section \ref{sec:independence} and previous works \cite{alemanno2023hopfield, theriault2024dense}. As in Section \ref{sec:independence}, decreasing the inference temperature $T$ too much can make it harder for the student to learn the teacher patterns (see Fig. \ref{fig:m_diagram_fixed_T_s}, top right panel).

\subsubsection{Permutation symmetry breaking transitions}
\label{sec:concentration}
As in Section \ref{sec:no_correlations}, the critical load marking the onset of learning (see Eq. \ref{eq:critical_load}) again does not depend on the number of hidden units $P$ of the student RBM. Despite this, a single wide RBM does not learn teacher patterns as would multiple separate RBMs with one hidden unit each.
Moreover, distinct phases emerge based on the level of correlation $c$ and data load $\alpha$, each characterized by a different learning strategy.

\begin{figure}
    \centering
    \includegraphics[width=0.495\linewidth]{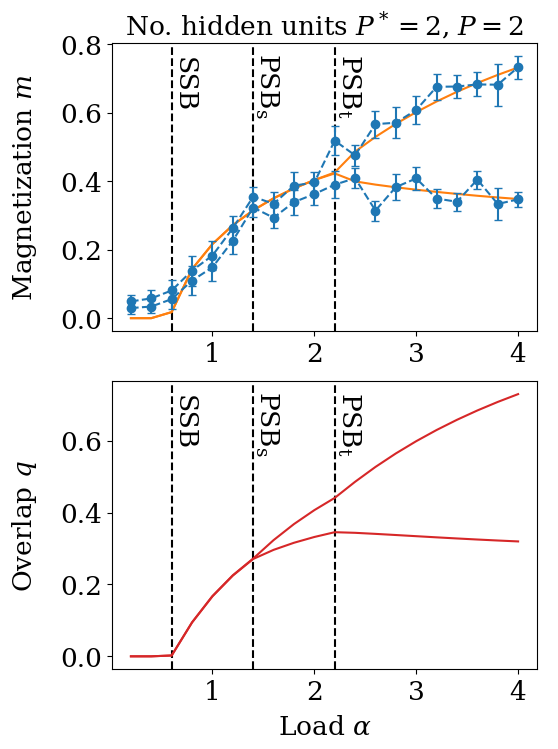}
    \includegraphics[width=0.495\linewidth]{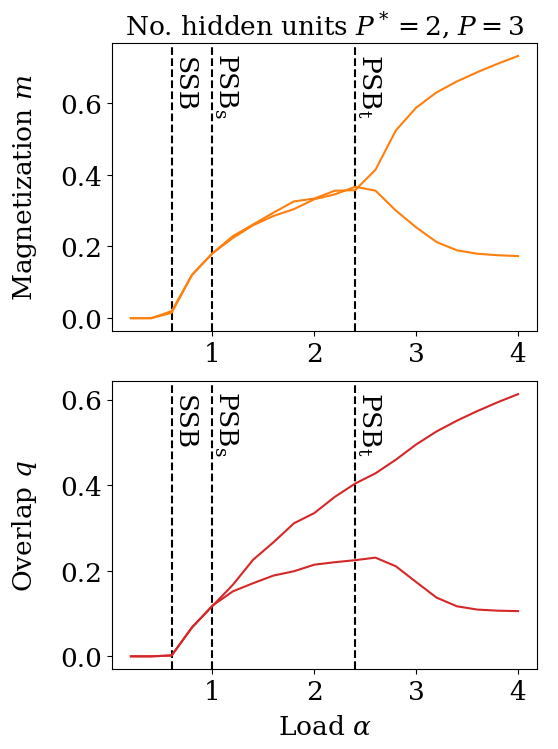}
    \caption{Mattis magnetization $m$ and SG overlap $q$ solving Eqs. (\ref{eq:RBM_saddle-point}), in orange and red, as a function of the load $\alpha$ and number of student patterns $P$ for $\beta = \beta^* = 1$, $P^* = 2$ and $c = 0.3$. The top and bottom branches of the plots are respectively the diagonal and off-diagonal coefficients of $m$ and $q$. In the top-left panel, $m$ is compared against $N = 512$ dimensional Monte Carlo simulations, in blue. The blue dots and error bars represent the means and standard deviations, respectively, of the diagonal and off-diagonal coefficients of the magnetization $m$ during the simulations. Simulation results do not agree with the predictions of the saddle-point equations shown in the right panel when $\alpha > \alpha_{\text{crit}}$ (see Eq. \ref{eq:critical_load}). Besides that, they are not very insightful, so we do not plot them for the sake of clarity.}
    \label{fig:magnetization_and_overlap_c=0.3}
\end{figure}

RBMs with $P=P^*=2$ hidden units learning correlated patterns were previously found to have three distinct ferromagnetic phases \cite{hou2019minimal}:
\begin{itemize}
    \item in the \textit{Spontaneous Symmetry Breaking} (SSB) phase, both student patterns converge to a single configuration that has the same overlap with the two teacher patterns. In other words, the student learns only the features that the teacher patterns have in common;
    \item in the \textit{student Permutation Symmetry Breaking} ($\text{PSB}_{\text{s}}$) phase, the student patterns converge to two distinct configurations that have the same overlap with the teacher patterns. In other words, the student is able to learn distinct features of the teacher patterns;
    \item finally, in the \textit{teacher Permutation Symmetry Breaking} ($\text{PSB}_{\text{t}}$) phase, the student can tell the two teacher patterns apart and learns them one-to-one as in the PSB phase of Section (\ref{sec:independence}).
\end{itemize}
These three phases can be identified by comparing the values of the order parameters $m^{\mu\nu}$ and $q^{\mu\nu}$ on and off the diagonal. As shown in the left panel of Fig. (\ref{fig:magnetization_and_overlap_c=0.3}), bifurcations occur as $\alpha$ increases, marking second-order phase transitions at the critical loads $\alpha^{\text{SSB}}_{\text{crit}} = \alpha_{\text{crit}}$, $\alpha^{\text{PSB}_{\text{s}}}_{\text{crit}} > \alpha^{\text{SSB}}_{\text{crit}}$ and $\alpha^{\text{PSB}_{\text{t}}}_{\text{crit}} > \alpha^{\text{PSB}_{\text{s}}}_{\text{crit}}$.
\begin{figure}
    \centering
    \includegraphics[width = \linewidth]{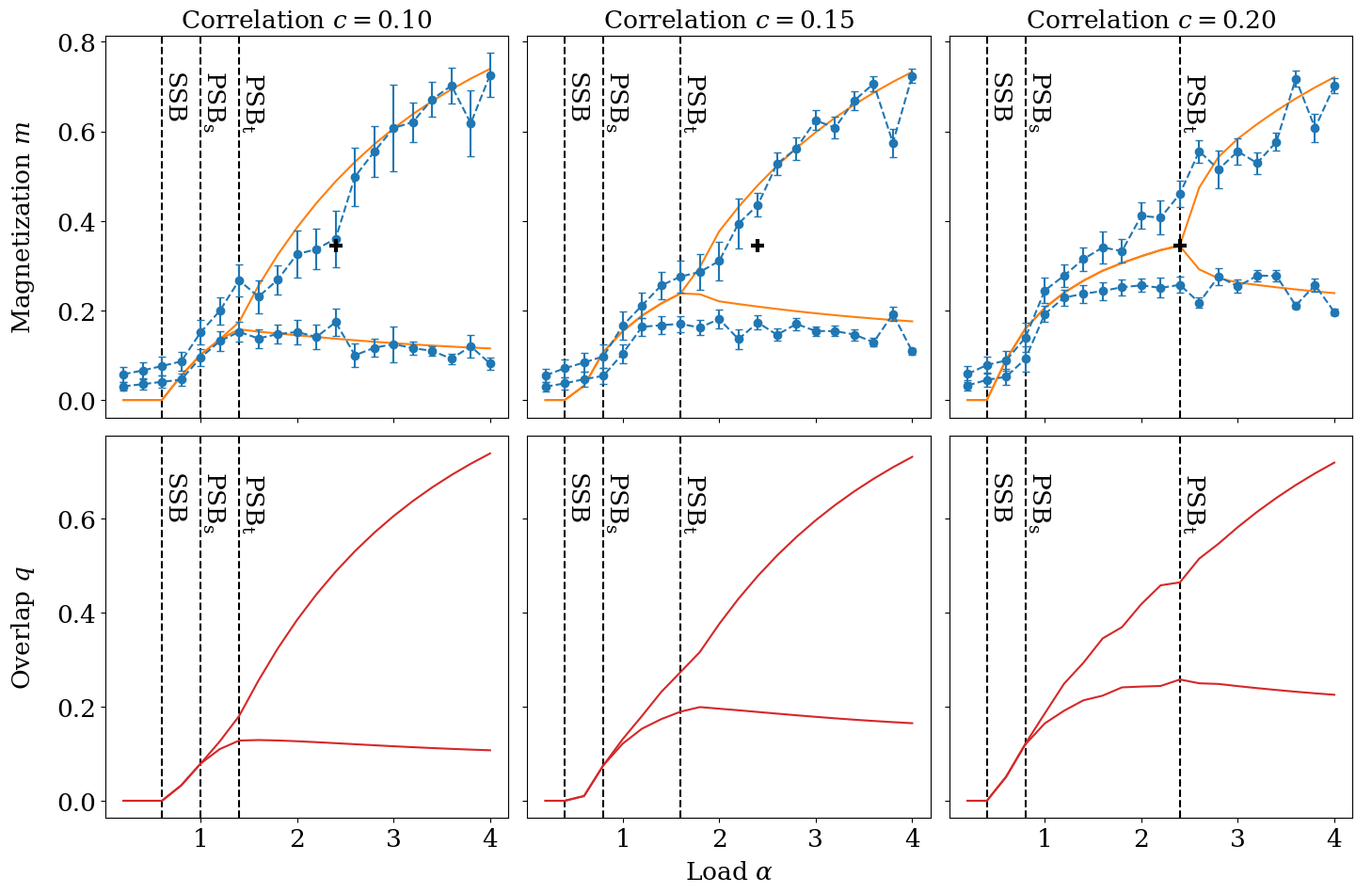}
    \caption{Mattis magnetization $m$ and SG overlap $q$ solving Eqs. (\ref{eq:RBM_saddle-point}), in orange and red, as a function of the load $\alpha$ and teacher pattern correlations $c$ for $\beta = \beta^* = 1$ and $P = P^* = 3$.
    The top and bottom branches of the plots are respectively the diagonal and off-diagonal coefficients of $m$ and $q$. $m$ is compared against $N = 512$ dimensional Monte Carlo simulations, in blue.
    The blue dots and error bars represent the means and standard deviations, respectively, of the diagonal and off-diagonal coefficients of the magnetization $m$ during the simulations.
    The dashed lines indicate the approximate locations of the SSB, $\text{PSB}_{\text{s}}$ and $\text{PSB}_{\text{t}}$ phase transitions of Eqs. (\ref{eq:RBM_saddle-point}). The black crosses in the top plots mark the $\alpha$ and $m$ at which the $\text{PSB}_{\text{t}}$ transition occurs in the right plots. It serves as a visual guide to show that the $m$ predicted by Eqs. (\ref{eq:RBM_saddle-point}) can sometimes decrease significantly with $c$, but not the $m$ of the simulations.
    }
    \label{fig:magnetization_and_overlap_P=3_P_t=3}
\end{figure}
In this context, the student does not know the level of correlation $c$ of the teacher patterns. As such, it uses a uniform prior $\Prob \left( \xi \right) \neq \Prob \left( \xi^* \right)$, and we are always outside the Nishimori regime. Despite this, the RS critical thresholds and magnetizations are in good agreement with Monte Carlo simulations (see Fig. \ref{fig:magnetization_and_overlap_c=0.3}, left panel).

As shown in Figs. (\ref{fig:magnetization_and_overlap_c=0.3}) and (\ref{fig:magnetization_and_overlap_P=3_P_t=3}), the same three phases also appear at larger $P$ and $P^*$. $\alpha^{\text{SSB}}_{\text{crit}}$ once again coincides with the onset of the ferromagnetic phase, which occurs at the same load $\alpha_{\text{crit}}$ regardless of the number of hidden units $P$ of the student. On the other hand, $\alpha^{\text{PSB}_{\text{s}}}_{\text{crit}}$ is smaller for $P = 3$ than for $P = 2$ (see Fig. \ref{fig:magnetization_and_overlap_c=0.3}). We think that the term $A_{\mu \mu} \left( q \right) = \sqrt{2 q^{\mu \mu} - \sum_{\eta = 1}^P q^{\mu \eta}}$ in $\mathcal{L}_{\lambda_1 \lambda_2}$ (see Eqs. \ref{eq:sym_L} and \ref{eq:diagonal_dominance}) penalizes values of $q$ with a large off-diagonal sum, i.e. $\sum_{\eta \neq \mu}^P q^{\mu \eta}$, compared to the diagonal $q^{\mu \mu}$. As the number of terms in the sum grows with $P$, the individual off-diagonal coefficients $q^{\mu\nu}$ may be encouraged to become smaller, pushing permutation symmetry breaking of the student patterns ($\text{PSB}_{\text{s}}$) to occur at a lower critical threshold $\alpha^{\text{PSB}_{\text{s}}}_{\text{crit}}$.

In \cite{hou2019minimal}, the critical load $\alpha^{\text{PSB}_{\text{t}}}_{\text{crit}}$ increases with the correlation strength $c$. As such, increasing $c$ can trigger a phase transition from $\text{PSB}_{\text{t}}$ to $\text{PSB}_{\text{s}}$ and thus decrease the magnetization.
In other words, large correlations in the teacher patterns can undermine the student's ability to learn them accurately. 
Based on these findings, Hou et al. \cite{hou2019minimal} formulated the hypothesis that the best $c$ for learning a relatively low load of data is nonzero but still relatively small.
In the case of $P, P^* \geq 3$ (see Fig. \ref{fig:magnetization_and_overlap_P=3_P_t=3}), we also find that $\alpha^{\text{PSB}_{\text{t}}}_{\text{crit}}$ grows with $c$.
However, unlike $P = P^* = 2$, Monte Carlo simulations of $P, P^* \geq 3$ do not completely agree with the RS approximation at large $c$. In particular, increasing $c$ does not seem to decrease the magnetization significantly in the simulations, even when it does so according to the RS approximation (see Fig. \ref{fig:magnetization_and_overlap_P=3_P_t=3}, black crosses).

\begin{figure}
    \centering
    \includegraphics[width = \textwidth]{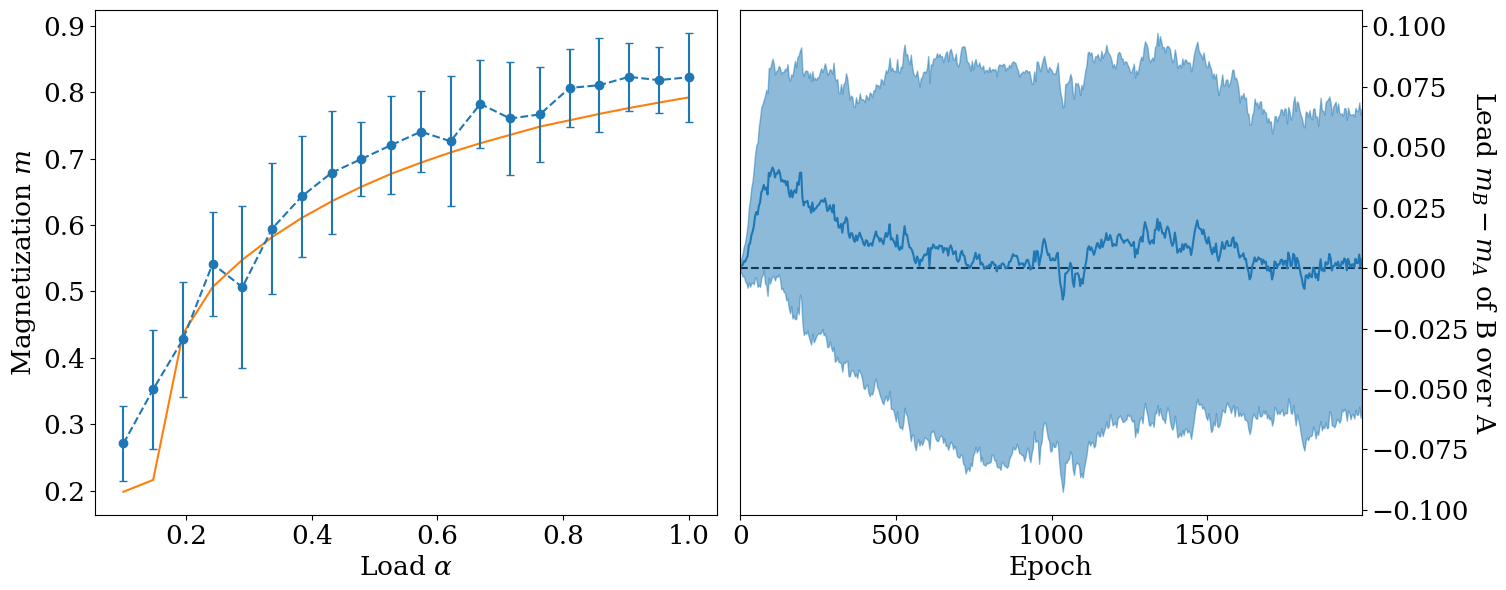}
    \caption{Results of the lottery ticket experiment of Section \ref{sec:lottery} when the teacher patterns have a uniform correlation matrix with $c = 0.05$ instead of being i.i.d. In the left panel, $N = 512$ dimensional Monte Carlo simulations of student B, in blue, are compared against the solution of Eqs. (\ref{eq:reduced_saddle-point}), in orange. The blue dots and error bars represent the means and standard deviations, respectively, of the diagonal of the magnetization $m$ during the simulations. The right panel shows the difference $m_B - m_A$ of the magnetizations of A and B as a function of the simulation epochs. The solid blue line and the shaded region represent the median of $m_A - m_B$ over $\alpha \in \left[ 0, 1 \right]$ and the corresponding mean absolute deviation around the median, respectively. $m_A - m_B$ goes to zero when the number of elapsed epochs is large, so student A converges to the solution of Eqs. (\ref{eq:reduced_normal_saddle-point}) like student B. The inverse temperature is set to $\beta^* = \beta = 4$.}
    \label{fig:winning_ticket_lead_c=0.05}
\end{figure}

Repeating the lottery ticket experiment of Section \ref{sec:lottery} for $c = 0.05$, we find that the lead of the winning ticket B over its randomly-initialized counterpart A has a smaller maximum (see Fig. \ref{fig:winning_ticket_lead_c=0.05}). In other words, small correlations in the data appear to make the student less sensitive to initial conditions, which also reduces the benefits of magnitude pruning (see Section \ref{sec:lottery}). As in Sections \ref{sec:independence_g} and \ref{sec:lottery}, the error in $m$ (see Fig. \ref{fig:winning_ticket_lead_c=0.05}, left panel) is useful for measuring the convergence of the learning algorithm, but poorly representative of the equilibrium distribution.

Outside the teacher-student setting, RBMs typically undergo a sequence of second-order phase transitions at the beginning of training \cite{decelle2017spectral, decelle2018thermodynamics, bachtis2025cascade, bereux2025fast}. After the first transition, they learn a rank-one matrix, which is reminiscent of our SSB phase. The second transition then breaks the rank-one symmetry like our $\text{PSB}_{\text{s}}$ transition. The subsequent phase transitions are harder to interpret from the point of view of permutation symmetry breaking, and it is not clear whether a transition analogous to $\text{PSB}_{\text{t}}$ can occur without an explicit teacher. $\text{PSB}_{\text{s}}$ phase transitions were also observed in Gaussian mixture models \cite{rose1990statistical, kloppenburg1997deterministic, akaho2000nonmonotonic, Decelle2021restricted} and modern Hopfield networks \cite{boukacem2024waddington}, which suggests that they could be common in machine learning models.

\subsection{Random correlations}
\label{sec:random_correlations}
In this section, we take the covariance matrix $\mathcal{Q}$ to be random.
By definition,
$\mathcal{Q}$ must be positive semi-definite. Moreover, when the teacher weights are binary, $\mathcal{Q}$ must have ones on the diagonal. We sample $\mathcal{Q}$ from the projected Wishart distribution $\mathcal{W} \left( c, D \right)$ defined in \ref{sec:projected_wishart} because it satisfies these two requirements. By the law of large numbers, $\mathcal{Q} \sim \mathcal{W} \left( c, D \right)$ approaches $\mathcal{C} = \delta_{\mu \nu} + \left( 1 - \delta_{\mu \nu} \right) c$ in probability as $D \rightarrow \infty$ (see \ref{sec:projected_wishart}). Therefore, $\lambda^{\mathcal{S}}_{\text{max}}$ and $\alpha_{\text{crit}}$ are the same as in Section \ref{sec:uniform_correlations} when $D$ is large.
At finite $D = P$, the dependence of $\alpha_{\text{crit}}$ on $T$ is still qualitatively similar to that of uniform correlations. However, the arithmetic mean $\overline{\alpha_{\text{crit}}}$ and the harmonic mean $\left[ \overline{1 / \lambda^{\mathcal{S}}_{\text{max}}} \right]^{-1}$ over many independent samples $\mathcal{Q} \sim \mathcal{W} \left( c, P = D \right)$ are very different from the $\alpha_{\text{crit}}$ and $\lambda^{\mathcal{S}}_{\text{max}}$ of uniform correlations as a function of $c$ and $P$ (see Figs. \ref{fig:alpha_crit} and \ref{fig:eigval_range}).
For instance, $\overline{\alpha_{\text{crit}}}$ decreases much more slowly with $c$ than the critical load of uniform correlations.
At small $c$ and high $T$, $\overline{\alpha_{\text{crit}}}$ tends to be smaller than the critical load of uniform correlations. Conversely, at large $c$ and low $T$, the critical load of uniform correlations is usually smaller.
At any given $c$, the entries of a random correlation matrix can sometimes be larger than $c$, which can make learning possible even when $T$ is too high for correlations of size $c$ or smaller to be picked up on by the student. However, this advantage disappears at larger $c$ and lower $T$ where correlations of size $c$ and smaller are no longer muddled in noise.
In summary, the behavior of the critical load as a function of $P$ and $c$ is in general very different for random correlations and for uniform correlations.

\section{Conclusion}
\label{sec:conclusion}
In this paper, we theoretically study the learning performance of Restricted Boltzmann Machines (RBM) \cite{ackley1985learning, smolensky1986information, freund1991unsupervised, hinton2002training} with a finite number of hidden units in the teacher-student setting \cite{huang2016unsupervised, huang2017statistical, barra2017phase, barra2018phase, huang2018role, hou2019minimal, decelle2021inverse} using the replica method under the Replica-Symmetric (RS) approximation \cite{charbonneau2022replica}. Given $M$ data points and $N$ visible units, we compute the critical data load $\alpha_{\text{crit}} = \frac{M}{N}$ above which learning becomes possible. Our findings validate the conjecture that the student's performance is independent of the number of hidden units when the patterns of the teacher are uncorrelated \cite{barra2017phase}, generalizing the results of \cite{hou2019minimal} to any number of hidden units $P$ and teacher patterns $P^*$ much smaller than $M$ and $N$. In particular, we confirm that an RBM with $P$ uncorrelated hidden units factorizes into $P$ RBMs with one hidden unit each \cite{hou2019minimal}.
Additionally, we show that the teacher-student setting without correlations has both a Permutation Symmetry Breaking (PSB) solution in which the student learns the teacher patterns one-to-one and a metastable partial PSB solutions in which multiple student patterns can converge to the same teacher pattern. We argue that the teacher-student setting complies with the lottery ticket hypothesis \cite{frankle2018lottery, zhou2019deconstructing, ramanujan2020hidden, eran2020proving, frankle2020linear}, and demonstrate that the student can be trained efficiently using a variant of the magnitude pruning algorithm \cite{frankle2018lottery, frankle2020linear}. Given teacher patterns with uniform correlations $c \in [0, 1)$, we find a closed-form expression for $\alpha_{\text{crit}}$, which generalizes the one found in \cite{hou2019minimal} for $P = P^* = 2$, and show that it decreases with both $c$ and $P^*$. Still in the case of uniform correlations, we find that the teacher-student setting undergoes a sequence of Spontaneous Symmetry Breaking (SSB) and permutation symmetry breaking transitions that generalizes the one found in \cite{hou2019minimal} to $P^*, P \geq 3$. Both with and without correlations, we find that decreasing the inference temperature $T$ too much can prevent the student from learning the teacher's solution even when the dataset is relatively large.
Throughout the paper, we compare key results against Monte Carlo simulations and observe that the RS ansatz is often a solid approximation even in the region outside the Nishimori line \cite{nishimori1980exact, iba1999nishimori, nishimori2001statistical, contucci2009spin} where the teacher and the student have the same temperature, but a different number of hidden units and a different prior on their patterns.

One could study the learning dynamics of the magnitude pruning experiment described in Section \ref{sec:lottery} or characterize the distribution of winning ticket initial conditions to gain additional insight into the lottery ticket hypothesis. More generally, it would be interesting to extend our study to the case where $P$ and $P^*$ are of order $N$. In fact, RBMs trained on real data typically have a large number of hidden units \cite{hinton2002training, leroux2011learning, tubiana2017emergence, kiviken2012multiple, fischer2014training}. One could also introduce correlations between the columns $\xi^*_i = \left\{ \xi^{* \mu}_i \right\}_{\mu = 1}^P$ of the teacher's weight matrix $\xi^* = \left\{ \xi^{* \mu}_i \right\}_{1 \leq i \leq N}^{1 \leq \mu \leq P}$ in addition to the correlations between patterns $\xi^{* \mu} = \left\{ \xi^{* \mu}_i \right\}_{i = 1}^N$ that we study in this work. Such a modification could be achieved by using a weight matrix with a low-rank SVD structure like in \cite{decelle2017spectral, decelle2018thermodynamics} and would allow the teacher to generate data with even more structure than in this paper. One could use this framework to investigate whether the distinction between $\text{PSB}_{\text{s}}$ and $\text{PSB}_{\text{t}}$ persists in real data and has a measurable effect on the critical slowing down patterns observed in RBM training \cite{decelle2017spectral, decelle2018thermodynamics, bachtis2025cascade, bereux2025fast}.
Another promising research avenue would be to study other generative models with a finite number of hidden units in the teacher-student setting, such as modern (a.k.a. dense) Hopfield networks \cite{hopfield1982neural, theriault2024dense, chen1986high, psaltis1986nonlinear, krotov2016dense}. In sum, the teacher-student setting still has a lot of untapped potential for studying generative models with many hidden units.

\section*{Data availability}
The code and hyperparameter values of the training algorithms
used to make the figures are available at the following public
repository \cite{theriault2025modellingsoftware}.

\section*{Acknowledgements}
I am grateful to Francesco Tosello, who made a significant contribution to this work. He wrote a first draft of the calculations presented in the Appendices, which Daniele Tantari and I then generalized and expanded upon to write this Chapter.

\begin{subappendices}

\section{Definitions}
\label{app:definitions}

\subsection{Binary random variables with a fixed covariance matrix}
\label{sec:binary_random}
We generate binary random variables $\xi^*_i \in \left\{ -1, +1 \right\}^{P^*}$ with mean $0$ and a fixed covariance matrix $\mathcal{Q}$ in two steps:
\begin{itemize}
    \item sample $x^*_i \sim \mathcal{N} \left( 0, \sin \left( \frac{\pi}{2} \mathcal{Q} \right) \right)$ from a multivariate Gaussian distribution with mean $0$ and covariance matrix $\sin \left( \frac{\pi}{2} \mathcal{Q} \right)$,
    \item set $\xi^*_i = \sign \left( x^*_i \right)$.
\end{itemize}
This sampling method is commonly known as the arcsine law and originates from \cite{van1966spectrum}. It implicitly defines a probability distribution $\Prob \left( \xi^* \right)$ for correlated binary teacher patterns $\xi^*$. 

\subsection{Projected Wishart distribution}
\label{sec:projected_wishart}
We generate random positive definite matrices $\mathcal{B}$ by sampling them from the Wishart distribution \cite{whishart1928generalized}. That is, we sample the columns of a $P \times D$ matrix $\mathcal{A}$ from the Gaussian distribution $\mathcal{N} \left( 0, \mathcal{C} \right)$, then set $\mathcal{B} = \mathcal{A} \mathcal{A}^T$.
Let $\mathcal{D}$ be the diagonal matrix with the same values as $\mathcal{B}$ on the diagonal. We obtain a positive definite matrix $\mathcal{Q}$ with ones on the diagonal by normalizing $\mathcal{B}$ according to $\mathcal{Q} = \mathcal{D}^{-1/2} \mathcal{B} \mathcal{D}^{-1/2}$. In other words, $\mathcal{B}$ is a generic covariance matrix, and the entries $\mathcal{Q}_{\mu \nu}$ of $\mathcal{Q}$ are the Pearson correlation coefficients corresponding to $\mathcal{B}_{\mu \nu}$.
For simplicity's sake, we take the covariance matrix $\mathcal{C}$ of the Gaussian distribution $\mathcal{N} \left( 0, \mathcal{C} \right)$ to be $\mathcal{C}_{\mu \nu} = \delta_{\mu \nu} + \left( 1 - \delta_{\mu \nu} \right) c$. In this paper, we say that $\mathcal{Q}$ is sampled from the projected Wishart distribution $\mathcal{W} \left( c, D \right)$.

\subsection{Effective Hamiltonian $\mathcal{L}_{\lambda_1 \lambda_2}$}
\label{sec:effective_hamiltonian}
The effective Hamiltonian $\mathcal{L}_{\lambda_1 \lambda_2}$ plays a crucial role in deriving the free entropy and the saddle-point equations of the problem studied in this work. In this Section, we define it to be
\begin{align*}
    \mathcal{L}_{\lambda_1, \lambda_2} \left( \xi, \xi^*, z ; m, s, q \right) &= \frac{1}{2} \left[ \lambda_2 \right]^2 \sum_{\mu, \nu = 1}^{P} \left( s^{\mu \nu} - \frac{q^{\mu \nu} + q^{\nu \mu}}{2} \right) \xi^{\mu} \xi^{\nu} + \lambda_1 \lambda_2 \sum_{\gamma = 1}^{P^*} \sum_{\mu = 1}^{P} m^{\gamma \mu} \xi^{* \gamma} \xi^{\mu} \\
    &\quad+ \lambda_2 \sum_{\mu, \nu = 1}^{P} \sqrt{q^{\mu \nu} + q^{\nu \mu} - \delta_{\mu \nu} \sum_{\eta = 1}^P \frac{q^{\mu \eta} + q^{\eta \mu}}{2}} z_{\mu \nu} \frac{\xi^{\mu} + \xi^{\nu}}{2}.
\end{align*}
where we set the diagonal of $s$ to $s^{\mu \mu} = 0$ to regroup $s^{\mu \nu}$ and $\frac{q^{\mu \nu} + q^{\nu \mu}}{2}$ in the same sum. In $\mathcal{L}^O$ (see Eq. \ref{hamiltonian_L_O}), the diagonal of $s$ is arbitrary because it does not affect the saddle-point equations (see Eqs. \ref{eq:RBM_saddle-point}). In $\mathcal{L}^C$ (see Eq. \ref{hamiltonian_L_C}), the diagonal of $\hat{s}$ always vanishes (also see Eqs. \ref{eq:RBM_saddle-point}).
When $q$ and $\hat{q}$ are symmetric, we obtain the simpler form
\begin{align*}
    \mathcal{L}_{\lambda_1, \lambda_2} \left( \xi, \xi^*, z ; m, s, q \right) &= \frac{1}{2} \left[ \lambda_2 \right]^2 \sum_{\mu, \nu} \left( s^{\mu \nu} - q^{\mu \nu} \right) \xi^{\mu} \xi^{\nu} + \lambda_1 \lambda_2 \sum_{\gamma, \mu} m^{\gamma \mu} \xi^{* \gamma} \xi^{\mu} \\
    &\quad+ \lambda_2 \sum_{\mu, \nu} \sqrt{2 q^{\mu \nu} - \delta_{\mu \nu} \sum_{\eta} q^{\mu \eta}} z_{\mu \nu} \frac{\xi^{\mu} + \xi^{\nu}}{2},
\end{align*}
which is the definition used in the main text (see Eq. \ref{eq:sym_L}). We symmetrize the initial conditions of $q^{\mu \nu}$ before solving the saddle-point equations numerically because any $q^{\mu \nu}$ solving them must be symmetric (see Eqs. \ref{eq:RBM_saddle-point} and \ref{app:numerical_methods}). Defining
\begin{align*}
    A_{\mu \nu} \left( q \right) &= \sqrt{2 q^{\mu \nu} - \delta_{\mu \nu} \sum_{\eta} q^{\mu \eta}},
\end{align*}
these two expressions can be written more compactly as
\begin{align*}
    \mathcal{L}_{\lambda_1, \lambda_2} \left( \xi, \xi^*, z ; m, s, q \right) &= \frac{1}{2} \left[ \lambda_2 \right]^2 \sum_{\mu, \nu} \left( s^{\mu \nu} - q^{\mu \nu} \right) \xi^{\mu} \xi^{\nu} + \lambda_1 \lambda_2 \sum_{\gamma, \mu} m^{\gamma \mu} \xi^{* \gamma} \xi^{\mu} \numberthis \label{eq:non_sym_L} \\
    &\quad+ \lambda_2 \sum_{\mu, \nu} A_{\mu \nu} \left( \frac{q + q^T}{2} \right) z_{\mu \nu} \frac{\xi^{\mu} + \xi^{\nu}}{2} \quad \text{for generic $q$ and} \\
    \mathcal{L}_{\lambda_1, \lambda_2} \left( \xi, \xi^*, z ; m, s, q \right) &= \frac{1}{2} \left[ \lambda_2 \right]^2 \sum_{\mu, \nu} \left( s^{\mu \nu} - q^{\mu \nu} \right) \xi^{\mu} \xi^{\nu} + \lambda_1 \lambda_2 \sum_{\gamma, \mu} m^{\gamma \mu} \xi^{* \gamma} \xi^{\mu} \numberthis \\
    &\quad+ \lambda_2 \sum_{\mu, \nu} A_{\mu \nu} \left( q \right) z_{\mu \nu} \frac{\xi^{\mu} + \xi^{\nu}}{2} \quad \text{for symmetric $q$.}
\end{align*}
This way of writing $\mathcal{L}_{\lambda_1, \lambda_2}$ is useful to make the derivation of the saddle-point equations more concise (see \ref{app:RBM_saddle-point}). The third term of $\mathcal{L}_{\lambda_1, \lambda_2}$ can also be written as
\begin{align*}
    \lambda_2 \sum_{\mu, \nu} A_{\mu \nu} \left( q \right) z_{\mu \nu} \frac{\xi^{\mu} + \xi^{\nu}}{2} &= \lambda_2 \sum_{\mu, \nu} A_{\mu \nu} \left( q \right) \frac{z_{\mu \nu} + z_{\nu \mu}}{2} \xi^{\mu}
\end{align*}
by interchanging summation indices. The two ways of writing the third term have different numerical implementations, but we did not see a significant difference in computational complexity and accuracy between them.

\section{Replicated partition function}
\label{app:RBM_partition_function}
In the following appendices, we omit the subscripts $\beta^*$ and $\beta$ to make notation lighter.
The sums over $\gamma$ or $\rho$ are from $1$ to $P^*$, and the sums over $\mu$ or $\nu$, from $1$ to $P$.
Given a set of $M$ examples $\boldsymbol{\sigma} = \left\{ \sigma^a \right\}_{a = 1}^M$, the probability distribution of a single replica $b$ takes the form
\begin{align*}
    \Prob \left( \xi^b \mid \boldsymbol{\sigma} \right) &= \mathcal{Z} \left( \boldsymbol{\sigma} \right)^{-1}\Prob \left( \xi^b \right) \prod_a \Prob \left( \sigma^a \mid \xi^b \right),
\end{align*}
by using Bayes' theorem (see Section \ref{sec:RBM_teacher-student}). The partition function is $\mathcal{Z} \left( \boldsymbol{\sigma} \right) = \mathbb{E}_{\xi^b} \left[ \prod_a \Prob \left( \sigma^a \mid \xi^b \right) \right]$, which leads to
\begin{align*}
    \mathbb{E}_{\xi^*,\boldsymbol{\sigma}} \left[ \mathcal{Z}^L \right] &= \sum_{\sigma} \mathbb{E}_{\xi^*} \left[ \prod_a \Prob \left( \sigma^a \mid \xi^* \right) \right] \mathcal{Z} \left( \boldsymbol{\sigma} \right)^L \\
    &= \sum_{\sigma} \mathbb{E}_{\xi^*} \left[ \prod_a \Prob \left( \sigma^a \mid \xi^* \right) \right] \mathbb{E}_{\xi} \left[ \prod_a \Prob \left( \sigma^a \mid \xi \right) \right] \\
    &= \mathbb{E}_{\xi^* \xi} \left[ \prod_a \sum_{\sigma^a} \Prob \left( \sigma^a \mid \xi^* \right) \Prob \left( \sigma^a \mid \xi \right) \right] \\
    &= \mathbb{E}_{\xi^* \xi} \left[ \left( \sum_{\sigma^a} \Prob \left( \sigma^a \mid \xi^* \right) \Prob \left( \sigma^a \mid \xi \right) \right)^M \right]
\end{align*}
where $\xi = \left\{ \xi^b \right\}_{b = 1}^L$ is a set of $L$ replicas and $\Prob \left( \sigma^a \mid \xi \right) = \prod_b \Prob \left( \sigma^a \mid \xi^b \right)$. As per Eq. (\ref{eq:RBM_direct_distribution}), $\Prob \left( \sigma^a \mid \xi^b \right)$ with binary units takes the form
\begin{align*}
    \Prob \left( \sigma^a \mid \xi^b \right) &= Z \left( \xi^b \right)^{-1} \psi \left( \sigma^a ; \xi^b \right),
\end{align*}
where $\psi \left( \sigma^a ; \xi^b \right) = \Prob \left( \sigma^a \right) \mathbb{E}_{\tau_b} \left[ \exp \left( \frac{\beta}{\sqrt{N}} \sum_{\mu} \tau_{b \mu} \sum_i \xi^{b \mu}_i \sigma^a_i \right) \right]$ and $Z \left( \xi^b \right) = \sum_{\sigma^a} \psi \left( \sigma^a ; \xi^b \right)$. In particular, we have
\begin{align*}
    \sum_{\sigma^a} \Prob \left( \sigma^a \mid \xi^* \right) \Prob \left( \sigma^a \mid \xi \right) &= \sum_{\sigma^a} Z \left( \xi^* \right)^{-1} \left[ \prod_b Z \left( \xi^b \right)^{-1} \right] \psi \left( \sigma^a ; \xi^* \right) \left[ \prod_b \psi \left( \sigma^a ; \xi^b \right) \right].
\end{align*}
We will now take $\Prob \left( \sigma^a \right)$ uniform and eliminate $\sigma^a$ from $Z \left( \xi^* \right) \prod_b Z \left( \xi^b \right)$ in order to factor it out of the sum over $\sigma^a$. We first rewrite $Z \left( \xi^b \right)$ as
\begin{align*}
    Z \left( \xi^b \right) &= 2^{-N} \sum_{\sigma^a} \mathbb{E}_{\tau_b} \left[ \exp \left( \frac{\beta}{\sqrt{N}} \sum_{\mu} \tau_{b \mu} \sum_i \xi^{b \mu}_i \sigma^a_i \right) \right] \\
    &= 2^{-N} \mathbb{E}_{\tau_b} \left[ \prod_i \sum_{\sigma^a_i} \exp \left( \frac{\beta}{\sqrt{N}} \sum_{\mu} \tau_{b \mu} \xi^{b \mu}_i \sigma^a_i \right) \right] \\
    &= \mathbb{E}_{\tau_b} \left[ \prod_i \cosh \left( \frac{\beta}{\sqrt{N}} \sum_{\mu} \tau_{b \mu} \xi^{b \mu}_i \right) \right] \\
    &= \mathbb{E}_{\tau_b} \left[ \exp \left( \sum_i \log \cosh \left[ \frac{\beta}{\sqrt{N}} \sum_{\mu} \tau_{b \mu} \xi^{b \mu}_i \right] \right) \right],
\end{align*}
then we expand the $\log \cosh$ function in small $\frac{P}{N}$ to obtain
\begin{align*}
    Z \left( \xi^b \right) &\approx \mathbb{E}_{\tau_b} \left[ \exp \left( \sum_i \frac{1}{2} \left[ \frac{\beta}{\sqrt{N}} \sum_{\mu} \tau_{b \mu} \xi^{b \mu}_i \right]^2 \right) \right] \\
    &= \mathbb{E}_{\tau_b} \left[ \exp \left( \frac{1}{2} \beta^2 \sum_{\mu, \nu} \tau_{b \mu} \tau_{b \nu} \frac{1}{N} \sum_i \xi^{b \mu}_i \xi^{b \nu}_i \right) \right].
\end{align*}
The equivalent expression for $Z \left( \xi^* \right)$ is identical except for the asterisk $*$ replacing the replica index $b$ and the sum running from $1$ to $P^*$ rather than $1$ to $P$. Once $Z \left( \xi^* \right) \prod_b Z \left( \xi^b \right)$ is factored out of the sum, $\sum_{\sigma^a} \psi \left( \sigma^a ; \xi^* \right) \prod_b \psi \left( \sigma^a ; \xi^b \right)$ simplifies in a similar way to
\begin{align*}
    \sum_{\sigma^a} \psi \left( \sigma^a ; \xi^* \right) \prod_b \psi \left( \sigma^a ; \xi^b \right) &\approx \mathbb{E}_{\tau_* \tau} \left[ \exp \left( \sum_i \frac{1}{2} \left[ \frac{\beta^*}{\sqrt{N}} \sum_{\gamma} \tau_{* \gamma} \xi^{* \gamma}_i + \frac{\beta}{\sqrt{N}} \sum_{\mu ; b} \tau_{b \mu} \xi^{b \mu}_i \right]^2 \right) \right] \\
    &= \mathbb{E}_{\tau_* \tau} \left[ \exp \left( \frac{1}{2} \left[ \beta^* \right]^2 \sum_{\gamma, \rho} \tau_{* \gamma} \tau_{* \rho} \frac{1}{N} \sum_i \xi^{* \gamma}_i \xi^{* \rho}_i \right. \right. \\
    &\quad+ \beta^* \beta \sum_{\gamma, \mu ; b} \tau_{* \gamma} \tau_{b \mu} \frac{1}{N} \sum_i \xi^{* \gamma}_i \xi^{b \mu}_i \\
    &\quad+ \left. \left. \frac{1}{2} \beta^2 \sum_{\mu, \nu ; b, c} \tau_{b \mu} \tau_{c \nu} \frac{1}{N} \sum_i \xi^{b \mu}_i \xi^{c \nu}_i \right) \right],
\end{align*}
up to an irrelevant multiplicative factor of $2^{-L N}$. Combining all these expressions, we get
\begin{align*}
    \mathbb{E} \left[ \mathcal{Z}^L \right] &= \mathbb{E}_{\xi^* \xi} \left[ \exp \left\{ M \left( \log \Bigg[ \mathbb{E}_{\tau_* \tau} \exp \Bigg( \left[ \beta^* \right]^2 \sum_{\gamma < \rho} \tau_{* \gamma} \tau_{* \rho} \frac{1}{N} \sum_i \xi^{* \gamma}_i \xi^{* \rho}_i \right. \right. \right. \\
    &\quad+ \beta^* \beta \sum_{\gamma, \mu ; b} \tau_{* \gamma} \tau_{b \mu} \frac{1}{N} \sum_i \xi^{* \gamma}_i \xi^{b \mu}_i + \beta^2 \sum_{\mu < \nu ; b} \tau_{b \mu} \tau_{b \nu} \frac{1}{N} \sum_i \xi^{b \mu}_i \xi^{b \nu}_i \\
    &\quad+ \beta^2 \sum_{\mu, \nu ; b < c} \tau_{b \mu} \tau_{c \nu} \frac{1}{N} \sum_i \xi^{b \mu}_i \xi^{c \nu}_i \Bigg) \Bigg] \\
    &\quad- \log \left[ \mathbb{E}_{\tau_*} \exp \left( \left[ \beta^* \right]^2 \sum_{\gamma < \rho} \tau_{* \gamma} \tau_{* \rho} \frac{1}{N} \sum_i \xi^{* \gamma}_i \xi^{* \rho}_i \right) \right] \\
    &\quad- \left. \left. \left. \log \left[ \mathbb{E}_{\tau} \exp \left( \beta^2 \sum_{\mu < \nu ; b} \tau_{b \mu} \tau_{b \nu} \frac{1}{N} \sum_i \xi^{b \mu}_i \xi^{b \nu}_i \right) \right] \right) \right\} \right].
\end{align*}
\section{RS free entropy}
\label{app:free_entropy}
In this Section, we introduce order parameters and conjugate order parameters for the various overlaps between patterns present in $\mathbb{E} \left[ \mathcal{Z}^L \right]$. We define
\begin{equation*}
\begin{aligned}
    &m^{b \gamma \mu} \hspace{-7pt} &\text{and} &\hspace{3pt} \hat{m}^{b \gamma \mu} &\text{for} \ \frac{1}{N} \sum_i \xi^{* \gamma}_i \xi^{b \mu}_i, \\
    &s^{b \mu \nu} \hspace{-7pt} &\text{and} &\hspace{3pt} \hat{s}^{b \mu \nu} &\text{for} \ \frac{1}{N} \sum_i \xi^{b \mu}_i \xi^{b \nu}_i \text{ where } \mu \neq \nu, \\
    &q^{b c \mu \nu} \hspace{-7pt} &\text{and} &\hspace{3pt} \hat{q}^{b c \mu \nu} &\text{for} \ \frac{1}{N} \sum_i \xi^{b \mu}_i \xi^{c \nu}_i \text{ where } b \neq c,
\end{aligned}
\end{equation*}
where $m^{b \gamma \mu}$, $s^{b \mu \nu}$ and $q^{b c \mu \nu}$ are the ordinary order parameters, $\hat{m}^{b \gamma \mu}$, $\hat{s}^{b \mu \nu}$ and $\hat{q}^{b c \mu \nu}$ are the conjugate order parameters and $s^{b \mu \nu}$ and $\hat{s}^{b \mu \nu}$ are symmetric in $\mu$ and $\nu$ by construction. We use a Fourier transform to rewrite $\mathbb{E} \left[ \mathcal{Z}^L \right]$ as
\begin{align*}
    \mathbb{E} \left[ \mathcal{Z}^L \right] &= \mathbb{E}_{\xi^* \xi} \left[ \int_{i \mathbb{R}} \prod_{\gamma, \mu ; b} d \hat{m}^{b \gamma \mu} \prod_{\mu < \nu ; b} d \hat{s}^{b \mu \nu} \prod_{\mu, \nu ; b < c} d \hat{q}^{b c \mu \nu} \right. \\
    &\quad \int_{\mathbb{R}} \prod_{\gamma, \mu ; b} d m^{b \gamma \mu} \prod_{\mu < \nu ; b} d s^{b \mu \nu} \prod_{\mu, \nu ; b < c} d q^{b c \mu \nu} \exp \left\{ \sum_{\gamma, \mu ; b} \hat{m}^{b \gamma \mu} \left( \sum_i \xi^{* \gamma}_i \xi^{b \mu}_i - N m^{b \gamma \mu} \right) \right\} \\
    &\quad \exp \left\{ \sum_{\mu < \nu ; b} \hat{s}^{b \mu \nu} \left( \sum_i \xi^{b \mu}_i \xi^{b \nu}_i - N s^{b \mu \nu} \right) + \sum_{\mu, \nu ; b < c} \hat{q}^{b c \mu \nu} \left( \sum_i \xi^{b \mu}_i \xi^{c \nu}_i - N q^{b c \mu \nu} \right) \right\} \\
    &\quad \exp \left\{ M \left( \log \Bigg[ \mathbb{E}_{\tau_* \tau} \exp \Bigg( \left[ \beta^* \right]^2 \sum_{\gamma < \rho} \tau_{* \gamma} \tau_{* \rho} \frac{1}{N} \sum_i \xi^{* \gamma}_i \xi^{* \rho}_i \right. \right. \\
    &\quad+ \beta^* \beta \sum_{\gamma, \mu ; b} m^{b \gamma \mu} \tau_{* \gamma} \tau_{b \mu} + \beta^2 \sum_{\mu < \nu ; b} s^{b \mu \nu} \tau_{b \mu} \tau_{b \nu} + \beta^2 \sum_{\mu, \nu ; b < c} q^{b c \mu \nu} \tau_{b \mu} \tau_{c \nu} \Bigg) \Bigg] \\
    &\quad- \log \left[ \mathbb{E}_{\tau_*} \exp \left( \left[ \beta^* \right]^2 \sum_{\gamma < \rho} \tau_{* \gamma} \tau_{* \rho} \frac{1}{N} \sum_i \xi^{* \gamma}_i \xi^{* \rho}_i \right) \right] \\
    &\quad- \left. \left. \left. \log \left[ \mathbb{E}_{\tau} \exp \left( \beta^2 \sum_{\mu < \nu ; b} s^{b \mu \nu} \tau_{b \mu} \tau_{b \nu} \right) \right] \right) \right\} \right].
\end{align*}
By hypothesis, the columns $\xi^*_i = \left\{ \xi^{* \gamma}_i \right\}_{\gamma = 1}^{P^*}$ of $\xi^* \sim \Prob \left( \xi^* \right)$ are i.i.d. random variables and their distribution $\Prob \left( \xi^*_i \right)$ has a well-defined $P^* \times P^*$ covariance matrix $\mathcal{Q}$ (see Section \ref{sec:RBM_introduction}). Therefore, by the law of large numbers, $\frac{1}{N} \sum_i \xi^{* \gamma}_i \xi^{* \rho}_i$ converges in probability to the covariance $\mathcal{Q}_{\gamma \rho}$ as $N \rightarrow \infty$.
$\mathbb{E} \left[ \mathcal{Z}^L \right]$ then simplifies to
\begin{align*}
    \mathbb{E} \left[ \mathcal{Z}^L \right] &= \int \prod_{\gamma, \mu ; b} d \hat{m}^{b \gamma \mu} d m^{b \gamma \mu} \prod_{\mu < \nu ; b} d \hat{s}^{b \mu \nu} d s^{b \mu \nu} \prod_{\mu, \nu ; b < c} d \hat{q}^{b c \mu \nu} d q^{b c \mu \nu} \\
    &\quad \exp \left\{ N \log \left[ \mathbb{E}_{\xi^*_i \xi_i} \exp \left\{ H_S \left( \xi_i, \xi^*_i ; \hat{m}, \hat{s}, \hat{q} \right) \right\} \right] - N H_Q \left( m, s, q, \hat{m}, \hat{s}, \hat{q} \right) \right\} \\
    &\quad \exp \left\{ \alpha N \log \left[ \left\langle \mathbb{E}_\tau \exp \left\{ H_E \left( \tau, \tau_* ; m, s, q \right) \right\} \right\rangle_{\mathcal{M}_*} \right] - \alpha N \log \left[ \mathcal{Z} \left( \mathcal{M} \right) \right] \right\}
\end{align*}
where the thermal average $\left\langle \cdot \right\rangle_{\mathcal{M}_*}$ and the partition function $\mathcal{Z} \left( \mathcal{M} \right)$ are defined in Section \ref{sec:core_results} using Eqs. (\ref{hamiltonian_M_s}) and (\ref{hamiltonian_M}), respectively, $\alpha = \frac{M}{N}$ and
\begin{align*}
    H_Q \left( m, s, q, \hat{m}, \hat{s}, \hat{q} \right) &= \sum_{\gamma, \mu ; b} \hat{m}^{b \gamma \mu} m^{b \gamma \mu} + \sum_{\mu < \nu ; b} \hat{s}^{b \mu \nu} s^{b \mu \nu} + \sum_{\mu, \nu ; b < c} \hat{q}^{b c \mu \nu} q^{b c \mu \nu}, \\
    H_S \left( \xi_i, \xi^*_i ; \hat{m}, \hat{s}, \hat{q} \right) &= \sum_{\gamma, \mu ; b} \hat{m}^{b \gamma \mu} \xi^{* \gamma}_i \xi^{b \mu}_i + \sum_{\mu < \nu ; b} \hat{s}^{b \mu \nu} \xi^{b \mu}_i \xi^{b \nu}_i + \sum_{\mu, \nu ; b < c} \hat{q}^{b c \mu \nu} \xi^{b \mu}_i \xi^{c \nu}_i, \\
    H_E \left( \tau, \tau_* ; m, s, q \right) &= \beta^* \beta \sum_{\gamma, \mu ; b} m^{b \gamma \mu} \tau_{* \gamma} \tau_{b \mu} + \beta^2 \sum_{\mu < \nu ; b} s^{b \mu \nu} \tau_{b \mu} \tau_{b \nu} + \beta^2 \sum_{\mu, \nu ; b < c} q^{b c \mu \nu} \tau_{b \mu} \tau_{c \nu}.
\end{align*}
We use the replica symmetry ansatz to simplify $\mathbb{E} \left[ \mathcal{Z}^L \right]$ further. To be more precise, we assume that
\begin{equation*}
\begin{aligned}
    &m^{b \gamma \mu} = m^{\gamma \mu} \hspace{-7pt} &\text{and} &\hspace{7pt} \hat{m}^{b \gamma \mu} = \hat{m}^{\gamma \mu} &\text{for all} \ b ; \gamma, \mu, \\
    &s^{b \mu \nu} = s^{\mu \nu} \hspace{-7pt} &\text{and} &\hspace{7pt} \hat{s}^{b \mu \nu} = \hat{s}^{\mu \nu} &\text{for all} \ b ; \mu \neq \nu, \\
    &q^{b c \mu \nu} = q^{\mu \nu} \hspace{-7pt} &\text{and} &\hspace{7pt} \hat{q}^{b c \mu \nu} = \hat{q}^{\mu \nu} &\text{for all} \ b \neq c ; \mu, \nu.
\end{aligned}
\end{equation*}
Under this hypothesis, the term $\sum_{\mu, \nu ; b < c} q^{\mu \nu} \tau_{b \mu} \tau_{c \nu}$ can be rewritten as
\begin{align*}
    \sum_{\mu, \nu ; b < c} q^{\mu \nu} \tau_{b \mu} \tau_{c \nu} &= \frac{1}{2} \sum_{\mu, \nu ; b, c} q^{\mu \nu} \tau_{b \mu} \tau_{c \nu} - \frac{1}{2} \sum_{\mu, \nu ; b} q^{\mu \nu} \tau_{b \mu} \tau_{b \nu} \\
    &= \frac{1}{4} \sum_{\mu, \nu} q^{\mu \nu} \left[ \sum_b \left( \tau_{b \mu} + \tau_{b \nu} \right) \right]^2 - \frac{1}{4} \sum_{\mu, \nu} q^{\mu \nu} \left[ \sum_b \tau_{b \mu} \right]^2 \\
    &\quad- \frac{1}{4} \sum_{\mu, \nu} q^{\mu \nu} \left[ \sum_b \tau_{b \nu} \right]^2 - \frac{1}{2} \sum_{\mu, \nu ; b} q^{\mu \nu} \tau_{b \mu} \tau_{b \nu} \\
    &= \sum_{\mu, \nu} \frac{q^{\mu \nu} + q^{\nu \mu}}{2} \left[ \sum_b \frac{\tau_{b \mu} + \tau_{b \nu}}{2} \right]^2 - \frac{1}{2} \sum_{\mu, \nu} \frac{q^{\mu \nu} + q^{\nu \mu}}{2} \left[ \sum_b \tau_{b \mu} \right]^2 \\
    &\quad- \frac{1}{2} \sum_{\mu, \nu ; b} q^{\mu \nu} \tau_{b \mu} \tau_{b \nu} \\
    &= \frac{1}{2} \sum_{\mu, \nu} \left[ q^{\mu \nu} + q^{\nu \mu} - \delta_{\mu \nu} \sum_{\Tilde{\eta}} \frac{q^{\Tilde{\mu} \Tilde{\eta}} + q^{\Tilde{\eta} \Tilde{\mu}}}{2} \right] \left[ \sum_b \frac{\tau_{b \mu} + \tau_{b \nu}}{2} \right]^2 \\
    &\quad- \frac{1}{2} \sum_{\mu, \nu ; b} q^{\mu \nu} \tau_{b \mu} \tau_{b \nu}.
\end{align*}
Subsequently, the Hubbard-Stratonovich transformation gives
\begin{align*}
    H_E \left( \tau, \tau_* ; m, s, q \right) &= \beta^* \beta \sum_{\gamma, \mu ; b} m^{\gamma \mu} \tau_{* \gamma} \tau_{b \mu} + \beta^2 \sum_{\mu < \nu ; b} s^{\mu \nu} \tau_{b \mu} \tau_{b \nu} + \beta^2 \sum_{\mu, \nu ; b < c} q^{\mu \nu} \tau_{b \mu} \tau_{c \nu} \\
    &= \log \left( \mathbb{E}_{z} \left[ \prod_{b} \exp \left\{ \mathcal{L}_{\beta^*, \beta} \left( \tau_b, \tau_*, z ; m, s, q \right) \right\} \right] \right)
\end{align*}
where $\mathcal{L}_{\lambda_1, \lambda_2} \left( \xi, \xi^*, z ; m, s, q \right)$ is defined in \ref{sec:effective_hamiltonian} and $z = \left[ z_{\mu \nu} \right]_{\mu, \nu = 1}^{P}$ is a matrix of independent standard Gaussian random variables $z_{\mu \nu}$.
Similarly, we get
\begin{align*}
    H_S \left( \xi, \xi^* ; \hat{m}, \hat{s}, \hat{q} \right) &= \sum_{\gamma, \mu ; b} \hat{m}^{\gamma \mu} \xi^{* \gamma} \xi^{b \Tilde{\mu}} + \sum_{\mu < \nu ; b} \hat{s}^{\mu \nu} \xi^{b \Tilde{\mu}} \xi^{b \Tilde{\nu}} + \sum_{\mu, \nu ; b < c} \hat{q}^{\mu \nu} \xi^{b \Tilde{\mu}} \xi^{c \nu} \\
    &= \log \left( \mathbb{E}_{z} \left[ \prod_{b} \exp \left\{ \mathcal{L}_{1, 1} \left( \xi^b, \xi^*, z ; \hat{m}, \hat{s}, \hat{q} \right) \right\} \right] \right).
\end{align*}
We then factor $\mathbb{E} \left[ \mathcal{Z}^L \right]$ over the replicas and take the limit of $L \rightarrow 0, N \rightarrow \infty$ to obtain
\begin{align*}
    \label{eq:func_free_entropy}
    f &= \Extr_{m, \hat{m}, q, \hat{q}, s, \hat{s}} f \left( m, \hat{m}, q, \hat{q}, s, \hat{s} \right) \numberthis \\
    &= \Extr_{m, \hat{m}, q, \hat{q}, s, \hat{s}} \Bigg\{ -\sum_{\gamma, \mu} m^{\gamma \mu} \hat{m}^{\gamma \mu} - \frac{1}{2} \sum_{\mu \neq \nu} s^{\mu \nu} \hat{s}^{\mu \nu} + \frac{1}{2} \sum_{\mu, \nu} q^{\mu \nu} \hat{q}^{\mu \nu} \\
    &\quad+ \mathbb{E}_{\xi^*} \mathbb{E}_{z} \log \left[ \mathcal{Z} \left( \mathcal{L}^C \right) \right] + \alpha \left\langle \mathbb{E}_z \log \left[ \mathcal{Z} \left( \mathcal{L}^O \right) \right] \right\rangle_{\mathcal{M}_*} - \alpha \log \left[ \mathcal{Z} \left( \mathcal{M} \right) \right] \Bigg\},
\end{align*}
where we used the replica trick $\lim_{L \rightarrow 0} \left( \frac{1}{L} \log  \mathbb{E}_z \left[ \mathcal{Z} \left( \mathcal{L}_{\lambda_1, \lambda_2} \right)^L \right] \right) = \mathbb{E}_z \log \left[ \mathcal{Z} \left( \mathcal{L}_{\lambda_1, \lambda_2} \right) \right]$ to simplify the expectations over $z$. The partition functions $\mathcal{Z} \left( \mathcal{L}^O \right)$ and $\mathcal{Z} \left( \mathcal{L}^C \right)$ are defined in Section \ref{sec:core_results} using Eqs. (\ref{hamiltonian_L_O}) and (\ref{hamiltonian_L_C}), respectively.

\section{Saddle-point equations}
\label{app:RBM_saddle-point}
Our goal is to find the values of the order parameters for which the derivatives of $f \left( m, \hat{m}, q, \hat{q}, s, \hat{s} \right)$ vanish (see Eq. \ref{eq:func_free_entropy}). For that purpose, we need to evaluate
\begin{align*}
    \partial_{m^{\rho \iota}} \log \left[ \mathcal{Z} \left( \mathcal{L}_{\lambda_1, \lambda_2} \right) \right] &= \left\langle \partial_{m^{\rho \iota}} \mathcal{L}_{\lambda_1, \lambda_2} \right\rangle_{\mathcal{L}_{\lambda_1, \lambda_2}} \\
    \partial_{s^{\iota \kappa}} \log \left[ \mathcal{Z} \left( \mathcal{L}_{\lambda_1, \lambda_2} \right) \right] &= \left\langle \partial_{s^{\iota \kappa}} \mathcal{L}_{\lambda_1, \lambda_2} \right\rangle_{\mathcal{L}_{\lambda_1, \lambda_2}} \\
    \partial_{q^{\iota \kappa}} \log \left[ \mathcal{Z} \left( \mathcal{L}_{\lambda_1, \lambda_2} \right) \right] &= \left\langle \partial_{q^{\iota \kappa}} \mathcal{L}_{\lambda_1, \lambda_2} \right\rangle_{\mathcal{L}_{\lambda_1, \lambda_2}},
\end{align*}
so we must calculate the partial derivatives inside the expectation values.
We use equation (\ref{eq:non_sym_L}) for $\mathcal{L}_{\lambda_1, \lambda_2}$ because we do not know that $q$ is symmetric before deriving the saddle-point equations (see \ref{sec:effective_hamiltonian}). We obtain
\begin{align*}
    \partial_{m^{\rho \iota}} \mathcal{L}_{\lambda_1, \lambda_2} &= \lambda_1 \lambda_2 \xi^{* \rho} \xi^{\iota} \\
    \partial_{s^{\iota \kappa}} \mathcal{L}_{\lambda_1, \lambda_2} &= \frac{1}{2} \left[ \lambda_2 \right]^2 \xi^{\iota} \xi^{\kappa} \\
    \partial_{q^{\iota \kappa}} \mathcal{L}_{\lambda_1, \lambda_2}
    &= \frac{1}{4} \lambda_2 \left( A_{\iota \kappa}^{-1} \left[ z_{\iota \kappa} + z_{\kappa \iota} \right] \left[ \xi^{\iota} + \xi^{\kappa} \right] - A_{\iota \iota}^{-1} z_{\iota \iota} \xi^{\iota} - A_{\kappa \kappa}^{-1} z_{\kappa \kappa} \xi^{\kappa} \right) \\
    &\quad- \frac{1}{2} \left[ \lambda_2 \right]^2 \xi^{\iota} \xi^{\kappa},
\end{align*}
where $A_{\mu \nu}$ is defined in \ref{sec:effective_hamiltonian}.
Using the first two equalities, we immediately get
\begin{align*}
    \hat{m}^{\gamma \mu} &= \beta^* \beta \alpha \ \left\langle \mathbb{E}_{z} \left[ \tau_{* \gamma} \left\langle \tau_{\mu} \right\rangle_{\mathcal{L}^O} \right] \right\rangle_{\mathcal{M}_*} \\
    \hat{s}^{\mu \nu} &= \beta^2 \alpha \left( \left\langle \mathbb{E}_{z} \left[ \left\langle \tau_{\mu} \tau_{\nu} \right\rangle_{\mathcal{L}^O} \right] \right\rangle_{\mathcal{M}_*} - \left\langle \tau_{\mu} \tau_{\nu} \right\rangle_{\mathcal{M}} \right) \\
    m^{\gamma \mu} &= \mathbb{E}_{\xi^*} \mathbb{E}_{z} \left[ \xi^{* \gamma} \left\langle \xi^{\mu} \right\rangle_{\mathcal{L}^C} \right] \\
    s^{\mu \nu} &= \mathbb{E}_{\xi^*} \mathbb{E}_{z} \left[ \left\langle \xi^{\mu} \xi^{\nu} \right\rangle_{\mathcal{L}^C} \right],
\end{align*}
where the thermal averages $\left\langle \cdot \right\rangle_{\mathcal{M}_*}$, $\left\langle \cdot \right\rangle_{\mathcal{M}}$, $\left\langle \cdot \right\rangle_{\mathcal{L}^O}$ and $\left\langle \cdot \right\rangle_{\mathcal{L}^C}$ are defined in Section \ref{sec:core_results} using Eqs. (\ref{hamiltonian_M_s}), (\ref{hamiltonian_M}), (\ref{hamiltonian_L_O}) and (\ref{hamiltonian_L_C}), respectively. The case of the third derivative is a bit more involved. We find that its expectation with respect to the Gaussian variables $z$ can be expressed as
\begin{align*}
    \mathbb{E}_z \left[ \left\langle \partial_{q^{\iota \kappa}} \mathcal{L}_{\lambda_1, \lambda_2} \right\rangle_{\mathcal{L}_{\lambda_1, \lambda_2}} \right]
    &= \mathbb{E}_z \bigg[ \frac{1}{4} \lambda_2 \Big( A_{\iota \kappa}^{-1} \left[ z_{\iota \kappa} + z_{\kappa \iota} \right] \left\langle \xi^{\iota} + \xi^{\kappa} \right\rangle_{\mathcal{L}_{\lambda_1, \lambda_2}} - A_{\iota \iota}^{-1} z_{\iota \iota} \left\langle \xi^{\iota} \right\rangle_{\mathcal{L}_{\lambda_1, \lambda_2}} \\
    &\quad- A_{\kappa \kappa}^{-1} z_{\kappa \kappa} \left\langle \xi^{\kappa} \right\rangle_{\mathcal{L}_{\lambda_1, \lambda_2}} \Big) - \frac{1}{2} \left[ \lambda_2 \right]^2 \langle \xi^{\iota} \xi^{\kappa} \rangle_{\mathcal{L}_{\lambda_1, \lambda_2}} \bigg] \\
    &= \mathbb{E}_z \bigg[ \frac{1}{4} \lambda_2 \Big( A_{\iota \kappa}^{-1} \Big[ \partial_{z_{\iota \kappa}} \left\langle \xi^{\iota} \right\rangle_{\mathcal{L}_{\lambda_1, \lambda_2}} + \partial_{z_{\kappa \iota}} \left\langle \xi^{\iota} \right\rangle_{\mathcal{L}_{\lambda_1, \lambda_2}} \\
    &\quad+ \partial_{z_{\iota \kappa}} \left\langle \xi^{\kappa} \right\rangle_{\mathcal{L}_{\lambda_1, \lambda_2}} + \partial_{z_{\kappa \iota}} \left\langle \xi^{\kappa} \right\rangle_{\mathcal{L}_{\lambda_1, \lambda_2}} \Big] - A_{\iota \iota}^{-1} \partial_{z_{\iota \iota}} \left\langle \xi^{\iota} \right\rangle_{\mathcal{L}_{\lambda_1, \lambda_2}} \\
    &\quad- A_{\kappa \kappa}^{-1} \partial_{z_{\kappa \kappa}} \left\langle \xi^{\kappa} \right\rangle_{\mathcal{L}_{\lambda_1, \lambda_2}} \Big) - \frac{1}{2} \left[ \lambda_2 \right]^2 \langle \xi^{\iota} \xi^{\kappa} \rangle_{\mathcal{L}_{\lambda_1, \lambda_2}} \bigg]
\end{align*}
by using integration by parts. As known from linear response theory \cite{chandler1987intro}, any Gibbs distribution with a generic Hamiltonian $\mathcal{H}$ verifies the identity
\begin{align}
    \label{eq:linear_response_identity}
    \partial_x \left\langle \theta \right\rangle_{\mathcal{H}} = \left\langle \theta \partial_x \mathcal{H} \right\rangle_{\mathcal{H}} - \left\langle \theta \right\rangle_{\mathcal{H}} \left\langle \partial_x \mathcal{H} \right\rangle_{\mathcal{H}}
\end{align}
for any order parameter $\theta$ that does not depend on $x$. We use this identity to get
\begin{align*}
    \partial_{z_{\iota \kappa}} \left\langle \xi^{\iota} \right\rangle_{\mathcal{L}_{\lambda_1, \lambda_2}} &= \left\langle \xi^{\iota} \partial_{z_{\iota \kappa}} \mathcal{L}_{\lambda_1, \lambda_2} \right\rangle_{\mathcal{L}_{\lambda_1, \lambda_2}} - \left\langle \xi^{\iota} \right\rangle_{\mathcal{L}_{\lambda_1, \lambda_2}} \left\langle \partial_{z_{\iota \kappa}} \mathcal{L}_{\lambda_1, \lambda_2} \right\rangle_{\mathcal{L}_{\lambda_1, \lambda_2}} \\
    &= \lambda_2 \left\langle A_{\iota \kappa} \xi^{\iota} \frac{\xi^{\iota} + \xi^{\kappa}}{2} \right\rangle_{\mathcal{L}_{\lambda_1, \lambda_2}} - \lambda_2 \left\langle \xi^{\iota} \right\rangle_{\mathcal{L}_{\lambda_1, \lambda_2}} \left\langle A_{\iota \kappa} \frac{\xi^{\iota} + \xi^{\kappa}}{2} \right\rangle_{\mathcal{L}_{\lambda_1, \lambda_2}} \\
    &= \frac{1}{2} \lambda_2 A_{\iota \kappa} \left( \left\langle \xi^{\iota} \xi^{\kappa} \right\rangle_{\mathcal{L}_{\lambda_1, \lambda_2}} - \left\langle \xi^{\iota} \right\rangle_{\mathcal{L}_{\lambda_1, \lambda_2}} \left\langle \xi^{\kappa} \right\rangle_{\mathcal{L}_{\lambda_1, \lambda_2}} \right) \\
    &\quad+ \frac{1}{2} \lambda_2 A_{\iota \kappa} \left( \left\langle \left[ \xi^{\iota} \right]^2 \right\rangle_{\mathcal{L}_{\lambda_1, \lambda_2}} - \left[ \left\langle \xi^{\iota} \right\rangle_{\mathcal{L}_{\lambda_1, \lambda_2}} \right]^2 \right),
\end{align*}
Coming back to the previous expression, we obtain
\begin{align*}
    \mathbb{E}_z \left[ \left\langle \partial_{q^{\iota \kappa}} \mathcal{L}_{\lambda_1, \lambda_2} \right\rangle_{\mathcal{L}_{\lambda_1, \lambda_2}} \right]
    &= -\frac{1}{2} \left[ \lambda_2 \right]^2 \mathbb{E}_z \bigg[ \left\langle \xi^{\iota} \right\rangle_{\mathcal{L}_{\lambda_1, \lambda_2}} \left\langle \xi^{\kappa} \right\rangle_{\mathcal{L}_{\lambda_1, \lambda_2}} \bigg].
\end{align*}
From this result, we find that the two remaining saddle-point equations are
\begin{align*}
    \hat{q}^{\mu \nu} &= \beta^2 \alpha \left\langle \mathbb{E}_{z} \left[ \left\langle \tau_{\mu} \right\rangle_{\mathcal{L}^O} \left\langle \tau_{\Tilde{\nu}} \right\rangle_{\mathcal{L}^O} \right] \right\rangle_{\mathcal{M}_*} \\
    q^{\mu \nu} &= \mathbb{E}_{\xi^*} \mathbb{E}_{z} \left[ \left\langle \xi^{\mu} \right\rangle_{\mathcal{L}^C} \left\langle \xi^{\nu} \right\rangle_{\mathcal{L}^C} \right].
\end{align*}

\section{Saddle-point equations for Gaussian $\xi$}
\label{app:gaussian_xi}
In this Appendix, we take the prior $\Prob \left( \xi \right)$ on the student pattern to be a standard Gaussian distribution. As per Section \ref{sec:core_results},
\begin{align*}
    \mathcal{P} \left[ \mathcal{L}^C \right] \left( \xi ; \xi^*, z \right) &= \mathcal{Z} \left( \mathcal{L}^C \right)^{-1} \frac{1}{\sqrt{\left( 2\pi \right)^{P}}} \exp \left( -\frac{1}{2} \sum_{\mu} \left[ \xi^{\mu} \right]^2 \right) \exp \left[ \mathcal{L}^C \left( \xi ; \xi^*, z \right) \right] \\
    &= \mathcal{Z} \left( \mathcal{L}^C \right)^{-1} \frac{1}{\sqrt{\left( 2\pi \right)^{P}}} \exp \left( \mathcal{L}^C \left( \xi ; \xi^*, z \right) - \frac{1}{2} \xi^T \xi \right),
\end{align*}
where the second line is written in matrix notation. This expression will allow us to simplify the thermal averages $\left\langle \cdot \right\rangle_{\mathcal{L}^C}$ in the saddle-point equations (Eqs. \ref{eq:RBM_saddle-point}). By completing the square, $\mathcal{L}^C \left( \xi ; \xi^*, z \right) - \frac{1}{2} \xi^T \xi$ can be written as
\begin{align*}
    \mathcal{L}^C \left( \xi, \xi^*, z \right) - \frac{1}{2} \xi^T \xi &= -\frac{1}{2} \left( \xi - \left[ I + \hat{q} - \hat{s} \right]^{-1} \left[ \hat{m}^T \xi^{*} + \diag \left\{ A \left( \hat{q} \right) \frac{z + z^T}{2} \right\} \right] \right)^T \\
    &\quad \left( I + \hat{q} - \hat{s} \right) \left( \xi - \left[ I + \hat{q} - \hat{s} \right]^{-1} \left[ \hat{m}^T \xi^{*} + \diag \left\{ A \left( \hat{q} \right) \frac{z + z^T}{2} \right\} \right] \right) \\
    &\quad+ \frac{1}{2} \left[ \hat{m}^T \xi^{*} + \diag \left\{ A \left( \hat{q} \right) \frac{z + z^T}{2} \right\} \right]^T \\
    &\quad \left[ I + \hat{q} - \hat{s} \right]^{-1} \left[ \hat{m}^T \xi^{*} + \diag \left\{ A \left( \hat{q} \right) \frac{z + z^T}{2} \right\} \right],
\end{align*}
where the function $\diag \left( \hat{q} \right)$ returns the diagonal of $\hat{q}$. We read out the mean and variance as
\begin{gather*}
    \left\langle \xi \right\rangle_{\mathcal{L}^C} = \left[ I + \hat{q} - \hat{s} \right]^{-1} \left[ \hat{m}^T \xi^{*} + \diag \left\{ A \left( \hat{q} \right) \frac{z + z^T}{2} \right\} \right], \\
    \left\langle \xi \xi^T \right\rangle_{\mathcal{L}^C} - \left\langle \xi \right\rangle_{\mathcal{L}^C} \left\langle \xi \right\rangle_{\mathcal{L}^C}^T = \left[ I + \hat{q} - \hat{s} \right]^{-1}.
\end{gather*}
By evaluating the thermal averages $\left\langle \cdot \right\rangle_{\mathcal{L}^C}$ of the saddle-point equations, we then obtain
\begin{align*}
    m &= \mathbb{E}_{\xi^*} \left[ \xi^{*} \xi^{* T} \hat{m} \left[ I + \hat{q} - \hat{s} \right]^{-1} \right] \\
    q &= \mathbb{E}_{\xi^*} \left[ \left[ I + \hat{q} - \hat{s} \right]^{-1} \hat{m}^T \xi^* \xi^{* T} \hat{m} \left[ I + \hat{q} - \hat{s} \right]^{-1} \right] \\
    &\quad+ \mathbb{E}_z \left[ \left[ I + \hat{q} - \hat{s} \right]^{-1} \diag \left\{ A \left( \hat{q} \right) \frac{z + z^T}{2} \right\} \diag \left\{ A \left( \hat{q} \right) \frac{z + z^T}{2} \right\}^T \left[ I + \hat{q} - \hat{s} \right]^{-1} \right] \\
    s &= \left[ I + \hat{q} - \hat{s} \right]^{-1} + \mathbb{E}_{\xi^*} \left[ \left[ I + \hat{q} - \hat{s} \right]^{-1} \hat{m}^T \xi^* \xi^{* T} \hat{m} \left[ I + \hat{q} - \hat{s} \right]^{-1} \right] \\
    &\quad+ \mathbb{E}_z \left[ \left[ I + \hat{q} - \hat{s} \right]^{-1} \diag \left\{ A \left( \hat{q} \right) \frac{z + z^T}{2} \right\} \diag \left\{ A \left( \hat{q} \right) \frac{z + z^T}{2} \right\}^T \left[ I + \hat{q} - \hat{s} \right]^{-1} \right].
\end{align*}
The expected value over $z$ simplifies to
\begin{align*}
    &\mathbb{E}_z \left[ \diag \left\{ A \left( \hat{q} \right) \frac{z + z^T}{2} \right\} \diag \left\{ A \left( \hat{q} \right) \frac{z + z^T}{2} \right\}^T \right]_{\mu \nu} \\
    &= \mathbb{E}_z \left[ \sum_{\Tilde{\kappa}} A_{\Tilde{\mu} \Tilde{\kappa}} \left( \hat{q} \right) \frac{z_{\Tilde{\kappa} \Tilde{\mu}} + z_{\Tilde{\mu} \Tilde{\kappa}}}{2} \sum_{\Tilde{\iota}} A_{\Tilde{\nu} \Tilde{\iota}} \left( \hat{q} \right) \frac{z_{\Tilde{\iota} \Tilde{\nu}} + z_{\Tilde{\nu} \Tilde{\iota}}}{2} \right] \\
    &= \frac{1}{4} \mathbb{E}_z \left[ \sum_{\Tilde{\kappa}, \Tilde{\iota}} A_{\Tilde{\mu} \Tilde{\kappa}} \left( \hat{q} \right) A_{\Tilde{\nu} \Tilde{\iota}} \left( \hat{q} \right) z_{\Tilde{\kappa} \Tilde{\mu}} z_{\Tilde{\iota} \Tilde{\nu}} \right] \\
    &\quad+ \frac{1}{4} \mathbb{E}_z \left[ \sum_{\Tilde{\kappa}, \Tilde{\iota}} A_{\Tilde{\mu} \Tilde{\kappa}} \left( \hat{q} \right) A_{\Tilde{\nu} \Tilde{\iota}} \left( \hat{q} \right) z_{\Tilde{\kappa} \Tilde{\mu}} z_{\Tilde{\nu} \Tilde{\iota}} \right] \\
    &\quad+ \frac{1}{4} \mathbb{E}_z \left[ \sum_{\Tilde{\kappa}, \Tilde{\iota}} A_{\Tilde{\mu} \Tilde{\kappa}} \left( \hat{q} \right) A_{\Tilde{\nu} \Tilde{\iota}} \left( \hat{q} \right) z_{\Tilde{\mu} \Tilde{\kappa}} z_{\Tilde{\iota} \Tilde{\nu}} \right] \\
    &\quad+ \frac{1}{4} \mathbb{E}_z \left[ \sum_{\Tilde{\kappa}, \Tilde{\iota}} A_{\Tilde{\mu} \Tilde{\kappa}} \left( \hat{q} \right) A_{\Tilde{\nu} \Tilde{\iota}} \left( \hat{q} \right) z_{\Tilde{\mu} \Tilde{\kappa}} z_{\Tilde{\nu} \Tilde{\iota}} \right] \\
    &= \frac{1}{2} \delta_{\mu \nu} \sum_{\Tilde{\iota}} \left[ A_{\Tilde{\mu} \Tilde{\iota}} \left( \hat{q} \right) \right]^2 + \frac{1}{2} \left[ A_{\mu \nu} \left( \hat{q} \right) \right]^2 \\
    &= \hat{q}^{\mu \nu}.
\end{align*}
In the end, we find
\begin{align*}
    \label{eq:normal_saddle-point}
    m &= \mathcal{Q} \hat{m} \left[ I + \hat{q} - \hat{s} \right]^{-1} \\
    q &= \left[ I + \hat{q} - \hat{s} \right]^{-1} \hat{m}^T \mathcal{Q} \hat{m} \left[ I + \hat{q} - \hat{s} \right]^{-1} \\
    &\quad+ \left[ I + \hat{q} - \hat{s} \right]^{-1} \hat{q} \left[ I + \hat{q} - \hat{s} \right]^{-1} \numberthis \\
    s &= \left[ I + \hat{q} - \hat{s} \right]^{-1} + \left[ I + \hat{q} - \hat{s} \right]^{-1} \hat{m}^T \mathcal{Q} \hat{m} \left[ I + \hat{q} - \hat{s} \right]^{-1} \\
    &\quad+ \left[ I + \hat{q} - \hat{s} \right]^{-1} \hat{q} \left[ I + \hat{q} - \hat{s} \right]^{-1}.
\end{align*}

\section{Critical load}
\label{app:critical_load}
In the paramagnetic phase, the order parameters all vanish. The paramagnetic to ferromagnetic phase transition of the student RBM is the line where the paramagnetic solution of the saddle-point equations (see Eq. \ref{eq:RBM_saddle-point}) becomes unstable to leading order in the order parameters (see Fig. \ref{fig:RBM_phase_diagram}). As a consequence of Eq. (\ref{eq:linear_response_identity}), any Hamiltonian of the form $\mathcal{H} \left( \xi \right) = \frac{1}{2} \sum_{\mu \neq \nu} J_{\mu \nu} \xi^{\mu} \xi^{\nu} + \sum_{\title{\mu}} h_{\mu} \xi^{\mu}$ has
\begin{align*}
    \left\langle \xi^{\mu} \right\rangle_{\mathcal{H}} \approx h_{\mu}
\end{align*}
to first order in the parameters $J_{\mu \nu}$ and $h_{\mu}$. Therefore, given that the prior on $\xi$ has a mean of zero, we have
\begin{align*}
    \mathbb{E}_{z} \left[ \xi^{* \gamma} \left\langle \xi^{\mu} \right\rangle_{\mathcal{L}_{\lambda_1, \lambda_2}} \right] &\approx \lambda_1 \lambda_2 \mathbb{E}_{z} \left[ \xi^{* \gamma} \sum_{\rho} m^{\rho \mu} \xi^{* \rho} \right] \\
    &= \lambda_1 \lambda_2 \sum_{\rho} m^{\rho \mu} \xi^{* \gamma} \xi^{* \rho}
\end{align*}
to first order in $m$, $s$ and $q$. The saddle-point equations for $\hat{m}$ and $m$ then simplify to
\begin{align*}
    \hat{m}^{\gamma \mu} &= \beta^* \beta \alpha \ \left\langle \mathbb{E}_{z} \left[ \tau_{* \gamma} \left\langle \tau_{\mu} \right\rangle_{\mathcal{L}^O} \right] \right\rangle_{\mathcal{M}_*} \\
    &= \left[ \beta^* \beta \right]^2 \alpha \sum_{\rho} \left\langle \tau_{* \gamma} \tau_{* \rho} \right\rangle_{\mathcal{M}_*} m^{\rho \mu}  \\
    &= \left[ \beta^* \beta \right]^2 \alpha \sum_{\rho} \mathcal{R}_{\mu \rho} m^{\rho \mu} \\
    m^{\gamma \mu} &= \mathbb{E}_{\xi^*} \mathbb{E}_{z} \left[ \xi^{* \gamma} \left\langle \xi^{\mu} \right\rangle_{\mathcal{L}^C} \right] \\
    &= \sum_{\rho} \mathbb{E}_{\xi^*} \left[ \xi^{* \gamma} \xi^{* \rho} \right] \hat{m}^{\rho \mu} \\
    &= \sum_{\rho} \mathcal{Q}_{\mu \rho} \hat{m}^{\rho \mu},
\end{align*}
where $\mathcal{Q}$ and $\mathcal{R}$ are the covariance matrices of $\xi^*$ and $\mathcal{M}_*$, respectively. We see that the behavior of $\hat{m}$ and $m$ is much simpler near criticality than at an arbitrary location in the phase diagram. In fact, the stationary values of $\hat{m}$ and $m$ do not depend on the other order parameters. We can even rewrite the equations for $\hat{m}$ and $m$ in the more compact form
\begin{align*}
    m^{\gamma \mu}
    &= \left[ \beta^* \beta \right]^2 \alpha \sum_{\iota, \kappa} \mathcal{Q}_{\mu \iota} \mathcal{R}_{\iota \kappa} m^{\kappa \mu}.
\end{align*}
Let the largest eigenvalue of $\mathcal{S}_{\mu \kappa} = \sum_{\iota} \mathcal{Q}_{\mu \iota} \mathcal{R}_{\iota \kappa}$ be $\lambda^{\mathcal{S}}_{\text{max}}$. As known from stability theory \cite{strogatz2018nonlinear}, the paramagnetic solution $m^{\gamma \mu} = 0$ is unstable when $\left[ \beta^* \beta \right]^2 \alpha \ \mathcal{S}$ has at least one eigenvalue larger than $1$. In other words, the student is able to learn the teacher patterns above a critical load of
\begin{align*}
    \alpha_{\text{crit}} = \frac{1}{\left[ \beta^* \beta \right]^2 \lambda^{\mathcal{S}}_{\text{max}}}.
\end{align*}

\section{Saddle-point equations in the absence of correlations}
\label{app:no_correlations}
In the absence of correlations, we have $\mathcal{Q} = \mathbf{I}$. The effective Hamiltonian $\mathcal{M_*}$ (see Eq. \ref{hamiltonian_M_s}) then simplifies to
\begin{align*}
    \mathcal{M}_* \left( \tau_* \right) = \frac{1}{2} P^* \left[ \beta^* \right]^2,
\end{align*}
so the expectation $\left\langle \cdot \right\rangle_{\mathcal{M}_*}$ is uniform. When $P \leq P^*$, i.e. the student has at most the same number of hidden units as the teacher, we make the ansatz
\begin{equation}
\label{eq:small_P_t_ansatz}
\begin{aligned}
    &m^{\gamma \mu} = \delta_{\gamma \mu} m, &\quad &\hat{m}^{\gamma \mu} = \delta_{\gamma \mu} \hat{m}, \\
    &s^{\mu \nu} = \delta_{\mu \nu}, &\quad &\hat{s}^{\mu \nu} = 0, \\
    &q^{\mu \nu} = \delta_{\mu \nu} q, &\quad &\hat{q}^{\mu \nu} = \delta_{\mu \nu} \hat{q},
\end{aligned}
\end{equation}
under which $\mathcal{M}$ (Eq. \ref{hamiltonian_M}) and $\mathcal{L}_{\lambda_1, \lambda_2}$ (Eq. \ref{eq:sym_L}) respectively simplify to
\begin{gather*}
    \mathcal{M} \left( \tau \right) = \frac{1}{2} P \beta^2 \quad \text{and} \\
    \mathcal{L}_{\lambda_1, \lambda_2} \left( \xi^*, \xi, z ; m, s, q \right) = \frac{1}{2} P \left[ \lambda_2 \right]^2 \left( 1 - q \right) + \sum_{\mu = 1}^{P} \left( \lambda_1 \lambda_2 m \xi^{* \mu} + \lambda_2 \sqrt{q} z_{\mu} \right) \xi^{\mu}.
\end{gather*}
The spins $\xi^{\mu}$ do not interact with one another in $\mathcal{L}_{\lambda_1, \lambda_2} \left( \xi^*, \xi, z ; m, s, q \right)$. In other words, they are independent with respect to the Gibbs distribution with Hamiltonian $\mathcal{L}_{\lambda_1, \lambda_2}$. Therefore, all of the spins with $\nu \neq \mu$ can be marginalized from the thermal average $\left\langle \xi^{\mu} \right\rangle_{\mathcal{L}_{\lambda_1, \lambda_2}}$. In particular, when the student patterns $\xi$ are binary random variables, we obtain
\begin{align*}
    \left\langle \xi^{\mu} \right\rangle_{\mathcal{L}_{\lambda_1, \lambda_2}} &= \sum_{\xi^{\mu} = \pm 1} \frac{\exp \left( \left[ \lambda_1 \lambda_2 m \xi^{* \mu} + \lambda_2 \sqrt{q} z_{\mu} \right] \xi^{\mu} \right) \xi^{\mu}}{\exp \left( \lambda_1 \lambda_2 m \xi^{* \mu} + \lambda_2 \sqrt{q} z_{\mu} \right) + \exp \left( -\lambda_1 \lambda_2 m \xi^{* \mu} - \lambda_2 \sqrt{q} z_{\mu} \right)} \\
    &= \tanh \left( \lambda_1 \lambda_2 m \xi^{* \mu} + \lambda_2 \sqrt{q} z_{\mu} \right).
\end{align*}
Similarly, the hidden unit spins $\tau_{\mu}$ are independent with respect to the Gibbs distribution with Hamiltonian $\mathcal{M} \left( \tau \right)$. By the independence of all these spins, the saddle-point equations (Eqs. \ref{eq:RBM_saddle-point}) with binary $\xi$ reduce to
\begin{align*}
    \hat{m}^{\gamma \mu} &= \beta^* \beta \alpha \ \delta_{\gamma \mu} \ \left\langle \mathbb{E}_{z} \left[ \tau_{* \mu} \tanh \left( \beta^* \beta m \tau_{* \mu} + \beta \sqrt{q} z_{\mu} \right) \right] \right\rangle_{\mathcal{M}_*} \\
    \hat{s}^{\mu \nu} &= 0 \\
    \hat{q}^{\mu \nu} &= \beta^2 \alpha \ \delta_{\mu \nu} \ \left\langle \mathbb{E}_{z} \left[ \tanh^2 \left( \beta^2 m \tau_{* \mu} + \beta \sqrt{q} z_{\mu} \right) \right] \right\rangle_{\mathcal{M}_*} \\
    m^{\gamma \mu} &= \delta_{\gamma \mu} \ \mathbb{E}_{\xi^*} \mathbb{E}_{z} \left[ \xi^{* \mu} \tanh \left( \hat{m} \xi^{* \mu} + \sqrt{\hat{q}} z_{\mu} \right) \right] \\
    s^{\mu \nu} &= 0 \\
    q^{\mu \nu} &= \delta_{\mu \nu} \ \mathbb{E}_{\xi^*} \mathbb{E}_{z} \left[ \tanh^2 \left( \hat{m} \xi^{* \mu} + \sqrt{\hat{q}} z_{\mu} \right) \right].
\end{align*}
Assume the teacher patterns $\xi^*$ are also binary. Since $\tanh$ is an odd function, we factor the spins out of it according to $\tanh \left( \xi^{* \mu} y \right) = \xi^{* \mu} \tanh \left( y \right)$ and use the change of variables $z = \xi^{* \mu} z_{\mu}$ to simplify the saddle-point equations to
\begin{align*}
    \hat{m} &= \beta^* \beta \alpha \ \mathbb{E}_{z} \left[ \tanh \left( \beta^* \beta m + \beta \sqrt{q} z \right) \right] \\
    \hat{q} &= \beta^2 \alpha \ \mathbb{E}_{z} \left[ \tanh^2 \left( \beta^2 m + \beta \sqrt{q} z \right) \right] \\
    m &= \mathbb{E}_{z} \left[ \tanh \left( \hat{m} + \sqrt{\hat{q}} z \right) \right] \\
    q &= \mathbb{E}_{z} \left[ \tanh^2 \left( \hat{m} + \sqrt{\hat{q}} z \right) \right].
\end{align*}
On the Nishimori line $\beta^* = \beta$, we have
\begin{align*}
    \hat{m} &= \beta^2 \alpha \ \mathbb{E}_{z} \left[ \tanh \left( \beta^2 m + \beta \sqrt{m} z \right) \right] \\
    m &= \mathbb{E}_{z} \left[ \tanh \left( \hat{m} + \sqrt{\hat{m}} z \right) \right].
\end{align*}
When $P > P^*$, the saddle-point equations are slightly different. We make the ansatz
\begin{equation}
\label{eq:large_P_t_ansatz}
\begin{aligned}
    &m^{\gamma \mu} = \delta_{\gamma \mu} m, &\quad &\hat{m}^{\gamma \mu} = \delta_{\gamma \mu} \hat{m}, \\
    &s^{\mu \nu} = \delta_{\mu \nu}, &\quad &\hat{s}^{\mu \nu} = 0, \\
    &q^{\mu \nu}
    = \begin{cases}
        \delta_{\mu \nu} q &\quad \mu, \nu \leq P^* \\
        \delta_{\mu \nu} g &\quad \text{otherwise},
    \end{cases} &\quad &\hat{q}^{\mu \nu}
    = \begin{cases}
        \delta_{\mu \nu} \hat{q} &\quad \mu, \nu \leq P^* \\
        \delta_{\mu \nu} \hat{g} &\quad \text{otherwise},
    \end{cases}
\end{aligned}
\end{equation}
and obtain
\begin{align*}
    \mathcal{L}_{\lambda_1, \lambda_2} \left( \xi^*, \xi, z ; m, s, q \right)
    &= \sum_{\mu = 1}^{P^*} \left( \lambda_1 \lambda_2 m \xi^{* \mu} + \lambda_2 \sqrt{q} z_{\mu} \right) \xi^{\mu} + \sum_{\mu = P^* + 1}^{P} \lambda_2 \sqrt{g} z_{\mu} \xi^{\mu}.
\end{align*}
Following the same steps as for $P \leq P^*$, we get
\begin{align*}
    \hat{m} &= \beta^* \beta \alpha \ \mathbb{E}_{z} \left[ \tanh \left( \beta^* \beta m + \beta \sqrt{q} z \right) \right] \\
    \hat{q} &= \beta^2 \alpha \ \mathbb{E}_{z} \left[ \tanh^2 \left( \beta^2 m + \beta \sqrt{q} z \right) \right] \\
    \hat{g} &= \beta^2 \alpha \ \mathbb{E}_{z} \left[ \tanh^2 \left( \beta \sqrt{g} z \right) \right] \\
    m &= \mathbb{E}_{z} \left[ \tanh \left( \hat{m} + \sqrt{\hat{q}} z \right) \right] \\
    q &= \mathbb{E}_{z} \left[ \tanh^2 \left( \hat{m} + \sqrt{\hat{q}} z \right) \right] \\
    g &= \mathbb{E}_{z} \left[ \tanh^2 \left( \sqrt{\hat{g}} z \right) \right].
\end{align*}
When $\beta = \beta^*$, these equations reduce to
\begin{align*}
    \hat{m} &= \beta^2 \alpha \ \mathbb{E}_{z} \left[ \tanh \left( \beta^2 m + \beta \sqrt{m} z \right) \right] \\
    \hat{g} &= \beta^2 \alpha \ \mathbb{E}_{z} \left[ \tanh^2 \left( \beta \sqrt{g} z \right) \right] \\
    m &= \mathbb{E}_{z} \left[ \tanh \left( \hat{m} + \sqrt{\hat{m}} z \right) \right] \\
    g &= \mathbb{E}_{z} \left[ \tanh^2 \left( \sqrt{\hat{g}} z \right) \right].
\end{align*}
When the student patterns $\xi$ are Gaussian random variables, we find instead
\begin{align*}
    \hat{m} &= \beta^* \beta \alpha \ \mathbb{E}_{z} \left[ \tanh \left( \beta^* \beta m + \beta \sqrt{q} z \right) \right] \\
    \hat{q} &= \beta^2 \alpha \ \mathbb{E}_{z} \left[ \tanh^2 \left( \beta^2 m + \beta \sqrt{q} z \right) \right] \\
    \hat{g} &= \beta^2 \alpha \ \mathbb{E}_{z} \left[ \tanh^2 \left( \beta \sqrt{g} z \right) \right] \\
    m &= \frac{\hat{m}}{1 + \hat{q}} \\
    q &= \frac{\hat{m}^2}{\left( 1 + \hat{q} \right)^2} + \frac{\hat{q}}{\left( 1 + \hat{q} \right)^2} \\
    g &= \frac{\hat{g}}{\left( 1 + \hat{g} \right)^2}.
\end{align*}
When $\beta = \beta^*$, these equations reduce to
\begin{align*}
    \hat{m} &= \beta^2 \alpha \ \mathbb{E}_{z} \left[ \tanh \left( \beta^2 m + \beta \sqrt{m} z \right) \right] \\
    \hat{g} &= \beta^2 \alpha \ \mathbb{E}_{z} \left[ \tanh^2 \left( \beta \sqrt{g} z \right) \right] \\
    m &= \frac{\hat{m}}{1 + \hat{m}} \\
    g &= \frac{\hat{g}}{\left( 1 + \hat{g} \right)^2}.
\end{align*}

\section{Effect of uniform correlations}
\label{app:uniform_correlations}
We introduce uniform correlations in the teacher patterns by fixing the covariance matrix $\mathcal{Q}$ of $\xi^*$ to $\mathcal{Q}_{\gamma \rho} = \delta_{\gamma \rho} + \left( 1 - \delta_{\gamma \rho} \right) c$, where $c \in \left[ 0, 1 \right)$. Given this particular $\mathcal{Q}$, the Hamiltonian $\mathcal{M}_*$ (see Eq. \ref{hamiltonian_M_s}) simplifies to
\begin{align}
    \label{eq:hamiltonian_M_s}
    \mathcal{M}_* \left( \tau_* \right) &= \frac{1}{2} \left[ \beta^* \right]^2 c \sum_{\gamma \neq \rho} \tau_{* \gamma} \tau_{* \rho} + \frac{1}{2} P^* \left[ \beta^* \right]^2.
\end{align}
The interaction between any two spins $\tau_{* \gamma}$ and $\tau_{* \rho}$ does not depend on the sites $\gamma \neq \rho$. Therefore, the covariance matrix $\mathcal{R}$ of $\tau_*$ has the same form as $\mathcal{Q}$, but with a different coefficient $d$ outside the diagonal. $\mathcal{S} = \mathcal{Q} \mathcal{R}$ then reduces to
\begin{align*}
    \mathcal{S}_{\gamma \rho} &= \sum_{\tau} \left( c + \left( 1 - c \right) \delta_{\gamma \tau} \right) \left( d + \left( 1 - d \right) \delta_{\tau \rho} \right) \\
    &= P^* c d + c \left( 1 - d \right) + \left( 1 - c \right) d + \left( 1 - c \right) \left( 1 - d \right) \delta_{\gamma \rho}
\end{align*}
Any $P^* \times P^*$ matrix of the form $\mathcal{A}_{\gamma \rho} = a + b \delta_{\gamma \rho}$ has eigenvalues
\begin{align*}
    \lambda^{\mathcal{A}}_{1} &= P^* a + b
    \text{ with corresponding eigenvector } e = \frac{1}{\sqrt{P^*}} \begin{bmatrix} 1 & ... & 1 \end{bmatrix}, \\
    \lambda^{\mathcal{A}}_{2} &= b \text{ with corresponding eigenspace } \left\{ x \in \mathbb{R}^{P^*} \bigg| \sum\nolimits_i x_i = 0 \right\}.
\end{align*}
Therefore, $\mathcal{S}$ has eigenvalues
\begin{align*}
    \lambda^{\mathcal{S}}_{1} &= P^* \left( P^* c d + c \left( 1 - d \right) + \left( 1 - c \right) d \right) + \left( 1 - c \right) \left( 1 - d \right), \\
    \lambda^{\mathcal{S}}_{2} &= \left( 1 - c \right) \left( 1 - d \right).
\end{align*}
$d$ is positive because the interaction between any two spins $\tau_{* \gamma}$ and $\tau_{* \rho}$ is positive (see Eq. \ref{eq:hamiltonian_M_s} and \cite{griffiths1967correlations}). Therefore, the largest eigenvalue $\lambda^{\mathcal{S}}_{\text{max}}$ of $\mathcal{S}$ is $\lambda^{\mathcal{S}}_1$. In sum,
\begin{align*}
    \lambda^{\mathcal{S}}_{\text{max}} = \lambda^{\mathcal{S}}_{1} = \left( P^* - 1 \right)^2 c d + \left( P^* - 1 \right) \left( c + d \right) + 1.
\end{align*}

\section{Numerical methods}
\label{app:numerical_methods}
We solve the saddle-point equations (Eqs. \ref{eq:RBM_saddle-point}) by numerical iteration. To be more specific, we iterate
\begin{align*}
    \hat{m}^{\gamma \mu} (t + 1) &= \hat{m}^{\gamma \mu} (t) + \Delta t \left( \beta^* \beta \alpha \ \mathbb{E}_{\mathcal{M}_*} \mathbb{E}_{z} \left[ \tau_{* \gamma} \left\langle \tau_{\mu} \right\rangle_{\mathcal{L}^O (t + 1)} \right] - \hat{m}^{\gamma \mu} (t) \right) \\
    \hat{s}^{\mu \nu} (t + 1) &= \hat{s}^{\mu \nu} (t) + \Delta t \left( \beta^2 \alpha \left( \mathbb{E}_{\mathcal{M}_*} \mathbb{E}_{z} \left[ \left\langle \tau_{\mu} \tau_{\Tilde{\nu}} \right\rangle_{\mathcal{L}^O (t)} \right] - \left\langle \tau_{\mu} \tau_{\Tilde{\nu}} \right\rangle_{\mathcal{M} (t)} \right) - \hat{s}^{\mu \nu} (t) \right) \\
    \hat{q}^{\mu \nu} (t + 1) &= \hat{q}^{\mu \nu} (t) + \Delta t \left( \beta^2 \alpha \mathbb{E}_{\mathcal{M}_*} \mathbb{E}_{z} \left[ \left\langle \tau_{\mu} \right\rangle_{\mathcal{L}^O (t)} \left\langle \tau_{\Tilde{\nu}} \right\rangle_{\mathcal{L}^O (t)} \right] - \hat{q}^{\mu \nu} (t) \right) \\
    m^{\gamma \mu} (t + 1) &= m^{\gamma \mu} (t) + \Delta \tau \left( \mathbb{E}_{\xi^*} \mathbb{E}_{z} \left[ \xi^{* \gamma} \left\langle \xi^{\mu} \right\rangle_{\mathcal{L}^C (t + 1)} \right] - m^{\gamma \mu} (t) \right) \\
    s^{\mu \nu} (t + 1) &= s^{\mu \nu} (t) + \Delta \tau \left( \mathbb{E}_{\xi^*} \mathbb{E}_{z} \left[ \left\langle \xi^{\mu} \xi^{\nu} \right\rangle_{\mathcal{L}^C (t + 1)} \right] - s^{\mu \nu} (t) \right) \\
    q^{\mu \nu} (t + 1) &= q^{\mu \nu} (t) + \Delta \tau \left( \mathbb{E}_{\xi^*} \mathbb{E}_{z} \left[ \left\langle \xi^{\mu} \right\rangle_{\mathcal{L}^C (t + 1)} \left\langle \xi^{\nu} \right\rangle_{\mathcal{L}^C (t + 1)} \right] - q^{\mu \nu} (t) \right),
\end{align*}
with time steps $\Delta t, \Delta \tau \in \left( 0, 1 \right]$. By construction, Eqs. (\ref{eq:RBM_saddle-point}) are a fixed point of the iteration. Empirically, the iteration is the most stable when one of the two time steps is equal to one and the other one is small. Even then, it is still occasionally unstable at large $\alpha$, large teacher pattern correlations and low $T$, which introduces a few easily identifiable spurious discontinuities in Figs. (\ref{fig:RBM_phase_diagram}). We symmetrize the initial conditions of $q$ before the iteration because any $q$ solving the saddle-point equations must be symmetric (see Eqs. \ref{eq:RBM_saddle-point}).

We use Monte Carlo integration over whitened samples to estimate the Gaussian expectations $\mathbb{E}_z \left[ \cdot \right]$. We enforce the sample means to be $0$ by symmetrizing around the origin each set of samples $z_{\mu \nu}$ that approximates the corresponding integral over $z_{\mu \nu}$. We then constrain the sample variances to $1$ using Cholesky whitening \cite{kessy2018optimal}.

$\mathcal{L}_{\lambda_1, \lambda_2} \left( \xi, \xi^*, z \right)$ is a complex-valued function because it involves the square root of a real number that is not necessarily positive. Therefore, the thermal averages $\left\langle \cdot \right\rangle_{\mathcal{L}^C}$ and $\left\langle \cdot \right\rangle_{\mathcal{L}^O}$, are also complex-valued.
By the symmetry of the Gaussian variables in the saddle point equations (see Eqs. \ref{eq:RBM_saddle-point}), the imaginary part of each thermal average is equal to $a$ and $-a$ with the same probability. Therefore, the exact Gaussian expectations must be real-valued. However, in practice, standard Monte Carlo integration for evaluating $\mathbb{E}_z \left[ g \left( z \right) \right]$ with a finite number of samples is very unlikely to randomly produce both $g \left( z \right)$ and its complex conjugate $\Bar{g} \left( z \right)$. Therefore, we replace $\mathbb{E}_z \left[ \cdot \right]$ by $\mathbb{E}_z \left[ \real \left( \cdot \right) \right]$ when solving the saddle-point equations numerically. This adjustment significantly increases the stability of the numerical iteration.

\section{Supplementary figures}
\label{app:supplementary_figures}

This Appendix contains a graph of the free entropy difference of the PSB and partial PSB solutions of Eqs. (\ref{eq:RBM_saddle-point}), as well as some plots of the Mattis magnetization $m$ and spin-glass overlap $q$ obtained for real-valued student patterns with a standard Gaussian prior. The former supports some claims made in Section \ref{sec:independence}, but is not strictly necessary to understand the paper. The latter are not shown in the main text because they look similar to the $m$ and $q$ obtained for binary student patterns with a uniform binary prior. We simplified the saddle-point equations (Eqs. \ref{eq:RBM_saddle-point}) according to \ref{app:gaussian_xi} in order to make them.

\begin{figure}
    \centering
    \includegraphics[width = \textwidth]{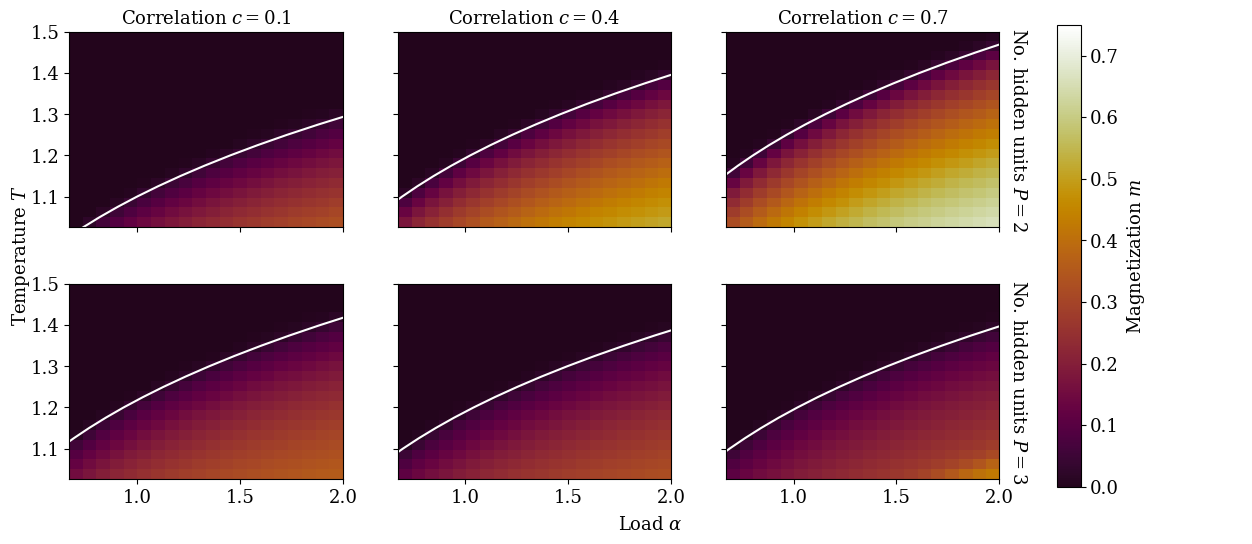}
    \caption{Mattis magnetization $m$ for $\beta = \beta^*$ and $P = P^*$ as a function of the number of hidden units $P$, the correlation $c$, the temperature $T$ and the data load $\alpha$. $m$ is obtained by solving Eqs. (\ref{eq:RBM_saddle-point}) for real-valued student patterns with a standard Gaussian prior and teacher pattern covariance $\mathcal{Q}_{\mu \nu} \sim \mathcal{W} \left( c, P \right)$, where $c \in \left[ 0, 1 \right)$ (see \ref{sec:projected_wishart}). The top and bottom rows feature $P = 2$ and $P = 3$, respectively. The white lines mark the phase transition of Eq. (\ref{eq:critical_load}) with $\lambda^{\mathcal{S}}_{\text{max}}$ given by Eq. (\ref{eq:max_eigval}).}
    \label{fig:normal_random_phase_diagram}
\end{figure}

\end{subappendices}

\chapter{Saddle hierarchy in dense associative memory}
\label{chap:DAM_paper}
\begin{table}[h!]
    \centering
    \begin{tabular}{cc}
        Based on the article \cite{theriault2026saddle}, & doi: \href{https://doi.org/10.1088/2632-2153/ae3051}{10.1088/2632-2153/ae3051} \\ available under the CC BY 4.0 license \includegraphics[width=0.1\linewidth]{cc_by_licence_badge.png} & \href{https://crossmark.crossref.org/dialog/?doi=10.1088/2632-2153/ae3051&domain=pdf&date_stamp=2026-1-6}{\includegraphics[width=0.2\linewidth]{crossmark_badge.png}}
    \end{tabular}
\end{table}
\section{Introduction}
\label{sec:intro}
Studying the stationary points of machine learning algorithms is crucial to understand how they work. For example, \cite{auffinger2013random, choromanska2015loss} demonstrated that local minima in the loss landscape of large artificial Neural Networks (NNs) are relatively close to the global minimum, explaining why they generalize well in practice. Moreover, \cite{razvan2014saddle, dauphin2014identifying} showed that saddle points are much more numerous than local minima in large NNs. These breakthroughs were made by establishing deep connections between the loss landscape of NNs and the energy landscape of disordered systems studied in statistical mechanics. Beyond the broad insights provided by these studies, and despite the progress made by \cite{FUKUMIZU2000local, fukuzumi2019semi-flat, simsek2021geometry, zhang2021embedding}, the classification of stationary points in machine learning algorithms remains an open problem. 

Energy-based models have been playing a central role in studies of NNs and their theoretical properties.
The Hopfield network, one of the most historically important energy-based models, was originally introduced
as a paradigmatic model of biological associative memory \cite{hopfield1982neural}.
Generalized Hopfield networks \cite{chen1986high, psaltis1986nonlinear} were then developed to improve upon the limited storage capacity of the original \cite{hopfield1982neural, amit1985storing}. The scale of the improvement was determined rigorously in following studies \cite{baldi1987number, gardner1987multiconnected, abbott1987storage}. A few years ago, these generalized networks, commonly referred to as Dense Associative Memory (DAM) or modern Hopfield networks, were made into trainable machine learning models capable of accurate pattern classification by Krotov and Hopfield (K \& H) \cite{krotov2016dense}. In a nutshell, K \& H's DAM learns prototypes of patterns in a trainable weight matrix. Each prototype casts a vote for a class, and the patterns awaiting classification are assigned based on the votes of the prototypes that most closely resemble them. The resulting classification scheme is considerably more adversarially robust \cite{krotov2018dense, theriault2024dense} and interpretable \cite{krotov2016dense} than that of feedforward NNs with ReLU activation functions. Since their debut as trainable machine learning architectures, deep connections have been made between modern Hopfield networks and other well-known and established machine learning paradigms, such as attention  \cite{ramsauer2020hopfield} and  generative diffusion \cite{hoover2023memory, ambrogioni2024search}. In particular, modern Hopfield networks were used to implement the attention mechanism of transformers \cite{ramsauer2020hopfield}, which has been attracting a lot of interest in fundamental and applied research. See \cite{krotov2023new, krotov2025modern} for a review of recent advances on the topic. Recently, it was observed that the trainable weights of K \& H's DAM are channeled toward minima by a low-dimensional network of valleys in the loss landscape \cite{boukacem2024waddington}. Moreover, the points where valleys branch out from one another were identified as saddles in the simple case where the DAM has two patterns to learn. In general, it is not straightforward to classify the stationary points of machine learning algorithms \cite{zhang2021embedding}. However, the results of \cite{boukacem2024waddington} and the interpretability of DAMs suggest that their stationary points are both fundamental to their learning dynamics and easier to characterize than that of generic NNs. With this goal in mind, we revisit dense associative memory for pattern classification \cite{krotov2016dense} using the framework of Boltzmann Machines (BMs) \cite{ackley1985learning, smolensky1986information, freund1991unsupervised, hinton2002training} and statistical mechanics.

The body of this paper is structured as follows. In Section \ref{sec:model}, we introduce the DAM model that we study and the analytical tools that we use throughout our work. In particular, in Section \ref{sec:DAM_model}, we derive the DAM in question from a BM template, and, in Section \ref{sec:teacher-student}, we explain the setting that we use to analyze DAM stationary points.

Section \ref{sec:theoretical_results} presents our theoretical contributions.  In Section \ref{sec:saddle-point}, we formulate saddle-point equations that characterize DAM stationary points, and, in Section \ref{sec:hierarchy}, we leverage these equations to establish a saddle-point hierarchy principle stating that weights learned by DAMs of a given width are embedded as saddle points in wider DAMs.

Section \ref{sec:empirical_results} explains how the theoretical insights of Section \ref{sec:theoretical_results} can be used to enhance training and explores how DAMs represent data once trained.
In Section \ref{sec:learning_eff_loss}, we introduce a regularization method that facilitates supervised learning. Next, in Section \ref{sec:interpretability}, we show that our DAM, although designed for supervised tasks, can discover interpretable solutions in both supervised and unsupervised classification settings. Finally, in Section \ref{sec:splitting}, we link our findings to the learning dynamics shaped by valleys and saddles studied in \cite{boukacem2024waddington}, and we implement a network-growing algorithm \cite{wu2019splitting} that exploits the saddle-point hierarchy to significantly reduce DAM training costs.

The code and hyperparameter configurations used in our experiments are available in the following public repository \cite{theriault2025saddlesoftware}.

\section{Model}
\label{sec:model}
A Boltzmann Machine is a canonical graphical model of correlations in discrete data \cite{ackley1985learning}. It is customary to partition BMs into a visible layer {$\mathbf{v} = \left\{ v_i \right\}_{i = 1}^N \in \mathbb{R}^N$} and a hidden layer {$\mathbf{h} = \left\{ h_\mu \right\}_{\mu = 1}^P \in \mathbb{R}^P$} such that connections between the two layers are allowed, but connections within them are prohibited \cite{smolensky1986information}. In this case, the visible layer represents concrete features of the data, whose mutual correlations are encoded in connections with the hidden layer. The Restricted Boltzmann Machine (RBM) obtained using this partition is much easier to train than a generic BM \cite{freund1991unsupervised, hinton2002training} and still has considerable generating power \cite{freund1991unsupervised, leroux2008representational}, making it more practical in machine learning applications \cite{Salakhutdinov2007restricted, kiviken2012multiple, srivastava2013modelling}. The visible and hidden units of an RBM follow the Gibbs distribution
\begin{align*}
    \Prob_\beta \left( \mathbf{v}, \mathbf{h} \big| \mathbf{J} \right) &= Z_\beta \left( \mathbf{J} \right)^{-1} \Prob_0 \left( \mathbf{v} \right) \Prob_0 \left( \mathbf{h} \right) \exp \left( -\beta H \left[ \mathbf{v}, \mathbf{h} ; \mathbf{J} \right] \right),
\end{align*}
where $\beta \geq 0$ is known as the inverse temperature, $\Prob_0 \left( \mathbf{v} \right)$ and $\Prob_0 \left( \mathbf{h} \right)$ are priors on $\mathbf{v}$ and $\mathbf{h}$, $H \left[ \mathbf{v}, \mathbf{h} ; \mathbf{J} \right] = -\sum_{i = 1}^N  \sum_{\mu = 1}^P J^{\mu}_i v_i  h_\mu$ is called the energy function or Hamiltonian, $\mathbf{J} = \left\{ J^\mu_i \right\}_{1 \leq i \leq N}^{1 \leq \mu \leq P}$ are trainable weights, and $Z_\beta \left( \mathbf{J} \right)$ is a normalization constant called the partition function. The inverse temperature $\beta$ represents the absolute strength of the RBM connections, or equivalently controls the amount of noise $T = 1/\beta$ in the RBM. In this regard, $\Prob_0 \left( \mathbf{v} \right)$ and $\Prob_0 \left( \mathbf{h} \right)$, which restrict the form of the Gibbs distribution to help the RBM represent the data, are the marginal laws of $\mathbf{v}$ and $\mathbf{h}$ when there are no connections, i.e. $\beta = 0$. Their contribution to the Gibbs distribution can be tuned with $\beta$, which can therefore be interpreted as a regularization parameter.

\subsection{A Dense Associative Memory (DAM) model}
\label{sec:DAM_model}

As mentioned in the Introduction, we will now derive a DAM model for classification from a BM template. We will explain why our model is a DAM at the end of this Section, once we have clearly defined it. We make three basic assumptions on the distribution of data to be classified:
\begin{enumerate}
    \item \label{assum:invariance} the data is scale invariant, i.e. for all positive scalars $c$, data points $\mathbf{x}$ and $c \mathbf{x}$ are equivalent;
    \item \label{assum:clustering} the data can be partitioned in disjoint clusters;
    \item \label{assum:labeling} the clusters can be grouped into mutually exclusive classes. 
\end{enumerate}
In order to exploit these three assumptions, we study a BM partitioned into three layers with different roles: the data layer $\mathbf{x}$, the hidden layer $\mathbf{h}$ and the class layer $\mathbf{q}$, which represent data, cluster membership and class membership, respectively. The corresponding energy is 
\begin{align*}
    \label{eq:RBM_energy}
    - H \left[ \mathbf{x}, \mathbf{q}, \mathbf{h} ; \mathbf{J} \right] =\sum_{i = 1}^N \sum_{\mu = 1}^P  w^\mu_i x_i h_\mu + \sum_{y = 1}^C \sum_{\mu = 1}^P   u^\mu_y q_y h_\mu + \sum_{\mu = 1}^P h_\mu b^\mu, \numberthis
\end{align*}
where $\mathbf{J} = \left\{ \mathbf{w}, \mathbf{u}, \mathbf{b} \right\}$ is the set of the trainable weights $\mathbf{w} = \left\{ w^\mu_i \right\}_{1 \leq i \leq N}^{1 \leq \mu \leq P}$, $\mathbf{u} = \left\{ u^\mu_i \right\}_{1 \leq i \leq C}^{1 \leq \mu \leq P}$ and $\mathbf{b} = \left\{ b^\mu \right\}_{\mu = 1}^P$.
There are no direct interactions between the visible layer and the class layer. In other words, conditional on the cluster layer, the visible layer and the class layer are independent. Therefore, this BM is a Deep Boltzmann machine (DBM) with 3 layers \cite{salakhutdinov2009deep}, which can also be thought of as an RBM whose visible layer $\mathbf{v}$ is further divided into $\mathbf{x}$ and $\mathbf{q}$.  

Since the data is scale invariant (Assumption \ref{assum:invariance}), the scale of individual data points contains no information about the classification task, so we normalize them by their (Euclidean) norm in the data layer. In other terms, we take the data units $x_i$ to be continuous variables with unit norm $\sqrt{\sum_{i = 1}^N \left( x_i \right)^2} = 1$. We assume no further knowledge about $\mathbf{x}$, so we take the prior $\Prob_0 \left( \mathbf{x} \right)$ to be the uniform distribution on the $N-1$ dimensional unit hypersphere $S^{N-1}$, {i.e. is the set of all $\mathbf{x}$ with $\sqrt{\sum_{i = 1}^N \left( x_i \right)^2} = 1$}. Data normalization is a very common practice in machine learning. For example, normalization by the Euclidean norm has been popular in text document clustering even since its introduction in the 1980s \cite{salton1983introduction}. Various types of normalization also occur in the brain and retina \cite{Carandini2012normalization}.

Since the hidden layer and the class layer aim to represent disjoint clusters and classes, respectively (Assumptions \ref{assum:clustering} and \ref{assum:labeling}), we take their respective units to be mutually exclusive binary variables, i.e. $\mathbf{h}\in \{0,1\}^P$ with $\sum_{\mu=1}^P h_\mu\in\{0,1\}$ and $\mathbf{q}\in \{0,1\}^C$ with $\sum_{y=1}^C q_y\in\{0,1\}$. At any given time, at most one unit per layer can take the value $1$, representing the fact that the clusters and classes are disjoint.
In other words, we take each of the two layers to be the vector representation of a single categorical (or Potts \cite{potts1952some,wu1982potts}) variable with $P+1$ and $C+1$ categories, respectively. As such, $\Prob_0 \left( \mathbf{h} \right)$ and $\Prob_0 \left( \mathbf{q} \right)$ simplify to probability mass functions $\Prob_0 \left( \mathbf{h} = \mathbf{e}_{\gamma} \right)$ and $\Prob_0 \left( \mathbf{q} = \mathbf{e}_{y} \right)$, where we introduce $\mathbf{e}_\gamma = \left\{ \delta_{\gamma \mu} \right\}_{\mu = 1}^P$ for $\gamma \in \left\{ 0,\ldots, P \right\}$ and define $\mathbf{e}_y\in\{0,1\}^C$ analogously for $y\in\{0,\ldots , C\}$. In particular $\mathbf{e}_0=\mathbf{0}$ represents a state outside the $P$ clusters or the $C$ classes.

These priors on the hidden layer and the class layer can also be obtained by introducing fixed inhibitory connections within the hidden layer and the class layer, respectively \cite{kappen1993using, KAPPEN1995deterministic}.
Since at most one hidden unit $h_\mu$ can be activated at once, the hidden layer is a very sparse representation of the visible layer. In machine learning, sparsity can improve interpretability \cite{mozer1988skeletonization}, generalization, computational efficiency \cite{hoefler2021sparsity}, and adversarial robustness \cite{guo2018sparse}. The sparsity of the brain suggests that it is also beneficial for biological neural networks \cite{friston2008hierarchical}.

Given these priors $\Prob_0 \left( \mathbf{x} \right)$, $\Prob_0 \left( \mathbf{h} \right)$ and $\Prob_0 \left( \mathbf{q} \right)$, we derive the marginal distribution of the visible layer $\left( \mathbf{x}, \mathbf{q} \right)$ (see Appendix \ref{app:model} for details). We start our derivation by showing that the conditional distribution of the data layer given the hidden layer has the form
\begin{align*}
    \label{eq:vmf_distribution}
    \Prob_{\beta} \left( \mathbf{x} | \mu, \mathbf{J} \right) &:= \Prob_{\beta} \left( \mathbf{x} | \mathbf{h} = \mathbf{e}_\mu, \mathbf{J} \right) \numberthis \\
    &\propto \exp \left( \beta \sum_{i = 1}^N w^\mu_i x_i \right) \quad \forall \ \mathbf{x}\in S^{N-1} \ \text{and} \ \mu \in \left\{ 1, ..., P \right\}.
\end{align*}
In other words, the probability density $\Prob_{\beta} \left( \mathbf{x} | \mu, \mathbf{J} \right)$ corresponding to each cluster $\mu > 0$ is a von Mises-Fisher (vMF) distribution centered on the direction $\mathbf{w}^\mu = \left\{ w^\mu_i \right\}_{i = 1}^N$ (see Appendix \ref{app:vmf_integration}). In order to interpret the $\mathbf{w}^\mu$ as centroids for their respective clusters, we assume that they belong to $S^{N - 1}$ like the data layer. Under this assumption, we find the normalization constant of $\Prob_{\beta} \left( \mathbf{x} | \mu, \mathbf{J} \right)$ to be $\Omega_N \left( \beta \right) = \frac{\left( 2\pi \right)^{N/2} I_{N/2 - 1} \left( \beta \right)}{\beta^{N/2 - 1}}$, where $I_n \left( x \right)$ is the modified Bessel function of the first kind of order $n$ (see Appendix \ref{app:vmf_integration}).

After slightly more work, we find the marginal distribution of the visible layer to be
\begin{align*}
    \label{eq:direct_distribution}
    \Prob_\beta \left( \mathbf{x}, y \big| \mathbf{w}, \mathbf{p} \right) &:= \Prob_\beta \left( \mathbf{x}, \mathbf{q}=\mathbf{e}_y \big| \mathbf{J} \right) \\
    &= \sum_{\mu = 1}^P p^\mu_y \frac{\exp \left( \beta \sum_{i = 1}^N w^\mu_i x_i \right)}{ \Omega_N \left( \beta \right)} + p^0_y \frac{1}{\Omega_N \left( 0 \right)} \quad \forall \ \mathbf{x}\in S^{N-1} \ \text{and} \ y \in \left\{ 0, ..., C \right\},\numberthis
\end{align*}
where $\mathbf{p} = \left\{ p^\gamma_y \right\}_{0 \leq y \leq C}^{0 \leq \gamma \leq P} = \left\{ \Prob_\beta\left(\mathbf{q}=\mathbf{e}_y,\mathbf{h}=\mathbf{e}_\gamma|\mathbf{J}\right) \right\}_{0 \leq y \leq C}^{0 \leq \gamma \leq P}$. $\Omega_N \left( 0 \right) = \frac{2 \pi^{N/2}}{\Gamma \left( N/2 \right)}$ is the surface area of $S^{N-1}$, where $\Gamma \left( x \right)$ is the Gamma function, so $\frac{1}{\Omega_N \left( 0 \right)}$ is the uniform distribution on $S^{N - 1}$. The uniform distribution term of Eq. (\ref{eq:direct_distribution}) encourages the model to ignore very noisy data during training, which may or may not be desirable depending on the application.

The detailed derivation of Eq. (\ref{eq:direct_distribution}) (in Appendix \ref{app:model}) is inspired by the derivation of Gaussian mixtures from RBMs presented in \cite{Decelle2021restricted} based on the framework of \cite{kappen1993using, KAPPEN1995deterministic}. In our case, the marginal distribution $\Prob_\beta \left( \mathbf{x} | \mathbf{w}, \mathbf{p} \right) = \sum_{y = 0}^C \Prob_\beta \left( \mathbf{x}, y | \mathbf{w}, \mathbf{p} \right)$ and conditional distributions $\Prob_\beta \left( \mathbf{x} | y ; \mathbf{w}, \mathbf{p} \right) = \frac{\Prob_\beta \left( \mathbf{x}, y | \mathbf{w}, \mathbf{p} \right)}{\sum_{\gamma = 0}^P \Prob_\beta \left( y, \gamma | \mathbf{w}, \mathbf{p} \right)}$ are convex mixtures between the uniform distribution on $S^{N - 1}$ and the vMF distributions $\Prob_\beta \left( \mathbf{x} | \mu, \mathbf{J} \right)$ (see Eq. \ref{eq:vmf_distribution}). Probabilistic models similar to $\Prob_\beta \left( \mathbf{x} | \mathbf{w}, \mathbf{p} \right)$ are notably used in text document clustering \cite{banerjee2005clustering}. Mixture distributions that have class-dependent weights like $\Prob_\beta \left( \mathbf{x} | y ; \mathbf{w}, \mathbf{p} \right)$ are also used in Gaussian mixture discriminant analysis \cite{hastie1996discriminant}.

The class weights $\mathbf{p}$ depend on the trainable parameters $\mathbf{u}$ and $\mathbf{b}$ of Eq. (\ref{eq:RBM_energy}) (see Appendix \ref{app:model}). Without loss of generality, we choose to directly study (and train) $\mathbf{p}$ instead of $\mathbf{u}$ and $\mathbf{b}$. Recall that $p^\gamma_y$ is a probability distribution, and in particular $p^\gamma_y \geq 0$ (see Eq. \ref{eq:direct_distribution}). We constrain the marginal $\sum_{y = 0}^C p^\gamma_y$, i.e. the fraction of data in each cluster, to a fixed distribution $p_{\mathbf{h}} \left( \gamma \right)$ and the marginal $\sum_{\gamma = 0}^P p^\gamma_y$, i.e. the proportion of data in each class $y$, to another fixed distribution $p_{\mathbf{q}} \left( y \right)$.
In sum, we end up with the constraints $p^\gamma_y \geq 0$, $\sum_{\gamma = 0}^P p^\gamma_y = p_{\mathbf{q}} \left( y \right)$ and $\sum_{y = 0}^C p^\gamma_y = p_{\mathbf{h}} \left( \gamma \right)$. Since each cluster belongs to a single class (Assumption \ref{assum:labeling}), we expect the $p^\gamma_y$ of a trained model to be close to $\sum_{y^\prime = 0}^C p^\gamma_{y^\prime}$ for a given $y$ and close to $0$ otherwise. 

Given a dataset of $P^*$ patterns $\left\{ \mathbf{x}^{* \mu} \right\}_{\mu = 1}^{P^*}$ with soft labels $q^{* \mu}_y$ \cite{Szegedy2016rethinking}, we can train the weights $\mathbf{w}$ and $\mathbf{p}$ by minimizing the negative log-likelihood loss
\begin{equation}
    \label{eq:loss}
    L \left( \mathbf{w}, \mathbf{p} \right) = -\frac{1}{P^*} \sum_{\mu = 1}^{P^*} \sum_{y = 0}^C q^{* \mu}_y \log \Prob_\beta \left( \mathbf{x}^{* \mu}, y | \mathbf{w}, \mathbf{p} \right),
\end{equation}
which is a form of maximum likelihood estimation. We do so using constrained Stochastic Gradient Descent with momentum (simply called SGD in this paper). We explain how the constraints on $\mathbf{w}$ and $\mathbf{p}$ are enforced in Appendix \ref{app:weight_normalization}, and we briefly discuss the initial conditions and the learning rate in Appendix \ref{app:initialization}.
\begin{figure}
    \centering
    \includegraphics[width=0.495\linewidth]{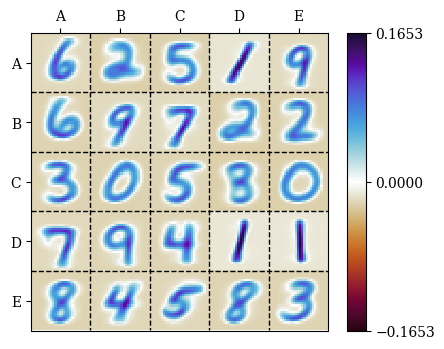}
    \caption{All of the $P = 25$ memories $\left\{ \mathbf{w}^\mu \right\}_{\mu = 1}^{25}$ learned by an instance of our model with $\beta = 16$ when it is trained on the MNIST dataset of handwritten digits \cite{lecun1998gradient} using constrained stochastic gradient descent (SGD) of the negative log-likelihood loss (Eq. \ref{eq:loss}). The hidden units are indexed using pairs of letters from A to E.}
    \label{fig:prelim_DAM_memories}
\end{figure}

The weights $\mathbf{w}^\mu$ learned by SGD of the loss (Eq. \ref{eq:loss}) \cite{lecun1998gradient} are interpretable prototypes, or memories, of the data (see Fig. \ref{fig:prelim_DAM_memories} in the case of the MNIST dataset of handwritten digits \cite{lecun1998gradient}), which is consistent with their role as cluster centroids. Once the model is trained, we can use its conditionals $\Prob_\beta \left( y | \mathbf{x}; \mathbf{w}, \mathbf{p} \right)$ and $\Prob_\beta \left( \mathbf{x} | y; \mathbf{w}, \mathbf{p} \right)$ to efficiently reconstruct $y$ from $\mathbf{x}$ or $\mathbf{x}$ from $y$, respectively. In particular, we can classify unseen patterns using the Bayes classification rule $y = \argmax_{y^\prime} \left\{ \log \Prob_\beta \left( y^\prime | \mathbf{x} ; \mathbf{w}, \mathbf{p} \right) \right\}$ and reconstruct patterns of a given class by finding the local minima of the effective energy $-\log \Prob_\beta \left( \mathbf{x} | y ; \mathbf{w}, \mathbf{p} \right)$ as a function of $\mathbf{x} \in S^{N - 1}$.

High-dimensional probabilistic models that store or learn prototypes, such as our model, can typically reconstruct a limited number of patterns with nontrivial accuracy. In other words, they have limited storage capacity. For instance, Hopfield's model of associative memory, the Hopfield network \cite{hopfield1982neural}, has a capacity of up to $\mathcal{O} \left( N \right)$ patterns \cite{hopfield1982neural, amit1985storing}. Dense Associative Memory (DAM) models, which are inspired by the Hopfield network, are a class of models with an asymptotically much higher capacity \cite{gardner1987multiconnected, krotov2016dense}. The effective energy $-\log \Prob_\beta \left( \mathbf{x} | y ; \mathbf{w}, \mathbf{p} \right)$ that we can minimize to reconstruct patterns with our model (see Eq. \ref{eq:direct_distribution}) is very similar to that of the DAM with exponential capacity introduced in \cite{ramsauer2020hopfield}, which uses energy minimization to implement the attention mechanism of transformers. In fact, according to \cite{lucibello2024exponential}, our model also belongs to the class of DAMs with exponential capacity.

\subsection{Teacher-student setting}
\label{sec:teacher-student}
Among the extensive research on the properties of artificial Neural Networks (NNs) from the perspective of statistical mechanics (see \cite{charbonneau2023spin} for a review), there have been many studies of simple RBMs trained by maximum likelihood estimation (a line of work pioneered in \cite{decelle2017spectral, decelle2018thermodynamics, Decelle2020gaussian}) or by averaging samples from the posterior distribution of the weights $\mathbf{J}$ given some observed data $\mathcal{D}$.
The latter approach has notably been used to characterize the fundamental limits of learning for many types of RBMs \cite{huang2016unsupervised, huang2017statistical, barra2017phase}, with different priors \cite{barra2018phase,huang2018role, manzan2025effect} and architectures \cite{hou2019minimal,theriault2025modeling}, in the \textit{teacher-student setting} \cite{gardner1989unfinished, decelle2021inverse, alemanno2023hopfield} where the data used to train $\mathbf{J}$ is sampled from another RBM with planted weights $\mathbf{J}^*$. This teacher-student setting can also be used to study our DAM. In this scenario, a \textit{teacher} DAM with weights $\mathbf{w}^*$ and $\mathbf{p}^*$ generates a large amount $M = \alpha N$ of noisy data $\mathcal{D} = \left\{ \mathbf{x}^c, y^c \right\}_{c = 1}^M$ and feeds them to a \textit{student} DAM, which then trains its weights $\mathbf{w}$ and $\mathbf{p}$ by averaging samples from the posterior distribution
\begin{align*}
    \label{eq:posterior}
    \Prob_\beta \left( \mathbf{w}, \mathbf{p} | \mathcal{D} \right) &= \mathcal{Z}_\beta \left( \mathcal{D} \right)^{-1} \Prob_0 \left( \mathbf{w}, \mathbf{p} \right) \prod_{c = 1}^M \Prob_\beta \left( \mathbf{x}^c, y^c | \mathbf{w}, \mathbf{p} \right), \numberthis
\end{align*}
where $\mathcal{Z}_\beta \left(\mathcal{D} \right) = \mathbb{E}_{\mathbf{w}, \mathbf{p}} \left[ \prod_{c = 1}^M \Prob_\beta \left( \mathbf{x}^c, y^c | \mathbf{w}, \mathbf{p} \right) \right]$ is the posterior partition function and $\Prob_0 \left( \mathbf{w}, \mathbf{p} \right)$ is the prior on $\mathbf{w}$ and $\mathbf{p}$, which for simplicity we choose as uniform over the sets in which $\mathbf{w}$ and $\mathbf{p}$ are constrained.

We use a statistical mechanics approach to derive a single set of \textit{saddle-point} equations that simultaneously characterize the weights that are stationary points of maximum likelihood estimation (Eq. \ref{eq:loss}) for generic data and the typical weight configurations obtained by averaging samples from Eq. (\ref{eq:posterior}) in the teacher-student setting. We assume that the student does not know the number of hidden units $P^*$ and the inverse temperature $\beta^*$ of the teacher, so it cannot match them with its own. In particular, we consider the case where the noise injected by the teacher in the data is relatively small, i.e. $\beta^* / N > 0$ as $N \rightarrow \infty$, while the student chooses a conservative inverse temperature $\beta \ll N$ to avoid overfitting. Moreover, we fix $\sum_{y = 0}^C p^{* 0}_y = p_{\mathbf{h}}^* \left( 0 \right) = 0$ and $\sum_{y = 0}^C p^{* \mu}_y = p_{\mathbf{h}}^* \left( \mu \right) = 1/P^*$ for all $\mu > 0$ so that each $\mathbf{g}^{* \mu} := \Prob_\beta \left( y | \mu ; \mathbf{w}, \mathbf{p} \right) = \mathbf{p}^{* \mu} / p_{\mathbf{h}}^* \left( \mu \right) = P^* \mathbf{p}^{* \mu}$ is a soft label for the corresponding $\mathbf{w}^{* \mu}$. On the contrary, we do not give $\sum_{\gamma = 0}^{P^*} p^{* \gamma}_y = p_{\mathbf{q}}^* \left( y \right)$ a restrictive form. In other words, $p_{\mathbf{q}}^* \left( y \right)$ is free to be any given probability mass function.

Table (\ref{tab:notation_table}) of Appendix \ref{app:model_summary} summarizes, for quick reference, the most important symbols introduced in this Section and their meaning.

\section{Theoretical results}
\label{sec:theoretical_results}

\subsection{Saddle-point equations}
\label{sec:saddle-point}
In this Section, we introduce a set of equations for the stationary points of the loss (Eq. \ref{eq:loss}), which we then relate to the saddle-point equations emerging from the statistical mechanics analysis of posterior sampling (see Eq. \ref{eq:posterior}) in the teacher-student setting. 

Let us first establish a few definitions that we will use frequently throughout the Section. Given two matrices $\mathbf{w}^*\in\mathbb{R}^{P^*\times N}$ and $\mathbf{w}\in\mathbb{R}^{P\times N}$, we define the overlap matrix $\mathbf{m}(\mathbf{w}^*,\mathbf{w})=\mathbf{w}^*\mathbf{w}^T \in\mathbb{R}^{P^*\times P}$. We write its entries as $m^{\mu_* \mu} \left( \mathbf{w}^*, \mathbf{w} \right) = \sum_{i = 1}^N w^{* \mu_*}_i w^\mu_i$ and its row vectors as $m^{\mu_*}\left( \mathbf{w}^*, \mathbf{w} \right)$, where $1 \leq \mu_* \leq P^*$ and $1 \leq \mu \leq P$. Moreover, for any matrix $\mathbf{m}\in \mathbb{R}^{P^*\times (P+1)}$ with entries $m^{\mu_* \gamma}$ and row vectors $m^{\mu_*}$, we use 
$$\sigma_\gamma(m^{\mu_*})= \frac{\exp \left( m^{\mu_* \gamma} \right)}{\sum_{\nu = 0}^P \exp \left( m^{\mu_* \nu} \right)}$$
to represent the entry number $\gamma \in \left\{ 0, 1, ..., P \right\}$ of the softmax function applied to the row vector $m^{\mu_*}$. In this context, $m^{\mu_*}$ has a zeroth component $m^{\mu_*0}$, and so does its softmax.

In Appendix \ref{app:stationarity}, we show that the stationary points of the negative log-likelihood loss (Eq. \ref{eq:loss}) satisfy
\begin{align*}
    \label{eq:stationarity}
    w^\mu_i &= \frac{\bar{w}^\mu_i}{\sqrt{\sum_{j = 1}^N \left[\bar{w}^\mu_j \right]^2}} \\
    p^\gamma_y &= \frac{\bar{p}^\gamma_y}{\zeta^\gamma_y \left( \mathbf{\bar{p}} ; p_{\mathbf{h}} \right)} \numberthis \\
    \text{with} \quad \bar{w}^{ \mu}_i &= \sum_{\mu_* = 1}^{P^*} x^{* \mu_*}_i \sum_{y = 0}^C q^{* \mu_*}_y \sigma_\mu \left( \beta m^{\mu_*} \left( \mathbf{x}^*, \mathbf{w} \right) + \log \left[ \mathbf{p}_y \right] \right) \\
    \bar{p}^\gamma_y &= \sum_{\mu_* = 1}^{P^*} q^{* \mu_*}_y \sigma_\gamma \left( \beta m^{\mu_*} \left( \mathbf{x}^*, \mathbf{w} \right) + \log \left[ \mathbf{p}_y \right] \right)
\end{align*}
for all $1 \leq \mu \leq P$ and $0 \leq \gamma \leq P$, where $m^{\mu_* 0} \left( \mathbf{x}^*, \mathbf{w} \right) = \frac{1}{\beta} \log \left[ \Omega_N \left( \beta \right) / \Omega_N \left( 0 \right) \right]$ and the normalization constant $\zeta^\gamma_y \left( \mathbf{\bar{p}} ; p_{\mathbf{h}} \right)$ is defined in Appendix \ref{app:weight_normalization}.

Similar equations arise naturally in our statistical mechanics analysis, which amounts to using the replica method \cite{nishimori2001statistical, charbonneau2022replica} to compute the limiting free entropy
\begin{align}
    \label{eq:free_entropy}
    f \left( \varrho, \upsilon, \beta, P^*, P \right) = \lim_{\beta^*, M,N\to\infty} \frac{1}{N} \mathbb{E}_{\mathbf{w}^*, \mathbf{p}^*, \mathcal{D}} \log \left[ \mathcal{Z}_\beta \left( \mathcal{D} \right) \right] ,
\end{align}
where $\upsilon = \beta^* / N$, $\varrho = \frac{M}{P^* N}$ and $\mathbb{E}_{\mathbf{w}^*, \mathbf{p}^*, \mathcal{D}}$ is the joint expectation over the distribution of examples $\mathcal{D} = \left\{ \mathbf{x}^c, y^c \right\}_{c = 1}^M$ generated by the teacher and the priors on the teacher weights. To be more precise, we show that the free entropy can be computed with a variational principle of the form
\begin{align}
  f \left( \varrho, \upsilon, \beta, P^*, P \right) = \operatorname{Extr}_{\mathbf{m},\hat{\mathbf{m}},\mathbf{p}} f(\mathbf{m},\hat{\mathbf{m}},\mathbf{p}),
\end{align}
whose extremizer $\mathbf{m} = \left\{ m^{\mu_* \mu} \right\}_{1 \leq \mu \leq P}^{1 \leq \mu_* \leq P^*}$ can be interpreted as the $N \rightarrow \infty$ limit of the expected value of the teacher-student overlaps $\mathbf{m}(\mathbf{w}^*,\mathbf{w})$. 
We show the derivation of the variational principle in Appendix \ref{app:partition_function} together with an explicit expression for the trial function $f(\mathbf{m},\hat{\mathbf{m}},\mathbf{p})$. If we assume that there are $P^* \ll N$ teacher memories and that the priors on $\mathbf{w}^{* \mu_*}$ and $\mathbf{g}^*$ are uniform like those of the student, we find that the expected teacher-student overlaps must satisfy the saddle-point equations
\begin{align*}
    \label{eq:uniform_saddle-point}
    m^{\mu_* \mu} &= \varsigma \left( 2 \beta_{\text{eff}} \varrho \sqrt{\sum_{\nu_* = 1}^{P^*} \left[ \hat{m}^{\nu_* \mu} \right]^2} \right) \frac{\hat{m}^{\mu_* \mu}}{\sqrt{\sum_{\nu_* = 1}^{P^*} \left[ \hat{m}^{\nu_* \mu} \right]^2}} \\
    p^\gamma_y &= \frac{\bar{p}^\gamma_y}{\zeta^\gamma_y \left( \mathbf{\bar{p}} ; p_{\mathbf{h}} \right)} \numberthis \\
    \text{with} \quad \hat{m}^{\mu_* \mu} &= \sum_{y = 0}^C p_{\mathbf{q}}^* \left( y \right) \sigma_\mu \left( \beta_{\text{eff}} m^{\mu_*} + \log \left[ \mathbf{p}_y \right] \right) \\
    \bar{p}^\gamma_y &= p_{\mathbf{q}}^* \left( y \right) \sum_{\mu_* = 1}^{P^*} \sigma_\gamma \left( \beta_{\text{eff}} m^{\mu_*} + \log \left[ \mathbf{p}_y \right] \right)
\end{align*}
for all $1 \leq \mu \leq P$ and $0 \leq \gamma \leq P$, where $\varsigma \left( x \right) = \frac{x}{\sqrt{x^2 + 1} + 1}$, $\beta_{\text{eff}} = \varsigma \left( 2 \upsilon \right) \beta$ and $m^{\mu_* 0} = \frac{1}{\beta_{\text{eff}}} \log \left[ \Omega_N \left( \beta \right) / \Omega_N \left( 0 \right) \right]$.

If we instead clamp the teacher weights $\mathbf{w}^{* \mu_*}$ and $\mathbf{g}^{* \mu_*}$ to fixed patterns $\mathbf{x}^{* \mu_*}$ and their corresponding soft labels $\mathbf{q}^{* \mu_*}$ \cite{Szegedy2016rethinking}, respectively, to mimic a more general distribution for the data, then Eqs. (\ref{eq:uniform_saddle-point}) become (see Appendix \ref{app:saddle-point})
\begin{align*}
    \label{eq:saddle-point}
    m^{\mu_* \mu} &= \varsigma \left( 2 \beta_{\text{eff}} \varrho \sqrt{\sum_{i = 1}^N \left[ \bar{x}^\mu_i \right]^2} \right) \frac{\sum_{i = 1}^N x^{* \mu_*}_i \bar{x}^\mu_i}{\sqrt{\sum_{i = 1}^N \left[ \bar{x}^\mu_i \right]^2}} \\
    p^\gamma_y &= \frac{\bar{p}^\gamma_y}{\zeta^\gamma_y \left( \mathbf{\bar{p}} ; p_{\mathbf{h}} \right)} \numberthis \\
    \text{with} \quad \bar{x}^\mu_i &= \sum_{\mu_* = 1}^{P^*} x^{* \mu_*}_i \sum_{y = 0}^C q^{* \mu_*}_y \sigma_{\mu} \left( \beta_{\text{eff}} m^{\mu_*} + \log \left[ \mathbf{p}_y \right] \right) \\
    \bar{p}^\gamma_y &= \sum_{\mu_* = 1}^{P^*} q^{* \mu_*}_y \sigma_\gamma \left( \beta_{\text{eff}} m^{\mu_*} + \log \left[ \mathbf{p}_y \right] \right). 
\end{align*}
for all $1 \leq \mu \leq P$ and $0 \leq \gamma \leq P$, where we recall that $m^{\mu_* 0} = \frac{1}{\beta_{\text{eff}}} \log \left[ \Omega_N \left( \beta \right) / \Omega_N \left( 0 \right) \right]$.

In the limit of $\varrho, \upsilon \rightarrow \infty$, which indicates a large number of examples and a low level of teacher noise, Eqs. (\ref{eq:saddle-point}) become equivalent to Eqs. (\ref{eq:stationarity}) if we make the identification $\bar{x}^\mu_i=\bar{w}^\mu_i$, i.e.
\begin{equation}\label{eq:identification}
w^\mu_i = \frac{\bar{x}^\mu_i}{\sqrt{\sum_{j = 1}^N \left[ \bar{x}^\mu_j \right]^2}}.
\end{equation}
Let us explain this limit step by step. For any $\upsilon$ and $\varrho$, the $M$ examples $\{\mathbf{x}^c\}_{c=1}^M$ given to the student are corrupted versions of the teacher patterns $\mathbf{x}^{*}$ with a noise level of $1/\upsilon$. Therefore, in the limit of $\upsilon \rightarrow \infty$ with finite $\varrho$, {each example $\mathbf{x}^c$} is one of the original patterns $\mathbf{x}^*$, as in Eqs. (\ref{eq:stationarity}). However, even in this case, the empirical distribution of the examples deviates from that of $\mathbf{x}^*$, of which it is merely a bootstrap sample. At fixed $P^*$ and $P$, this mismatch disturbs the accurate learning of $\mathbf{x}^*$ when $\varrho = \frac{M}{P^* N}$ is finite, or equivalently the expected number of repetitions $M / P^*$ of each pattern is not sufficiently large compared to $N$ (see Eq. \ref{eq:saddle-point}). This influence progressively weakens as $\varrho$ grows and the $\varsigma$ function approaches $1$, reflecting the convergence of the empirical distribution of the examples to that of the teacher patterns.

In the limit of $\varrho \rightarrow \infty$ with finite $\upsilon$, Eq. (\ref{eq:saddle-point}) is still very similar to Eq. (\ref{eq:stationarity}). In that case, the only difference between them is that Eq. (\ref{eq:saddle-point}) has $\beta_{\text{eff}}$ instead of $\beta$ in the argument of $\sigma$ and in the denominator of the definition of $m^{\mu_* 0}$. Using $\varsigma$ as a shorthand for $\varsigma \left( 2 \upsilon \right)$, we find that the fixed points of Eqs. (\ref{eq:saddle-point}) with $\varrho \rightarrow \infty$ are related, through the identification made in Eq. (\ref{eq:identification}), to the stationary points of the effective loss
\begin{gather*}
    \label{eq:effective_loss}
    \mathcal{L} \left( \mathbf{w}, \mathbf{p} \right) = -\frac{1}{P^*} \sum_{\mu = 1}^{P^*} \sum_{y = 0}^C q^{* \mu}_y \log \mathcal{P}_{\beta, \varsigma} \left( \mathbf{x}^{* \mu}, y | \mathbf{w}, \mathbf{p} \right), \numberthis \\
    \text{where} \quad \mathcal{P}_{\beta, \varsigma} \left( \mathbf{x}, y \big| \mathbf{w}, \mathbf{p} \right) = \sum_{\mu = 1}^P p^\mu_y \frac{\exp \left( \varsigma \beta \sum_{i = 1}^N w^\mu_i x_i \right)}{ \Omega_N \left( \beta \right)} + p^0_y \frac{1}{\Omega_N \left( 0 \right)}.
\end{gather*}
What distinguishes this equation from the standard loss (Eq. \ref{eq:loss}) is that $\mathcal{P}_{\beta, \varsigma} \left( \mathbf{x}, y \big| \mathbf{w}, \mathbf{p} \right)$ has $\beta_{\text{eff}} = \varsigma \beta$ in the argument of the exponential function instead of $\beta$. As a consequence, $\mathcal{P}_{\beta, \varsigma} \left( \mathbf{x}, y \big| \mathbf{w}, \mathbf{p} \right)$ is not a probability distribution unless $\varsigma = 1$, in the limit of $\upsilon \rightarrow \infty$ (see Appendix \ref{app:vmf_integration}). A value of $\varsigma$ less than one in the effective loss (Eq. \ref{eq:effective_loss}) is reminiscent of the presence of noise in the data generation process, so we propose to use it as a regularizer for the weights. We discuss this point in more detail in Section \ref{sec:learning_eff_loss}.

\subsection{Saddle-point hierarchy}
\label{sec:hierarchy}
\begin{figure}
    \centering
    \includegraphics[width=0.9\linewidth]{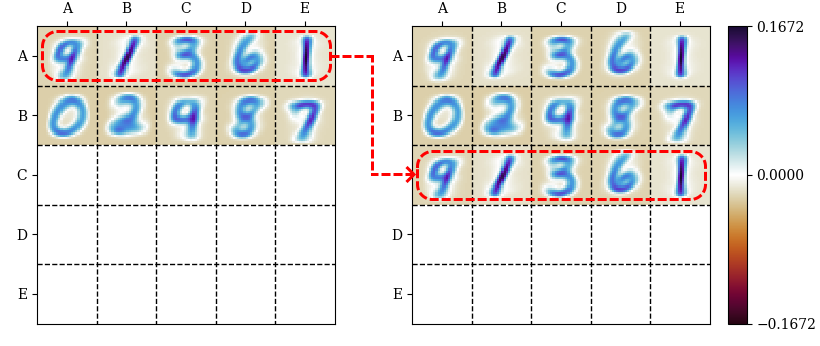}
    
    \hspace{-20pt} $\bar{x}^{\text{fixed}, \mu}_i$ \hspace{146pt} $\bar{x}^{\text{dupli}, \mu}_i$
    
    \caption{Illustration of the relationship  between $\bar{x}^{\text{fixed}, \mu}_i$ and $\bar{x}^{\text{dupli}, \mu}_i$ stated in Eq. (\ref{eq:fixed_point}). The left panel represents the fixed point parameters $\bar{x}^{\text{fixed}, \mu}_i$ of Eq. (\ref{eq:saddle-point}) with $P = 10$, and the right panel represents the fixed point parameters $\bar{x}^{\text{dupli}, \mu}_i$ of Eq. (\ref{eq:saddle-point}) with $P = 15$. In this example, $R = 5$, which means that the first $5$ entries of $\bar{x}^{\text{fixed}, \mu}_i$ are repeated twice in $\bar{x}^{\text{dupli}, \mu}_i$, while the remaining ones are repeated only once. The first $10$ entries of $\bar{x}^{\text{dupli}, \mu}_i$ are identical to $\bar{x}^{\text{fixed}, \mu}_i$, and the dashed red lines highlights that the first $5$ entries of $\bar{x}^{\text{fixed}, \mu}_i$ are repeated a second time at the end of $\bar{x}^{\text{dupli}, \mu}_i$.}
    \label{fig:hierarchy_diagram}
\end{figure}

As shown in \cite{zhang2021embedding}, the loss landscape of any NN with unconstrained weights contains the stationary points of narrower NNs with the same architecture. In Appendix \ref{app:fixed-point}, we show that this result also applies to the teacher-student setting with $\varrho \rightarrow \infty$ and any nonzero $\upsilon$. To be more precise, we show that, if the parameters $\bar{x}^{\text{fixed}, \mu}_i$, $\bar{p}^{\text{fixed}, \gamma}_y$, $m^{\text{fixed}, \mu_* \gamma}$, $p^{\text{fixed}, \gamma}_y$ with hidden unit prior $p_{\mathbf{h}}^{\text{given}} \left( \gamma \right)$ are a fixed point of Eqs. (\ref{eq:saddle-point}) with $P$ hidden units, then the duplicated parameters
{\allowdisplaybreaks
\begin{align*}
    \label{eq:fixed_point}
    \bar{x}^{\text{dupli}, \mu}_i
    &= \begin{cases}
        \bar{x}^{\text{fixed}, \mu}_i &\quad 0 < \mu \leq P \\
        \bar{x}^{\text{fixed}, \mu - P}_i &\quad P < \mu \leq P + R \\
    \end{cases} \\
    \bar{p}^{\text{dupli}, \gamma}_y
    &= \begin{cases}
        \bar{p}^{ \text{fixed}, 0}_y &\quad \gamma = 0 \\
        \frac{1}{2} \bar{p}^{ \text{fixed}, \gamma}_y &\quad 0 < \gamma \leq R \\
        \bar{p}^{ \text{fixed}, \gamma}_y &\quad R < \gamma \leq P \\
        \frac{1}{2} \bar{p}^{ \text{fixed}, \gamma - P}_y &\quad P < \gamma \leq P + R \\
    \end{cases} \\
    m^{\text{dupli}, \mu_* \gamma}
    &= \begin{cases}
        m^{\text{fixed}, \mu_* 0} &\quad \gamma = 0 \\
        m^{\text{fixed}, \mu_* \gamma} &\quad 0 < \gamma \leq P \numberthis \\
        m^{\text{fixed}, \mu_*, \gamma - P} &\quad P < \gamma \leq P + R \\
    \end{cases} \\
    p^{\text{dupli}, \gamma}_y
    &= \begin{cases}
        p^{\text{fixed}, 0}_y &\quad \gamma = 0 \\
        \frac{1}{2} p^{\text{fixed}, \gamma}_y &\quad 0 < \gamma \leq R \\
        p^{\text{fixed}, \gamma}_y &\quad R < \gamma \leq P \\
        \frac{1}{2} p^{\text{fixed}, \gamma - P}_y &\quad P < \gamma \leq P + R \\
    \end{cases}
\end{align*}
\begin{align*}
    \text{along with} \quad p_{\mathbf{h}} \left( \gamma \right)
    &= \begin{cases}
        p_{\mathbf{h}}^{\text{given}} \left( 0 \right) &\quad \gamma = 0 \\
        \frac{1}{2} p_{\mathbf{h}}^{\text{given}} \left( \gamma \right) &\quad 0 < \gamma \leq R \\
        p_{\mathbf{h}}^{\text{given}} \left( \gamma \right) &\quad R < \gamma \leq P \\
        \frac{1}{2} p_{\mathbf{h}}^{\text{given}} \left( \gamma - P \right) &\quad P < \gamma \leq P + R,
    \end{cases}
\end{align*}}%
are a fixed point of the same saddle-point equations with $P + R \in \left\{ P, ..., 2 P \right\}$ hidden units (See Appendix \ref{app:fixed-point} for a detailed derivation {and Fig. \ref{fig:hierarchy_diagram} for a illustrative example}). In other words, we duplicate some of the weights solving Eqs. (\ref{eq:saddle-point}) to construct a fixed point for a wider network. In that sense, wide DAMs contain the fixed points of narrower DAMs. In particular (see Section \ref{sec:saddle-point}), this property also holds for the stationary points of both the standard loss (Eq. \ref{eq:loss}) and the effective loss (Eq. \ref{eq:effective_loss}).

The saddle-point equations (Eq. \ref{eq:saddle-point}) with duplicated order parameters (Eqs. \ref{eq:fixed_point}) are invariant to the permutation of any hidden unit $\gamma \in \left\{ 1, ..., R \right\}$ and its duplicate $\gamma + P$. This kind of symmetry can be spontaneously broken if it leads to a higher free entropy (see Eqs. \ref{eq:free_entropy} and \ref{eq:var_free_entropy}). However, symmetry-breaking transitions can be prohibitively slow when the symmetric state is stable to local perturbations \cite{zdeborova2016statistical}. Interestingly, the DAM introduced in \cite{krotov2016dense} quickly undergoes many successive permutation symmetry-breaking bifurcations during training \cite{boukacem2024waddington}. This observation and additional empirical evidence suggest that the symmetric states are unstable, or in other words that they are saddles, which was verified analytically for DAMs with only two data points to memorize \cite{boukacem2024waddington}. In Appendix \ref{app:fixed-point}, we prove that, if $\beta$ is large enough, Eqs. (\ref{eq:fixed_point}) is a saddle, which is a major step toward explaining why permutation symmetry breaking is relatively fast in DAMs trained with a large number of data points. We call this result \textit{the saddle-point hierarchy principle}.

\section{Empirical results}
\label{sec:empirical_results}
In this Section, we use our theoretical results to improve training, and we show empirically that our DAM learns interpretable solutions to both supervised and unsupervised classification problems. Unless explicitly stated otherwise, we perform our numerical experiments on the MNIST dataset of handwritten digits \cite{lecun1998gradient}. The code and hyperparameter values used in our numerical experiments are available in the following public repository \cite{theriault2025saddlesoftware}.

\subsection{Learning by minimizing the effective loss}
\label{sec:learning_eff_loss}
\begin{figure}
    \centering
    \includegraphics[width=0.495\linewidth]{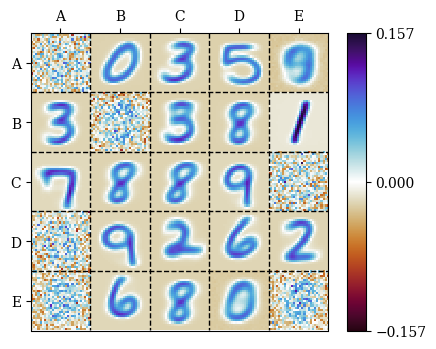}
    \includegraphics[width=0.495\linewidth]{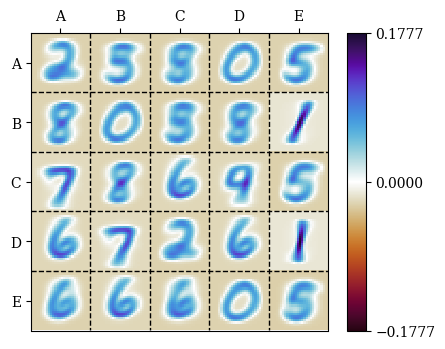}
    \caption{$25$ of the $P = 1000$ memories $\mathbf{w}^\mu$ learned by two instances of our dense associative memory (DAM) model with different values of $\beta$. Both networks are trained on the MNIST dataset of handwritten digits \cite{lecun1998gradient} using constrained stochastic gradient descent (SGD) of the negative log-likelihood loss (Eq. \ref{eq:loss}). The left-panel model has $\beta = 18$, and the right-panel one $\beta = 6$. DAMs with $18 > \beta > 6$ learn memories that interpolate between these two pictures. The hidden units are indexed using pairs of letters from A to E.}
    \label{fig:suboptimal_memories}
\end{figure}

As the number of hidden units $P$ increases, standard maximum likelihood estimation with fixed inverse temperature $\beta$ becomes progressively less apt to train our DAM to its full potential. At high $P$ and $\beta$, many memories stay stuck in noisy states that do not contribute to classification, which is wasteful (see Fig. \ref{fig:suboptimal_memories}). Using a lower $\beta$ helps the memories converge, but also reduces their resolution and diversity, possibly because it also lowers the DAM's capacity \cite{lucibello2024exponential}. This change is far from being only cosmetic. In fact, it comes with a gradual reduction in classification accuracy (see Table \ref{tab:accuracy_table}).

Optimization algorithms with a parameter analogous to $\beta$ often converge to better solutions when this parameter is increased from a small value during optimization. For example, annealing schedules \cite{kirkpatrick1983optimization, rose1990statistical} can be incorporated into minimization algorithms to help them find deeper local minima than they could otherwise. This observation suggests that our DAM would learn better memories if $\beta$ increased during training. It is tempting to do so by SGD of Eq. (\ref{eq:loss}) with respect to $\beta$, but it makes $\beta$ increase so quickly that many memories still stay stuck in noisy states.

We find that optimizing the effective loss (Eq. \ref{eq:effective_loss}) with a suitable $\varsigma$ slows down the evolution of $\beta$ enough to let the DAM learn much cleaner memories (see Fig. \ref{fig:DAM_memories}, top panel) and achieve much better classification accuracy (see Table \ref{tab:accuracy_table}) than with standard training (Eq. \ref{eq:loss}). Looking back at Section \ref{sec:saddle-point}, we propose to interpret $\varsigma$ as a regularization parameter that helps the DAM take noise from the data into account during training. There is no obvious theoretically motivated way to find the best value of $\varsigma$ for a generic dataset, so we choose it by hand. Despite this limitation, we believe that the simplicity and interpretability of our method still make it an interesting alternative to annealing schedules.

When we use this training method, we compute the DAM classification accuracy from the effective predictions
\begin{gather*}
    \label{eq:effective_preds}
    \mathcal{P}_{\beta, \varsigma} \left( y \big| \mathbf{x} ; \mathbf{w}, \mathbf{p} \right) = \frac{\mathcal{P}_{\beta, \varsigma} \left( \mathbf{x}, y \big| \mathbf{w}, \mathbf{p} \right)}{\mathcal{P}_{\beta, \varsigma} \left( \mathbf{x} \big| \mathbf{w}, \mathbf{p} \right)} \numberthis \\
    \text{where} \quad \mathcal{P}_{\beta, \varsigma} \left( \mathbf{x} \big| \mathbf{w}, \mathbf{p} \right) = \sum_{y = 0}^C \mathcal{P}_{\beta, \varsigma} \left( \mathbf{x}, y \big| \mathbf{w}, \mathbf{p} \right),
\end{gather*}
instead of the true predictions $\Prob_{\beta} \left( y \big| \mathbf{x} ; \mathbf{w}, \mathbf{p} \right) = \Prob_{\beta} \left( \mathbf{x}, y \big| \mathbf{w}, \mathbf{p} \right) / \Prob_{\beta} \left( \mathbf{x} \big| \mathbf{w}, \mathbf{p} \right)$.
This approach allows us to calculate both the accuracy and the loss through a single evaluation of $\mathcal{P}_{\beta, \varsigma} \left( y \big| \mathbf{x} ; \mathbf{w}, \mathbf{p} \right)$, which is more efficient than computing $\mathcal{P}_{\beta, \varsigma} \left( \mathbf{x}, y \big| \mathbf{w}, \mathbf{p} \right)$ and $\Prob_{\beta} \left( \mathbf{x}, y \big| \mathbf{w}, \mathbf{p} \right)$ separately when monitoring the progress of training.

\begin{table}
    \centering
    \begin{tabular}{ |c|c| }
        \hline
        Inverse temperature $\beta$ & Classification accuracy \\
        \hline\hline
        6 & 79\% \\
        \hline
        10 & 85\% \\
        \hline
        14 & 89\% \\
        \hline
        18 & 91\% \\
        \hline
        Trained with $\varsigma = 0.25$ & 96\% \\
        \hline
    \end{tabular}
    \caption{DAM classification accuracy (rounded down to two significant figures) for $P = 1000$ and various values of $\beta$. The last line is for $\beta$ trained by SGD of the effective loss (Eq. \ref{eq:effective_loss}) with $\varsigma = 0.25$.}
    \label{tab:accuracy_table}
\end{table}

\subsection{Dense associative memory is interpretable, even in unsupervised classification}
\label{sec:interpretability}
\begin{figure}[!ht]
    \centering
    \includegraphics[width=0.495\linewidth]{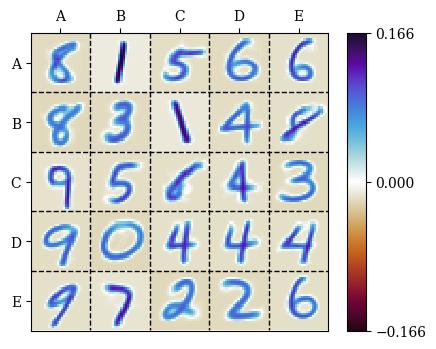}
    \includegraphics[width=\linewidth]{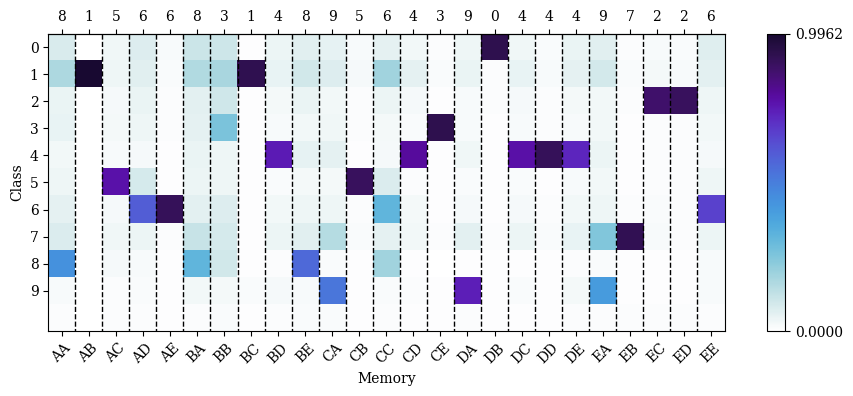}
    \caption{In the top panel, $25$ of the $P = 1000$ memories $\mathbf{w}^\mu$ learned by an instance of our dense associative memory (DAM) model trained on the MNIST dataset of handwritten digits \cite{lecun1998gradient} using constrained stochastic gradient descent (SGD) of the effective loss (Eq. \ref{eq:loss}) with $\varsigma = 0.25$. In the bottom panel, the corresponding rescaled class weights $\mathbf{p}^\mu / p_{\mathbf{h}} \left( \mu \right)$, where $p_{\mathbf{h}} \left( \gamma \right) = \frac{1}{P + 1}$ for all $0 \leq \gamma \leq P$. The hidden units are indexed using pairs of letters from A to E, and the column-wise maxima of the class weights are the classes of the memories with the corresponding letter indices. Rescaled class weights learned with $p_{\mathbf{h}} \left( \gamma \right) \neq \frac{1}{P + 1}$ are qualitatively similar to the ones shown in this figure. Approximately 98\% of test digits fed to the DAM are given the class of the memory that resembles them the most. For example, a digit that looks like the memory \#AA is given the class 8.}
    \label{fig:DAM_memories}
\end{figure}

Now that we understand how to train our DAM reasonably well, we investigate one of the most interesting properties of the solutions that it learns: their interpretability. As advertised in the Introduction, we will explain how to further improve training in the following Section. We already mentioned that the regularization parameter $\varsigma$ of the effective loss is interpretable (see Section \ref{sec:learning_eff_loss} and Eq. \ref{eq:effective_loss}). Here, we point out that the learned weights are as well. In fact, each learned $\mathbf{p}^\mu / p_{\mathbf{h}} \left( \mu \right)$ can be interpreted as a soft label for the corresponding $\mathbf{w}^\mu$ (see Fig. \ref{fig:DAM_memories}). This property was also observed by \cite{boukacem2024waddington} in K \& H's DAM for pattern classification \cite{krotov2016dense}.

At test time, we observe that our DAM classifies approximately 98\% of the test data points into the class $y = \argmax_{y^\prime} \left\{ p^\mu_{y^\prime} \right\}$ of the memory $\mathbf{w}^\mu$ to which they are the most similar. In other words, a 1-nearest neighbor classifier \cite{fix1989discriminatory} conditioned on the memories $\mathbf{w}^\mu$ and their soft labels $\mathbf{p}^\mu / p_{\mathbf{h}} \left( \mu \right)$ approximates the classification of our model with 98\% fidelity. This behavior is reminiscent of K \& H's DAM \cite{krotov2016dense}, where only a few memories participate in the classification of each data point. The 1-nearest neighbor classifier approximates DAM classification more faithfully for correctly classified data points (approximately 99\% fidelity) than incorrectly classified data points (approximately 70\% fidelity).

\begin{figure}
    \centering
    \includegraphics[width=0.495\linewidth]{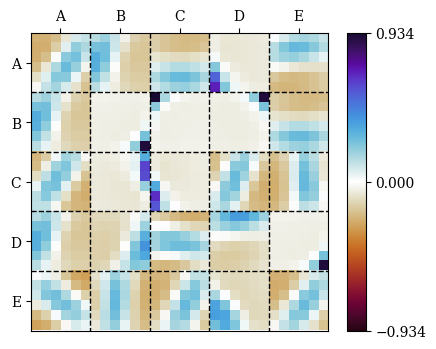}
    \includegraphics[width=\linewidth]{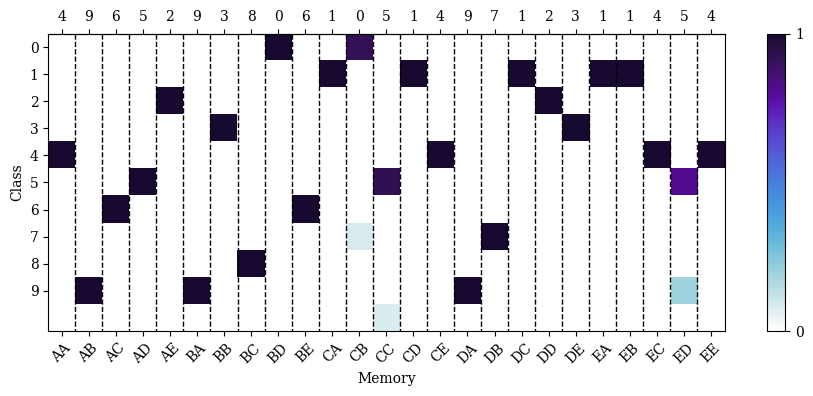}
    \caption{In the top panel, $25$ of the $P = 100$ memories $\mathbf{w}^\mu$ learned by an instance of our dense associative memory (DAM) model trained in an unsupervised way (Eq. \ref{eq:unsupervised_loss}) on $6 \times 6$ patches of the MNIST dataset of handwritten digits \cite{lecun1998gradient} while assuming $C = 10$ latent classes and $\varsigma = 0.6$. In the bottom panel, the corresponding rescaled class weights $\mathbf{p}^\mu / p_{\mathbf{h}} \left( \mu \right)$, where $p_{\mathbf{h}} \left( \gamma \right) = \frac{1}{P + 1}$ for all $0 \leq \gamma \leq P$. The hidden units are indexed using pairs of letters from A to E, and the column-wise maxima of the class weights are the classes of the memories with the corresponding letter indices.}
    \label{fig:unsupervised_DAM_memories}
\end{figure}

\begin{figure}
    \centering
    \includegraphics[width=0.495\linewidth]{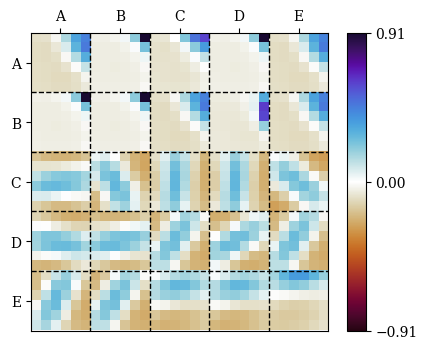}
    \includegraphics[width=\linewidth]{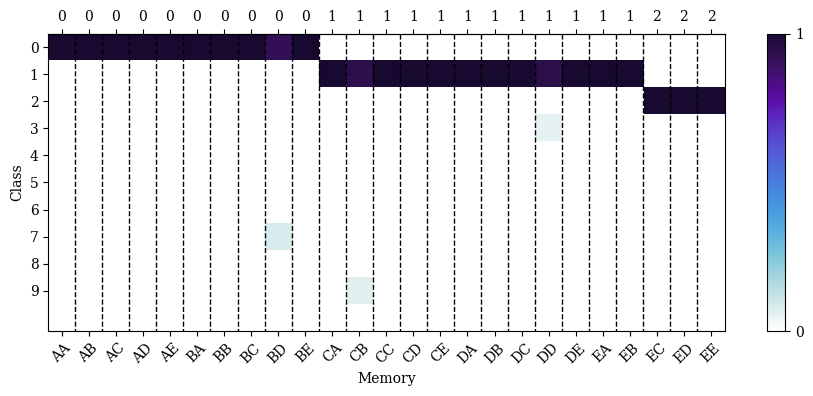}
    \caption{In the top panel, $25$ of the $P = 100$ memories $\mathbf{w}^\mu$ learned by an instance of our dense associative memory (DAM) model trained in an unsupervised way (Eq. \ref{eq:unsupervised_loss}) on $6 \times 6$ patches of the MNIST dataset of handwritten digits \cite{lecun1998gradient} while assuming $C = 10$ latent classes and $\varsigma = 0.6$. In the bottom panel, the corresponding rescaled class weights $\mathbf{p}^\mu / p_{\mathbf{h}} \left( \mu \right)$, where $p_{\mathbf{h}} \left( \gamma \right) = \frac{1}{P + 1}$ for all $0 \leq \gamma \leq P$. The hidden units are indexed using pairs of letters from A to E, and the column-wise maxima of the class weights are the classes of the memories with the corresponding letter indices. $\mathbf{w}^\mu$ and $\mathbf{p}^\mu$ are sorted in increasing $y = \argmax_{y^\prime} \left\{ p^\mu_{y^\prime} \right\}$, and this figure shows $1 \leq \mu \leq 25$.}
    \label{fig:unsupervised_DAM_memories_1}
\end{figure}

\begin{figure}
    \centering
    \includegraphics[width=0.495\linewidth]{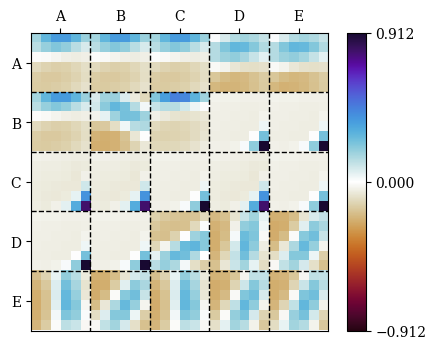}
    \includegraphics[width=\linewidth]{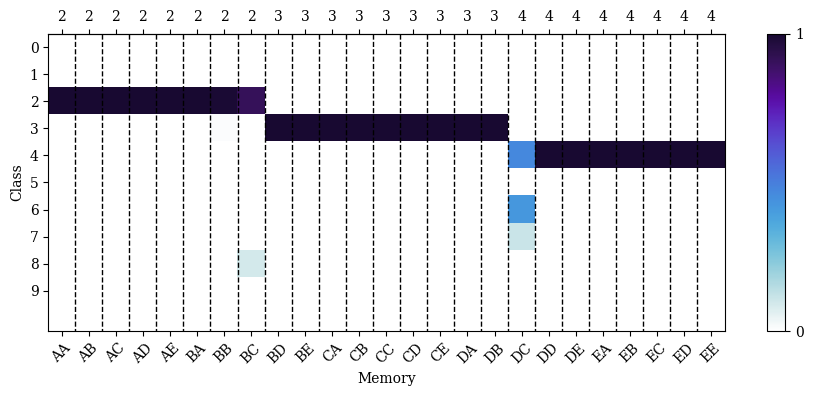}
    \caption{In the top panel, $25$ of the $P = 100$ memories $\mathbf{w}^\mu$ learned by an instance of our dense associative memory (DAM) model trained in an unsupervised way (Eq. \ref{eq:unsupervised_loss}) on $6 \times 6$ patches of the MNIST dataset of handwritten digits \cite{lecun1998gradient} while assuming $C = 10$ latent classes and $\varsigma = 0.6$. In the bottom panel, the corresponding rescaled class weights $\mathbf{p}^\mu / p_{\mathbf{h}} \left( \mu \right)$, where $p_{\mathbf{h}} \left( \gamma \right) = \frac{1}{P + 1}$ for all $0 \leq \gamma \leq P$. The hidden units are indexed using pairs of letters from A to E, and the column-wise maxima of the class weights are the classes of the memories with the corresponding letter indices. $\mathbf{w}^\mu$ and $\mathbf{p}^\mu$ are sorted in increasing $y = \argmax_{y^\prime} \left\{ p^\mu_{y^\prime} \right\}$, and this figure shows $26 \leq \mu \leq 50$.}
    \label{fig:unsupervised_DAM_memories_2}
\end{figure}

We now show that the notion of interpretability described in this Section extends beyond supervised learning. Given a dataset of $P^*$ unlabeled patterns $\{\mathbf{x}^{* \mu}\}_{\mu=1}^{P^*}$, we find that we can train our DAM for unsupervised classification by replacing the soft labels $q^{* \mu}_y$ in the effective loss (Eq. \ref{eq:effective_loss}) with the softened DAM predictions $\left( 1 - \varepsilon \right) \mathcal{P}_{\beta, \varsigma} \left( y | \mathbf{x}^{* \mu} ; \mathbf{w}, \mathbf{p} \right) + \varepsilon \frac{1}{C + 1}$, where $\varepsilon \in \left[ 0, 1 \right]$ and $\mathcal{P}_{\beta, \varsigma} \left( y | \mathbf{x}^{* \mu} ; \mathbf{w}, \mathbf{p} \right)$ is defined in Eq. (\ref{eq:effective_preds}). In this scenario, we are minimizing
\begin{equation}
    \label{eq:unsupervised_loss}
    \mathcal{L}_{\text{unsup}} \left( \mathbf{w}, \mathbf{p} \right) = -\frac{1}{P^*} \sum_{\mu = 1}^{P^*} \sum_{y = 0}^C \left[ \left( 1 - \varepsilon \right) \mathcal{P}_{\beta, \varsigma} \left( y | \mathbf{x}^{* \mu} ; \mathbf{w}, \mathbf{p} \right) + \varepsilon \frac{1}{C + 1} \right] \log \mathcal{P}_{\beta, \varsigma} \left( \mathbf{x}^{* \mu}, y | \mathbf{w}, \mathbf{p} \right).
\end{equation}
As before, we do so using constrained SGD (see Appendix \ref{app:weight_normalization}) with the initialization and learning rate described in Appendix \ref{app:initialization}. Equivalently, we can also view this algorithm as minimizing the combined loss
\begin{gather*}
    \label{eq:total_loss}
    \mathcal{L}_{\text{total}} \left( \mathbf{w}, \mathbf{p} \right) = \mathcal{L}_{\text{margin}} \left( \mathbf{w}, \mathbf{p} \right) + \lambda \mathcal{L}_{\text{cond}} \left( \mathbf{w}, \mathbf{p} \right), \numberthis \\
    \text{where} \quad \mathcal{L}_{\text{margin}} \left( \mathbf{w}, \mathbf{p} \right) = -\frac{1}{P^*} \sum_{\mu = 1}^{P^*} \log \mathcal{P}_{\beta, \varsigma} \left( \mathbf{x}^{* \mu} | \mathbf{w}, \mathbf{p} \right) \\
    \text{and} \quad \mathcal{L}_{\text{cond}} \left( \mathbf{w}, \mathbf{p} \right) = -\frac{1}{P^*} \sum_{\mu = 1}^{P^*} \sum_{y = 0}^C \mathcal{P}_{\beta, \varsigma} \left( y | \mathbf{x}^{* \mu} ; \mathbf{w}, \mathbf{p} \right) \log \mathcal{P}_{\beta, \varsigma} \left( y | \mathbf{x}^{* \mu} ; \mathbf{w}, \mathbf{p} \right),
\end{gather*}
where $\mathcal{P}_{\beta, \varsigma} \left( \mathbf{x}^{* \mu} | \mathbf{w}, \mathbf{p} \right) = \sum_{y = 0}^C \mathcal{P}_{\beta, \varsigma} \left( \mathbf{x}^{* \mu}, y | \mathbf{w}, \mathbf{p} \right)$ (see Eq. \ref{eq:effective_preds}). $\mathcal{L}_{\text{cond}} \left( \mathbf{w}, \mathbf{p} \right)$ is called the minimum entropy regularization term in unsupervised machine learning \cite{grandvalet2004semi}.
Intuitively, it encourages the DAM to learn different class weights $p^\gamma_y$ for each class $y$, which is not possible by minimizing only $\mathcal{L}_{\text{margin}} \left( \mathbf{w}, \mathbf{p} \right)$ because $\mathcal{P}_{\beta, \varsigma} \left( \mathbf{x}^{* \mu} | \mathbf{w}, \mathbf{p} \right)$ does not depend on $y$. Exploiting this characteristic, we train our DAM on patches of MNIST digits \cite{lecun1998gradient} and find that it learns reasonable latent classes $y = \argmax_{y^\prime} \left\{ p^\mu_{y^\prime} \right\}$ for the memories $\mathbf{w}^\mu$ (see Fig. \ref{fig:unsupervised_DAM_memories} and Figs. \ref{fig:unsupervised_DAM_memories_1}, \ref{fig:unsupervised_DAM_memories_2}, \ref{fig:unsupervised_DAM_memories_3} and \ref{fig:unsupervised_DAM_memories_4} of Appendix \ref{app:unsupervised_weights}). This approach could potentially be useful for feature extraction \cite{khalid2014survey}.

\subsection{Fast training with splitting steepest descent}
\label{sec:splitting}
We now explain how to further improve the training method of Section \ref{sec:learning_eff_loss}. More specifically, we use the saddle-point hierarchy derived in \ref{sec:hierarchy} to accelerate training.

\begin{figure}
    \centering
    \includegraphics[width=\linewidth]{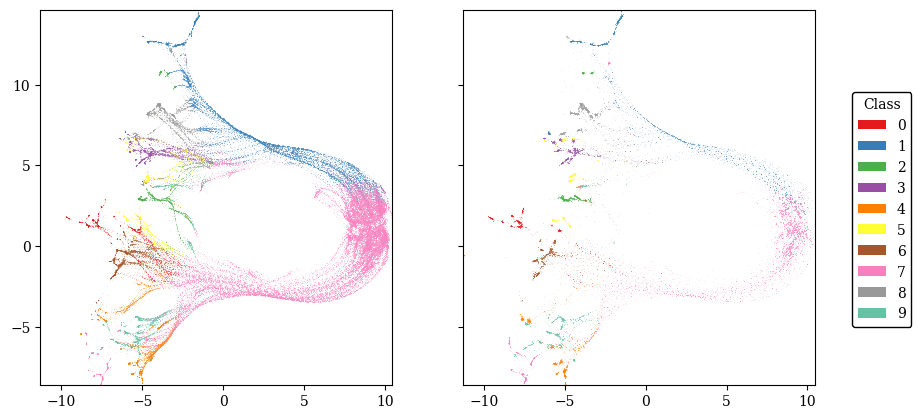}
    \caption{Overlaps $m^{\mu_* \mu} \left( \mathbf{x}^*, \mathbf{w} \right) = \sum_{i = 1}^N x^{* \mu_*}_i w^\mu_i$ between the first 1000 digits $\mathbf{x}^{* \mu_*} = \left\{ x^{* \mu_*}_i \right\}_{i = 1}^N$ of the MNIST training set \cite{lecun1998gradient} and the memories $\mathbf{w}^\mu = \left\{ w^\mu_i \right\}_{i = 1}^N$ of our dense associative memory (DAM) model while it is learning them. Each point is one of the high-dimensional magnetization vectors $m^{\cdot \mu} \left( \mathbf{x}^*, \mathbf{w} \right) = \left\{ m^{\mu_* \mu} \left( \mathbf{x}^*, \mathbf{w} \right) \right\}_{\mu_* = 1}^{P^*}$ projected onto a two-dimensional plane using the UMAP algorithm \cite{McInnes2018UMAP}. On the left, the $m^{\cdot \mu} \left( \mathbf{x}^*, \mathbf{w} \right)$ found during training without splitting steepest descent. On the right, the $m^{\cdot \mu} \left( \mathbf{x}^*, \mathbf{w} \right)$ found with splitting steepest descent (Alg. \ref{alg:splitting_descent}). In both cases, UMAP is trained on the $m^{\cdot \mu} \left( \mathbf{x}^*, \mathbf{w} \right)$ found without splitting steepest descent. The classes $y = \argmax_{y^\prime} \left\{ p^\mu_{y^\prime} \right\}$ of the memories $\mathbf{w}^\mu$ are {color coded} in the legend. 7 and 1 are the most numerous classes, so they have the top two largest entries in $\sum_{\gamma = 0}^P p^\mu_{y} = p_{\mathbf{q}} \left( y \right)$ (see Appendix \ref{app:initialization}), which is why the memories are classified as either 7 or 1 at the beginning of training.
    }
    \label{fig:umap_tree}
\end{figure}

Machine learning models that experience permutation symmetry breaking as described in Section \ref{sec:hierarchy}---such as K \& H's DAM \cite{boukacem2024waddington}, RBMs with binary units \cite{hou2019minimal, theriault2025modeling} and Gaussian mixtures \cite{kappen1993using, KAPPEN1995deterministic, rose1990statistical}
---have characteristic tree-shaped learning curves. Intuitively, permutation symmetry-breaking transitions are points where model parameters differentiate from each other, so they correspond to the bifurcations of the tree. In the left panel of Fig. (\ref{fig:umap_tree}), we show that the learning dynamics of our DAM also follows a tree of permutation symmetry-breaking transitions.

The saddle-point hierarchy principle introduced in Section \ref{sec:hierarchy} suggests the following idea to accelerate learning: train a relatively narrow DAM for cheap, then repeatedly duplicate (or ``split'') some of the hidden units, escape the corresponding saddle point and continue training. Intuitively, we expect to save a lot of computing resources by using relatively few hidden units at the start of training, when the learning dynamics follows a few branches close to the root of the learning dynamics tree. The splitting steepest descent algorithm introduced in \cite{wu2019splitting} formalizes this idea in an efficient and theoretically motivated way. We implement a variant of this algorithm that takes the constraint $\mathbf{w}^\mu \in S^{N - 1}$ into account, but is otherwise very similar to the original version (see Alg. \ref{alg:splitting_descent}), with fast splitting implemented as in \cite{wang2019energy}. The constraint $\mathbf{w}^\mu \in S^{N - 1}$ only matters in steps 5 and 11, which are also arguably the most conceptually difficult parts of Alg. (\ref{alg:splitting_descent}), so we explain them in detail in Appendix \ref{app:splitting}.
Fig. (\ref{fig:umap_tree}) shows that the learning dynamics tree of splitting steepest descent has a more sparsely populated trunk than that of SGD without splitting (see Fig. \ref{fig:umap_tree}). In other words, by using a relatively small number $P$ of memories at the beginning of training, we reduce the total number of values $P \times T$ that they take over a period of time $T$, which is consistent with our intuition that it can save computational resources. The situation could have been different if using fewer memories slowed down training.
\begin{algorithm}[H]
    \caption{Splitting steepest descent \cite{wu2019splitting, wang2019energy}}
    \label{alg:splitting_descent}
    \begin{algorithmic}[1]
        \State Preallocate space for a DAM with $P_{\text{max}}$ hidden units and the corresponding weights $\mathbf{w}$ and $\mathbf{p}$
        \State Initialize the weights $\mathbf{w}^\mu$ and $\mathbf{p}^\mu$ connected to the $P_{\text{cur}}$ first hidden units $\mu \in \left\{ 1, ..., P_{\text{cur}} \right\}$, as well as $\mathbf{p}^0$
        \State min $L \left( \mathbf{w}, \mathbf{p} \right)$ with SGD
        \While{$P_{\text{cur}} < P_{\text{max}}$}
        \State Identify a subset $\boldsymbol{\mu}_{\text{copy}} \subseteq \left\{ 1, ..., P_{\text{cur}} \right\}$ of $R \leq P_{\text{max}} - P_{\text{cur}}$ hidden units to split, \Return if empty
        \State Let $\boldsymbol{\mu}_{\text{paste}} = \left\{ P_{\text{cur}} + 1, ..., P_{\text{cur}} + R \right\}$
        \State Build weights $\mathbf{w}^{\boldsymbol{\mu}_{\text{paste}}} = \mathbf{w}^{\boldsymbol{\mu}_{\text{copy}}}$ for $\boldsymbol{\mu}_{\text{paste}}$
        \State Rescale $\mathbf{p}^{\boldsymbol{\mu}_{\text{copy}}} \gets \mathbf{p}^{\boldsymbol{\mu}_{\text{copy}}} / 2$ and $p_{\mathbf{h}} \left( {\boldsymbol{\mu}_{\text{split}}} \right) \gets p_{\mathbf{h}} \left( \boldsymbol{\mu}_{\text{split}} \right) / 2$
        \State Build weights $\mathbf{p}^{\boldsymbol{\mu}_{\text{paste}}} = \mathbf{p}^{\boldsymbol{\mu}_{\text{copy}}}$ and $p_{\mathbf{h}} \left( \boldsymbol{\mu}_{\text{paste}} \right) = p_{\mathbf{h}} \left( \boldsymbol{\mu}_{\text{copy}} \right)$ for $\boldsymbol{\mu}_{\text{paste}}$
        \State Update $P_{\text{cur}} \gets P_{\text{cur}} + R$
        \State Escape the saddle point by $2^{\text{nd}}$ order descent of $L \left( \mathbf{w}, \mathbf{p} \right)$ w.r.t. $\mathbf{w}$
        \State min $L \left( \mathbf{w}, \mathbf{p} \right)$ with SGD
        \EndWhile
        \State \Return \Comment{\colorbox{lightgray}{See Appendix \ref{app:splitting} and \cite{wu2019splitting, wang2019energy} for details about steps 5 and 11}}
    \end{algorithmic}
\end{algorithm}
We now establish a methodology that we will use to compare the training times of DAMs trained on MNIST \cite{lecun1998gradient} {and Fashion-MNIST \cite{xiao2017fashion}} using the effective loss (Eq. \ref{eq:effective_loss}) with and without splitting steepest descent. It is more interesting and meaningful to compare the training times of NNs with similar performance. Therefore, we pick hyperparameters such that DAMs trained with and without splitting have similar classification accuracy. For that purpose, we find the general region of hyperparameter space where DAMs trained without splitting steepest descent have the best classification accuracy. The accuracy does not change much in this region, so we pick generic hyperparameters inside. Next, we manually tune the hyperparameters of splitting steepest descent so that the resulting DAMs have comparable accuracy. The accuracy obtained for Fashion-MNIST is generically slightly higher with splitting steepest descent than
without it. In contrast, splitting steepest descent does not significantly affect the accuracy obtained for MNIST.

Once the hyperparameters are set, we collect statistics of the accuracy and training time. We run our experiment on a CPU and manually set the seed of pseudorandom number generation to make training deterministic and reproducible. Two DAMs with the same hyperparameters and the same seed are guaranteed to be trained using the same number of computer operations and to have the same classification accuracy after training. However, background processes and other external factors can change between two runs with the same seed, which adds noise to the bare training time that we want to measure. As such, we approximate the bare training time as the minimum over many runs with the same seed. We then calculate the average and standard deviation of the classification accuracy and bare training time over multiple seeds.

\begin{figure}
    \centering
    \hspace{7pt}\includegraphics[width=0.85\linewidth]{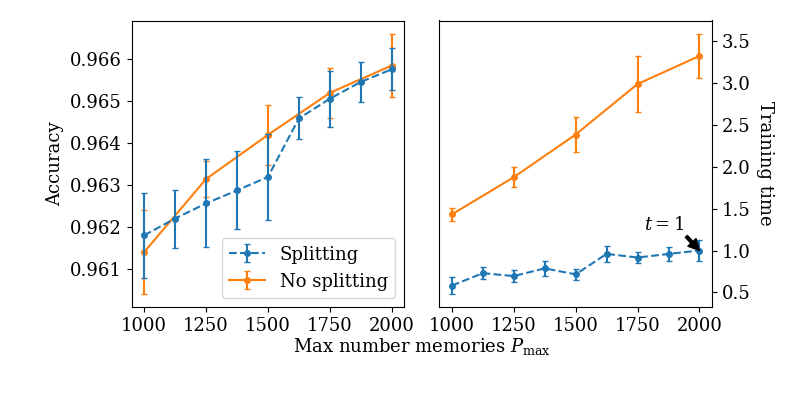}
    \includegraphics[width=0.85\linewidth]{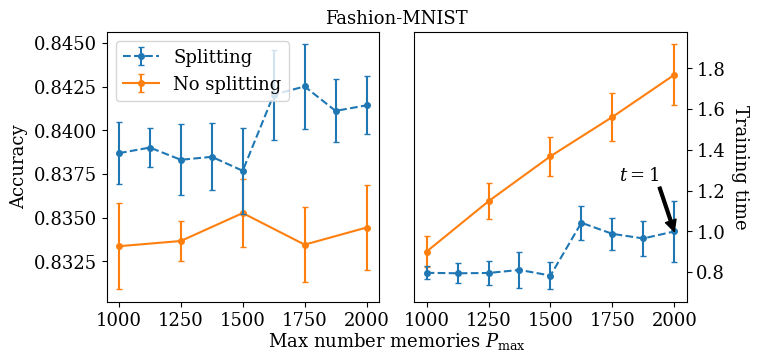}
    \caption{The classification accuracy and training time of dense associative memory (DAM) networks trained on MNIST \cite{lecun1998gradient} and Fashion-MNIST \cite{xiao2017fashion} with and without splitting as a function of the maximum number of hidden unit $P_{\text{max}}$. The round markers and their corresponding error bars show the means and standard deviations of the measurements made at each $P_{\text{max}}$. Statistics are collected from $10$ different random seeds. In the left panels, we verify that the DAMs with and without splitting have a similar accuracy. In the right panels, we compare their training times $t$. To facilitate this comparison, we normalize the $t$-axis such that the data point indicated by the arrow is at $t = 1$.}
    \label{fig:performance}
\end{figure}

In Fig. (\ref{fig:performance}), we compare DAM training time and accuracy with and without splitting steepest descent, using round markers and error bars to show their respective means and standard deviations as a function of the maximum number of hidden units $P_{\text{max}}$. Splitting stops at $P_{\text{max}}$ hidden units, and DAMs without splitting have $P_{\text{max}}$ hidden units from the start. All hidden units are split in each while-loop iteration of Alg. (\ref{alg:splitting_descent}), except for the last iteration, where exactly $P_{\text{max}} - P_{\text{cur}}$ hidden units are split. As wanted, DAMs have similar accuracy with and without splitting (left panels). Moreover, the reasonably small error bars of the training time indicate that the residual noise of the background processes is controlled and that the seed has a limited impact on the training speed (right panels). Splitting steepest descent has a significant speed advantage that scales very advantageously with $P_{\text{max}}$ (right panels). Without splitting, the training time is proportional to $P_{\text{max}}$. With splitting, it appears to be piecewise constant. The clearest sign that it increases with $P_{\text{max}}$ is the jump between $P_{\text{max}} = 1500$ and $P_{\text{max}} = 1625$, where the number of while-loop iterations in Alg. (\ref{alg:splitting_descent}) increases by 1. In other words, the run time of Alg. (\ref{alg:splitting_descent}) is better explained as proportional to $\left\lceil \log P_{\text{max}} \right\rceil$ than $P_{\text{max}}$ in the range of $P_{\text{max}}$ of our numerical experiment, which is a dramatic improvement over the run time without splitting. The benefit of the algorithm becomes evident primarily when the accuracy improves consistently across a wide range of hidden unit counts, i.e. when training a substantially large network is beneficial. This pattern, for example, does not apply to Fashion MNIST (see Fig. \ref{fig:performance}, bottom plot), where the accuracy gain from training a larger network is marginal. As a result, the need for strategies that accelerate training becomes less critical, although such acceleration still occurs.

\section{Conclusion}
In this work, we study a Dense Associative Memory model based on the framework of Boltzmann machines that is capable of learning interpretable solutions to both supervised and unsupervised problems.

We derive two sets of equations that respectively characterize the stationary points of DAMs trained on real data using maximum likelihood estimation and the fixed points of DAMs trained on synthetic data in the teacher-student setting.
Guided by their similarity, we then establish that the maximum likelihood equations are a special case of the teacher-student equations.
Building on this equivalence, we introduce an effective loss function that makes training significantly more stable thanks to a regularization parameter that mimics the noise present in the data generation process.

We show that the stationary points of the effective loss of DAMs with a given number of hidden units are saddle points of larger DAMs, and we considerably accelerate training using a custom implementation of the splitting steepest descent network-growing algorithm \cite{wu2019splitting} inspired by this \textit{saddle-point hierarchy principle}. In our numerical experiments, the training time of splitting steepest descent is approximately logarithmic in the number of hidden units, which is a huge improvement over the linear training time that we observe without splitting steepest descent. Further research is needed to clearly determine the asymptotic functional form of the training time.

In Section \ref{sec:learning_eff_loss}, we briefly discuss how our DAM is interpretable. The ingredients needed to make an interpretable neural network are not well understood, and we believe that our DAM offers a promising opportunity to explore that area. Studying the teacher-student setting with teacher inverse temperature $\beta^*$ less than order $N$ is another interesting research direction that we leave to future work. Beyond the framework of our model, \cite{rose1993constrained, rose1998deterministic} have designed a technique that trains Gaussian mixture models by replicating mixture components. In contrast with our work, it uses a handpicked cooling schedule and breaks permutation symmetries with random noise. By design, splitting steepest descent finds the directions in the loss landscape that allow NNs to escape saddle points with permutation symmetries as quickly as possible, but finding these directions (as described in Appendix \ref{app:splitting} and \cite{wu2019splitting, wang2019energy}) is more expensive than generating random noise. It would be interesting to study this tradeoff and identify the regimes where each method is preferable over the other. More generally, one could investigate how permutation symmetry breaking takes place in energy models related to DAMs, such as attentional BMs \cite{ota2023attention}, and study its relationship with the permutation symmetry-breaking \cite{hou2019minimal, theriault2025modeling} and dynamical transitions \cite{decelle2017spectral, decelle2018thermodynamics, bachtis2025cascade} found in other types of RBMs, whose careful analysis was also used to improve training \cite{bereux2025fast}. Recent works \cite{boukacem2024waddington, grishechkin2025mathematical} have made connections between DAM learning dynamics and cellular differentiation. Similarly, it could be possible to investigate whether there are similarities between cellular division and splitting steepest descent in our model. Finally, it would be interesting to see if splitting steepest descent and our proposed regularization technique can be used to improve the training of transformers and generative diffusion.

\section*{Data availability}
The code for the numerical experiments presented in this work is available at this public repository \cite{theriault2025saddlesoftware}.

\section*{Acknowledgements}
I am grateful to Carlo Lucibello and Matteo Negri for insightful discussions.

\begin{subappendices}

\section{Derivation of the model}
\label{app:model}

Consider the RBM Hamiltonian
\begin{align*}
    -\beta H \left[ \mathbf{x}, \mathbf{q}, \mathbf{h} ; \mathbf{J} \right] = \beta \sum_{\mu = 1}^P h_\mu \sum_{i = 1}^N w^\mu_i x_i + \beta \sum_{\mu = 1}^P h_\mu \sum_{y = 1}^C u^\mu_y q_y + \beta \sum_{\mu = 1}^P h_\mu b^\mu.
\end{align*}
where $\mathbf{J} = \left\{ \mathbf{w}, \mathbf{u}, \mathbf{b} \right\}$ is the set of all RBM parameters, i.e. $\mathbf{w} = \left\{ w^\mu_i \right\}_{1 \leq i \leq N}^{1 \leq \mu \leq P}$, $\mathbf{u} = \left\{ u^\mu_i \right\}_{1 \leq i \leq C}^{1 \leq \mu \leq P}$ and $\mathbf{b} = \left\{ b^\mu \right\}_{\mu = 1}^P$. The prior densities $\Prob_0 \left( \mathbf{h} \right)$ and $\Prob_0 \left( \mathbf{q} \right)$ are nonzero only when $\mathbf{h} \in \left\{ \mathbf{e}_{\gamma} \right\}_{\gamma = 0}^P$ and $\mathbf{q} \in \left\{ \mathbf{e}_y \right\}_{y = 0}^C$, in which case we have
\begin{align*}
    -\beta H \left[ \mathbf{x}, \mathbf{e}_y, \mathbf{e}_{\gamma}; \mathbf{J} \right]
    &= \begin{cases}
        0 &\quad \gamma = 0 \\
        \beta \sum_{i = 1}^N w^\gamma_i x_i + a^\gamma_y &\quad 0 < \gamma \leq P
    \end{cases} \\
    \text{where} \quad a^\mu_y
    &= \begin{cases}
        \beta b^\mu &\quad y = 0 \\
        \beta u^\mu_y + \beta b^\mu &\quad 0 < y \leq C
    \end{cases}
\end{align*}
For convenience, we write the corresponding Gibbs distribution $\Prob_\beta \left( \mathbf{x}, \mathbf{q} = \mathbf{e}_y, \mathbf{h} = \mathbf{e}_\gamma | \mathbf{J} \right)$ as $\Prob_\beta \left( \mathbf{x}, y, \mu | \mathbf{J} \right)$, and the priors $\Prob_0 \left( \mathbf{q} = \mathbf{e}_y \right)$ and $\Prob_0 \left( \mathbf{h} = \mathbf{e}_\gamma \right)$ as $\pi_{\mathbf{q}} \left( y \right)$ and $\pi_{\mathbf{h}} \left( \gamma \right)$, respectively. Given a uniform prior on $\mathbf{x}$, we find
\begin{align*}
    \Prob_{\beta} \left( \mathbf{x}, y, \mu | \mathbf{J} \right) = \frac{1}{Z_\beta \left( \mathbf{J} \right)} \pi_{\mathbf{q}} \left( y \right) \begin{cases}
        \pi_{\mathbf{h}} \left( 0 \right) &\quad \gamma = 0 \\
        \pi_{\mathbf{h}} \left( \gamma \right) \exp \left( a^\gamma_y \right) \exp \left( \beta \sum_{i = 1}^N w^\gamma_i x_i \right) &\quad 0 < \gamma \leq P,
    \end{cases}
\end{align*}
where $Z_\beta \left( \mathbf{J} \right)$ is an unknown normalization constant. The data $\mathbf{x}$ belonging to each cluster $\mu > 0$ follows a von Mises-Fisher (vMF) distribution $\Prob_{\beta} \left( \mathbf{x} | \mu, \mathbf{J} \right) \propto \exp \left( \beta \sum_{i = 1}^N w^\mu_i x_i \right)$ centered on $\mathbf{w}^\mu = \left\{ w^\mu_i \right\}_{i = 1}^N$ (see Appendix \ref{app:vmf_integration}). We assume that $\sqrt{\sum_{i = 1}^N \left( w^\mu_i \right)^2} = 1$ so that the corresponding cluster centroids $\mathbf{w}^\mu$ lie on the hypersphere $S^{N - 1}$ like the data $\mathbf{x} = \left\{ x_i \right\}_{i = 1}^N$. We marginalize $\Prob_{\beta} \left( \mathbf{x}, y, \mu | \mathbf{J} \right)$ over the hidden units $\mu$ and get
\begin{align*}
    \Prob_{\beta} \left( \mathbf{x}, y | \mathbf{J} \right) &= \frac{1}{Z_\beta \left( \mathbf{J} \right)} \pi_{\mathbf{q}} \left( y \right) \left[ \sum_{\mu = 1}^P \pi_{\mathbf{h}} \left( \mu \right) \exp \left( a^\mu_y \right) \exp \left( \beta \sum_{i = 1}^N w^\mu_i x_i \right) + \pi_{\mathbf{h}} \left( 0 \right) \right].
\end{align*}
We will now find the normalization constant $Z_\beta \left( \mathbf{J} \right)$. Marginalizing $\Prob_\beta \left( \mathbf{x}, y | \mathbf{J} \right)$ over $\mathbf{x}$, we obtain
\begin{align*}
    \Prob_{\beta} \left( y | \mathbf{J} \right) &= \int_{S^{N - 1}} d\mathbf{x} \Prob_{\beta} \left( \mathbf{x}, y | \mathbf{J} \right) \\
    &= \frac{1}{Z_\beta \left( \mathbf{J} \right)} \pi_{\mathbf{q}} \left( y \right) \int_{S^{N - 1}} d\mathbf{x} \left[ \sum_{\mu = 1}^P \pi_{\mathbf{h}} \left( \mu \right) \exp \left( a^\mu_y \right) \exp \left( \beta \sum_{i = 1}^N w^\mu_i x_i \right) + \pi_{\mathbf{h}} \left( 0 \right) \right] \\
    &= \frac{1}{Z_\beta \left( \mathbf{J} \right)} \pi_{\mathbf{q}} \left( y \right) \Omega_N \left( \beta \right) \sum_{\mu = 1}^P \pi_{\mathbf{h}} \left( \mu \right) \exp \left( a^\mu_y \right) + \Omega_N \left( 0 \right) \pi_{\mathbf{h}} \left( 0 \right) \\
    &= \frac{1}{Z_\beta \left( \mathbf{J} \right)} \pi_{\mathbf{q}} \left( y \right) \Omega_N \left( \beta \right) \left[ \sum_{\mu = 1}^P \pi_{\mathbf{h}} \left( \mu \right) \exp \left( a^\mu_y \right) + \pi_{\mathbf{h}} \left( 0 \right) \exp \left( a^0 \right) \right],
\end{align*}
where $a^0 = \log \left[ \Omega_N \left( 0 \right) / \Omega_N \left( \beta \right) \right]$. Normalization of $\Prob_\beta \left( y | \mathbf{J} \right)$ requires that
\begin{align*}
    \Prob_{\beta} \left( y | \mathbf{J} \right) &= \frac{\pi_{\mathbf{q}} \left( y \right) \left[ \sum_{\mu = 1}^P \pi_{\mathbf{h}} \left( \mu \right) \exp \left( a^\mu_y \right) + \pi_{\mathbf{h}} \left( 0 \right) \exp \left( a^0 \right) \right]}{\sum_{y^\prime = 0}^C \pi_{\mathbf{q}} \left( y^\prime \right) \left[ \sum_{\nu = 1}^P \pi_{\mathbf{h}} \left( \nu \right) \exp \left( a^{\nu}_{y^\prime} \right) + \pi_{\mathbf{h}} \left( 0 \right) \exp \left( a^0 \right) \right]},
\end{align*}
so we deduce that $Z_\beta \left( \mathbf{J} \right) = \Omega_N \left( \beta \right) \sum_{y = 0}^C \pi_{\mathbf{q}} \left( y \right) \left[ \sum_{\mu = 1}^P \pi_{\mathbf{h}} \left( \mu \right) \exp \left( a^\mu_y \right) + \pi_{\mathbf{h}} \left( 0 \right) \exp \left( a^0 \right) \right]$. We define
\begin{gather*}
    p^\gamma_y
    = \begin{cases}
        \frac{\pi_{\mathbf{q}} \left( y \right) \pi_{\mathbf{h}} \left( 0 \right) \exp \left( a^0 \right)}{\sum_{y^\prime = 0}^C \pi_{\mathbf{q}} \left( y^\prime \right) \left[ \sum_{\nu = 1}^P \pi_{\mathbf{h}} \left( \nu \right) \exp \left( a^{\nu}_{y^\prime} \right) + \pi_{\mathbf{h}} \left( 0 \right) \exp \left( a^0 \right) \right]} \quad \gamma = 0 \\
        \frac{\pi_{\mathbf{q}} \left( y \right) \pi_{\mathbf{h}} \left( \mu \right) \exp \left( a^\mu_y \right)}{\sum_{y^\prime = 0}^C \pi_{\mathbf{q}} \left( y^\prime \right) \left[ \sum_{\nu = 1}^P \pi_{\mathbf{h}} \left( \nu \right) \exp \left( a^{\nu}_{y^\prime} \right) + \pi_{\mathbf{h}} \left( 0 \right) \exp \left( a^0 \right) \right]} \quad 0 < \gamma \leq P,
    \end{cases}
\end{gather*}
from which we obtain
\begin{align*}
    \Prob_{\beta} \left( \mathbf{x}, y | \mathbf{J} \right) &= \sum_{\mu = 1}^P p^\mu_y \frac{\exp \left( \beta \sum_{i = 1}^N w^\mu_i x_i \right)}{\Omega_N \left( \beta \right)} + p^0_y \frac{1}{\Omega_N \left( 0 \right)}.
\end{align*}
The conditional distributions $\Prob_{\beta} \left( \mathbf{x} | y, \mathbf{J} \right)$ and their marginal $\Prob_{\beta} \left( \mathbf{x} | \mathbf{J} \right) = \sum_{y = 0}^C \Prob_{\beta} \left( \mathbf{x} | y, \mathbf{J} \right) \Prob_\beta \left( y | \mathbf{J} \right)$ are von Mises-Fisher mixtures \cite{banerjee2005clustering} with weights $\frac{p^\gamma_y}{\sum_{\nu = 0}^P p^\nu_y}$ and $\sum_{y = 0}^C p^\gamma_y$, respectively. As explained in Section \ref{sec:model}, we constrain the marginals $\sum_{y = 0}^C p^\gamma_y = \Prob_\beta \left( \gamma | \mathbf{J} \right)$ and $\sum_{\gamma = 0}^P p^\gamma_y = \Prob_\beta \left( y | \mathbf{J} \right)$ to be equal to fixed distributions $p_{\mathbf{h}} \left( \gamma \right)$ and $p_{\mathbf{q}} \left( y \right)$, respectively. Formally, this means that $\mathbf{p} = \left\{ p^\gamma_y \right\}_{0 \leq y \leq C}^{0 \leq \gamma \leq C}$ belongs to the transportation polytope with sum constraints $\sum_{\gamma = 0}^P p^\gamma_y = p_{\mathbf{q}} \left( y \right)$ and $\sum_{y = 0}^C p^\gamma_y = p_{\mathbf{h}} \left( \gamma \right)$ \cite{deloera2013combinatorics}.

\section{Integration of the von Mises-Fisher density}
\label{app:vmf_integration}
The von Mises-Fisher (vMF) distribution \cite{mardia1999directional} is an isotropic Gaussian distribution restricted to the $N - 1$ dimensional unit sphere $S^{N - 1}$. It takes the form
\begin{align*}
    p \left( \mathbf{x} \right) = \Omega \left( r \right)^{-1} \exp \left( \sum_{i = 1}^N r_i x_i \right),
\end{align*}
where $r = \sqrt{\sum_{i} \left[ r_i \right]^2}$ is called the concentration parameter. When $r = 0$, it reduces to the uniform distribution on the unit sphere, whose surface surface area $\frac{2 \pi^{N/2}}{\Gamma \left( N/2 \right)}$ is thus also the normalization constant $\Omega_N \left( 0 \right)$. When $r > 0$, we define the mean direction $\hat{\mathbf{r}} = \left\{ r_i / r \right\}_{i = 1}^N$ and find the normalization constant to be
\begin{align*}
    \label{eq:vmf_integration_constant}
    \Omega_N \left( r \right) &= \int_{S^{N - 1}} d \mathbf{x} \exp \left( \sum_{i = 1}^N r_i x_i \right) \numberthis \\
    &= \int_{S^{N - 1}} d \mathbf{x} \exp \left( r \sum_{i = 1}^N \hat{r}_i x_i \right) \\
    &= \Omega_{N-1} \left( 0 \right) \int_{-1}^{1} d u \left( 1 - u^2 \right)^{(N - 3)/2} \exp \left( r u \right) \\
    &= \Omega_N \left( 0 \right) \left( \frac{r}{2} \right)^{1 - N / 2} \Gamma \left( \frac{N}{2} \right) I_{N / 2 - 1} \left( r \right),
\end{align*}
where $I_n \left( x \right)$ is the modified Bessel function of the first kind of order $n$. The third line of (\ref{eq:vmf_integration_constant}) comes from the change of variables $u = \sum_{i = 1}^N \hat{r}_i x_i$, and the fourth line is a consequence of Poisson's Bessel function integral \cite{NIST:DLMF_0}. In the limit of large $N$ with $\rho = \frac{r}{N - 2} \approx \frac{r}{N}$, we find
\begin{gather*}
    \begin{aligned}
        \int_{S^{N - 1}} d \mathbf{x} \exp \left( \sum_{i = 1}^N r_i x_i \right)
        &\approx \frac{\Omega_N \left( 0 \right)}{\sqrt{2 \pi \left( N/2 - 1 \right)}} \left( \frac{N}{2} - 1 \right)^{1 - N / 2} \left( 1 + \left( 2 \rho \right)^2 \right)^{-1/4} \\
        &\quad \Gamma \left( \frac{N}{2} \right) \exp \left[ \left( \frac{N}{2} - 1 \right) \left( \eta \left( 2 \rho \right) + 1 \right) \right]
    \end{aligned} \\
    \text{where} \ \eta \left( x \right) = \left( 1 + x^2 \right)^{1/2} - 1 - \log \left[ 1 + \left( 1 + x^2 \right)^{1/2} \right] + \log 2,
\end{gather*}
by exploiting the large $N$ asymptotic expansion of $I_{N/2 - 1} \left( \left[ N/2 - 1 \right] \cdot 2\rho \right)$ found in \cite{Fröhlich1981transition, NIST:DLMF_1}. We use Stirling's approximation
\begin{align*}
    \Gamma \left( \frac{N}{2} \right)
    &\approx \sqrt{2 \pi \left( N/2 - 1 \right)} \left( \frac{N}{2} - 1 \right)^{N / 2 - 1} \exp \left( -\frac{N}{2} + 1 \right)
\end{align*}
to simplify it to
\begin{gather*}
    \label{eq:thermodynamic_approx_to_vmf}
    \begin{aligned}
        \int_{S^{N - 1}} d \mathbf{x} \exp \left( \sum_{i = 1}^N r_i x_i \right) &\approx \Omega_N \left( 0 \right) \left( 1 + \left( 2 \rho \right)^2 \right)^{-1/4} \exp \left[ \left( \frac{N}{2} - 1 \right) \eta \left( 2 \rho \right) \right]
    \end{aligned} \numberthis \\
    \text{where} \ \eta \left( x \right) = \left( 1 + x^2 \right)^{1/2} - 1 - \log \left[ 1 + \left( 1 + x^2 \right)^{1/2} \right] + \log 2.
\end{gather*}

\section{Normalization of the weights}
\label{app:weight_normalization}
In order to enforce $\sqrt{\sum_{i = 1}^N \left( w^\mu_i \right)^2} = 1$ at each SGD step, we project $\mathbf{w}^\mu$ and the gradient of the loss with respect to $\mathbf{w}^\mu$ onto the unit sphere $S^{N - 1}$ and the tangent space of $\mathbf{w}^\mu$, respectively. We divide $w^\mu_i$ by its norm to project it onto $S^{N - 1}$, and we multiply the gradient by $\delta_{j k} - w^\mu_j w^\mu_k$ to project it onto the tangent space. Projecting the gradient onto the tangent space is a mathematically sound way to obtain the gradient of a function restricted to a manifold embedded in $\mathbb{R}^N$ \cite{barilari2023lecture}.

The tasks of deriving the saddle-point equations for $\bar{p}^\gamma_y$ (Eqs. \ref{eq:saddle-point}), finding the stationarity condition of the loss with respect to the class weights $p^\gamma_y$ (Eqs. \ref{eq:stationarity}) and efficiently training $p^\gamma_y$ all amount to solving the extremization problem
\begin{align*}
    \label{eq:extremization}
    \Extr_{\mathbf{p}, \omega, \lambda} \left\{ f \left( \mathbf{p} \right) + \sum_{y = 0}^C \lambda_y \left( \sum_{\gamma = 0}^P p^\gamma_y - p_{\mathbf{q}} \left( y \right) \right) + \sum_{\gamma = 0}^P \omega^{\gamma} \left( \sum_{y = 0}^C p^\gamma_y - p_{\mathbf{h}} \left( \gamma \right) \right) \right\}, \numberthis
\end{align*}
where $\lambda_y$ and $\omega^{\gamma}$ are Lagrange multipliers that enforce the constraints $\sum_\gamma p^\gamma_y = p_{\mathbf{q}} \left( y \right)$ and $\sum_y p^\gamma_y = p_{\mathbf{h}} \left( \gamma \right)$. In the former task, $f \left( \mathbf{p} \right)$ is the free entropy (Eq. \ref{eq:var_free_entropy}) at fixed $\mathbf{m}$ and $\mathbf{\hat{m}}$. In the latter two, it is the negative log-likelihood loss (Eq. \ref{eq:loss}) at a given $\mathbf{w}$. To find the saddle-point equations and the stationarity conditions of the loss, we derive an implicit solution of Eq. (\ref{eq:extremization}) that is useful in analytical calculations. On the other hand, to train $p^\gamma_y$, we design an algorithm to quickly compute a numerical solution of Eq. (\ref{eq:extremization}). We start by noting that, since the extrema of Eq. (\ref{eq:extremization}) are the points where the gradient vanishes, they take the form
\begin{align*}
    \label{eq:nonlinear_system}
    \partial_{p^\gamma_y} f \left( \mathbf{p} \right) &= \lambda_y + \omega^{\gamma} \\
    \sum_{\gamma = 0}^P p^\gamma_y &= p_{\mathbf{q}} \left( y \right) \numberthis \\
    \sum_{y = 0}^C p^\gamma_y &= p_{\mathbf{h}} \left( \gamma \right).
\end{align*}
Define $\bar{p}^{\gamma}_y = \partial_{p^\gamma_y} \exp \left( f \left( \mathbf{p} \right) \right) = p^\gamma_y \partial_{p^\gamma_y} f \left( \mathbf{p} \right)$, then
\begin{align*}
    \frac{\bar{p}^{\gamma}_y}{p^\gamma_y} &= \lambda_y + \omega^{\gamma},
\end{align*}
along with the previously established row and column sum constraints on $\mathbf{p}$. Rearranging terms, we get
\begin{align*}
    p^\gamma_y &= \frac{1}{\lambda_y + \omega^\gamma} \bar{p}^{\gamma}_y.
\end{align*}
Using the row and column constraints $\sum_{\gamma = 0}^P p^\gamma_y = p_{\mathbf{q}} \left( y \right)$ and $\sum_{y = 0}^C p^\gamma_y = p_{\mathbf{h}} \left( \gamma \right)$, we find that the Lagrange multipliers $\lambda_y$ and $\omega^\gamma$ solve the nonlinear equations
\begin{align*}
    \label{eq:nonlinear}
    \lambda_y &= \frac{1}{p_{\mathbf{q}} \left( y \right)} \sum_{\gamma = 0}^P \frac{\lambda_y}{\lambda_y + \omega^\gamma} \bar{p}^{\gamma}_y \\
    \omega^\gamma &= \frac{1}{p_{\mathbf{h}} \left( \gamma \right)} \sum_{y = 0}^C \frac{\omega^\gamma}{\lambda_y + \omega^\gamma} \bar{p}^{\gamma}_y, \numberthis
\end{align*}
For conciseness, we define $\zeta^\gamma_y \left( \mathbf{\bar{p}} ; p_{\mathbf{h}} \right) = \lambda_y \left( \mathbf{\bar{p}} ; p_{\mathbf{h}} \right) + \omega^\gamma \left( \mathbf{\bar{p}} ; p_{\mathbf{h}} \right)$, where
$\lambda_y \left( \mathbf{\bar{p}} ; p_{\mathbf{h}} \right)$ and $\omega^\gamma \left( \mathbf{\bar{p}} ; p_{\mathbf{h}} \right)$ are the $\lambda_y$ and $\omega^\gamma$ solving Eqs. (\ref{eq:nonlinear}) at given $P$ and $p_{\mathbf{h}}$. Using these definitions, we find the implicit solution $p^\gamma_y = \bar{p}^{\gamma}_y / \zeta^\gamma_y \left( \mathbf{\bar{p}} ; p_{\mathbf{h}} \right)$. As wanted, this equation is useful in analytical calculations involving the saddle-point equations and the stationarity conditions of the loss. However, it is also quite slow to solve numerically for a given $\mathbf{\bar{p}}$, as reported in multiple studies \cite{uribe1966information, hewings1980exchanging, MCNEIL1985note}. Therefore, although we can train $p^\gamma_y$ by iterating $p^\gamma_y = \bar{p}^{\gamma}_y / \zeta^\gamma_y \left( \mathbf{\bar{p}} ; p_{\mathbf{h}} \right)$, it is not a very efficient method.

In order to efficiently train $p^\gamma_y$, we devise a faster way to solve Eqs. (\ref{eq:nonlinear_system}) than through Eq. (\ref{eq:nonlinear}). Exponentiating both sides of the first line of Eqs. (\ref{eq:nonlinear_system}), we find
\begin{align*}
    \exp \left( \eta \partial_{p^\gamma_y} f \left( \mathbf{p} \right) \right) \exp \left( -\eta \lambda_y \right) \exp \left( -\eta \omega^\gamma \right) &= 1 \\
    \exp \left( -\eta \omega^\gamma \right) p^\gamma_y \exp \left( \eta \partial_{p^\gamma_y} f \left( \mathbf{p} \right) \right) \exp \left( -\eta \lambda_y \right) &= p^\gamma_y,
\end{align*}
where $\eta$ is an arbitrary scalar that will play the role of a learning rate. Fix $k^\gamma_y = p^\gamma_y \exp \left( \eta \partial_{p^\gamma_y} f \left( \mathbf{p}  \right) \right)$. By the Sinkhorn-Knopp theorem \cite{sinkhorn1964relationship, knopp1967concerning}, there is a rescaled matrix of the form $p^{\prime \gamma}_y = D^\gamma_L \left( \mathbf{k} \right) k^\gamma_y D^y_R \left( \mathbf{k} \right)$ that satisfies the same constraints $p^{\prime \gamma}_y \geq 0$, $\sum_{\gamma = 0}^P p^{\prime \gamma}_y = p_{\mathbf{q}} \left( y \right)$ and $\sum_{y = 0}^C p^{\prime \gamma}_y = p_{\mathbf{h}} \left( \gamma \right)$ as $p^\gamma_y$ if some technical conditions are satisfied \cite{idel2016review}. Moreover, if $p^{\prime \gamma}_y$ exists, then it is unique \cite{MENON1969spectrum, Hershkowitz1988classifications}, and we can quickly compute suitable scaling factors $D^\gamma_L \left( \mathbf{k} \right)$ and $D^\gamma_R \left( \mathbf{k} \right)$ for $\mathbf{k} = \left\{ k^\gamma_y \right\}^{0 \leq \gamma \leq P}_{0 \leq y \leq C}$ using the Sinkhorn-Knopp algorithm first proposed in \cite{deming1940least} and theoretically justified in subsequent works \cite{sinkhorn1964relationship, knopp1967concerning}. See \cite{idel2016review} for a concise review of all these points. $p^{\prime \gamma}_y$ is generally not equal to $p^\gamma_y$. However, we observe that the iteration
\begin{align*}
    \label{eq:sinkhorn_iteration}
    k^\gamma_y \left( t \right) &= p^\gamma_y \left( t \right) \exp \left( \eta \partial_{p^\gamma_y \left( t \right)} f \left( \mathbf{p} \left( t \right) \right) \right) \\
    p^\gamma_y \left( t + 1 \right) &= D^\gamma_L \left( \mathbf{k} \left( t \right) \right) k^\gamma_y \left( t \right) D^y_R \left( \mathbf{k} \left( t \right) \right) \numberthis
\end{align*}
converges at small $\eta$ (for example $\sim 0.1 / P$), which means that it must converge to a solution of Eqs. (\ref{eq:nonlinear_system}). Since the Sinkhorn-Knopp algorithm is fast, it is significantly more efficient to iterate Eqs. (\ref{eq:sinkhorn_iteration}) than $p^\gamma_y = \bar{p}^{\gamma}_y / \zeta^\gamma_y \left( \mathbf{\bar{p}} ; p_{\mathbf{h}} \right)$ to solve Eqs. (\ref{eq:nonlinear_system}).

In practice, we train $p^\gamma_y$ using a stochastic variant of Eqs. (\ref{eq:sinkhorn_iteration}) where the gradient is estimated over small batches of data and smoothed with a momentum hyperparameter.
Furthermore, we multiply the gradient by $\delta_{y y^\prime} - p^\gamma_y / p_{\mathbf{h}} \left( \gamma \right)$ to reduce the size of the components of $\exp \left( \eta \partial_{p^\gamma_y \left( t \right)} f \left( \mathbf{p} \left( t \right) \right) \right)$ that move $p^\gamma_y$ away from the constraints $\sum_{\gamma = 0}^P p^\gamma_y = p_{\mathbf{q}} \left( y \right)$ and $\sum_{y = 0}^C p^\gamma_y = p_{\mathbf{h}} \left( \gamma \right)$.
This step is a simple approximation of the gradient projection proposed in \cite{douik2019manifold}.

\section{Initialization and learning rate}
\label{app:initialization}
To make all our figures, except Fig. (\ref{fig:umap_tree}), we initialize the memories $\mathbf{w}^\mu$ uniformly at random on the unit hypersphere $S^{N - 1}$ \cite{muller1959note}. To make Fig. \ref{fig:umap_tree}, we instead use the algorithm of \cite{pinzon2023fast} to sample the initial memories $\mathbf{w}^\mu$ from a vMF distribution (see Appendix \ref{app:vmf_integration}) with mean direction $\tilde{\mathbf{x}}^* = \bar{\mathbf{x}}^* / \| \bar{\mathbf{x}}^* \|$, where $\bar{\mathbf{x}}^* = \frac{1}{N} \sum_{\mu_* = 1}^{P^*} \mathbf{x}^{* \mu_*}$ is the mean of the patterns $\mathbf{x}^{* \mu_*}$ and $\| \bar{\mathbf{x}}^* \| = \sqrt{\sum_{i = 1}^N \left[ \bar{x}^*_i \right]^2}$. By construction, the mean direction $\Tilde{\mathbf{x}}^*$ is the ordered solution of Eq. (\ref{eq:saddle-point}) with the most permutation symmetries, or in other words the root of the learning dynamics tree shown in Fig. \ref{fig:umap_tree}, so initializing the memories around it helps reveal the tree structure.

We use a learning rate of $\eta \sim 0.1$ to train the memories $\mathbf{w}^\mu$. Based on our experience, $\mathbf{p}$ trains well when its own learning rate is approximately $\left( 1 + \frac{p_{\mathbf{h}} \left( 0 \right)}{1 - p_{\mathbf{h}} \left( 0 \right)} \right) P$ times smaller. Without this rescaling, the multiplicative update factor $\exp \left( \eta \partial_{p^\gamma_y} L \left( \mathbf{w}, \mathbf{p} \right) \right)$ described in Appendix \ref{app:weight_normalization} can be relatively large even when $\eta$ is relatively small, making it ill behaved. Equivalently, we can also train $g^\gamma_y = \left( 1 + \frac{p_{\mathbf{h}} \left( 0 \right)}{1 - p_{\mathbf{h}} \left( 0 \right)} \right) P \mathbf{p}$ with the same learning rate $\eta$ as $\mathbf{w}$. We adopt this approach in our code available at \cite{theriault2025saddlesoftware}.

Let $P_0 = \frac{p_{\mathbf{h}} \left( 0 \right)}{1 - p_{\mathbf{h}} \left( 0 \right)} P$ so that $g^\gamma_y = \left( P + P_0 \right) p^\gamma_y$. In terms of $g^{\text{fixed}, \gamma}_y = \left( P + P_0 \right) p^{\text{fixed}, \gamma}_y$, the duplicated parameters (Eq. \ref{eq:fixed_point}) of the saddle-point hierarchy principle are
{\allowdisplaybreaks
\begin{align*}
    \bar{x}^{\text{dupli}, \mu}_i
    &= \begin{cases}
        \bar{x}^{\text{fixed}, \mu}_i &\quad 0 < \mu \leq P \\
        \bar{x}^{\text{fixed}, \mu - P}_i &\quad P < \mu \leq P + R \\
    \end{cases} \\
    \bar{g}^{\text{dupli}, \gamma}_y
    &= \frac{P + R}{P} \begin{cases}
        \bar{g}^{ \text{fixed}, 0}_y &\quad \gamma = 0 \\
        \frac{1}{2} \bar{g}^{ \text{fixed}, \gamma}_y &\quad 0 < \gamma \leq R \\
        \bar{g}^{ \text{fixed}, \gamma}_y &\quad R < \gamma \leq P \\
        \frac{1}{2} \bar{g}^{ \text{fixed}, \gamma - P}_y &\quad P < \gamma \leq P + R \\
    \end{cases} \\
    m^{\text{dupli}, \mu_* \gamma}
    &= \begin{cases}
        m^{\text{fixed}, \mu_* 0} &\quad \gamma = 0 \\
        m^{\text{fixed}, \mu_* \gamma} &\quad 0 < \gamma \leq P \\
        m^{\text{fixed}, \mu_*, \gamma - P} &\quad P < \gamma \leq P + R \\
    \end{cases} \\
    g^{\text{dupli}, \gamma}_y
    &= \frac{P + R}{P} \begin{cases}
        g^{\text{fixed}, 0}_y &\quad \gamma = 0 \\
        \frac{1}{2} g^{\text{fixed}, \gamma}_y &\quad 0 < \gamma \leq R \\
        g^{\text{fixed}, \gamma}_y &\quad R < \gamma \leq P \\
        \frac{1}{2} g^{\text{fixed}, \gamma - P}_y &\quad P < \gamma \leq P + R \\
    \end{cases}
\end{align*}
\begin{align*}
    \text{along with} \quad g_{\mathbf{h}} \left( \gamma \right)
    &= \frac{P + R}{P} \begin{cases}
        g_{\mathbf{h}}^{\text{given}} \left( 0 \right) &\quad \gamma = 0 \\
        \frac{1}{2} g_{\mathbf{h}}^{\text{given}} \left( \gamma \right) &\quad 0 < \gamma \leq R \\
        g_{\mathbf{h}}^{\text{given}} \left( \gamma \right) &\quad R < \gamma \leq P \\
        \frac{1}{2} g_{\mathbf{h}}^{\text{given}} \left( \gamma - P \right) &\quad P < \gamma \leq P + R,
    \end{cases}
\end{align*}}%
where $g_{\mathbf{h}}^{\text{given}} \left( \gamma \right) = \left( P + P_0 \right) p_{\mathbf{h}}^{\text{given}} \left( \gamma \right)$ and $g_{\mathbf{h}} \left( \gamma \right) = \left( P + P_0 \right) p_{\mathbf{h}} \left( \gamma \right)$. The newly introduced scaling factor of $\frac{P + R}{P}$ must be taken into account in our implementation of splitting steepest descent (Alg. \ref{alg:splitting_descent}), which gives Alg. (\ref{alg:rescaled_splitting_descent}), shown below.
\begin{algorithm}[H]
    \caption{Rescaled splitting steepest descent \cite{wu2019splitting, wang2019energy}}
    \label{alg:rescaled_splitting_descent}
    \begin{algorithmic}[1]
        \State Preallocate space for a DAM with $P_{\text{max}}$ hidden units and the corresponding weights $\mathbf{w}$ and $\mathbf{g}$
        \State Initialize the weights $\mathbf{w}^\mu$ and $\mathbf{g}^\mu$ connected to the $P_{\text{cur}}$ first hidden units $\mu \in \left\{ 1, ..., P_{\text{cur}} \right\}$, as well as $\mathbf{g}^0$
        \State min $L \left( \mathbf{w}, \mathbf{g} \right)$ with SGD
        \While{$P_{\text{cur}} < P_{\text{max}}$}
        \State Identify a subset $\boldsymbol{\mu}_{\text{copy}} \subseteq \left\{ 1, ..., P_{\text{cur}} \right\}$ of $R \leq P_{\text{max}} - P_{\text{cur}}$ hidden units to split, \Return if empty
        \State Let $\boldsymbol{\mu}_{\text{paste}} = \left\{ P_{\text{cur}} + 1, ..., P_{\text{cur}} + R \right\}$ and $\boldsymbol{\mu}_{\text{dupli}} = \left\{ 1, ..., P_{\text{cur}} + R \right\}$
        \State Build weights $\mathbf{w}^{\boldsymbol{\mu}_{\text{paste}}} = \mathbf{w}^{\boldsymbol{\mu}_{\text{copy}}}$ for $\boldsymbol{\mu}_{\text{paste}}$
        \State Rescale $\mathbf{g}^{\boldsymbol{\mu}_{\text{copy}}} \gets \mathbf{g}^{\boldsymbol{\mu}_{\text{copy}}} / 2$ and $g_{\mathbf{h}} \left( \boldsymbol{\mu}_{\text{split}} \right) \gets g_{\mathbf{h}} \left( \boldsymbol{\mu}_{\text{split}} \right) / 2$
        \State Build weights $\mathbf{g}^{\boldsymbol{\mu}_{\text{paste}}} = \mathbf{g}^{\boldsymbol{\mu}_{\text{copy}}}$ and $g \left( \boldsymbol{\mu}_{\text{paste}} \right) = g \left( \boldsymbol{\mu}_{\text{copy}} \right)$ for $\boldsymbol{\mu}_{\text{paste}}$
        \State Rescale $\mathbf{g}^{\boldsymbol{\mu}_{\text{dupli}}} \gets \frac{P_{\text{cur}} + R}{P_{\text{cur}}} \mathbf{g}^{\boldsymbol{\mu}_{\text{dupli}}}$ and $g_{\mathbf{h}} \left( \boldsymbol{\mu}_{\text{dupli}} \right) \gets \frac{P_{\text{cur}} + R}{P_{\text{cur}}} g_{\mathbf{h}} \left( \boldsymbol{\mu}_{\text{dupli}} \right)$
        \State Update $P_{\text{cur}} \gets P_{\text{cur}} + R$
        \State Escape the saddle point by $2^{\text{nd}}$ order descent of $L \left( \mathbf{w}, \mathbf{g} \right)$ w.r.t. $\mathbf{w}$
        \State min $L \left( \mathbf{w}, \mathbf{g} \right)$ with SGD
        \EndWhile
        \State \Return \Comment{\colorbox{lightgray}{See Appendix \ref{app:splitting} and \cite{wu2019splitting, wang2019energy} for details about steps 5 and 12}}
    \end{algorithmic}
\end{algorithm}
We initialize the weights $\mathbf{g}^\gamma$ and the hidden unit distribution $g_{\mathbf{h}} \left( \gamma \right)$ according to
\begin{gather*}
    g^\gamma_y
    = \begin{cases}
        P_0 p_{\mathbf{q}} \left( y \right) &\quad \gamma = 0 \\
        p_{\mathbf{q}} \left( y \right) &\quad \gamma > 0
    \end{cases} \\
    \text{and} \quad g_{\mathbf{h}} \left( \gamma \right)
    = \begin{cases}
        \frac{P_0}{P + P_0} &\quad \gamma = 0 \\
        \frac{1}{P + P_0} &\quad \gamma > 0.
    \end{cases}
\end{gather*}
When we train our DAM without splitting steepest descent, we set $P_0 = 1$. As discussed in \cite{BARVINOK2010WHAT, good1963maximum}, the resulting weights are, in some sense, the most ``typical'' for the constraints $g^\gamma_y \geq 0$, $\sum_{\gamma = 0}^P g^\gamma_y = \left( P + 1 \right) p_{\mathbf{q}} \left( y \right)$ and $\sum_{y = 0}^C g^\gamma_y = \left( P + 1 \right) p_{\mathbf{h}} \left( \gamma \right) = 1$ (see Section \ref{sec:model}). With splitting steepest descent, we set $P_0 = P_{\text{cur}}/P_{\text{final}}$ to ensure that $g_{\mathbf{h}} \left( 0 \right) = \sum_{y = 0}^C g^0_y$ is approximately the same size as the other entries of $g_{\mathbf{h}} \left( \gamma \right)$ after training. For simplicity, we set $p_{\mathbf{q}} \left( y \right)$ to the proportions of classes in the data during supervised training. For unsupervised training (see Section \ref{sec:interpretability}), we break the permutation symmetry between the different classes by using different values for all the entries of $p_{\mathbf{q}} \left( y \right)$. Otherwise, class weights with unbroken permutation symmetries stay stuck at their initial conditions.
\pagebreak

\section{Important symbols introduced in the Model Section}
\label{app:model_summary}
The following table summarizes the important symbols introduced in Section \ref{sec:model} and their meaning.
\begin{table}[h!]
    \centering
    \begin{tabular}{ |c|c| }
        \hline
        Symbol & Meaning \\
        \hline\hline
        $\mathbf{x} = \left\{ x_i \right\}_{i = 1}^N$ & Data layer \\
        \hline
        $\mathbf{h} = \left\{ h_\mu \right\}_{\mu = 1}^P$ & Hidden layer \\
        \hline
        $\mathbf{q} = \left\{ q_y \right\}_{y = 1}^C$ & Class layer \\
        \hline
        $N$ & \# data units \\
        \hline
        $P$ & \# hidden units (and clusters) \\
        \hline
        $C$ & \# class units (and classes) \\
        \hline
        $\mathbf{J}$ & Set of all trainable weights \\
        \hline
        $\mathbf{w} = \left\{ w_i^\mu \right\}_{1 \leq i \leq N}^{1 \leq \mu \leq P}$ & Weight matrix of the data layer \\
        \hline
        $\mathbf{p} = \left\{ p^\gamma_y \right\}_{0 \leq y \leq C}^{0 \leq \gamma \leq P}$ & Weight matrix of the class layer \\
        \hline
        $p_{\mathbf{q}} \left( y \right)$ & Marginal distribution  $\sum_{\gamma = 0}^P p^\gamma_y $ \\
        \hline
        $p_{\mathbf{h}} \left( \gamma \right)$ & Marginal distribution $\sum_{y = 0}^C p^\gamma_y $ \\
        \hline
        $\beta$ & Inverse temperature \\
        \hline
        $\Omega_N \left( \beta \right)$ & Normalization constant of the vMF distribution \\
        \hline
        $\mathbf{x}^{* \mu} = \left\{ x^{* \mu}_i \right\}_{i = 1}^N$ & Pattern from a given dataset \\
        \hline
        $\mathbf{q}^{* \mu} = \left\{ q^{* \mu}_y \right\}_{y = 0}^C$ & Soft label of the pattern \\
        \hline
        $P^*$ & \# hidden units (and clusters) of the teacher \\
        \hline
        $\mathbf{w}^* = \left\{ w_i^{* \mu} \right\}_{1 \leq i \leq N}^{1 \leq \mu \leq P}$ & Weight matrix of the data layer of the teacher DAM \\
        \hline
        $\mathbf{p}^* = \left\{ p^{* \gamma}_y \right\}_{0 \leq y \leq C}^{0 \leq \gamma \leq P}$ & Weight matrix of the class layer of the teacher DAM \\
        \hline
        $\mathbf{g}^*$ & Rescaled weight matrix $\mathbf{g}^* = P^* \mathbf{p}^*$ of the class layer of the teacher DAM \\
        \hline
        $p_{\mathbf{q}}^* \left( y \right)$ & Marginal distribution  $\sum_{\gamma = 0}^P p^{* \gamma}_y $ \\
        \hline
        $p_{\mathbf{h}}^* \left( \gamma \right)$ & Marginal distribution  $\sum_{y = 0}^C p^{* \gamma}_y $ \\
        \hline
        $\beta^*$ & Inverse temperature of the teacher \\
        \hline
        $\mathcal{D} = \left\{ \mathbf{x}^c, y^c \right\}_{c = 1}^M$ & Set of data points $\mathbf{x}^c$ and labels $y^c$ generated by the teacher \\
        \hline
        $M$ & Number of data points generated by the teacher \\
        \hline
        $\alpha = M/N$ & Load of data generated by the teacher \\
        \hline
    \end{tabular}
    \caption{Summary of the important symbols introduced in Section \ref{sec:model} and their meaning.}
    \label{tab:notation_table}
\end{table}

\section{Stationarity conditions of the loss}
\label{app:stationarity}
Since we constrain the memories $\mathbf{w}^\mu$ to have unit norm (see Section \ref{sec:model}), the method of Lagrange multipliers tells us any set of memories $\mathbf{w}$ that minimizes Eq. (\ref{eq:loss}) must solve the extremization problem
\begin{align*}
    \Extr_{\mathbf{w}, \varphi} \left\{ L \left( \mathbf{w}, \mathbf{p} \right) + \frac{1}{2} \sum_{\mu = 1}^P \varphi^\mu \left( \sum_{i = 1}^N \left[ w^\mu_i \right]^2 - 1 \right) \right\}.
\end{align*}
The extrema are the points where the gradient vanishes, so they take the form
\begin{gather*}
    \partial_{w^\mu_i} L \left( \mathbf{w}, \mathbf{p} \right) + \varphi^\mu w^\mu_i = 0 \\
    \sum_{i = 1}^N \left[ w^\mu_i \right]^2 = 1
\end{gather*}
for all $1 \leq \mu \leq P$. We calculate $\partial_{w^\mu_i} L \left( \mathbf{w}, \mathbf{p} \right)$ and write the solution in the more explicit form
\begin{align*}
   \bar{w}^\mu_i &= \sum_{\mu_* = 1}^{P^*} x^{* \mu_*}_i \sum_{y = 0}^C q^{* \mu_*}_y \sigma_\mu \left( \beta m^{\mu_*} \left( \mathbf{x}^*, \mathbf{w} \right) + \log \left[ \mathbf{p}_y \right] \right) \\
    w^\mu_i &= \frac{\bar{w}^\mu_i}{\sqrt{\sum_{j = 1}^N \left[\bar{w}^\mu_j \right]^2}},
\end{align*}
for all $1 \leq \mu \leq P$, where $\sigma_\gamma \left( x^{\gamma_*} \right) = \frac{\exp \left( x^{\gamma_* \gamma} \right)}{\sum_{\nu = 0}^P \exp \left( x^{\gamma_* \nu} \right)}$, $m^{\mu_* \mu} \left( \mathbf{x}^*, \mathbf{w} \right) = \sum_{i = 1}^N x^{* \mu_*}_i w^\mu_i$ for $1 \leq \mu \leq P$ and $m^{\mu_* 0} \left( \mathbf{x}, \mathbf{w} \right) = \frac{1}{\beta} \log \left[ \Omega_N \left( \beta \right) / \Omega_N \left( 0 \right) \right]$.

Similarly, any set of class weights $\mathbf{p}$ that minimizes Eq. (\ref{eq:loss}) must solve the extremization problem
\begin{align*}
    \Extr_{\mathbf{p}, \omega, \lambda} \left\{ L \left( \mathbf{w}, \mathbf{p} \right) + \sum_{y = 0}^C \lambda_y \left( \sum_{\gamma = 0}^P p^\gamma_y - p_{\mathbf{q}} \left( y \right) \right) + \sum_{\gamma = 0}^P \omega^{\gamma} \left( \sum_{y = 0}^C p^\gamma_y - p_{\mathbf{h}} \left( \gamma \right) \right) \right\}.
\end{align*}
In Appendix \ref{app:weight_normalization}, we show that its solution is
\begin{align*}
    \bar{p}^\gamma_y &= \sum_{\mu_* = 1}^{P^*} q^{* \mu_*}_y \sigma_\gamma \left( \beta m^{\mu_*} \left( \mathbf{x}, \mathbf{w} \right) + \log \left[ \mathbf{p}_y \right] \right) \\
    p^\gamma_y &= \frac{\bar{p}^\gamma_y}{\zeta^\gamma_y \left( \mathbf{\bar{p}} ; p_{\mathbf{h}} \right)}
\end{align*}
for all $0 \leq \gamma \leq P$, where the normalization constant $\zeta^\gamma_y \left( \mathbf{\bar{p}} ; p_{\mathbf{h}} \right)$ is defined using Eqs. (\ref{eq:nonlinear}) of Appendix \ref{app:weight_normalization}. Combining these equations with the ones for $\mathbf{w}$, we find the stationarity conditions
\begin{align*}
   \bar{w}^\mu_i &= \sum_{\mu_* = 1}^{P^*} x^{* \mu_*}_i \sum_{y = 0}^C q^{* \mu_*}_y \sigma_\mu \left( \beta m^{\mu_*} \left( \mathbf{x}^*, \mathbf{w} \right) + \log \left[ \mathbf{p}_y \right] \right) \\
    \bar{p}^\gamma_y &= \sum_{\mu_* = 1}^{P^*} q^{* \mu_*}_y \sigma_\gamma \left( \beta m^{\mu_*} \left( \mathbf{x}^*, \mathbf{w} \right) + \log \left[ \mathbf{p}_y \right] \right) \\
    w^\mu_i &= \frac{\bar{w}^\mu_i}{\sqrt{\sum_{j = 1}^N \left[\bar{w}^\mu_j \right]^2}} \\
    p^\gamma_y &= \frac{\bar{p}^\gamma_y}{\zeta^\gamma_y \left( \mathbf{\bar{p}} ; p_{\mathbf{h}} \right)}
\end{align*}
for all $1 \leq \mu \leq P$ and $0 \leq \gamma \leq P$.

\section{Replicated partition function and free entropy}
\label{app:partition_function}
Suppose a student DAM model is trained using a dataset $\mathcal{D} = \left\{ \mathbf{x}^c, y^c \right\}_{c = 1}^M$ of $M$ i.i.d. examples $\mathbf{x}^c$ with labels $y^c$. By Bayes' theorem, the student weights $\mathbf{w}$ and $\mathbf{p}$ follow the distribution
\begin{align*}
    \Prob_\beta \left( \mathbf{w}, \mathbf{p} | \mathcal{D} \right) &= \mathcal{Z} \left( \mathcal{D} \right)^{-1} \Prob \left( \mathbf{w}, \mathbf{p} \right) \prod_{c = 1}^M \Prob_\beta \left( \mathbf{x}^c, y^c \big| \mathbf{w}, \mathbf{p} \right),
\end{align*}
where $\mathcal{Z} \left( \mathcal{D} \right) = \mathbb{E}_{\mathbf{w}, \mathbf{p}} \left[ \prod_{c = 1}^M \Prob_\beta \left( \mathbf{x}^c, y^c | \mathbf{w}, \mathbf{p} \right) \right]$. Assuming the examples are sampled from a teacher DAM with weights $\mathbf{w}^*$ and $\mathbf{p}^*$, the average replicated partition takes the form
\begin{align*}
    \left\langle \mathcal{Z}^L \right\rangle &= \sum_{\left\{ y^c \right\}_{c = 1}^M} \int
    \left[ \prod_{c = 1}^M d \mathbf{x}^c \right] \ \mathbb{E}_{\mathbf{w}^*, \mathbf{p}^*} \left[ \prod_{c = 1}^M \Prob_\beta \left( \mathbf{x}^c, y^c | \mathbf{w}^*, \mathbf{p}^* \right) \right] \mathcal{Z} \left( \mathcal{D} \right)^L \\
    &= \sum_{\left\{ y^c \right\}_{c = 1}^M} \int
    \left[ \prod_{c = 1}^M d \mathbf{x}^c \right] \ \mathbb{E}_{\mathbf{w}^*, \mathbf{p}^*} \left[ \prod_{c = 1}^M \Prob_\beta \left( \mathbf{x}^c, y^c | \mathbf{w}^*, \mathbf{p}^* \right) \right] \prod_{a = 1}^L \mathbb{E}_{\mathbf{w}^a, \mathbf{p}^a} \left[ \prod_{c = 1}^M \Prob_\beta \left( \mathbf{x}^c, y^c | \mathbf{w}^a, \mathbf{p}^a \right) \right] \\
    &= \mathbb{E}_{\mathbf{w}^*, \mathbf{w}} \mathbb{E}_{\mathbf{p}^*, \mathbf{p}} \left[ \prod_{c = 1}^M \sum_{y^c = 0}^C \int_{S^{N - 1}} d \mathbf{x}^c \Prob_\beta \left( \mathbf{x}^c, y^c | \mathbf{w}^*, \mathbf{p}^* \right) \prod_{a = 1}^L \Prob_\beta \left( \mathbf{x}^c, y^c | \mathbf{w}^a, \mathbf{p}^a \right) \right] \\
    &= \mathbb{E}_{\mathbf{w}^*, \mathbf{w}} \mathbb{E}_{\mathbf{p}^*, \mathbf{p}} \left[ \left( \sum_{y = 0}^C \int_{S^{N - 1}} d \mathbf{x} \Prob_\beta \left( \mathbf{x}, y | \mathbf{w}^*, \mathbf{p}^* \right) \prod_{a = 1}^L \Prob_\beta \left( \mathbf{x}, y | \mathbf{w}^a, \mathbf{p}^a \right) \right)^M \right],
\end{align*}
where we redefined $\mathbf{w} = \left\{ \mathbf{w}^a \right\}_{a = 1}^L$ and $\mathbf{p} = \left\{ \mathbf{p}^a \right\}_{a = 1}^L$. In order to simplify the argument of $\left( \cdot \right)^M$, it is convenient to write Eq. (\ref{eq:direct_distribution}) in the form
\begin{gather*}
    \label{eq:alt_direct_distribution}
    \Prob_\beta \left( \mathbf{x}, y | \mathbf{w}^a, \mathbf{p}^a \right) = \sum_{\gamma = 0}^P \Omega_N \left( \beta \left[ 1 - \delta_{\gamma 0} \right] \right)^{-1} p^{a \gamma}_y \exp \left( \beta \left[ 1 - \delta_{\gamma 0} \right] \sum_{i = 1}^N w^{a \gamma}_i x_i \right), \numberthis
\end{gather*}
where the value of $\mathbf{w}^{a 0}$ is arbitrary. Until the end of this Section, all sums over the hidden units will include the index $0$ unless explicitly indicated otherwise. Define $I_y \left( \mathbf{x} \right) = \Prob_\beta \left( \mathbf{x}, y | \mathbf{w}^*, \mathbf{p}^* \right) \prod_{a = 1}^L \Prob_\beta \left( \mathbf{x}, y | \mathbf{w}^a, \mathbf{p}^a \right)$, then
\begin{align*}
    I_y \left( \mathbf{x} \right) &= \left[ \sum_{\gamma_* = 0}^{P^*} \Omega_N \left( \beta^* \left[ 1 - \delta_{\gamma_* 0} \right] \right)^{-1} p^{* \gamma_*}_y \exp \left( \beta^* \left[ 1 - \delta_{\gamma_* 0} \right] \sum_{i = 1}^N w^{* \gamma_*}_i x_i \right) \right] \\
    &\quad \prod_{a = 1}^L \left[ \sum_{\gamma = 0}^P \Omega_N \left( \beta \left[ 1 - \delta_{\gamma 0} \right] \right)^{-1} p^{a \gamma}_y \exp \left( \beta \left[ 1 - \delta_{\gamma 0} \right] \sum_{i = 1}^N w^{a \gamma}_i x_i \right) \right] \\
    &= \sum_{\gamma_* \gamma_1 ... \gamma_L} \Omega_N \left( \beta^* \left[ 1 - \delta_{\gamma_* 0} \right] \right)^{-1} \left[ \prod_{a} \Omega_N \left( \beta \left[ 1 - \delta_{\gamma_a 0} \right] \right) \right]^{-1} \\
    &\quad p^{* \gamma_*}_y \left[ \prod_a p^{a \gamma_a}_y \right] \exp \left( \beta^* \left[ 1 - \delta_{\gamma_* 0} \right] \sum_{i} w^{* \gamma_*}_i x_i + \beta \sum_{a} \left[ 1 - \delta_{\gamma_a 0} \right] \sum_{i} w^{a \gamma_a}_i x_i \right).
\end{align*}
We will now evaluate the integral $\int_{S^{N - 1}} d \mathbf{x} \ I_y \left( \mathbf{x} \right)$. Using Eq. (\ref{eq:thermodynamic_approx_to_vmf}) of Appendix \ref{app:vmf_integration} with $\rho = \rho_{\gamma_* \gamma} := \frac{1}{N} \sqrt{\sum_i \left[ \beta^{*} \left[ 1 - \delta_{\gamma_* 0} \right] w^{* \gamma_*}_i + \beta \sum_{a} \left[ 1 - \delta_{\gamma_a 0} \right] w^{a \gamma_a}_i \right]^2}$, we get
\begin{align*}
    &\int_{S^{N - 1}} d \mathbf{x} \exp \left( \beta^* \left[ 1 - \delta_{\gamma_* 0} \right] \sum_{i} w^{* \gamma_*}_i x_i + \beta \sum_{a} \left[ 1 - \delta_{\gamma_a 0} \right] \sum_{i} w^{a \gamma_a}_i x_i \right) \\
    &\approx \Omega_N \left( 0 \right) \left( 1 + \left( 2 \rho_{\gamma_* \gamma} \right)^2 \right)^{-1/4} \exp \left[ \left( \frac{N}{2} - 1 \right) \eta \left( 2 \rho_{\gamma_* \gamma} \right) \right],
\end{align*}
in the limit of large $N$. Since $\sqrt{\sum_i \left( w^{* \gamma_*}_i \right)^2} = 1$, the square of $\rho_{\gamma_* \gamma}$ simplifies to
\begin{align*}
    \left( \rho_{\gamma_* \gamma} \right)^2 &= \left( \frac{1}{N} \sqrt{\sum_i \left[ \beta^{*} \left[ 1 - \delta_{\gamma_* 0} \right] w^{* \gamma_*}_i + \beta \sum_{a} \left[ 1 - \delta_{\gamma_a 0} \right] w^{a \gamma_a}_i \right]^2} \right)^2 \\
    &= \upsilon^2 \left[ 1 - \delta_{\gamma_* 0} \right] + 2 \frac{\beta}{N} \upsilon \left[ 1 - \delta_{\gamma_* 0} \right] \sum_a \left[ 1 - \delta_{\gamma_a 0} \right] \sum_i w^{* \gamma_*}_i w^{a \gamma_a}_i \\
    &\quad+ \frac{\beta^2}{N^2} \sum_{a, b} \left[ 1 - \delta_{\gamma_a 0} \right] \left[ 1 - \delta_{\gamma_b 0} \right] \sum_i w^{a \gamma_a}_i w^{b \gamma_b}_i,
\end{align*}
where $\upsilon = \frac{\beta^*}{N}$.  To leading order in $\frac{\beta}{N}$ we find
\begin{align*}
    \left( 1 + \left( 2 \rho_{\gamma_* \gamma} \right)^2 \right)^{-1/4} &\approx \left( 1 + \left( 2 \upsilon \left[ 1 - \delta_{\gamma_* 0} \right] \right)^2 \right)^{-1/4} \\
    \text{and} \quad \left( \frac{N}{2} - 1 \right) \eta \left( 2 \rho_{\gamma_* \gamma_a} \right) &\approx \left( \frac{N}{2} - 1 \right) \eta \left( 2 \upsilon \left[ 1 - \delta_{\gamma_* 0} \right] \right) + \beta_{\text{eff}} \left[ 1 - \delta_{\gamma_* 0} \right] \sum_a \left[ 1 - \delta_{\gamma_a 0} \right] \sum_i w^{* \gamma_*}_i w^{a \gamma_a}_i \\
    &\quad+ \frac{\left[ \beta \xi_{\gamma_*} \right]^2}{2 N} \sum_{a, b} \left[ 1 - \delta_{\gamma_a 0} \right] \left[ 1 - \delta_{\gamma_b 0} \right] \sum_i w^{a \gamma_a}_i w^{b \gamma_b}_i,
\end{align*}
where $\beta_{\text{eff}} = \frac{2 \upsilon}{\sqrt{\left[ 2 \upsilon \right]^2 + 1} + 1} \beta$ and $\xi_{\gamma_*} = \delta_{\gamma_* 0} + \sqrt{\frac{2}{\sqrt{\left[ 2 \upsilon \right]^2 + 1} + 1}} \left[ 1 - \delta_{\gamma_* 0} \right]$. Assuming that $\upsilon \gg 1/N$ and $\beta \ll N$, we drop the last term. Finally, we use Eq. (\ref{eq:thermodynamic_approx_to_vmf}) backwards with $\rho = \upsilon \left[ 1 - \delta_{\gamma_* 0} \right]$ and obtain 
\begin{align*}
    &\Omega_N \left( \beta^* \left[ 1 - \delta_{\gamma_* 0} \right] \right)^{-1} \int_{S^{N - 1}} d \mathbf{x} \exp \left( \beta^* \left[ 1 - \delta_{\gamma_* 0} \right] \sum_{i} w^{* \gamma_*}_i x_i + \beta \sum_{a} \left[ 1 - \delta_{\gamma_a 0} \right] \sum_{i} w^{a \gamma_a}_i x_i \right) \\
    &\approx \exp \left( \beta_{\text{eff}} \left[ 1 - \delta_{\gamma_* 0} \right] \sum_a \left[ 1 - \delta_{\gamma_a 0} \right] \sum_i w^{* \gamma_*}_i w^{a \gamma_a}_i \right),
\end{align*}
from which we conclude that
\begin{align*}
    \sum_{y = 0}^C \int_{S^{N - 1}} d \mathbf{x} \ I_y \left( \mathbf{x} \right) &\approx \sum_{\gamma_* \gamma_1 ... \gamma_L} \sum_{y} p^{* \gamma_*}_y \left[ \prod_{a} p^{a \gamma_a}_y \right] \left[ \prod_{a} \Omega_N \left( \beta \left[ 1 - \delta_{\gamma_a 0} \right] \right) \right]^{-1} \\
    &\quad \exp \left( \beta_{\text{eff}} \left[ 1 - \delta_{\gamma_* 0} \right] \sum_a \left[ 1 - \delta_{\gamma_a 0} \right] \sum_i w^{* \gamma_*}_i w^{a \gamma_a}_i \right).
\end{align*}
We define $\alpha = M / N$, introduce the order parameter
\begin{equation}
\label{eq:alt_order_parameters}
    m^{a \gamma_* \nu} \hspace{-7pt} \quad \text{for} \ \sum_i w^{* \gamma_*}_i w^{a \nu}_i,
\end{equation}
and insert into $\left\langle \mathcal{Z}^L \right\rangle$ the identity operator
\begin{align*}
    1 &= \int_{\mathbb{R}} \prod_{\gamma_*, \nu ; a} d m^{a \gamma_* \nu}
    \delta \left( m^{a \gamma_* \nu} - \sum_i w^{* \gamma_*}_i w^{a \nu}_i \right) \\
    &= \int_{\mathbb{R}} \prod_{\gamma_*, \nu ; a} d m^{a \gamma_* \nu}
    \int_{i \mathbb{R}} \prod_{\gamma_*, \nu ; a} d \hat{m}^{a \gamma_* \nu} \exp \left\{ \frac{\beta_{\text{eff}} \alpha}{P^*} \sum_{\gamma_*, \nu ; a} \hat{m}^{a \gamma_* \nu} \left( \sum_i w^{* \gamma_*}_i w^{a \nu}_i - m^{a \gamma_* \nu} \right) \right\}
\end{align*}
so that we can rewrite it as
\begin{gather*}
\label{eq:replicated_partition_function}
\begin{aligned}
    \left\langle \mathcal{Z}^L \right\rangle &= \int \prod_{\gamma_*, \nu ; a} d \hat{m}^{a \gamma_* \nu} d m^{a \gamma_* \nu} \mathbb{E}_{\mathbf{w}^*, \mathbf{w}} \exp \left\{ N H_{S} \left( \mathbf{w}, \mathbf{w}^* ; \mathbf{\hat{m}} \right) \right\}
    \exp \left\{ -N H_Q \left( \mathbf{m}, \mathbf{\hat{m}} \right) \right\} \\
    &\quad \mathbb{E}_{\mathbf{p}^*, \mathbf{p}} \exp \left\{ \alpha N \log \left[ \sum_{\gamma_* \gamma_1 ... \gamma_L} \sum_{y} p^{* \gamma_*}_y \left[ \prod_{a} p^{a \gamma_a}_y \right] \exp \left\{ H_E \left( \gamma, \gamma_* ; \mathbf{m} \right) \right\} \right] \right\},
\end{aligned} \numberthis \\[12pt]
\begin{aligned}
    \text{where} \quad H_{S} \left( \mathbf{w}, \mathbf{w}^* ; \mathbf{\hat{m}} \right) &= \frac{\beta_{\text{eff}} \alpha}{P^*} \sum_{\gamma_*, \nu ; a} \hat{m}^{a \gamma_* \nu} \sum_i w^{* \gamma_*}_i w^{a \nu}_i \\
    H_Q \left( \mathbf{m}, \mathbf{\hat{m}} \right) &= \frac{\beta_{\text{eff}} \alpha}{P^*} \sum_{\gamma_*, \nu ; a} \hat{m}^{a \gamma_* \nu} m^{a \gamma_* \nu}, \\
    H_E \left( \gamma, \gamma_* ; \mathbf{m} \right) &= -\sum_a \log \left[ \Omega_N \left( \beta \left[ 1 - \delta_{\gamma_a 0} \right] \right) \right] + \beta_{\text{eff}} \left[ 1 - \delta_{\gamma_* 0} \right] \sum_a \left[ 1 - \delta_{\gamma_a 0} \right] m^{a \gamma_* \gamma_a}.
\end{aligned}
\end{gather*}
Using Eq. (\ref{eq:thermodynamic_approx_to_vmf}) of Appendix \ref{app:vmf_integration}, the exponential of $H_{S}$ integrates to
\begin{align*}
    &\mathbb{E}_{\mathbf{w}} \exp \left\{ N H_{S} \left( \mathbf{w}, \mathbf{w}^* ; \mathbf{\hat{m}} \right) \right\} \\
    &= \Omega_N \left( 0 \right)^{-P L} \int_{S^{N - 1}} d \mathbf{w} \exp \left( \frac{\beta_{\text{eff}} \alpha}{P^*} \sum_{\gamma_*, \nu ; a} \hat{m}^{a \gamma_* \nu} \sum_i w^{* \gamma_*}_i w^{a \nu}_i \right) \\
    &\approx \prod_{\nu ; a} \exp \left[ -\frac{1}{4} \log \left( 1 + \left[ \frac{2 \beta_{\text{eff}} \alpha}{P^*} \right]^2 \sum_i \left[ \sum_{\gamma_*} \hat{m}^{a \gamma_* \nu} w^{* \gamma_*}_i \right]^2 \right) \right] \\
    &\quad \exp \left[ \left( \frac{N}{2} - 1 \right) \eta \left( \frac{2 \beta_{\text{eff}} \alpha}{P^*} \sqrt{\sum_i \left[ \sum_{\gamma_*} \hat{m}^{a \gamma_* \nu} w^{* \gamma_*}_i \right]^2} \right) \right].
\end{align*}
In order to continue the calculations, we make the replica-symmetric ansatz
\begin{gather*}
    m^{a \gamma_* \nu} = m^{\gamma_* \nu} \hspace{-7pt} \quad \text{and} \ \hat{m}^{a \gamma_* \nu} = \hat{m}^{\gamma_* \nu} \ \text{for all} \ a ; \gamma_*, \nu \\
    p^{a \nu}_y = p^{\nu}_y \ \text{for all} \ a ; \nu
\end{gather*}
so that we can use the replica trick to simplify
\begin{align*}
    &\frac{1}{L} \log \left[ \sum_{\gamma_* \gamma_1 ... \gamma_L} \sum_{y} p^{* \gamma_*}_y \left[ \prod_{a} p^{a \gamma_a}_y \right] \exp \left\{ H_E \left( \gamma, \gamma_* ; \mathbf{m} \right) \right\} \right] \\
    &= \frac{1}{L} \log \left[ \sum_{\gamma_*} \sum_{y} p^{* \gamma_*}_y \sum_{\gamma_1 ... \gamma_L} \left[ \prod_{a} p^{a \gamma_a}_y \right] \exp \left\{ H_E \left( \gamma, \gamma_* ; \mathbf{m} \right) \right\} \right] \\
    &\approx \sum_{\gamma_*} \sum_{y} p^{* \gamma_*}_y \log \Bigg[ \sum_{\gamma} p^\gamma_y \Omega_N \left( \beta \left[ 1 - \delta_{\gamma 0} \right] \right)^{-1} \exp \left( \beta_{\text{eff}} \left[ 1 - \delta_{\gamma_* 0} \right] \left[ 1 - \delta_{\gamma 0} \right] m^{\gamma_* \gamma} \right) \Bigg],
\end{align*}
as we take $L$ to zero. Similarly, we get
\begin{align*}
    \frac{1}{L} \log \left[ \mathbb{E}_{\mathbf{w}} \exp \left\{ N H_{S} \left( \mathbf{w}, \mathbf{w}^* ; \mathbf{\hat{m}} \right) \right\} \right] &= -\frac{1}{4} \log \left( 1 + \left[ \frac{2 \beta_{\text{eff}} \alpha}{P^*} \right]^2 \sum_i \left[ \sum_{\gamma_*} \hat{m}^{\gamma_* \nu} w^{* \gamma_*}_i \right]^2 \right) \\
    &\quad+ \left( \frac{N}{2} - 1 \right) \eta \left( \frac{2 \beta_{\text{eff}} \alpha}{P^*} \sqrt{\sum_i \left[ \sum_{\gamma_*} \hat{m}^{\gamma_* \nu} w^{* \gamma_*}_i \right]^2} \right).
\end{align*}
From now on, we assume for simplicity that $\sum_y p^{* 0}_y = p_{\mathbf{h}}^* \left( 0 \right) = 0$ and $\sum_y p^{* \mu_*}_y = p_{\mathbf{h}}^* \left( \mu_* \right) = 1/P^*$ for all $\mu_* > 0$. Defining $g^{* \mu_*}_y = P^* p^{* \mu_*}_y = \Prob_\beta \left( y | \mu_*, \mathbf{J} \right)$ for all $\mu_* > 0$, $\mathbf{g}^* = \left\{ g^{* \gamma_*}_y \right\}_{0 \leq y \leq C}^{1 \leq \mu_* \leq P^*}$ and $\varrho = \frac{\alpha}{P^*} = \frac{M}{P^* N}$, the free entropy $f$ then takes the form
\begin{align*}
    \label{eq:var_free_entropy}
    &f \approx \Extr_{\mathbf{m}, \mathbf{\hat{m}}, \mathbf{p}} \left\{ f \left( \mathbf{m}, \mathbf{\hat{m}}, \mathbf{p} \right) \right\} \quad \text{such that} \quad \sum_{\gamma = 0}^P p^\gamma_y = p_{\mathbf{q}} \left( y \right) \quad \text{and} \quad \sum_{y = 0}^C p^\gamma_y = p_{\mathbf{h}} \left( \gamma \right) \numberthis \\
    &\text{with} \quad f \left( \mathbf{m}, \mathbf{\hat{m}}, \mathbf{p} \right) = -\beta_{\text{eff}} \varrho \sum_{\gamma_*, \gamma = 0}^{P^*, P} \hat{m}^{\gamma_* \gamma} m^{\gamma_* \gamma} + \frac{1}{2} \mathbb{E}_{\mathbf{w}^*} \left[ \sum_{\gamma = 0}^P \eta \left( 2\beta_{\text{eff}} \varrho \sqrt{\sum_{i = 1}^N \left[ \sum_{\gamma_* = 0}^{P^*} \hat{m}^{\gamma_* \gamma} w^{* \gamma_*}_i \right]^2} \right) \right] \\
    &\quad+ \varrho \sum_{\mu_* = 1}^{P^*} \sum_{y = 0}^C \mathbb{E}_{\mathbf{g}^*} \left[ g^{* \mu_*}_y \right] \log \left[ \sum_{\gamma = 0}^P p^\gamma_y \Omega_N \left( \beta \left[ 1 - \delta_{\gamma 0} \right] \right)^{-1} \exp \left( \beta_{\text{eff}} \left[ 1 - \delta_{\gamma 0} \right] m^{\mu_* \gamma} \right) \right]
\end{align*}
where we approximated the expectation over $\mathbf{p}$ as a (constrained) extremization problem using Laplace's method. By inspection, the order parameters $m^{0 \gamma}$ and $\hat{m}^{0 \gamma}$ always vanish, so we ignore them in the incoming derivation of the saddle-point equations.

\section{Saddle-point equations}
\label{app:saddle-point}
In this Appendix, we adopt the convention $1 \leq \mu_* \leq P^*$, $0 \leq \gamma \leq P$ and $1 \leq \mu \leq P$. By the Lagrange multiplier theorem, any set of class weights $\mathbf{p}$ that extremizes Eq. (\ref{eq:var_free_entropy}) must solve the extremization problem
\begin{align*}
    \Extr_{\mathbf{p}, \omega, \lambda} \left\{ f \left( \mathbf{m}, \mathbf{\hat{m}}, \mathbf{p} \right) + \sum_{y = 0}^C \lambda_y \left( \sum_{\gamma = 0}^P p^\gamma_y - p_{\mathbf{q}} \left( y \right) \right) + \sum_{\gamma = 0}^P \omega^{\gamma} \left( \sum_{y = 0}^C p^\gamma_y - p_{\mathbf{h}} \left( \gamma \right) \right) \right\}.
\end{align*}
In Appendix \ref{app:weight_normalization}, we show that its solution is
\begin{align*}
    \bar{p}^\gamma_y &= \sum_{\mu_* = 1}^{P^*} \mathbb{E}_{\mathbf{g}^*} \left[ g^{* \mu_*}_y \right] \sigma_\gamma \left( \beta_{\text{eff}} \left[ 1 - \delta_{\gamma 0} \right] m^{\mu_*} - \log \left[ \Omega_N \left( \beta \left[ 1 - \delta_{\gamma 0} \right] \right) \right] + \log \left[ \mathbf{p}_y \right] \right) \\
    p^\gamma_y &= \frac{\bar{p}^\gamma_y}{\zeta^\gamma_y \left( \mathbf{\bar{p}} ; p_{\mathbf{h}} \right)},
\end{align*}
for all $0 \leq \gamma \leq P$, where $\zeta^\gamma_y \left( \mathbf{\bar{p}} ; p_{\mathbf{h}} \right)$ is defined using Eqs. (\ref{eq:nonlinear}) of Appendix \ref{app:weight_normalization} and $\sigma_{\gamma} \left( x^{\mu_*} \right) = \frac{\exp \left( x^{\mu_* \gamma} \right)}{\sum_{\kappa = 0}^P \exp \left( x^{\mu_* \kappa} \right)}$ is the softmax function. We extremize Eq. (\ref{eq:var_free_entropy}) with respect to the remaining parameters by solving for the gradient equal to zero. $\partial_{m^{\mu_* \mu}} f \left( \mathbf{m}, \mathbf{\hat{m}}, \mathbf{p} \right) = 0$ immediately yields
\begin{align*}
    \label{eq:prelim_saddle_point}
    \hat{m}^{\mu_* \gamma} &= \left[ 1 - \delta_{\gamma 0} \right] \sum_{y = 0}^C \mathbb{E}_{\mathbf{g}^*} \left[ g^{* \mu_*}_y \right] \frac{p^\gamma_y \Omega_N \left( \beta \left[ 1 - \delta_{\gamma 0} \right] \right)^{-1} \exp \left( \beta_{\text{eff}} \left[ 1 - \delta_{\gamma 0} \right] m^{\mu_* \gamma} \right)}{\sum_{\kappa = 0}^P p^\kappa_y \Omega_N \left( \beta \left[ 1 - \delta_{\kappa 0} \right] \right)^{-1} \exp \left( \beta_{\text{eff}} \left[ 1 - \delta_{\kappa 0} \right] m^{\mu_* \kappa} \right)}. \numberthis
\end{align*}
On the other hand, $\partial_{\hat{m}^{\mu_* \gamma}} f \left( \mathbf{m}, \mathbf{\hat{m}}, \mathbf{p} \right) = 0$ gives an equation that depends on the choice of prior for the teacher memories $\mathbf{w}^{* \mu_*}$. We first investigate the case where the teacher memories $\mathbf{w}^{* \mu_*}$ and the class weights $\mathbf{g}^*$ are distributed uniformly at random over the sets to which they are constrained (see Section \ref{sec:teacher-student}). In this scenario, we have $\mathbb{E}_{\mathbf{g}^{*}} \left[ g^{* \mu_*}_y \right] = p_{\mathbf{q}}^* \left( y \right)$. Moreover, assuming that $P^* \ll N$, random vectors on the unit sphere are orthonormal with high probability, which means that
\begin{align*}
    \sum_{i = 1}^N \left[ \sum_{\mu_* = 1}^{P^*} \hat{m}^{\mu_* \gamma} w^{* \mu_*}_i \right]^2 &= \sum_{i = 1}^N \sum_{\mu_*, \nu_* = 1}^{P^*} \hat{m}^{\mu_* \gamma} \hat{m}^{\nu_* \gamma} w^{* \mu_*}_i w^{* \nu_*}_i \\
    &= \sum_{\mu_* = 1}^{P^*} \left[ \hat{m}^{\mu_* \gamma} \right]^2 + \sum_{\mu_* \neq \nu_*} \hat{m}^{\mu_* \gamma} \hat{m}^{\nu_* \gamma} \sum_{i = 1}^N w^{* \mu_*}_i w^{* \nu_*}_i \\
    &\approx \sum_{\mu_* = 1}^{P^*} \left[ \hat{m}^{\mu_* \gamma} \right]^2.
\end{align*}
Therefore, $\partial_{\hat{m}^{\mu_* \gamma}} f \left( \mathbf{m}, \mathbf{\hat{m}}, \mathbf{p} \right) = 0$ gives
\begin{align*}
    0 &= \partial_{\hat{m}^{\mu_* \gamma}} f \left( \mathbf{m}, \mathbf{\hat{m}}, \mathbf{p} \right) \\
    0 &= -\beta_{\text{eff}} \varrho m^{\mu_* \gamma} + \frac{1}{2} \partial_{\hat{m}^{\mu_* \gamma}} \sum_{\kappa = 0}^P \eta \left( 2 \beta_{\text{eff}} \varrho \sqrt{\sum_{\nu_* = 1}^{P^*} \left[ \hat{m}^{\nu_* \kappa} \right]^2} \right) \\
    m^{\mu_* \gamma} &= \varsigma \left( 2 \beta_{\text{eff}} \varrho \sqrt{\sum_{\nu_* = 1}^{P^*} \left[ \hat{m}^{\nu_* \gamma} \right]^2} \right) \frac{\hat{m}^{\mu_* \gamma}}{\sqrt{\sum_{\nu_* = 1}^{P^*} \left[ \hat{m}^{\nu_* \gamma} \right]^2}},
\end{align*}
where $\varsigma \left( x \right) = \frac{x}{\sqrt{x^2 + 1} + 1}$. $\hat{m}^{\mu_* \gamma}$ vanishes (see Eqs. \ref{eq:prelim_saddle_point}) and $m^{\mu_* \gamma}$ is arbitrary (see Eqs. \ref{eq:alt_direct_distribution} and \ref{eq:alt_order_parameters}) when $\gamma = 0$. Therefore, we may update only $\hat{m}^{\mu_* \mu}$ and $m^{\mu_* \mu}$ with $1 \leq \mu \leq P$ when solving for $\nabla f = 0$ by fixed-point iteration. Defining $m^{\mu_* 0} = \frac{1}{\beta_{\text{eff}}} \log \left[ \Omega_N \left( \beta \right) / \Omega_N \left( 0 \right) \right]$, we then obtain the saddle-point equations
\begin{align*}
    \hat{m}^{\mu_* \mu} &= \sum_{y = 0}^C p_{\mathbf{q}}^* \left( y \right) \sigma_\mu \left( \beta_{\text{eff}} m^{\mu_*} + \log \left[ \mathbf{p}_y \right] \right) \\
    m^{\mu_* \mu} &= \varsigma \left( 2 \beta_{\text{eff}} \varrho \sqrt{\sum_{\nu_* = 1}^{P^*} \left[ \hat{m}^{\nu_* \mu} \right]^2} \right) \frac{\hat{m}^{\mu_* \mu}}{\sqrt{\sum_{\nu_* = 1}^{P^*} \left[ \hat{m}^{\nu_* \mu} \right]^2}},
\end{align*}
for all $1 \leq \mu \leq P$, where we simplified the argument of the softmax function $\sigma_{\mu} \left( x^{\mu_*} \right) = \frac{\exp \left( x^{\mu_* \mu} \right)}{\sum_{\kappa = 0}^P \exp \left( x^{\mu_* \kappa} \right)}$ using $m^{\mu_* 0} = \frac{1}{\beta_{\text{eff}}} \log \left[ \Omega_N \left( \beta \right) / \Omega_N \left( 0 \right) \right]$. Putting the equations for $m^{\mu_* \mu}$ and $p^\gamma_y$ together, we find
\begin{align*}
    \hat{m}^{\mu_* \mu} &= \sum_{y = 0}^C p_{\mathbf{q}}^* \left( y \right) \sigma_\mu \left( \beta_{\text{eff}} m^{\mu_*} + \log \left[ \mathbf{p}_y \right] \right) \\
    \bar{p}^\gamma_y &= p_{\mathbf{q}}^* \left( y \right) \sum_{\mu = 1}^{P^*} \sigma_\gamma \left( \beta_{\text{eff}} m^{\mu_*} + \log \left[ \mathbf{p}_y \right] \right) \\
    m^{\mu_* \mu} &= \varsigma \left( 2 \beta_{\text{eff}} \varrho \sqrt{\sum_{\nu_* = 1}^{P^*} \left[ \hat{m}^{\nu_* \mu} \right]^2} \right) \frac{\hat{m}^{\mu_* \mu}}{\sqrt{\sum_{\nu_* = 1}^{P^*} \left[ \hat{m}^{\nu_* \mu} \right]^2}} \\
    p^\gamma_y &= \frac{\bar{p}^\gamma_y}{\zeta^\gamma_y \left( \mathbf{\bar{p}} ; p_{\mathbf{h}} \right)}
\end{align*}
for all $1 \leq \mu \leq P$ and $0 \leq \gamma \leq P$. If we instead clamp $\mathbf{w}^{* \mu_*}$ and $\mathbf{g}^{* \mu_*}$ to some fixed patterns $\mathbf{x}^{* \mu_*}$ and their soft labels $\mathbf{q}^{* \mu_*}$, respectively, then the solutions of Eq. (\ref{eq:var_free_entropy}) take the form
\begin{align*}
    \hat{m}^{\mu_* \mu} &= \sum_{y = 0}^C q^{* \mu_*}_y \sigma_\mu \left( \beta_{\text{eff}} m^{\mu_*} + \log \left[ \mathbf{p}_y \right] \right) \\
    \bar{p}^\gamma_y &= \sum_{\mu_* = 1}^{P^*} q^{* \mu_*}_y \sigma_\gamma \left( \beta_{\text{eff}} m^{\mu_*} + \log \left[ \mathbf{p}_y \right] \right) \\
    m^{\mu_* \mu} &= \varsigma \left( 2 \beta_{\text{eff}} \varrho \sqrt{\sum_{i = 1}^N \left[ \sum_{\nu_* = 1}^{P^*} \hat{m}^{\nu_* \mu} x^{* \nu_*}_i \right]^2} \right) \frac{\sum_{i = 1}^N x^{* \nu_*}_i \sum_{\nu_* = 1}^{P^*} \hat{m}^{\nu_* \mu} x^{* \nu_*}_i}{\sqrt{\sum_{i = 1}^N \left[ \sum_{\nu_* = 1}^{P^*} \hat{m}^{\nu_* \mu} x^{* \nu_*}_i \right]^2}} \\
    p^\gamma_y &= \frac{\bar{p}^\gamma_y}{\zeta^\gamma_y \left( \mathbf{\bar{p}} ; p_{\mathbf{h}} \right)}.
\end{align*}
Defining $\bar{x}^\mu_i = \sum_{\mu_*} \hat{m}^{\mu_* \mu} x^{* \mu_*}_i$, we thus find
\begin{align*}
    \bar{x}^\mu_i &= \sum_{\mu_* = 1}^{P^*} x^{* \mu_*}_i \sum_{y = 0}^C q^{* \mu_*}_y \sigma_{\mu} \left( \beta_{\text{eff}} m^{\mu_*} + \log \left[ \mathbf{p}_y \right] \right) \\
    \bar{p}^\gamma_y &= \sum_{\mu_* = 1}^{P^*} q^{* \mu_*}_y \sigma_\gamma \left( \beta_{\text{eff}} m^{\mu_*} + \log \left[ \mathbf{p}_y \right] \right) \\
    m^{\mu_* \mu} &= \varsigma \left( 2 \beta_{\text{eff}} \varrho \sqrt{\sum_{i = 1}^N \left[ \bar{x}^\mu_i \right]^2} \right) \frac{\sum_{i = 1}^N x^{* \mu_*}_i \bar{x}^\mu_i}{\sqrt{\sum_{i = 1}^N \left[ \bar{x}^\mu_i \right]^2}} \\
    p^\gamma_y &= \frac{\bar{p}^\gamma_y}{\zeta^\gamma_y \left( \mathbf{\bar{p}} ; p_{\mathbf{h}} \right)},
\end{align*}
for all $1 \leq \mu \leq P$ and $0 \leq \gamma \leq P$.

\section{Saddle-point hierarchy}
\label{app:fixed-point}
Suppose that the set of parameters $\bar{x}^{\text{fixed}, \mu}_i$, $\bar{p}^{\text{fixed}, \gamma}_y$, $m^{\text{fixed}, \mu_* \gamma}$, $p^{\text{fixed}, \gamma}_y$ with hidden unit prior $p_{\mathbf{h}}^{\text{given}} \left( \gamma \right)$ is a fixed point of Eqs. (\ref{eq:saddle-point}) with $P$ hidden units. Substitute into the same saddle-point equations with $P + R \in \left\{ P, ..., 2 P \right\}$ hidden units the duplicated order parameters
\begin{align*}
    \bar{x}^{\text{dupli}, \mu}_i
    &= \begin{cases}
        \bar{x}^{\text{fixed}, \mu}_i &\quad 0 < \mu \leq P \\
        \bar{x}^{\text{fixed}, \mu - P}_i &\quad P < \mu \leq P + R \\
    \end{cases} \\
    \bar{p}^{\text{dupli}, \gamma}_y
    &= \begin{cases}
        \bar{p}^{ \text{fixed}, 0}_y &\quad \gamma = 0 \\
        \frac{1}{2} \bar{p}^{ \text{fixed}, \gamma}_y &\quad 0 < \gamma \leq R \\
        \bar{p}^{\text{fixed}, \gamma}_y &\quad R < \gamma \leq P \\
        \frac{1}{2} \bar{p}^{ \text{fixed}, \gamma - P}_y &\quad P < \gamma \leq P + R \\
    \end{cases} \\
    m^{\text{dupli}, \mu_* \gamma}
    &= \begin{cases}
        m^{\text{fixed}, \mu_* 0} &\quad \gamma = 0 \\
        m^{\text{fixed}, \mu_* \gamma} &\quad 0 < \gamma \leq P \\
        m^{\text{fixed}, \mu_*, \gamma - P} &\quad P < \gamma \leq P + R \\
    \end{cases} \\
    p^{\text{dupli}, \gamma}_y
    &= \begin{cases}
        p^{\text{fixed}, 0}_y &\quad \gamma = 0 \\
        \frac{1}{2} p^{\text{fixed}, \gamma}_y &\quad 0 < \gamma \leq R \\
        p^{\text{fixed}, \gamma}_y &\quad R < \gamma \leq P \\
        \frac{1}{2} p^{\text{fixed}, \gamma - P}_y &\quad P < \gamma \leq P + R \\
    \end{cases}
\end{align*}
\begin{align*}
    \text{along with} \quad p_{\mathbf{h}} \left( \gamma \right)
    &= \begin{cases}
        p_{\mathbf{h}}^{\text{given}} \left( 0 \right) &\quad \gamma = 0 \\
        \frac{1}{2} p_{\mathbf{h}}^{\text{given}} \left( \gamma \right) &\quad 0 < \gamma \leq R \\
        p_{\mathbf{h}}^{\text{given}} \left( \gamma \right) &\quad R < \gamma \leq P \\
        \frac{1}{2} p_{\mathbf{h}}^{\text{given}} \left( \gamma - P \right) &\quad P < \gamma \leq P + R,
    \end{cases}
\end{align*}
where the hidden units $\gamma \in \left\{ P + 1, ..., P + R \right\}$ and their corresponding order parameters are duplicates, or copies, of $\gamma \in \left\{ 1, ..., R \right\}$. $\gamma = 0$ can also be duplicated, but the result is less interesting. By definition (see Eqs. \ref{eq:nonlinear}), $\zeta^\gamma_y \left( \mathbf{\bar{p}}^{\text{dupli}} ; p_{\mathbf{h}} \right) = \lambda_y \left( \mathbf{\bar{p}}^{\text{dupli}} ; p_{\mathbf{h}} \right) + \omega^\gamma \left( \mathbf{\bar{p}}^{\text{dupli}} ; p_{\mathbf{h}} \right)$, where $\lambda_y \left( \mathbf{\bar{p}}^{\text{dupli}} ; p_{\mathbf{h}} \right)$ and $\omega^\gamma \left( \mathbf{\bar{p}}^{\text{dupli}} ; p_{\mathbf{h}} \right)$ are the $\lambda_y$ and $\omega^\gamma$ solving
\begin{align*}
    \omega^\gamma &= \frac{1}{p_{\mathbf{h}} \left( \gamma \right)} \sum_{y = 0}^C \frac{\omega^\gamma}{\lambda_y + \omega^\gamma} \bar{p}^{\text{dupli}, \gamma}_y \\
    &= \frac{1}{p_{\mathbf{h}}^{\text{given}} \left( \gamma \right)} \sum_{y = 0}^C \frac{\omega^\gamma}{\lambda_y + \omega^\gamma} \bar{p}^{\text{fixed}, \gamma}_y \\
    \lambda_y &= \frac{1}{p_{\mathbf{q}} \left( y \right)} \sum_{\gamma = 0}^{P + R} \frac{\lambda_y}{\lambda_y + \omega^\gamma} \bar{p}^{ \text{dupli}, \gamma}_y \\
    &= \frac{1}{p_{\mathbf{q}} \left( y \right)} \sum_{\gamma = 0}^{P} \frac{\lambda_y}{\lambda_y + \omega^\gamma} \bar{p}^{ \text{fixed}, \gamma}_y.
 \end{align*}
Therefore, $\zeta^\gamma_y \left( \mathbf{\bar{p}}^{ \text{dupli}} ; p_{\mathbf{h}} \right) = \zeta^\gamma_y \left( \mathbf{\bar{p}}^{ \text{fixed}} ; p_{\mathbf{h}}^{\text{given}} \right)$, and the saddle-point equations simplify to
\begin{align*}
    \label{eq:replicated_saddle-point}
    \bar{x}^{\text{fixed}, \mu}_i &= \frac{1}{2} \left( 1 + \mathbb{I} \left( \mu > R \right) \right) \sum_{\mu_* = 1}^{P^*} x^{* \mu_*}_i \sum_{y = 0}^C q^{* \mu_*}_y \sigma_{\mu} \left( \beta_{\text{eff}} m^{\text{fixed}, \mu_*} + \log \left[ \mathbf{p}^{\text{fixed}}_y \right] \right) \\
    \bar{p}^{ \text{fixed}, \gamma}_y &= \sum_{\mu_* = 1}^{P^*} q^{* \mu_*}_y \sigma_{\gamma} \left( \beta_{\text{eff}} m^{\text{fixed}, \mu_*} + \log \left[ \mathbf{p}^{\text{fixed}}_y \right] \right) \\
    m^{\text{fixed}, \mu_* \mu} &= \varsigma \left( 2 \beta_{\text{eff}} \varrho \sqrt{\sum_{i = 1}^N \left[ \mathbf{\bar{x}}^{* \text{fixed}, \mu}_i \right]^2} \right) \frac{\sum_{i = 1}^N x^{* \mu_*}_i \mathbf{\bar{x}}^{* \text{fixed}, \mu}_i}{\sqrt{\sum_{i = 1}^N \left[ \mathbf{\bar{x}}^{* \text{fixed}, \mu}_i \right]^2}} \numberthis \\
    p^{\text{fixed}, \gamma}_y &= \frac{\bar{p}^{ \text{fixed}, \gamma}_y}{\zeta^\gamma_y \left( \mathbf{\bar{p}}^{ \text{fixed}} ; p_{\mathbf{h}}^{\text{given}} \right)},
\end{align*}
where $\mathbb{I} \left( \mu > R \right)$ is the indicator function equal to $1$ when $\mu > R$ and $0$ otherwise. Assume that $\varrho \rightarrow \infty$ so that $\varsigma \left( 2 \beta_{\text{eff}} \varrho \sqrt{\sum_{i = 1}^N \left[ \bar{x}^{\mu}_i \right]^2} \right) \rightarrow \mathbb{I} \left( \sqrt{\sum_{i = 1}^N \left[ \bar{x}^{\mu}_i \right]^2} > 0 \right)$, then the saddle-point equations are the same no matter how the prefactor of $\frac{1}{2} \left( 1 + \mathbb{I} \left( \mu > R \right) \right)$ affects the norm of $\bar{x}^{\mu}_i$, and Eqs. (\ref{eq:fixed_point}) are a fixed point of the saddle-point equations with $P + R$ hidden units. In particular, Eq. (\ref{eq:fixed_point}) is a stationary point of the loss (Eq. \ref{eq:loss}) when $\beta_{\text{eff}} = \beta$.

For the rest of this Appendix, all sums are understood as having the same bounds as Eqs. (\ref{eq:replicated_saddle-point}) and (\ref{eq:saddle-point}). The stability of any fixed point of the form $\mathbf{x} = \mathbf{F} (\mathbf{x})$, such as those of Eqs. (\ref{eq:saddle-point}), can be evaluated using the Jacobian matrix $\mathbf{J}$ of $\mathbf{F}$. If all eigenvalues $\lambda$ of the Jacobian satisfy $| \lambda | < 1$, then the fixed point is stable. Conversely, if the Jacobian has an eigenvalue $\lambda$ with $| \lambda | > 1$, then the fixed point is unstable. In particular, if the quadratic form $\mathbf{v}^T \mathbf{J} \mathbf{v}$ is larger than $1$ for some $\mathbf{v}$ with $\| \mathbf{v} \| = 1$, then the fixed point is unstable. For the rest of this Appendix, we evaluate the stability of Eqs. (\ref{eq:saddle-point}) with duplicated order parameters. Keeping $\mathbf{\bar{p}}$ and $\mathbf{p}$ fixed, the Jacobian of the saddle-point equation for $\mathbf{\bar{x}}$ is
\begin{align*}
    \partial_{\bar{x}^{\nu}_k} \bar{x}^{\mu}_j
    &= \sum_{\mu_*} x^{* \mu_*}_j \partial_{\bar{x}^{\nu}_k} m^{\mu_* \nu} \sum_{y} q^{* \mu_*}_y \partial_{m^{\mu_* \nu}} \sigma_{\mu} \left( \beta_{\text{eff}} m^{\mu_*} + \log \left[ \mathbf{p}_y \right] \right) \\
    &= \beta_{\text{eff}} \sum_{\mu_*} x^{* \mu_*}_j \partial_{\bar{x}^{\nu}_k} m^{\mu_* \nu} \sum_{y} q^{* \mu_*}_y \sigma_{\mu} \left( \beta_{\text{eff}} m^{\mu_*} + \log \left[ \mathbf{p}_y \right] \right) \left( \delta_{\mu \nu} - \sigma_{\nu} \left( \beta_{\text{eff}} m^{\mu_*} + \log \left[ \mathbf{p}_y \right] \right) \right) \\
    &= \beta_{\text{eff}} \sum_{\mu_*} x^{* \mu_*}_j \frac{1}{\sqrt{\sum_i \left[ \bar{x}^{\nu}_i \right]^2}} \left( x^{* \mu_*}_k - \left[ \sum_{i} x^{* \mu_*}_i \tilde{x}^{\nu}_i \right] \tilde{x}^{\nu}_k \right) \\
    &\quad \sum_{y} q^{* \mu_*}_y \sigma_{\mu} \left( \beta_{\text{eff}} m^{\mu_*} + \log \left[ \mathbf{p}_y \right] \right) \left( \delta_{\mu \nu} - \sigma_{\nu} \left( \beta_{\text{eff}} m^{\mu_*} + \log \left[ \mathbf{p}_y \right] \right) \right),
\end{align*}
where $\tilde{x}^\nu_j = \frac{\bar{x}^{\nu}_j}{\sqrt{\sum_i \left[ \bar{x}^{\nu}_i \right]^2}}$. Suppose that $\bar{x}^{\mu}_i$, $\bar{p}^\gamma_y$, $m^{\mu_* \gamma}$ and $p^{ \gamma}_y$ are the duplicated parameters of Eq. (\ref{eq:fixed_point}) with $0 < \mu, \nu \leq R$ or $P < \mu, \nu \leq P + R$, then
\begin{gather*}
    \begin{aligned}
    \partial_{\bar{x}^{\nu}_k} \bar{x}^{\mu}_j &= \beta_{\text{eff}} \sum_{\mu_*} x^{* \mu_*}_j \frac{1}{\sqrt{\sum_i \left[ \bar{x}^{\nu}_i \right]^2}} \left( x^{* \mu_*}_k - \left[ \sum_{i} x^{* \mu_*}_i \tilde{x}^{\nu}_i \right] \tilde{x}^{\nu}_k \right) \\
    &\quad \sum_{y} q^{* \mu_*}_y \frac{1}{2} \sigma_{\theta \left( \mu \right) } \left( \beta_{\text{eff}} m^{\text{fixed}, \mu_*} + \log \left[ \mathbf{p}^{\text{fixed}}_y \right] \right) \left( \delta_{\mu \nu} - \frac{1}{2} \sigma_{\theta \left( \nu \right)} \left( \beta_{\text{eff}} m^{\text{fixed}, \mu_*} + \log \left[ \mathbf{p}^{\text{fixed}}_y \right] \right) \right),
    \end{aligned} \\
    \text{where} \quad \theta \left( \mu \right) = \begin{cases}
        \mu &\quad 0 < \mu \leq R \\
        \mu - P &\quad 0 < \mu \leq P + R.
    \end{cases}
\end{gather*}
If $\beta_{\text{eff}}$ is relatively large (for instance of order $\sqrt{N}$), then the softmax $\sigma_{\theta \left( \mu \right)}$ splits the indices $\mu_*$ into a cover $\mathcal{S}$ of sets $\mathcal{S} \left( \mu \right)$ such that $\sigma_{\theta \left( \mu \right)} \left( \beta_{\text{eff}} m^{\text{fixed}, \mu_*} + \log \left[ \mathbf{p}_y \right] \right) \approx \mathbb{I} \left( \mu_* \in \mathcal{S} \left( \mu \right) \right)$. Therefore, we have
\begin{align*}
    \partial_{\bar{x}^{\nu}_k} \bar{x}^{\mu}_j &\approx \beta_{\text{eff}} \sum_{\mu_*} x^{* \mu_*}_j \frac{1}{\sqrt{\sum_i \left[ \bar{x}^{\nu}_i \right]^2}} \left( x^{* \mu_*}_k - \left[ \sum_{i} x^{* \mu_*}_i \tilde{x}^{\nu}_i \right] \tilde{x}^{\nu}_k \right) \\
    &\quad \frac{1}{2} \mathbb{I} \left( \mu_* \in \mathcal{S} \left( \mu \right) \right) \left( \delta_{\mu \nu} - \frac{1}{2} \mathbb{I} \left( \mu_* \in \mathcal{S} \left( \nu \right) \right) \right).
\end{align*}
The subcovers $\mathcal{S} \left( 0 < \mu \leq P \right)$ and $\mathcal{S} \left( R < \mu \leq P + R \right)$ are both partitions of the indices $\mu_*$. As such, $\left[ \partial_{\bar{x}^{\nu}_k} \bar{x}^{\mu}_j \right]_{\mu, \nu = 1}^{P + R}$ is block diagonal with $2 \times 2$ blocks coupling the indices $0 < \mu \leq R$ and $\nu = \mu + P$. Without loss of generality, we investigate the stability of the block
\begin{align*}
    &\begin{bmatrix}
        \partial_{\bar{x}^{1}_k} \bar{x}^{1}_j & \partial_{\bar{x}^{P + 1}_k} \bar{x}^{1}_j \\
        \partial_{\bar{x}^{1}_k} \bar{x}^{P + 1}_j & \partial_{\bar{x}^{P + 1}_k} \bar{x}^{P + 1}_j
    \end{bmatrix} \\
    &= \frac{1}{2} \frac{\beta_{\text{eff}}}{\sqrt{\sum_i \left[ x^{\prime 1}_i \right]^2}} \sum_{\mu_* \in \mathcal{S} (1)}
    \begin{bmatrix}
        x^{* \mu_*}_j \left( x^{* \mu_*}_k - \left[ \sum_{i} x^{* \mu_*}_i \tilde{x}^{1}_i \right] \tilde{x}^{1}_k \right) & -x^{* \mu_*}_j \left( x^{* \mu_*}_k - \left[ \sum_{i} x^{* \mu_*}_i \tilde{x}^{1}_i \right] \tilde{x}^{1}_k \right) \\
        -x^{* \mu_*}_j \left( x^{* \mu_*}_k - \left[ \sum_{i} x^{* \mu_*}_i \tilde{x}^{1}_i \right] \tilde{x}^{1}_k \right) & x^{* \mu_*}_j \left( x^{* \mu_*}_k - \left[ \sum_{i} x^{* \mu_*}_i \tilde{x}^{1}_i \right] \tilde{x}^{1}_k \right)
    \end{bmatrix}.
\end{align*}
where $x^{\prime 1}_i = \sum_{\mu_* \in \mathcal{S} \left( 1 \right)} x^{* \mu_*}_i$. Let $\mathbf{u}^{1}$ be an arbitrary vector orthogonal to $\mathbf{\Tilde{x}}^{1}$ and define
\begin{align*}
    v^{\mu}_j
    &= \frac{1}{\sqrt{2}} \begin{cases}
        u^{1}_j &\quad \mu = 1 \\
        -u^{1}_j &\quad \mu = P + 1,
    \end{cases}
\end{align*}
then we have the quadratic form
\begin{align*}
    \sum_{\nu \in \left\{ 1, P + 1 \right\}} \sum_{k = 1}^N v^{\nu}_k \ \partial_{\bar{x}^{\nu}_k} \bar{x}^{\mu}_j &=
    \frac{\beta_{\text{eff}}}{\sqrt{\sum_i \left[ x^{\prime 1}_i \right]^2}} \sum_{\mu_* \in \mathcal{S} (1)} \frac{1}{\sqrt{2}}
    \begin{bmatrix}
        x^{* \mu_*}_j \sum_k x^{* \mu_*}_k u^{1}_k \\
        -x^{* \mu_*}_j \sum_k x^{* \mu_*}_k u^{1}_k
    \end{bmatrix} \\
    \sum_{\mu, \nu \in \left\{ 1, P + 1 \right\}} \sum_{j, k = 1}^N v^{\mu}_j v^{\nu}_k \ \partial_{\bar{x}^{\nu}_k} \bar{x}^{\mu}_j &= \frac{\beta_{\text{eff}}}{\sqrt{\sum_i \left[ x^{\prime 1}_i \right]^2}} \sum_{\mu_* \in \mathcal{S} (1)} \left[ \sum_j x^{* \mu_*}_j u^{1}_j \right]^2.
\end{align*}
If $\mathbf{u}^1$ is orthogonal to all the $\mathbf{x}^{* \mu_*}$ such that $\mu_* \in \mathcal{S} \left( 1 \right)$, and in particular if $\mathcal{S} \left( 1 \right)$ contains a single pattern $\mathbf{x}^{* \mu_*} = \mathbf{\Tilde{x}}^{1}$, then $\sum_{\mu_* \in \mathcal{S} (1)} \left[ \sum_j x^{* \mu_*}_j u^{1}_j \right]^2 = 0$, so the quadratic form vanishes. In this case, $\mathbf{u}^{1}$ is a stable direction of the saddle-point equations. Otherwise, the quadratic form does not vanish, so there is a $\beta_{\text{split}}$ such that the direction $\mathbf{u}^1$ is unstable when $\beta_{\text{eff}} > \beta_{\text{split}}$.

\section{Weights learned with unsupervised training}
\label{app:unsupervised_weights}
This Appendix contains plots of DAM weights learned in an unsupervised way (see Section \ref{sec:interpretability}) and sorted in increasing $y = \argmax_{y^\prime} \left\{ p^\mu_{y^\prime} \right\}$. They are not in the main text because they take a lot of space.

\begin{figure}
    \centering
    \includegraphics[width=0.495\linewidth]{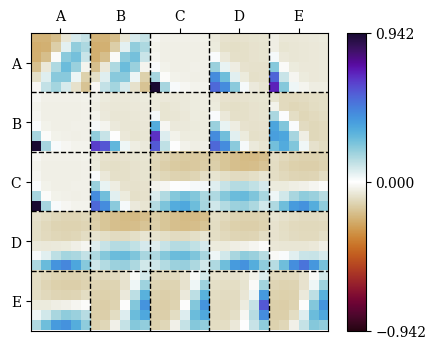}
    \includegraphics[width=\linewidth]{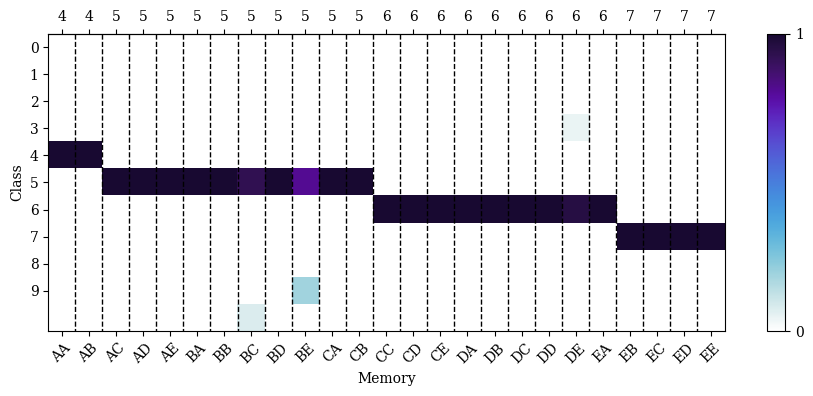}
    \caption{In the top panel, $25$ of the $P = 100$ memories $\mathbf{w}^\mu$ learned by an instance of our dense associative memory (DAM) model trained in an unsupervised way (Eq. \ref{eq:unsupervised_loss}) on $6 \times 6$ patches of the MNIST dataset of handwritten digits \cite{lecun1998gradient} while assuming $C = 10$ latent classes and $\varsigma = 0.6$. In the bottom panel, the corresponding rescaled class weights $\mathbf{p}^\mu / p_{\mathbf{h}} \left( \mu \right)$, where $p_{\mathbf{h}} \left( \gamma \right) = \frac{1}{P + 1}$ for all $0 \leq \gamma \leq P$. The hidden units are indexed using pairs of letters from A to E, and the column-wise maxima of the class weights are the classes of the memories with the corresponding letter indices. $\mathbf{w}^\mu$ and $\mathbf{p}^\mu$ are sorted in increasing $y = \argmax_{y^\prime} \left\{ p^\mu_{y^\prime} \right\}$, and this figure shows $51 \leq \mu \leq 75$.}
    \label{fig:unsupervised_DAM_memories_3}
\end{figure}

\begin{figure}
    \centering
    \includegraphics[width=0.495\linewidth]{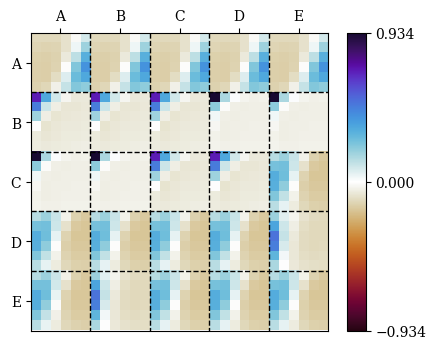}
    \includegraphics[width=\linewidth]{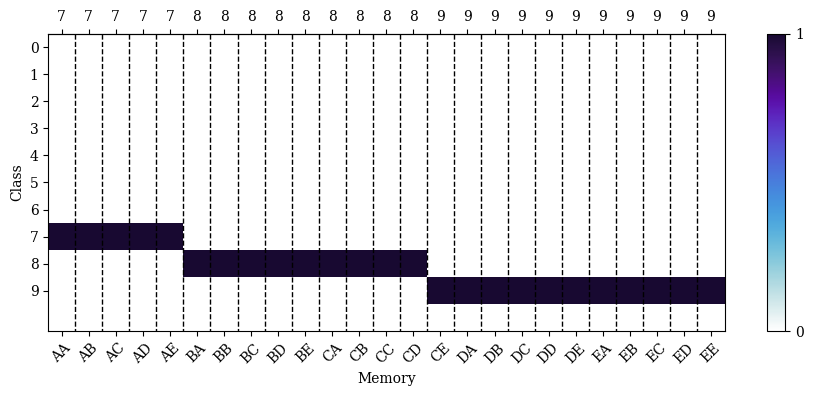}
    \caption{In the top panel, $25$ of the $P = 100$ memories $\mathbf{w}^\mu$ learned by an instance of our dense associative memory (DAM) model trained in an unsupervised way (Eq. \ref{eq:unsupervised_loss}) on $6 \times 6$ patches of the MNIST dataset of handwritten digits \cite{lecun1998gradient} while assuming $C = 10$ latent classes and $\varsigma = 0.6$. In the bottom panel, the corresponding rescaled class weights $\mathbf{p}^\mu / p_{\mathbf{h}} \left( \mu \right)$, where $p_{\mathbf{h}} \left( \gamma \right) = \frac{1}{P + 1}$ for all $0 \leq \gamma \leq P$. The hidden units are indexed using pairs of letters from A to E, and the column-wise maxima of the class weights are the classes of the memories with the corresponding letter indices. $\mathbf{w}^\mu$ and $\mathbf{p}^\mu$ are sorted in increasing $y = \argmax_{y^\prime} \left\{ p^\mu_{y^\prime} \right\}$, and this figure shows $76 \leq \mu \leq 100$.}
    \label{fig:unsupervised_DAM_memories_4}
\end{figure}
\pagebreak

\section{Splitting steepest descent}
\label{app:splitting}
Steps 5 and 11 of splitting steepest descent (Alg. \ref{alg:splitting_descent}) involve the splitting matrices $\mathcal{S}_\mu \left( \mathbf{w}, \mathbf{p} \right)$ derived in \cite{wu2019splitting}. In this appendix, we explain their role in the algorithm. Consult \cite{wu2019splitting} for a detailed explanation of their interpretation and theoretical underpinnings.

Define the thresholds $\tau_{\text{thres}} \in \left( 0, 1 \right]$ and $\lambda_{\text{thres}} \leq 0$. At step 5 of Alg. (\ref{alg:splitting_descent}), we duplicate the hidden units corresponding to the $R \leq \min \left\{ \tau_{\text{thres}} P_{\text{cur}}, P_{\text{max}} - P_{\text{cur}} \right\}$ most negative minimum eigenvalues $\lambda_{\text{min}}^\mu$ of the splitting matrices $\mathcal{S}_\mu \left( \mathbf{w}, \mathbf{p} \right)$ such that $\lambda_{\text{min}}^\mu \leq \lambda_{\text{thres}}$ \cite{wu2019splitting}. To make Fig. (\ref{fig:performance}), we pick $\tau_{\text{thres}} = 1$ and $\lambda_{\text{thres}} \approx 0$. Our splitting matrices are
\begin{align*}
    \mathcal{S}_\mu \left( \mathbf{w}, \mathbf{p} \right) = -\mathbb{E}_{\mathbf{x}^*, y^*} \left[ \frac{p^\mu_y \Bar{\nabla}_{\mathbf{w}^\mu} \Bar{\nabla}_{\mathbf{w}^\mu}^T \exp \left( \beta_{\text{eff}} \sum_{i = 1}^N w^\mu_i x_i \right)}{\sum_{\nu = 1}^P p^{\nu}_y \exp \left( \beta_{\text{eff}} \sum_{i = 1}^N w^\nu_i x_i \right) + p^0_y \frac{\Omega_N \left( \beta \right)}{\Omega_N \left( 0 \right)}} \right],
\end{align*}
where $\Bar{\nabla}_{\bm{\theta}}$ is the gradient constrained to the unit hypersphere $S^{N - 1}$ (see Appendix \ref{app:weight_normalization}) and $\beta_{\text{eff}}$ can be written explicitly in terms of $\beta$ as $\beta_{\text{eff}} = \varsigma \beta$ (see Eqs. \ref{eq:effective_loss} and \ref{eq:saddle-point}). At step 11 of Alg. (\ref{alg:splitting_descent}), we break permutation symmetries in the memories $\mathbf{w}^\mu$ by descending along the eigenvectors $\mathbf{u}^\mu \in S^{N - 1}$ corresponding to the eigenvalues $\lambda_{\text{min}}^\mu$. To be more precise, we update the memories $\mathbf{w}^\mu$ and their duplicates $\mathbf{w}^{\text{dupli}, \mu}$ according to $\mathbf{w}^\mu \gets \mathbf{w}^\mu + \delta \mathbf{u}^\mu$ and $\mathbf{w}^{\text{dupli}, \mu} \gets \mathbf{w}^{\text{dupli}, \mu} - \delta \mathbf{u}^\mu$, respectively, where $\delta$ is a relatively small learning rate \cite{wu2019splitting}. In physics terminology, the eigenvectors $\mathbf{u}^\mu$ are excitation modes that break the permutation symmetries of the memories. $N$ and $P$ are generally large, so it is prohibitively expensive to store $\mathcal{S}_\mu \left( \mathbf{w}, \mathbf{p} \right)$ explicitly for all hidden units $1 \leq \mu \leq P$. Therefore, as proposed in \cite{wang2019energy}, we find the eigenvectors $\mathbf{u}^\mu$ and their eigenvalues $\lambda_{\text{min}}^\mu$ by minimizing the Rayleigh quotients $Q_\mu \left[ \mathbf{w}, \mathbf{p} \right] : S^{N - 1} \ni \mathbf{u}^\mu \mapsto \left[ \mathbf{u}^\mu \right]^T \mathcal{S}_\mu \left( \mathbf{w}, \mathbf{p} \right) \mathbf{u}^\mu$, which can be evaluated without constructing $\mathcal{S}_\mu \left( \mathbf{w}, \mathbf{p} \right)$ explicitly. To derive $Q_\mu \left[ \mathbf{w}, \mathbf{p} \right]$, we first compute $\Bar{\nabla}_{\mathbf{w}^\mu} \Bar{\nabla}_{\mathbf{w}^\mu}^T \exp \left( \beta_{\text{eff}} \sum_i w^\mu_i x_i \right)$. Given $f \left( \bm{\theta} \right) = \sum_i \theta_i x_i$, we find
\begin{align*}
    \Bar{\nabla}_{\bm{\theta}} \Bar{\nabla}_{\bm{\theta}}^T \exp \left( \beta_{\text{eff}} f \left( \bm{\theta} \right) \right) &= \exp \left( \beta_{\text{eff}} f \left( \bm{\theta} \right) \right) \left( \beta_{\text{eff}} \Bar{\nabla}_{\bm{\theta}} \Bar{\nabla}_{\bm{\theta}}^T f \left( \bm{\theta} \right) + \beta_{\text{eff}}^2 \Bar{\nabla}_{\bm{\theta}} f \left( \bm{\theta} \right) \Bar{\nabla}_{\bm{\theta}}^T f \left( \bm{\theta} \right) \right).
\end{align*}
As mentioned at the beginning of Appendix \ref{app:weight_normalization}, $\Bar{\nabla}_{\bm{\theta}} f \left( \bm{\theta} \right)$ is the unrestricted gradient of $f \left( \bm{\theta} \right)$ projected onto the tangent space of $\bm{\theta}$. It remains to calculate $\Bar{\nabla}_{\bm{\theta}} \Bar{\nabla}_{\bm{\theta}}^T f \left( \bm{\theta} \right)$. Following \cite{barilari2023lecture}, we find it to be the iterated projected gradient
\begin{align*}
    \left[ \Bar{\nabla}_{\bm{\theta}} \Bar{\nabla}_{\bm{\theta}}^T f \left( \bm{\theta} \right) \right]_{i \ell} &= \sum_k \left( \delta_{k \ell} - \theta_k \theta_\ell \right) \partial_{\theta_i} \left( \sum_j \left( \delta_{j k} - \theta_j \theta_k \right) \partial_{\theta_j} \left[ \sum_h \theta_h x_h \right] \right) \\ &= \sum_k \left( \delta_{k \ell} - \theta_k \theta_\ell \right) \partial_{\theta_i} \left( x_k - \left[ \sum_j \theta_j x_j \right] \theta_k \right) \\
    &= -\sum_k \left( \delta_{k \ell} - \theta_k \theta_\ell \right) \left( x_i \theta_k + \left[ \sum_j \theta_j x_j \right] \delta_{i k} \right) \\
    &= -\left[ \sum_j \theta_j x_j \right] \left( \delta_{i \ell} - \theta_i \theta_\ell \right),
\end{align*}
so we obtain
\begin{align*}
    Q_\mu \left[ \mathbf{w}, \mathbf{p} \right] \left( \mathbf{u}^\mu \right) &= -\mathbb{E}_{\mathbf{x}^*, y^*} \left[ \frac{p^\mu_y \exp \left( \beta_{\text{eff}} \sum_{i = 1}^N w^\mu_i x_i \right)}{\sum_{\nu = 1}^P p^{\nu}_y \exp \left( \beta_{\text{eff}} \sum_{i = 1}^N w^\nu_i x_i \right) + p^0_y \frac{\Omega_N \left( \beta \right)}{\Omega_N \left( 0 \right)}} F \left( \mathbf{u}^\mu ; \mathbf{w}^\mu, \mathbf{x} \right) \right] \\
    \text{where} \quad F \left( \bm{\varphi} ; \bm{\theta}, \mathbf{x} \right) &= \bm{\varphi}^T \left[ \beta_{\text{eff}} \bar{\nabla}_{\bm{\theta}} \bar{\nabla}_{\bm{\theta}}^T f \left( \bm{\theta} \right) + \beta_{\text{eff}}^2 \bar{\nabla}_{\bm{\theta}} f \left( \bm{\theta} \right) \bar{\nabla}_{\bm{\theta}}^T f \left( \bm{\theta} \right) \right] \bm{\varphi} \\
    &= \beta_{\text{eff}}^2 \left( \sum_{k} \varphi_k x_k - \left[ \sum_{k} \varphi_k \theta_k \right] \left[ \sum_{j} \theta_j x_j \right]\right)^2 \\
    &\quad+ \beta_{\text{eff}} \left[ \sum_j \theta_j x_j \right] \left[ \sum_{i} \varphi_i \theta_i - 1 \right] \left[ \sum_{\ell} \varphi_\ell \theta_\ell + 1 \right].
\end{align*}
We can directly minimize $Q \left[ \mathbf{w}, \mathbf{p} \right] \left( \mathbf{u} \right) = \sum_\mu Q_\mu \left[ \mathbf{w}, \mathbf{p} \right] \left( \mathbf{u}^\mu \right)$ to find the set of all eigenvectors $\mathbf{u}^\mu$ simultaneously. However, it is more convenient to integrate the eigenvector calculations into the DAM architecture. As such, we define the modified loss
\begin{align*}
    \mathcal{L}_\epsilon \left( \mathbf{w}, \mathbf{p}, \mathbf{u} \right) &= -\log \left[ \sum_{\mu = 1}^P p^\mu_y \exp \left( \varsigma \beta \sum_{i = 1}^N w^\mu_i x_i \right) \left[ 1 + \epsilon F \left( \mathbf{u}^\mu ; \mathbf{w}^\mu, \mathbf{x} \right) \right] + p^0_y \frac{\Omega_N \left( \beta \right)}{\Omega_N \left( 0 \right)} \right] \\
    &= -\log \left[ \sum_{\mu = 1}^P p^\mu_y \exp \left( \varsigma \beta \sum_{i = 1}^N w^\mu_i x_i + \log \left[ 1 + \epsilon F \left( \mathbf{u}^\mu ; \mathbf{w}^\mu, \mathbf{x} \right) \right] \right) + p^0_y \frac{\Omega_N \left( \beta \right)}{\Omega_N \left( 0 \right)} \right],
\end{align*}
so that the four gradients used to trained the DAM can be calculated using the equations
\begin{align*}
    \nabla_{\mathbf{u}} \mathcal{Q} \left[ \mathbf{w}, \mathbf{p} \right] \left( \mathbf{u} \right) &= \lim_{\epsilon \rightarrow 0} \left\{ \frac{1}{\epsilon} \nabla_{\mathbf{u}} \mathcal{L}_{\epsilon} \left( \mathbf{w}, \mathbf{p}, \mathbf{u} \right) \right\}, \\
    \nabla_{\mathbf{w}} \mathcal{L} \left( \mathbf{w}, \mathbf{p} \right) &= \nabla_{\mathbf{w}} \mathcal{L}_0 \left( \mathbf{w}, \mathbf{p}, \mathbf{u} \right) \\
    \quad \nabla_{\mathbf{p}} \mathcal{L} \left( \mathbf{w}, \mathbf{p} \right) &= \nabla_{\mathbf{p}} \mathcal{L}_0 \left( \mathbf{w}, \mathbf{p}, \mathbf{u} \right) \\
    \text{and} \quad \nabla_{\beta} \mathcal{L} \left( \mathbf{w}, \mathbf{p} \right) &= \nabla_{\beta} \mathcal{L}_0 \left( \mathbf{w}, \mathbf{p}, \mathbf{u} \right).
\end{align*}
This technique is based on the automatic differentiation trick proposed in \cite{wang2019energy}. We numerically implement the limit in the first equation by setting $\epsilon = 0$ during loss evaluation and $\epsilon = 1$ during gradient computation. When we optimize $\mathcal{Q} \left[ \mathbf{w}, \mathbf{p} \right] \left( \mathbf{u} \right)$, the minima of the different Rayleigh quotients $Q_\mu \left[ \mathbf{w}, \mathbf{p} \right] \left( \mathbf{u}^\mu \right)$ can span multiple orders of magnitude, so we normalize the gradients $\nabla_{\mathbf{u}^\mu} Q_\mu \left[ \mathbf{w}, \mathbf{p} \right] \left( \mathbf{u}^\mu \right)$ by a running average of their magnitudes to facilitate convergence. This is inspired by the use of RMSProp in \cite{wang2019energy}. Moreover, we constrain the eigenvectors $\mathbf{u}^\mu$ and the gradients $\nabla_{\mathbf{u}^\mu} Q_\mu \left[ \mathbf{w}, \mathbf{p} \right] \left( \mathbf{u}^\mu \right)$ to the unit hypersphere as we do for $\mathbf{w}^\mu$ (see Appendix \ref{app:weight_normalization}).
To make Fig. (\ref{fig:performance}), we train the eigenvectors for $1$ epoch, during which we monitor $\min_\mu \left\{ \frac{1}{2} \left[ \mathbf{u}^\mu \right]^T \nabla_{\mathbf{u}^\mu} \mathcal{Q} \left[ \mathbf{w}, \mathbf{p} \right] \left( \mathbf{u}^\mu \right) \right\} = \min_\mu \left\{ \mathcal{Q} \left[ \mathbf{w}, \mathbf{p} \right] \left( \mathbf{u}^\mu \right) \right\} \sim \min_\mu \left\{ \lambda_{\text{min}}^\mu \right\}$ as a metric. The Rayleigh quotients converge very quickly, and further training of the eigenvectors is not necessary.

\end{subappendices}

\chapter*{Conclusion}
\label{chap:conclusion}
In this thesis, we study various kinds of dense associative memory (DAM) models and restricted Boltzmann machines (RBMs) in the teacher-student setting.

In Chapter \ref{chap:dense_HN_paper}, we study a DAM called dense Hopfield network (dense HN). On the Nishimori line, we show that the phase transition where student dense HNs with a single pattern become capable of learning a dataset with $M$ samples generated by a teacher dense HN with a single pattern coincides with the spin-glass transition of dense HNs with $M$ random patterns. Outside the Nishimori line, we study the resistance of dense HNs to noise and adversarial attacks. In particular, we derive a formula quantifying the adversarial robustness of dense HNs at zero temperature, and we clarify why the adversarial robustness of dense HNs depends on the learning regime, as observed in \cite{krotov2018dense}.

In Chapter \ref{chap:RBM_paper}, we study RBMs with any finite number of hidden units. When the patterns incident to the $P^*$ hidden units of the teacher RBM are uncorrelated, we show that a student RBM with $P$ hidden units learns the data in the same way as $P$ separate RBMs with one hidden unit each, thus validating a conjecture formulated in \cite{barra2017phase}. Moreover, we show that student RBMs with more hidden units than their teacher learn a representation of the data where exactly $P^*$ of their patterns align themselves with those of the teacher while the $P - P^*$ remaining ones freeze in spin-glass states. We then argue that such RBMs can be used as toy models to study the lottery ticket hypothesis \cite{frankle2018lottery}. When the teacher patterns are correlated, we show that the student can adopt different learning strategies depending on the hyperparameters of the teacher-student setting and the number of samples in the training dataset.

In Chapter \ref{chap:DAM_paper}, we study a kind of DAM that fits into the framework of RBMs and is capable of both supervised and unsupervised classification. Based on insights from our theoretical analysis of the teacher-student setting, we propose a novel regularized loss function that makes training significantly more stable. Moreover, we show that the fixed points of relatively small DAMs are saddle points of larger DAMs, and we leverage this saddle-point hierarchy to considerably accelerate training using the splitting steepest descent algorithm introduced in \cite{wu2019splitting}.

A natural continuation of this work would be to study adversarial attacks in DAMs with many hidden units, such as the model studied in Chapter \ref{chap:DAM_paper} \cite{theriault2026saddle}. In particular, it would be interesting to see if the methods that we use to study adversarial attacks in Chapter \ref{chap:dense_HN_paper} can be generalized and used for this purpose in combination with those used to study adversarial attacks in linear models \cite{tanner2024high, vilucchio2024geometry, vilucchio2025existence}.

One could also extend this work by studying DAM and RBM training dynamics with analytical calculations, the statistical mechanics approach of \cite{decelle2017spectral, decelle2018thermodynamics} being a possible starting point. First, it would be interesting to build a theoretical model of the training dynamics of the lottery ticket experiment presented in Chapter \ref{chap:RBM_paper} \cite{theriault2025modeling}, as it could provide crucial insight into the lottery ticket hypothesis. In particular, it could help us to understand if the initial conditions that make certain subnetworks of large NNs converge especially quickly have common properties that can be exploited to design novel initialization schemes leading to faster training. Second, it would also be interesting to study analytically the learning dynamics of the DAM studied in Chapter \ref{chap:DAM_paper} \cite{theriault2026saddle}, particularly near saddle points, for a rigorous comparison of the standard training time with that of splitting steepest descent.

To further our study of implicitly low-dimensional learning, it would be interesting to investigate models with low-rank weights, such as in \cite{decelle2017spectral, decelle2018thermodynamics}. For example, one could study the teacher-student setting in which a student RBM (or DAM) with weights constrained to have a low rank learns data generated by a teacher whose weights are structured in a similar way for representing data lying on a low-dimensional manifold \cite{goldt2020modelling, gerace2021generalization}. This kind of study could be used to investigate the benefits of low-rank constraints in exploiting the structure of data to improve the speed and quality of learning.

Finally, the interpretable learning dynamics \cite{decelle2017spectral, decelle2018thermodynamics, boukacem2024waddington} of RBMs and DAMs make them promising surrogate models of biological systems. It was notably shown that DAM learning dynamics is similar to cellular differentiation \cite{boukacem2024waddington}, and it would be interesting to see if the splitting steepest descent algorithm used to train DAMs in Chapter \ref{chap:DAM_paper} \cite{theriault2026saddle} can be related to cellular division in an analogous way.
\pagebreak

\section*{Funding information}
This work was partially supported by project SERICS (PE00000014) under the MUR National Recovery and Resilience Plan funded by the European Union - NextGenerationEU. The work was also supported by the project PRIN22TANTARI "Statistical Mechanics of Learning Machines: from algorithmic and information-theoretical limits to new biologically inspired paradigms" 20229T9EAT – CUP J53D23003640001. 

\printbibliography

\end{document}